\documentclass[11pt]{article}

    \usepackage[breakable]{tcolorbox}
    \usepackage{parskip} % Stop auto-indenting (to mimic markdown behaviour)

    \usepackage{graphicx}
    \usepackage{caption}
    \DeclareCaptionFormat{nocaption}{}
    \usepackage{authblk}
    \usepackage{float}
    \usepackage{xcolor} % Allow colors to be defined
    \usepackage{enumerate} % Needed for markdown enumerations to work
    \usepackage{geometry} % Used to adjust the document margins
    \usepackage{amsmath} % Equations
    \usepackage{amssymb} % Equations
    \usepackage{textcomp} % defines textquotesingle
    \AtBeginDocument{%
        \def\PYZsq{\textquotesingle}% Upright quotes in Pygmentized code
    }
    \usepackage{upquote} % Upright quotes for verbatim code
    \usepackage{eurosym} % defines \euro

    \usepackage{iftex}
    \ifPDFTeX
        \usepackage[T1]{fontenc}
        \IfFileExists{alphabeta.sty}{
              \usepackage{alphabeta}
          }{
              \usepackage[mathletters]{ucs}
              \usepackage[utf8x]{inputenc}
          }
    \else
        \usepackage{fontspec}
        \usepackage{unicode-math}
    \fi

    \usepackage{fancyvrb} % verbatim replacement that allows latex
    \usepackage{grffile} % extends the file name processing of package graphics 
    \makeatletter % fix for old versions of grffile with XeLaTeX
    \@ifpackagelater{grffile}{2019/11/01}
    {
    }
    {
      \def\Gread@@xetex#1{%
        \IfFileExists{"\Gin@base".bb}%
        {\Gread@eps{\Gin@base.bb}}%
        {\Gread@@xetex@aux#1}%
      }
    }
    \makeatother
    \usepackage[Export]{adjustbox} % Used to constrain images to a maximum size
    \adjustboxset{max size={0.9\linewidth}{0.9\paperheight}}

    \usepackage{hyperref}
    \usepackage{titling}
    \usepackage{longtable} % longtable support required by pandoc >1.10
    \usepackage{booktabs}  % table support for pandoc > 1.12.2
    \usepackage{array}     % table support for pandoc >= 2.11.3
    \usepackage{calc}      % table minipage width calculation for pandoc >= 2.11.1
    \usepackage[inline]{enumitem} % IRkernel/repr support (it uses the enumerate* environment)
    \usepackage[normalem]{ulem} % ulem is needed to support strikethroughs (\sout)
    \usepackage{mathrsfs}

    \definecolor{urlcolor}{rgb}{0,.145,.698}
    \definecolor{linkcolor}{rgb}{.71,0.21,0.01}
    \definecolor{citecolor}{rgb}{.12,.54,.11}

    \definecolor{ansi-black}{HTML}{3E424D}
    \definecolor{ansi-black-intense}{HTML}{282C36}
    \definecolor{ansi-red}{HTML}{E75C58}
    \definecolor{ansi-red-intense}{HTML}{B22B31}
    \definecolor{ansi-green}{HTML}{00A250}
    \definecolor{ansi-green-intense}{HTML}{007427}
    \definecolor{ansi-yellow}{HTML}{DDB62B}
    \definecolor{ansi-yellow-intense}{HTML}{B27D12}
    \definecolor{ansi-blue}{HTML}{208FFB}
    \definecolor{ansi-blue-intense}{HTML}{0065CA}
    \definecolor{ansi-magenta}{HTML}{D160C4}
    \definecolor{ansi-magenta-intense}{HTML}{A03196}
    \definecolor{ansi-cyan}{HTML}{60C6C8}
    \definecolor{ansi-cyan-intense}{HTML}{258F8F}
    \definecolor{ansi-white}{HTML}{C5C1B4}
    \definecolor{ansi-white-intense}{HTML}{A1A6B2}
    \definecolor{ansi-default-inverse-fg}{HTML}{FFFFFF}
    \definecolor{ansi-default-inverse-bg}{HTML}{000000}

    \definecolor{outerrorbackground}{HTML}{FFDFDF}

    \providecommand{\tightlist}{%
      \setlength{\itemsep}{0pt}\setlength{\parskip}{0pt}}
    \DefineVerbatimEnvironment{Highlighting}{Verbatim}{commandchars=\\\{\}}
    \let\Oldtex\TeX
    \let\Oldlatex\LaTeX
    \renewcommand{\TeX}{\textrm{\Oldtex}}
    \renewcommand{\LaTeX}{\textrm{\Oldlatex}}
    \title{Tutorials on Stance Detection using Pre-trained Language Models: Fine-tuning BERT and Prompting Large Language Models}
	
	\author{Yun-Shiuan Chuang$^{1,2}$\\
	\small $^{1}$Department of Psychology, $^{2}$Department of Computer Science\\
	\small yunshiuan.chuang@wisc.edu}
        
    \date{}    
    
\makeatletter
\def\PY@reset{\let\PY@it=\relax \let\PY@bf=\relax%
    \let\PY@ul=\relax \let\PY@tc=\relax%
    \let\PY@bc=\relax \let\PY@ff=\relax}
\def\PY@tok#1{\csname PY@tok@#1\endcsname}
\def\PY@toks#1+{\ifx\relax#1\empty\else%
    \PY@tok{#1}\expandafter\PY@toks\fi}
\def\PY@do#1{\PY@bc{\PY@tc{\PY@ul{%
    \PY@it{\PY@bf{\PY@ff{#1}}}}}}}
\def\PY#1#2{\PY@reset\PY@toks#1+\relax+\PY@do{#2}}

\@namedef{PY@tok@w}{\def\PY@tc##1{\textcolor[rgb]{0.73,0.73,0.73}{##1}}}
\@namedef{PY@tok@c}{\let\PY@it=\textit\def\PY@tc##1{\textcolor[rgb]{0.24,0.48,0.48}{##1}}}
\@namedef{PY@tok@cp}{\def\PY@tc##1{\textcolor[rgb]{0.61,0.40,0.00}{##1}}}
\@namedef{PY@tok@k}{\let\PY@bf=\textbf\def\PY@tc##1{\textcolor[rgb]{0.00,0.50,0.00}{##1}}}
\@namedef{PY@tok@kp}{\def\PY@tc##1{\textcolor[rgb]{0.00,0.50,0.00}{##1}}}
\@namedef{PY@tok@kt}{\def\PY@tc##1{\textcolor[rgb]{0.69,0.00,0.25}{##1}}}
\@namedef{PY@tok@o}{\def\PY@tc##1{\textcolor[rgb]{0.40,0.40,0.40}{##1}}}
\@namedef{PY@tok@ow}{\let\PY@bf=\textbf\def\PY@tc##1{\textcolor[rgb]{0.67,0.13,1.00}{##1}}}
\@namedef{PY@tok@nb}{\def\PY@tc##1{\textcolor[rgb]{0.00,0.50,0.00}{##1}}}
\@namedef{PY@tok@nf}{\def\PY@tc##1{\textcolor[rgb]{0.00,0.00,1.00}{##1}}}
\@namedef{PY@tok@nc}{\let\PY@bf=\textbf\def\PY@tc##1{\textcolor[rgb]{0.00,0.00,1.00}{##1}}}
\@namedef{PY@tok@nn}{\let\PY@bf=\textbf\def\PY@tc##1{\textcolor[rgb]{0.00,0.00,1.00}{##1}}}
\@namedef{PY@tok@ne}{\let\PY@bf=\textbf\def\PY@tc##1{\textcolor[rgb]{0.80,0.25,0.22}{##1}}}
\@namedef{PY@tok@nv}{\def\PY@tc##1{\textcolor[rgb]{0.10,0.09,0.49}{##1}}}
\@namedef{PY@tok@no}{\def\PY@tc##1{\textcolor[rgb]{0.53,0.00,0.00}{##1}}}
\@namedef{PY@tok@nl}{\def\PY@tc##1{\textcolor[rgb]{0.46,0.46,0.00}{##1}}}
\@namedef{PY@tok@ni}{\let\PY@bf=\textbf\def\PY@tc##1{\textcolor[rgb]{0.44,0.44,0.44}{##1}}}
\@namedef{PY@tok@na}{\def\PY@tc##1{\textcolor[rgb]{0.41,0.47,0.13}{##1}}}
\@namedef{PY@tok@nt}{\let\PY@bf=\textbf\def\PY@tc##1{\textcolor[rgb]{0.00,0.50,0.00}{##1}}}
\@namedef{PY@tok@nd}{\def\PY@tc##1{\textcolor[rgb]{0.67,0.13,1.00}{##1}}}
\@namedef{PY@tok@s}{\def\PY@tc##1{\textcolor[rgb]{0.73,0.13,0.13}{##1}}}
\@namedef{PY@tok@sd}{\let\PY@it=\textit\def\PY@tc##1{\textcolor[rgb]{0.73,0.13,0.13}{##1}}}
\@namedef{PY@tok@si}{\let\PY@bf=\textbf\def\PY@tc##1{\textcolor[rgb]{0.64,0.35,0.47}{##1}}}
\@namedef{PY@tok@se}{\let\PY@bf=\textbf\def\PY@tc##1{\textcolor[rgb]{0.67,0.36,0.12}{##1}}}
\@namedef{PY@tok@sr}{\def\PY@tc##1{\textcolor[rgb]{0.64,0.35,0.47}{##1}}}
\@namedef{PY@tok@ss}{\def\PY@tc##1{\textcolor[rgb]{0.10,0.09,0.49}{##1}}}
\@namedef{PY@tok@sx}{\def\PY@tc##1{\textcolor[rgb]{0.00,0.50,0.00}{##1}}}
\@namedef{PY@tok@m}{\def\PY@tc##1{\textcolor[rgb]{0.40,0.40,0.40}{##1}}}
\@namedef{PY@tok@gh}{\let\PY@bf=\textbf\def\PY@tc##1{\textcolor[rgb]{0.00,0.00,0.50}{##1}}}
\@namedef{PY@tok@gu}{\let\PY@bf=\textbf\def\PY@tc##1{\textcolor[rgb]{0.50,0.00,0.50}{##1}}}
\@namedef{PY@tok@gd}{\def\PY@tc##1{\textcolor[rgb]{0.63,0.00,0.00}{##1}}}
\@namedef{PY@tok@gi}{\def\PY@tc##1{\textcolor[rgb]{0.00,0.52,0.00}{##1}}}
\@namedef{PY@tok@gr}{\def\PY@tc##1{\textcolor[rgb]{0.89,0.00,0.00}{##1}}}
\@namedef{PY@tok@ge}{\let\PY@it=\textit}
\@namedef{PY@tok@gs}{\let\PY@bf=\textbf}
\@namedef{PY@tok@gp}{\let\PY@bf=\textbf\def\PY@tc##1{\textcolor[rgb]{0.00,0.00,0.50}{##1}}}
\@namedef{PY@tok@go}{\def\PY@tc##1{\textcolor[rgb]{0.44,0.44,0.44}{##1}}}
\@namedef{PY@tok@gt}{\def\PY@tc##1{\textcolor[rgb]{0.00,0.27,0.87}{##1}}}
\@namedef{PY@tok@err}{\def\PY@bc##1{{\setlength{\fboxsep}{\string -\fboxrule}\fcolorbox[rgb]{1.00,0.00,0.00}{1,1,1}{\strut ##1}}}}
\@namedef{PY@tok@kc}{\let\PY@bf=\textbf\def\PY@tc##1{\textcolor[rgb]{0.00,0.50,0.00}{##1}}}
\@namedef{PY@tok@kd}{\let\PY@bf=\textbf\def\PY@tc##1{\textcolor[rgb]{0.00,0.50,0.00}{##1}}}
\@namedef{PY@tok@kn}{\let\PY@bf=\textbf\def\PY@tc##1{\textcolor[rgb]{0.00,0.50,0.00}{##1}}}
\@namedef{PY@tok@kr}{\let\PY@bf=\textbf\def\PY@tc##1{\textcolor[rgb]{0.00,0.50,0.00}{##1}}}
\@namedef{PY@tok@bp}{\def\PY@tc##1{\textcolor[rgb]{0.00,0.50,0.00}{##1}}}
\@namedef{PY@tok@fm}{\def\PY@tc##1{\textcolor[rgb]{0.00,0.00,1.00}{##1}}}
\@namedef{PY@tok@vc}{\def\PY@tc##1{\textcolor[rgb]{0.10,0.09,0.49}{##1}}}
\@namedef{PY@tok@vg}{\def\PY@tc##1{\textcolor[rgb]{0.10,0.09,0.49}{##1}}}
\@namedef{PY@tok@vi}{\def\PY@tc##1{\textcolor[rgb]{0.10,0.09,0.49}{##1}}}
\@namedef{PY@tok@vm}{\def\PY@tc##1{\textcolor[rgb]{0.10,0.09,0.49}{##1}}}
\@namedef{PY@tok@sa}{\def\PY@tc##1{\textcolor[rgb]{0.73,0.13,0.13}{##1}}}
\@namedef{PY@tok@sb}{\def\PY@tc##1{\textcolor[rgb]{0.73,0.13,0.13}{##1}}}
\@namedef{PY@tok@sc}{\def\PY@tc##1{\textcolor[rgb]{0.73,0.13,0.13}{##1}}}
\@namedef{PY@tok@dl}{\def\PY@tc##1{\textcolor[rgb]{0.73,0.13,0.13}{##1}}}
\@namedef{PY@tok@s2}{\def\PY@tc##1{\textcolor[rgb]{0.73,0.13,0.13}{##1}}}
\@namedef{PY@tok@sh}{\def\PY@tc##1{\textcolor[rgb]{0.73,0.13,0.13}{##1}}}
\@namedef{PY@tok@s1}{\def\PY@tc##1{\textcolor[rgb]{0.73,0.13,0.13}{##1}}}
\@namedef{PY@tok@mb}{\def\PY@tc##1{\textcolor[rgb]{0.40,0.40,0.40}{##1}}}
\@namedef{PY@tok@mf}{\def\PY@tc##1{\textcolor[rgb]{0.40,0.40,0.40}{##1}}}
\@namedef{PY@tok@mh}{\def\PY@tc##1{\textcolor[rgb]{0.40,0.40,0.40}{##1}}}
\@namedef{PY@tok@mi}{\def\PY@tc##1{\textcolor[rgb]{0.40,0.40,0.40}{##1}}}
\@namedef{PY@tok@il}{\def\PY@tc##1{\textcolor[rgb]{0.40,0.40,0.40}{##1}}}
\@namedef{PY@tok@mo}{\def\PY@tc##1{\textcolor[rgb]{0.40,0.40,0.40}{##1}}}
\@namedef{PY@tok@ch}{\let\PY@it=\textit\def\PY@tc##1{\textcolor[rgb]{0.24,0.48,0.48}{##1}}}
\@namedef{PY@tok@cm}{\let\PY@it=\textit\def\PY@tc##1{\textcolor[rgb]{0.24,0.48,0.48}{##1}}}
\@namedef{PY@tok@cpf}{\let\PY@it=\textit\def\PY@tc##1{\textcolor[rgb]{0.24,0.48,0.48}{##1}}}
\@namedef{PY@tok@c1}{\let\PY@it=\textit\def\PY@tc##1{\textcolor[rgb]{0.24,0.48,0.48}{##1}}}
\@namedef{PY@tok@cs}{\let\PY@it=\textit\def\PY@tc##1{\textcolor[rgb]{0.24,0.48,0.48}{##1}}}

\def\PYZbs{\char`\\}
\def\PYZus{\char`\_}
\def\PYZob{\char`\{}
\def\PYZcb{\char`\}}

\def\PYZgt{\char`\>}
\def\PYZsh{\char`\#}
\def\PYZpc{\char`\%}
\def\PYZdl{\char`\$}
\def\PYZhy{\char`\-}
\def\PYZsq{\char`\'}
\def\PYZdq{\char`\"}

\makeatother

    \makeatletter
        \newbox\Wrappedcontinuationbox 
        \newbox\Wrappedvisiblespacebox 
        \newcommand*\Wrappedvisiblespace {\textcolor{red}{\textvisiblespace}} 
        \newcommand*\Wrappedcontinuationsymbol {\textcolor{red}{\llap{\tiny$\m@th\hookrightarrow$}}} 
        \newcommand*\Wrappedcontinuationindent {3ex } 
        \newcommand*\Wrappedafterbreak {\kern\Wrappedcontinuationindent\copy\Wrappedcontinuationbox} 
        \newcommand*\Wrappedbreaksatspecials {% 
            \def\PYGZus{\discretionary{\char`\_}{\Wrappedafterbreak}{\char`\_}}% 
            \def\PYGZob{\discretionary{}{\Wrappedafterbreak\char`\{}{\char`\{}}% 
            \def\PYGZcb{\discretionary{\char`\}}{\Wrappedafterbreak}{\char`\}}}% 
            \def\PYGZca{\discretionary{\char`\^}{\Wrappedafterbreak}{\char`\^}}% 
            \def\PYGZam{\discretionary{\char`\&}{\Wrappedafterbreak}{\char`\&}}% 
            \def\PYGZlt{\discretionary{}{\Wrappedafterbreak\char`\<}{\char`\<}}% 
            \def\PYGZgt{\discretionary{\char`\>}{\Wrappedafterbreak}{\char`\>}}% 
            \def\PYGZsh{\discretionary{}{\Wrappedafterbreak\char`\#}{\char`\#}}% 
            \def\PYGZpc{\discretionary{}{\Wrappedafterbreak\char`\%}{\char`\%}}% 
            \def\PYGZdl{\discretionary{}{\Wrappedafterbreak\char`\$}{\char`\$}}% 
            \def\PYGZhy{\discretionary{\char`\-}{\Wrappedafterbreak}{\char`\-}}% 
            \def\PYGZsq{\discretionary{}{\Wrappedafterbreak\textquotesingle}{\textquotesingle}}% 
            \def\PYGZdq{\discretionary{}{\Wrappedafterbreak\char`\"}{\char`\"}}% 
            \def\PYGZti{\discretionary{\char`\~}{\Wrappedafterbreak}{\char`\~}}% 
        } 
        \newcommand*\Wrappedbreaksatpunct {% 
            \lccode`\~`\.\lowercase{\def~}{\discretionary{\hbox{\char`\.}}{\Wrappedafterbreak}{\hbox{\char`\.}}}% 
            \lccode`\~`\,\lowercase{\def~}{\discretionary{\hbox{\char`\,}}{\Wrappedafterbreak}{\hbox{\char`\,}}}% 
            \lccode`\~`\;\lowercase{\def~}{\discretionary{\hbox{\char`\;}}{\Wrappedafterbreak}{\hbox{\char`\;}}}% 
            \lccode`\~`\:\lowercase{\def~}{\discretionary{\hbox{\char`\:}}{\Wrappedafterbreak}{\hbox{\char`\:}}}% 
            \lccode`\~`\?\lowercase{\def~}{\discretionary{\hbox{\char`\?}}{\Wrappedafterbreak}{\hbox{\char`\?}}}% 
            \lccode`\~`\!\lowercase{\def~}{\discretionary{\hbox{\char`\!}}{\Wrappedafterbreak}{\hbox{\char`\!}}}% 
            \lccode`\~`\/\lowercase{\def~}{\discretionary{\hbox{\char`\/}}{\Wrappedafterbreak}{\hbox{\char`\/}}}% 
            \catcode`\.\active
            \catcode`\,\active 
            \catcode`\;\active
            \catcode`\:\active
            \catcode`\?\active
            \catcode`\!\active
            \catcode`\/\active 
            \lccode`\~`\~ 	
        }
    \makeatother

    \let\OriginalVerbatim=\Verbatim
    \makeatletter
    \renewcommand{\Verbatim}[1][1]{%
        \sbox\Wrappedcontinuationbox {\Wrappedcontinuationsymbol}%
        \sbox\Wrappedvisiblespacebox {\FV@SetupFont\Wrappedvisiblespace}%
        \def\FancyVerbFormatLine ##1{\hsize\linewidth
            \vtop{\raggedright\hyphenpenalty\z@\exhyphenpenalty\z@
                \doublehyphendemerits\z@\finalhyphendemerits\z@
                \strut ##1\strut}%
        }%
        \def\FV@Space {%
            \nobreak\hskip\z@ plus\fontdimen3\font minus\fontdimen4\font
            \discretionary{\copy\Wrappedvisiblespacebox}{\Wrappedafterbreak}
            {\kern\fontdimen2\font}%
        }%
        
        \Wrappedbreaksatspecials
        \OriginalVerbatim[#1,codes*=\Wrappedbreaksatpunct]%
    }
    \makeatother

    \definecolor{incolor}{HTML}{303F9F}
    \definecolor{outcolor}{HTML}{D84315}
    \definecolor{cellborder}{HTML}{CFCFCF}
    \definecolor{cellbackground}{HTML}{F7F7F7}
    
    \makeatletter
    \newcommand{\boxspacing}{\kern\kvtcb@left@rule\kern\kvtcb@boxsep}
    \makeatother
    \newcommand{\prompt}[4]{
        {\ttfamily\llap{{\color{#2}[#3]:\hspace{3pt}#4}}\vspace{-\baselineskip}}
    }

    \hypersetup{
      breaklinks=true,  % so long urls are correctly broken across lines
      colorlinks=true,
      urlcolor=urlcolor,
      linkcolor=linkcolor,
      citecolor=citecolor,
      }
\begin{document}
    
    \maketitle

\begin{abstract}
This paper presents two self-contained tutorials on stance detection in Twitter data using BERT fine-tuning and prompting large language models (LLMs). The first tutorial explains BERT architecture and tokenization, guiding users through training, tuning, and evaluating standard and domain-specific BERT models with HuggingFace transformers. The second focuses on constructing prompts and few-shot examples to elicit stances from ChatGPT and open-source FLAN-T5 without fine-tuning. Various prompting strategies are implemented and evaluated using confusion matrices and macro F1 scores. The tutorials provide code, visualizations, and insights revealing the strengths of few-shot ChatGPT and FLAN-T5 which outperform fine-tuned BERTs. By covering both model fine-tuning and prompting-based techniques in an accessible, hands-on manner, these tutorials enable learners to gain applied experience with cutting-edge methods for stance detection.
\end{abstract}

\section{Part 1: Stance Detection on Tweets with fine-tuning BERT}\label{stance-detection-on-tweets-using-nlp-methods---part-1}

Note: This tutorial consists of two separate Python notebooks. This
notebook is the first one. The second notebook can be found
\href{https://colab.research.google.com/drive/1IFr6Iz1YH9XBWUKcWZyTU-1QtxgYqrmX?usp=sharing}{here}.
I recommend that you go through the first notebook before the second one
as the second notebook builds on top of the first one.

\begin{enumerate}
\def\labelenumi{\arabic{enumi}.}
\tightlist
\item
  First notebook (this one): Fine-tuning BERT models: include standard
  BERT and domain-specific BERT
\end{enumerate}

\begin{itemize}
\tightlist
\item
  link:
  \url{https://colab.research.google.com/drive/1nxziaKStwRnSyOLI6pLNBaAnB_aB6IsE?usp=sharing}
\end{itemize}

\begin{enumerate}
\def\labelenumi{\arabic{enumi}.}
\setcounter{enumi}{1}
\tightlist
\item
  Second notebook: Prompting large language models (LLMs): include
  ChatGPT, FLAN-T5 and different prompt types (zero-shot, few-shot,
  chain-of-thought)
\end{enumerate}

\begin{itemize}
\tightlist
\item
  link:
  \url{https://colab.research.google.com/drive/1IFr6Iz1YH9XBWUKcWZyTU-1QtxgYqrmX?usp=sharing}
\end{itemize}

    \hypertarget{getting-started-overview-prerequisites-and-setup}{%
\subsection{Getting Started: Overview, Prerequisites, and
Setup}\label{getting-started-overview-prerequisites-and-setup}}

\textbf{Objective of the tutorial}: This tutorial will guide you through
the process of stance detection on tweets using two main approaches:
fine-tuning a BERT model and using large language models (LLMs).

\textbf{Prerequisites}:

\begin{itemize}
\tightlist
\item
  If you want to run the tutorial without editting the codes but want to
  understand the content

  \begin{itemize}
  \tightlist
  \item
    Basic Python skills: functions, classes, pandas, etc.
  \item
    Basic ML knowledge: train-validation-test split, F1 score, forward
    pass, backpropagation etc.
  \end{itemize}
\item
  Familiarity with NLP concepts is a plus, particularly with
  transformers. However, if you're not familiar with them, don't worry.
  I'll provide brief explanations in the tutorial, as well as links to
  fantastic in-depth resources throughout the text.
\end{itemize}

    \textbf{Acknowledgements}

\begin{itemize}
\tightlist
\item
  While the application of BERT on stance detection is my own work, some
  part of this tutorials, e.g., transformer and BERT, are inspired by
  the following tutorials. Some of the figures are also modified from
  the images in these tutorials. I highly recommend you to check them
  out if you want to learn more about transformers and BERT.

  \begin{itemize}
  \tightlist
  \item
    http://jalammar.github.io/illustrated-transformer/
  \item
    http://jalammar.github.io/illustrated-bert/
  \end{itemize}
\item
  This tutorial was created with the assistance of ChatGPT (GPT-4), a
  cutting-edge language model developed by OpenAI. The AI-aided writing
  process involved an iterative approach, where I provided the model
  with ideas for each section and GPT-4 transformed those ideas into
  well-structured paragraphs. Even the outline itself underwent a
  similar iterative process to refine and improve the tutorial
  structure. Following this, I fact-checked and revised the generated
  content, asking GPT-4 to make further revisions based on my
  evaluation, until I took over and finalized the content.
\end{itemize}

    \textbf{Setup}

    \begin{enumerate}
\def\labelenumi{\arabic{enumi}.}
\tightlist
\item
  Before we begin with Google Colab, please ensure that you have
  selected the GPU runtime. To do this, go to \texttt{Runtime}
  -\textgreater{} \texttt{Change\ runtime\ type} -\textgreater{}
  \texttt{Hardware\ accelerator} -\textgreater{} \texttt{GPU}. This will
  ensure that the note will run more efficiently and quickly.
\end{enumerate}

    \begin{enumerate}
\def\labelenumi{\arabic{enumi}.}
\setcounter{enumi}{1}
\tightlist
\item
  Now, let's download the content of this tutorial and install the
  necessary libraries by running the following cell.
\end{enumerate}

    \begin{tcolorbox}[breakable, size=fbox, boxrule=1pt, pad at break*=1mm,colback=cellbackground, colframe=cellborder]
\prompt{In}{incolor}{1}{\boxspacing}
\begin{Verbatim}[commandchars=\\\{\}]
\PY{k+kn}{from} \PY{n+nn}{os}\PY{n+nn}{.}\PY{n+nn}{path} \PY{k+kn}{import} \PY{n}{join}
\PY{n}{ON\PYZus{}COLAB} \PY{o}{=} \PY{k+kc}{True}
\PY{k}{if} \PY{n}{ON\PYZus{}COLAB}\PY{p}{:}
  \PY{o}{!}git\PY{+w}{ }clone\PY{+w}{ }\PYZhy{}\PYZhy{}single\PYZhy{}branch\PY{+w}{ }\PYZhy{}\PYZhy{}branch\PY{+w}{ }colab\PY{+w}{ }https://github.com/yunshiuan/prelim\PYZus{}stance\PYZus{}detection.git
  \PY{o}{!}python\PY{+w}{ }\PYZhy{}m\PY{+w}{ }pip\PY{+w}{ }install\PY{+w}{ }pandas\PY{+w}{ }datasets\PY{+w}{ }openai\PY{+w}{ }accelerate\PY{+w}{ }transformers\PY{+w}{ }transformers\PY{o}{[}sentencepiece\PY{o}{]}\PY{+w}{ }\PY{n+nv}{torch}\PY{o}{=}\PY{o}{=}\PY{l+m}{1}.12.1+cu113\PY{+w}{ }\PYZhy{}f\PY{+w}{ }https://download.pytorch.org/whl/torch\PYZus{}stable.html\PY{+w}{ }emoji\PY{+w}{ }\PYZhy{}q
  \PY{o}{\PYZpc{}}\PY{k}{cd} /content/prelim\PYZus{}stance\PYZus{}detection/scripts
\PY{k}{else}\PY{p}{:}
  \PY{c+c1}{\PYZsh{} if you are not on colab, you have to set up the environment by yourself. You would also need a machine with GPU.}
  \PY{o}{\PYZpc{}}\PY{k}{cd} scripts
\end{Verbatim}
\end{tcolorbox}

    \begin{Verbatim}[commandchars=\\\{\}]
Cloning into 'prelim\_stance\_detection'{\ldots}
remote: Enumerating objects: 513, done.
remote: Counting objects: 100\% (36/36), done.
remote: Compressing objects: 100\% (24/24), done.
remote: Total 513 (delta 21), reused 24 (delta 12), pack-reused 477
Receiving objects: 100\% (513/513), 58.56 MiB | 12.29 MiB/s, done.
Resolving deltas: 100\% (254/254), done.
     \textcolor{ansi-black-intense}{------------------------------------------------------------------------------------------------------------------------} \textcolor{ansi-green}{492.4/492.4
kB} \textcolor{ansi-red}{2.7 MB/s} eta \textcolor{ansi-cyan}{0:00:00}
     \textcolor{ansi-black-intense}{------------------------------------------------------------------------------------------------------------------------} \textcolor{ansi-green}{73.6/73.6 kB}
\textcolor{ansi-red}{2.4 MB/s} eta \textcolor{ansi-cyan}{0:00:00}
     \textcolor{ansi-black-intense}{------------------------------------------------------------------------------------------------------------------------} \textcolor{ansi-green}{244.2/244.2 kB}
\textcolor{ansi-red}{14.6 MB/s} eta \textcolor{ansi-cyan}{0:00:00}
     \textcolor{ansi-black-intense}{------------------------------------------------------------------------------------------------------------------------} \textcolor{ansi-green}{7.4/7.4 MB}
\textcolor{ansi-red}{43.5 MB/s} eta \textcolor{ansi-cyan}{0:00:00}
     \textcolor{ansi-black-intense}{------------------------------------------------------------------------------------------------------------------------} \textcolor{ansi-green}{1.8/1.8 GB}
\textcolor{ansi-red}{472.8 kB/s} eta \textcolor{ansi-cyan}{0:00:00}
     \textcolor{ansi-black-intense}{------------------------------------------------------------------------------------------------------------------------} \textcolor{ansi-green}{361.8/361.8 kB}
\textcolor{ansi-red}{28.6 MB/s} eta \textcolor{ansi-cyan}{0:00:00}
  Installing build dependencies {\ldots} done
  Getting requirements to build wheel {\ldots} done
  Preparing metadata (pyproject.toml) {\ldots} done
     \textcolor{ansi-black-intense}{------------------------------------------------------------------------------------------------------------------------} \textcolor{ansi-green}{115.3/115.3 kB}
\textcolor{ansi-red}{10.2 MB/s} eta \textcolor{ansi-cyan}{0:00:00}
     \textcolor{ansi-black-intense}{------------------------------------------------------------------------------------------------------------------------} \textcolor{ansi-green}{212.5/212.5 kB}
\textcolor{ansi-red}{13.5 MB/s} eta \textcolor{ansi-cyan}{0:00:00}
     \textcolor{ansi-black-intense}{------------------------------------------------------------------------------------------------------------------------} \textcolor{ansi-green}{134.8/134.8
kB} \textcolor{ansi-red}{2.5 MB/s} eta \textcolor{ansi-cyan}{0:00:00}
     \textcolor{ansi-black-intense}{------------------------------------------------------------------------------------------------------------------------} \textcolor{ansi-green}{268.8/268.8 kB}
\textcolor{ansi-red}{18.5 MB/s} eta \textcolor{ansi-cyan}{0:00:00}
     \textcolor{ansi-black-intense}{------------------------------------------------------------------------------------------------------------------------} \textcolor{ansi-green}{7.8/7.8 MB}
\textcolor{ansi-red}{48.2 MB/s} eta \textcolor{ansi-cyan}{0:00:00}
     \textcolor{ansi-black-intense}{------------------------------------------------------------------------------------------------------------------------} \textcolor{ansi-green}{1.3/1.3 MB}
\textcolor{ansi-red}{59.0 MB/s} eta \textcolor{ansi-cyan}{0:00:00}
     \textcolor{ansi-black-intense}{------------------------------------------------------------------------------------------------------------------------} \textcolor{ansi-green}{1.3/1.3 MB}
\textcolor{ansi-red}{63.9 MB/s} eta \textcolor{ansi-cyan}{0:00:00}
  Building wheel for emoji (pyproject.toml) {\ldots} done
\textcolor{ansi-red}{ERROR: pip's dependency resolver does not currently take into account all
the packages that are installed. This behaviour is the source of the following
dependency conflicts.
torchaudio 2.0.2+cu118 requires torch==2.0.1, but you have torch 1.12.1+cu113
which is incompatible.
torchdata 0.6.1 requires torch==2.0.1, but you have torch 1.12.1+cu113 which is
incompatible.
torchtext 0.15.2 requires torch==2.0.1, but you have torch 1.12.1+cu113 which is
incompatible.
torchvision 0.15.2+cu118 requires torch==2.0.1, but you have torch 1.12.1+cu113
which is incompatible.}\textcolor{ansi-red}{
}/content/prelim\_stance\_detection/scripts
    \end{Verbatim}

    \begin{tcolorbox}[breakable, size=fbox, boxrule=1pt, pad at break*=1mm,colback=cellbackground, colframe=cellborder]
\prompt{In}{incolor}{ }{\boxspacing}
\begin{Verbatim}[commandchars=\\\{\}]
\PY{c+c1}{\PYZsh{} a helper function to load images in the notebook}
\PY{k+kn}{from} \PY{n+nn}{IPython}\PY{n+nn}{.}\PY{n+nn}{display} \PY{k+kn}{import} \PY{n}{display}
\PY{k+kn}{from} \PY{n+nn}{PIL} \PY{k+kn}{import} \PY{n}{Image} \PY{k}{as} \PY{n}{PILImage}
\PY{k+kn}{from} \PY{n+nn}{parameters\PYZus{}meta} \PY{k+kn}{import} \PY{n}{ParametersMeta} \PY{k}{as} \PY{n}{par}
\PY{n}{PATH\PYZus{}IMAGES} \PY{o}{=} \PY{n}{join}\PY{p}{(}\PY{n}{par}\PY{o}{.}\PY{n}{PATH\PYZus{}ROOT}\PY{p}{,} \PY{l+s+s2}{\PYZdq{}}\PY{l+s+s2}{images}\PY{l+s+s2}{\PYZdq{}}\PY{p}{)}


\PY{k}{def} \PY{n+nf}{display\PYZus{}resized\PYZus{}image\PYZus{}in\PYZus{}notebook}\PY{p}{(}\PY{n}{file\PYZus{}image}\PY{p}{,} \PY{n}{scale}\PY{o}{=}\PY{l+m+mi}{1}\PY{p}{,} \PY{n}{use\PYZus{}default\PYZus{}path}\PY{o}{=}\PY{k+kc}{True}\PY{p}{)}\PY{p}{:}
\PY{+w}{    }\PY{l+s+sd}{\PYZdq{}\PYZdq{}\PYZdq{} Display an image in a notebook.}
\PY{l+s+sd}{    \PYZdq{}\PYZdq{}\PYZdq{}}
    \PY{c+c1}{\PYZsh{} \PYZhy{} https://stackoverflow.com/questions/69654877/how\PYZhy{}to\PYZhy{}set\PYZhy{}image\PYZhy{}size\PYZhy{}to\PYZhy{}display\PYZhy{}in\PYZhy{}ipython\PYZhy{}display}
    \PY{k}{if} \PY{n}{use\PYZus{}default\PYZus{}path}\PY{p}{:}
        \PY{n}{file\PYZus{}image} \PY{o}{=} \PY{n}{join}\PY{p}{(}\PY{n}{PATH\PYZus{}IMAGES}\PY{p}{,} \PY{n}{file\PYZus{}image}\PY{p}{)}
    \PY{n}{image} \PY{o}{=} \PY{n}{PILImage}\PY{o}{.}\PY{n}{open}\PY{p}{(}\PY{n}{file\PYZus{}image}\PY{p}{)}
    \PY{n}{display}\PY{p}{(}\PY{n}{image}\PY{o}{.}\PY{n}{resize}\PY{p}{(}\PY{p}{(}\PY{n+nb}{int}\PY{p}{(}\PY{n}{image}\PY{o}{.}\PY{n}{width} \PY{o}{*} \PY{n}{scale}\PY{p}{)}\PY{p}{,} \PY{n+nb}{int}\PY{p}{(}\PY{n}{image}\PY{o}{.}\PY{n}{height} \PY{o}{*} \PY{n}{scale}\PY{p}{)}\PY{p}{)}\PY{p}{)}\PY{p}{)}
\end{Verbatim}
\end{tcolorbox}

    \begin{center}\rule{0.5\linewidth}{0.5pt}\end{center}

    \hypertarget{what-is-stance-detection-and-why-is-it-important}{%
\subsection{What is Stance Detection and Why is it
Important?}\label{what-is-stance-detection-and-why-is-it-important}}

    Stance detection is an essential task in natural language processing
that aims to determine the attitude expressed by an author towards a
specific target, such as an entity, topic, or claim. The output of
stance detection is typically a categorical label, such as ``in-favor,''
``against,'' or ``neutral,'' indicating the stance of the author in
relation to the target. This task is critical for studying human belief
dynamics, e.g., how people influence each other's opinions and how
beliefs change over time. To better understand the complexities involved
in stance detection, let's consider an example related to the topic of
``abortion legalization''.

    For example, consider the following tweet:

``\textbf{\emph{A pregnancy, planned or unplanned, brings spouses,
families \& everyone closer to each other. \#Life is beautiful!
\#USA}}''

In this case, the stance expressed towards the topic of abortion
legalization might be inferred as against, but the clues indicating the
author's attitude are implicit and subtle - notice that it does not
explicitly mention abortion, making it challenging to determine the
stance without careful examination and contextual understanding.

    There are two key challenges in stance detection, especially when
working with large datasets like Twitter data. First, as illustrated
above, the underlying attitude expressed in the text is often subtle,
which requires domain knowledge and context to correctly label the
stance. Second, the corpus can be very large, with millions of tweets,
making it impractical to manually annotate all of them.

To address these challenges, we will leverage advanced natural language
processing (NLP) techniques including two paradigms, 1) fine-tuning BERT
model, and 2) prompting large language models (LLMs). I will elaborte
the details of these two approaches in the following sections.

    Before discussing the two paradigms for addressing the challenges in
stance detection, it's essential to understand the difference between
\textbf{sentiment analysis} and \textbf{stance detection}, as these two
tasks are often confused.

Sentiment analysis involves identifying the overall emotional tone
expressed in a piece of text, usually categorized as positive, negative,
or neutral. In contrast, stance detection aims to determine the specific
attitude of an author towards a target. While sentiment analysis focuses
on the general emotional valence of the text, stance detection requires
a deeper understanding of the author's position concerning the target
topic.

To illustrate that sentiment and stance are orthogonal concepts,
consider the following four examples, each representing a combination of
two stance types (against and in-favor) and two sentiments (positive and
negative):

    \begin{tcolorbox}[breakable, size=fbox, boxrule=1pt, pad at break*=1mm,colback=cellbackground, colframe=cellborder]
\prompt{In}{incolor}{ }{\boxspacing}
\begin{Verbatim}[commandchars=\\\{\}]
\PY{n}{display\PYZus{}resized\PYZus{}image\PYZus{}in\PYZus{}notebook}\PY{p}{(}\PY{l+s+s2}{\PYZdq{}}\PY{l+s+s2}{stance\PYZus{}vs\PYZus{}sentiment.png}\PY{l+s+s2}{\PYZdq{}}\PY{p}{,}\PY{l+m+mf}{0.7}\PY{p}{)}
\end{Verbatim}
\end{tcolorbox}

    \begin{center}
    \adjustimage{max size={0.9\linewidth}{0.9\paperheight}}{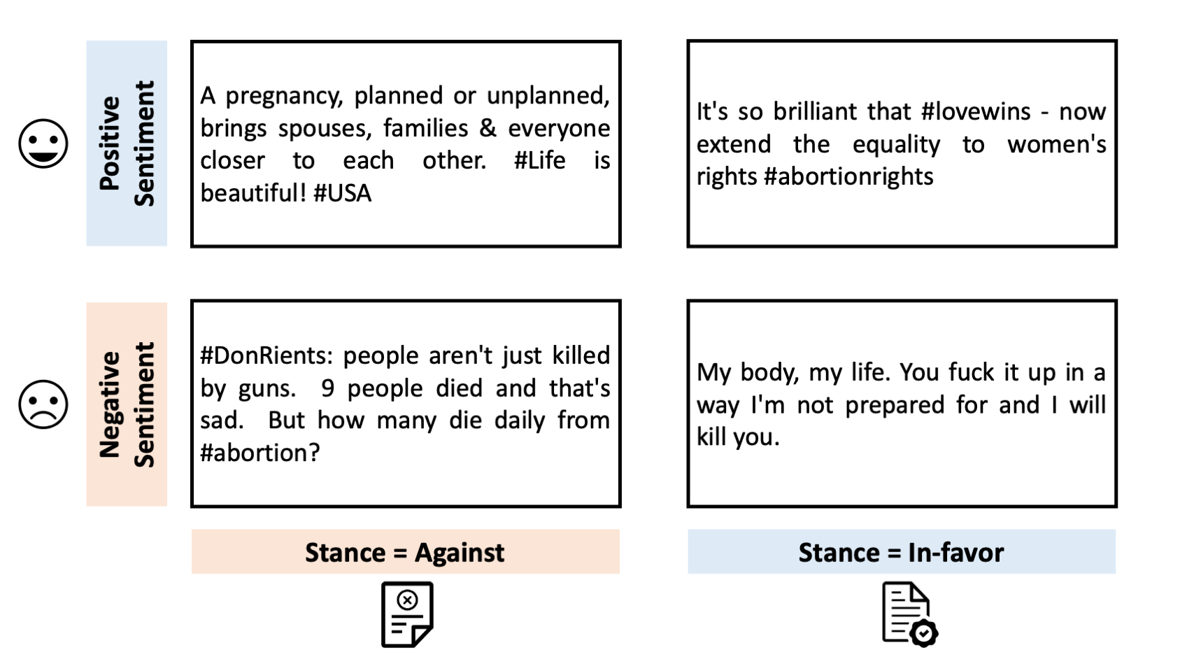}
    \end{center}
    { \hspace*{\fill} \\}
    
    In this tutorial, we will focus on stance detection in the context of
the ``Abortion'' topic using the SemEval-2016 dataset (data is publicly
available
\href{https://alt.qcri.org/semeval2016/task6/index.php?id=data-and-tools}{here}).
We chose the abortion topic because it is currently a hotly debated
issue, and it is important to understand public opinion on this matter.
We will analyze a dataset containing tweets about abortion, with each
tweet labeled as either in-favor, against, or neutral with respect to
the topic. My goal is to develop a model that can accurately identify
the stance expressed in these tweets.

    \begin{tcolorbox}[breakable, size=fbox, boxrule=1pt, pad at break*=1mm,colback=cellbackground, colframe=cellborder]
\prompt{In}{incolor}{ }{\boxspacing}
\begin{Verbatim}[commandchars=\\\{\}]
\PY{n}{display\PYZus{}resized\PYZus{}image\PYZus{}in\PYZus{}notebook}\PY{p}{(}\PY{l+s+s2}{\PYZdq{}}\PY{l+s+s2}{dataset\PYZus{}semeval\PYZus{}abortion.png}\PY{l+s+s2}{\PYZdq{}}\PY{p}{,}\PY{n}{scale} \PY{o}{=} \PY{l+m+mf}{0.25}\PY{p}{)}
\end{Verbatim}
\end{tcolorbox}

    \begin{center}
    \adjustimage{max size={0.9\linewidth}{0.9\paperheight}}{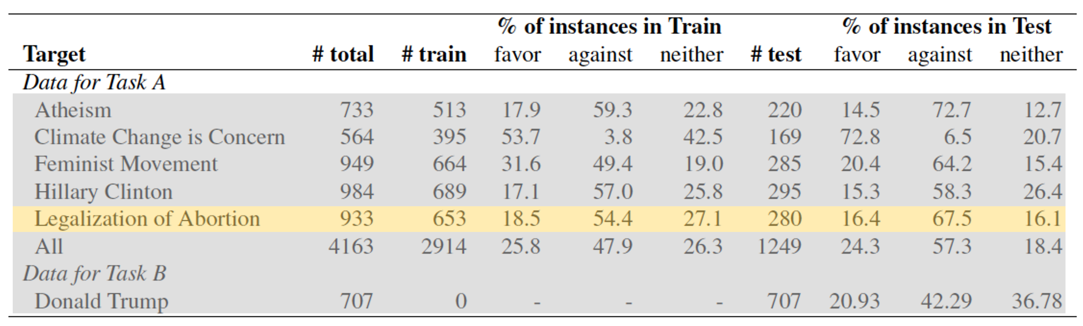}
    \end{center}
    { \hspace*{\fill} \\}
    
    \begin{quote}
Note: The SemEval-2016 dataset contains tweets related to six different
topics: Abortion, Atheism, Climate Change, Feminist Movement, Hillary
Clinton, and Legalization of Abortion. In this tutorial, we will focus
on the Abortion topic only. However, you can easily extend the tutorial
to other topics. For an interactive visualization of the SemEval-2016
dataset, please visit
\href{https://www.saifmohammad.com/WebPages/StanceDataset.htm}{here}.
\end{quote}

    \begin{center}\rule{0.5\linewidth}{0.5pt}\end{center}

    \hypertarget{two-stance-detection-paradigms}{%
\subsection{Two Stance Detection
Paradigms}\label{two-stance-detection-paradigms}}

    \begin{tcolorbox}[breakable, size=fbox, boxrule=1pt, pad at break*=1mm,colback=cellbackground, colframe=cellborder]
\prompt{In}{incolor}{ }{\boxspacing}
\begin{Verbatim}[commandchars=\\\{\}]
\PY{n}{display\PYZus{}resized\PYZus{}image\PYZus{}in\PYZus{}notebook}\PY{p}{(}\PY{l+s+s2}{\PYZdq{}}\PY{l+s+s2}{stance\PYZus{}detection\PYZus{}two\PYZus{}paradigm.png}\PY{l+s+s2}{\PYZdq{}}\PY{p}{,} \PY{l+m+mf}{0.7}\PY{p}{)}
\end{Verbatim}
\end{tcolorbox}

    \begin{center}
    \adjustimage{max size={0.9\linewidth}{0.9\paperheight}}{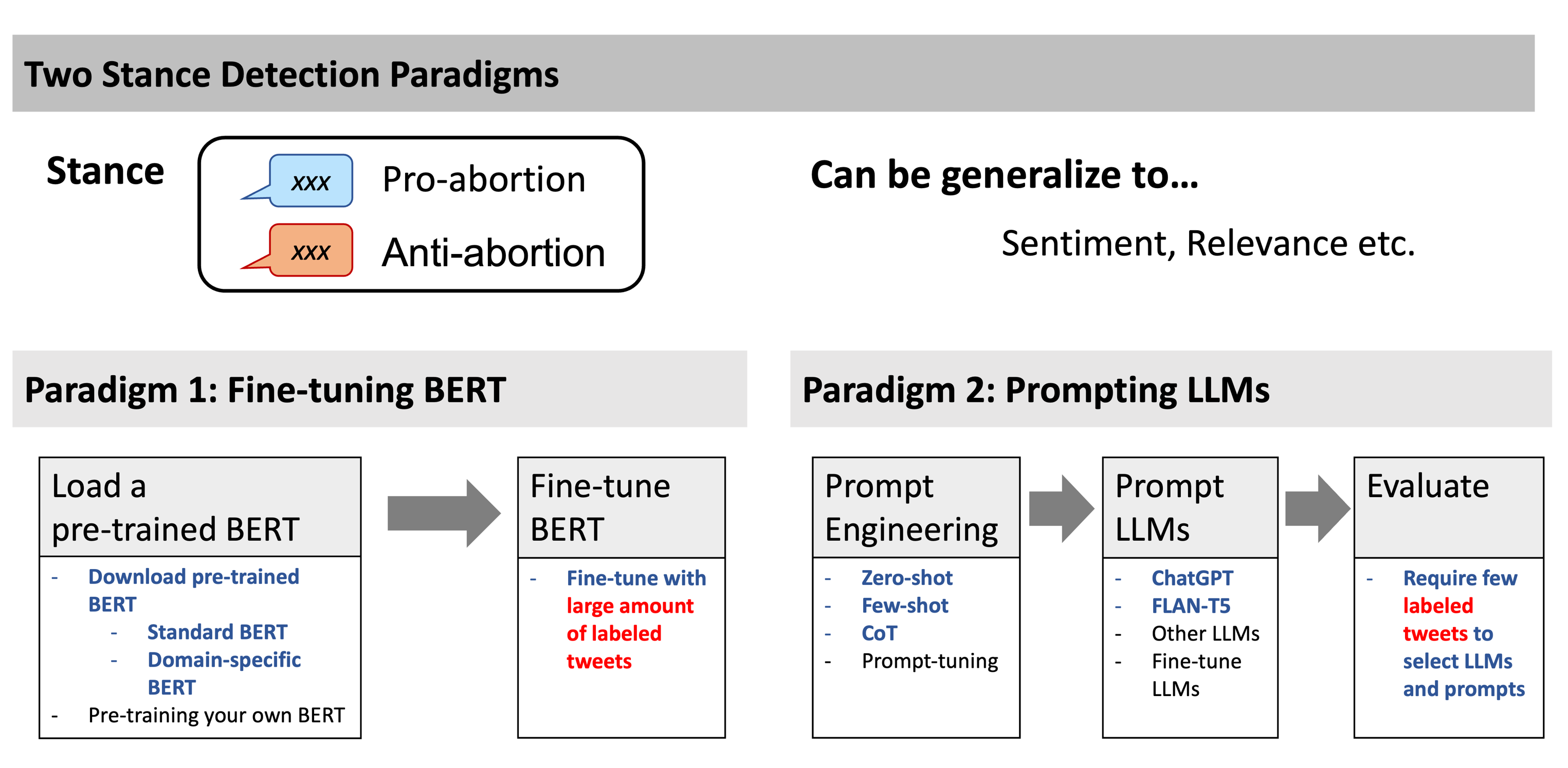}
    \end{center}
    { \hspace*{\fill} \\}
    
    \begin{quote}
The diagram above illustrates the two paradigms for stance detection:
(1) Fine-tuning a BERT model and (2) Prompting Large Language Models
(LLMs). The red text highlights the key practical difference between the
two approaches, which is the need for large labeled data when
fine-tuning a BERT model. The blue texts indicates the parts covered in
these two tutorials. While the black parts are not covered in these
tutorials, they are important to consider when applying these two
paradigms in practice.
\end{quote}

    In this tutorial, we are exploring two different paradigms for stance
detection: 1) fine-tuning a BERT model, and 2) prompting large language
models (LLMs) like ChatGPT.

Fine-tuning a BERT model involves training the model on a specific task
using a labeled dataset, which adapts the model's pre-existing knowledge
to the nuances of the task. This approach can yield strong performance
but typically requires a substantial amount of labeled data for the
target task.

On the other hand, prompting LLMs involves crafting carefully designed
input prompts that guide the model to generate desired outputs based on
its pre-trained knowledge. This method does not require additional
training, thus significantly reducing the amount of labeled data needed.
Note that some labeled data is still required to evaluate the
performance.

In this first tutorial, we will focus on the first paradigm: fine-tuning
a BERT model, including domain-specific BERT which may be more suitable
for our task. In
\href{https://colab.research.google.com/drive/1IFr6Iz1YH9XBWUKcWZyTU-1QtxgYqrmX?usp=sharing}{the
second tutorial}, we will explore the second paradigm: prompting LLMs.

    \begin{center}\rule{0.5\linewidth}{0.5pt}\end{center}

    \hypertarget{paradigm-1-using-bert-for-stance-detection}{%
\section{Paradigm 1: using BERT for stance
detection}\label{paradigm-1-using-bert-for-stance-detection}}

In this section, I will briefly introduce BERT, a powerful NLP model
that has been widely used in many NLP tasks. we will explain what BERT
is, how it is trained, and how it can be used for stance detection. we
will also show you how to fine-tune BERT for stance detection using
python.

    \hypertarget{what-is-bert-and-how-it-works}{%
\subsection{What is BERT and how it
works}\label{what-is-bert-and-how-it-works}}

BERT, which stands for \emph{\textbf{B}idirectional \textbf{E}ncoder
\textbf{R}epresentations from \textbf{T}ransformers}, is a
groundbreaking natural language processing (NLP) model that has taken
the world by storm. \href{https://arxiv.org/pdf/1810.04805.pdf}{Created
by researchers at google in 2018}, BERT is designed to learn useful
representations for words from unlabeled text, which can then be
tailored, or, ``fine-tuned'' for a wide range of NLP tasks, such as
stance detection, sentiment analysis, question-answering, among many.

    \begin{quote}
Note: ``Unlabeled text'' means that the text does not have any labels,
such as the sentiment, or stance, of a tweet. This is in contrast to
supervised learning, where the training data is labeled. In supervised
learning, the model learns to predict the labels of the text of the
training data. When pre-training BERT on unlabeled data, it learns to
predict the randomly masked out words in a sentence (explained in
details below).
\end{quote}

    In a nutshell, BERT is a powerful NLP model that leverages 1) the
transformer architecture and 2) the pre-training and fine-tuning
approach. we will explain these two concepts in more details below.

    \begin{quote}
Note: In this tutorial, my primary focus is on applying NLP models for
stance detection, and I won't be elabortating all the details of BERT.
If you're interested in learning more about BERT, I highly recommend
checking out the excellent interactive tutorial available at
http://jalammar.github.io/illustrated-bert/. This tutorial provides a
thorough and visually engaging explanation of BERT's inner workings.
Some of the plots in my tutorial are borrowed from this resource.
\end{quote}

    \hypertarget{bidirectional-context-understanding-context-in-both-directions}{%
\subsubsection{Bidirectional Context: Understanding Context in Both
Directions}\label{bidirectional-context-understanding-context-in-both-directions}}

Language is complex, and understanding it is no simple task. Traditional
NLP models (e.g., RNN; no worries if you don't know what RNN is) have
focused on reading text in one direction (e.g.,from left-to-right),
making it difficult for them to grasp the full context when trying to
understand a word. BERT, however, is designed to process text in both
directions, allowing it to understand the meaning of words based on the
words that come before and after them.

To explain how this is possible, we first need to understand what a
transformer is, and specifically, the critical ``self-attention
mechanism'' component that makes it possible for BERT to understand
context in both directions.

    \begin{center}\rule{0.5\linewidth}{0.5pt}\end{center}

    \hypertarget{a-powerful-backbone-architecture-transformers-with-self-attention-mechanism}{%
\subsubsection{A Powerful Backbone Architecture: Transformers with
Self-Attention
Mechanism}\label{a-powerful-backbone-architecture-transformers-with-self-attention-mechanism}}

BERT is built upon the \textbf{transformer architecture}, the critical
backbone of many state-of-the-art NLP models (including both BERT and
the LLMs described in the second tutorial), was introduced by Vaswani et
al.~in their 2017 paper,
``\href{https://proceedings.neurips.cc/paper/2017/file/3f5ee243547dee91fbd053c1c4a845aa-Paper.pdf}{Attention
Is All You Need.}''

The key component of the architecture is the ``\textbf{self-attention
mechanism}'', which helps the model identify important parts of the
input text and understand the relationships between words.

    Let's use the concrete example below to illustrate the self-attention
mechanism.

    \begin{tcolorbox}[breakable, size=fbox, boxrule=1pt, pad at break*=1mm,colback=cellbackground, colframe=cellborder]
\prompt{In}{incolor}{ }{\boxspacing}
\begin{Verbatim}[commandchars=\\\{\}]
\PY{n}{display\PYZus{}resized\PYZus{}image\PYZus{}in\PYZus{}notebook}\PY{p}{(}\PY{l+s+s2}{\PYZdq{}}\PY{l+s+s2}{example\PYZus{}sentence\PYZus{}self\PYZus{}attention.png}\PY{l+s+s2}{\PYZdq{}}\PY{p}{,}\PY{l+m+mf}{0.2}\PY{p}{)}
\end{Verbatim}
\end{tcolorbox}

    \begin{center}
    \adjustimage{max size={0.9\linewidth}{0.9\paperheight}}{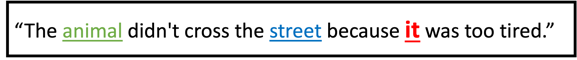}
    \end{center}
    { \hspace*{\fill} \\}
    
    In this example, what does ``\textbf{it}'' refer to? Does it refer to
the animal or the street?

As humans, we understand that ``it'' refers to the ``animal''. However,
for a machine, determining the correct reference is not a simple task,
especially given that the word ``street'' is closer to ``it'' than
``animal'' in the sentence. A naive machine might assume that ``it''
refers to the ``street'' because the word ``street'' is closer to ``it''
than ``animal''.

We, as humans, know that ``it'' refers to the ``animal'' because we
understand that animals can get tired while streets cannot. We also
recognize that being too tired is a legitimate reason for not crossing
the street. In summary, we can comprehend the meaning of the word ``it''
by taking into account other words in the sentence, or, in technical
terms, the ``context''.

    With the help of the self-attention mechanism, a transformer model takes
into account of the ``\textbf{context}'' of a word to understand its
meaning.

Let's use a diagram to show how this works. The figure below visualizes
how this work. On the left-hand side, the sentence is the input to the
self-attention mechanism, while on the right-hand side, the output is
also the same sentence (hence the name ``self-attention''). The lines
between the input and the output dipicts the ``attention weight'' of
each word. In this example, there are two ``attention heads'', the green
one and the orange one. Each head represents a different way of
understanding the meaning of the word ``it''.

Let's focus on the green one (``Head 1'') now. This attention head has a
high weight on the word ``tired'', which means that the attention weight
of the word ``tired'' is higher than other words when the model is
trying to understand the meaning of the word ``it''.

    \begin{tcolorbox}[breakable, size=fbox, boxrule=1pt, pad at break*=1mm,colback=cellbackground, colframe=cellborder]
\prompt{In}{incolor}{ }{\boxspacing}
\begin{Verbatim}[commandchars=\\\{\}]
\PY{n}{display\PYZus{}resized\PYZus{}image\PYZus{}in\PYZus{}notebook}\PY{p}{(}\PY{l+s+s2}{\PYZdq{}}\PY{l+s+s2}{self\PYZus{}attention\PYZus{}head\PYZus{}1.png}\PY{l+s+s2}{\PYZdq{}}\PY{p}{,}\PY{l+m+mf}{0.3}\PY{p}{)}
\end{Verbatim}
\end{tcolorbox}

    \begin{center}
    \adjustimage{max size={0.9\linewidth}{0.9\paperheight}}{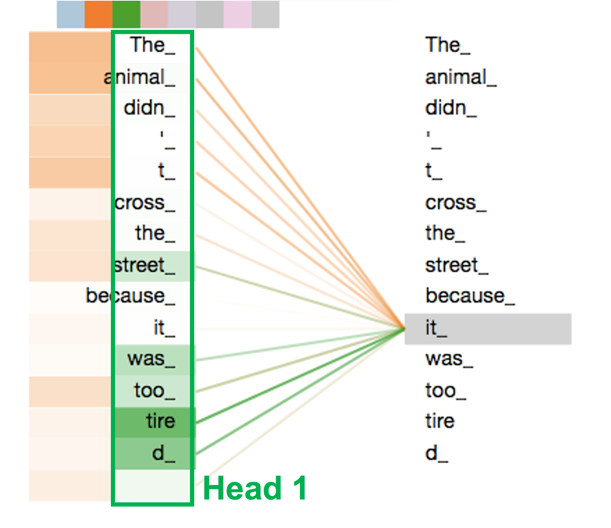}
    \end{center}
    { \hspace*{\fill} \\}
    
    \begin{quote}
Image modified from: http://jalammar.github.io/illustrated-transformer/
\end{quote}

    Let's now focus on the orange one (``Head 2''). This attention head has
a high weight on the word ``animal'', indicating that this attention
head cares more about the word ``animal'' when trying to understand the
meaning of the word ``it''.

    \begin{tcolorbox}[breakable, size=fbox, boxrule=1pt, pad at break*=1mm,colback=cellbackground, colframe=cellborder]
\prompt{In}{incolor}{ }{\boxspacing}
\begin{Verbatim}[commandchars=\\\{\}]
\PY{n}{display\PYZus{}resized\PYZus{}image\PYZus{}in\PYZus{}notebook}\PY{p}{(}\PY{l+s+s2}{\PYZdq{}}\PY{l+s+s2}{self\PYZus{}attention\PYZus{}head\PYZus{}2.png}\PY{l+s+s2}{\PYZdq{}}\PY{p}{,}\PY{l+m+mf}{0.3}\PY{p}{)}
\end{Verbatim}
\end{tcolorbox}

    \begin{center}
    \adjustimage{max size={0.9\linewidth}{0.9\paperheight}}{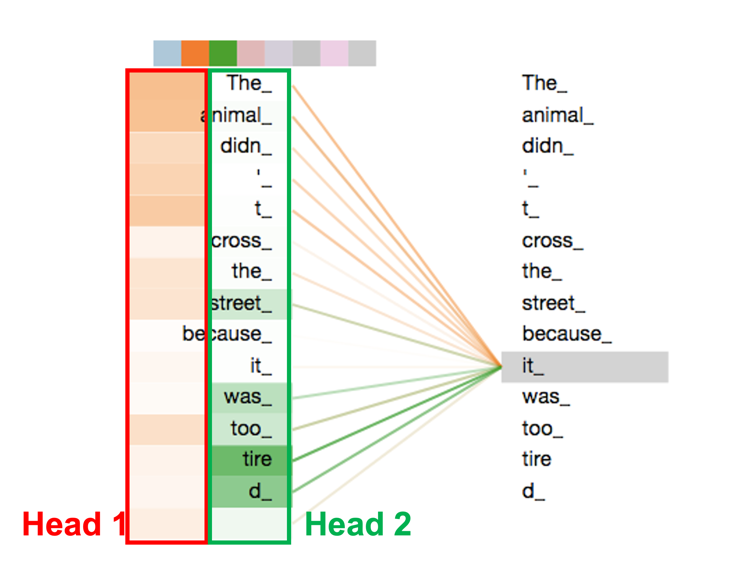}
    \end{center}
    { \hspace*{\fill} \\}
    
    \begin{quote}
Image modified from: http://jalammar.github.io/illustrated-transformer/
\end{quote}

    In the actual BERT model, there are 12 attention heads, meaning that the
model has 12 different ways of understanding the meaning of any word in
a sentence. After we combine the outputs of all 12 attention heads, we
then get the representation of the word ``it'' in the sentence after
this ``multi-head attention layer''. This multi-head attention layer,
along with other components (as shown below in the graph below), is
called an ``encoder''.

In the BERT model, for any given input sentence, this attention
mechanism is repeated 12 times (i.e., 12 encoders). Intuitively
speaking, every time the vector goes through an encoder, it learns a
more ``abstract'' relationship between words in a sentence.

The final product after these 12 layers is the
``\textbf{representation}'' of the input sentence. In total, there are
about 110 million trainable parameters in the BERT model.

To make a prediction (e.g., the stance) based on the
\textbf{representation}, this representation vector is then fed into a
linear layer to produce the final output of the model. In the case of
BERT, the representation is a vector of 768 numbers (the ``hidden
units'').

    \begin{quote}
Note: the number of layers, the number of attention heads, the number of
encoders, are based on the \texttt{bert-base} model, which is the
smaller variant of BERT. The larger variant, the \texttt{bert-large}
model, has 24 encoders, 16 attention heads, and 1024 hidden units,
amounting to about 340 million trainable parameters. In this tutorial,
we will be using the \texttt{bert-base} model.
\end{quote}

    \begin{tcolorbox}[breakable, size=fbox, boxrule=1pt, pad at break*=1mm,colback=cellbackground, colframe=cellborder]
\prompt{In}{incolor}{ }{\boxspacing}
\begin{Verbatim}[commandchars=\\\{\}]
\PY{n}{display\PYZus{}resized\PYZus{}image\PYZus{}in\PYZus{}notebook}\PY{p}{(}\PY{l+s+s2}{\PYZdq{}}\PY{l+s+s2}{bert\PYZus{}model\PYZus{}architecture.png}\PY{l+s+s2}{\PYZdq{}}\PY{p}{,}\PY{l+m+mf}{0.7}\PY{p}{)}
\end{Verbatim}
\end{tcolorbox}

    \begin{center}
    \adjustimage{max size={0.9\linewidth}{0.9\paperheight}}{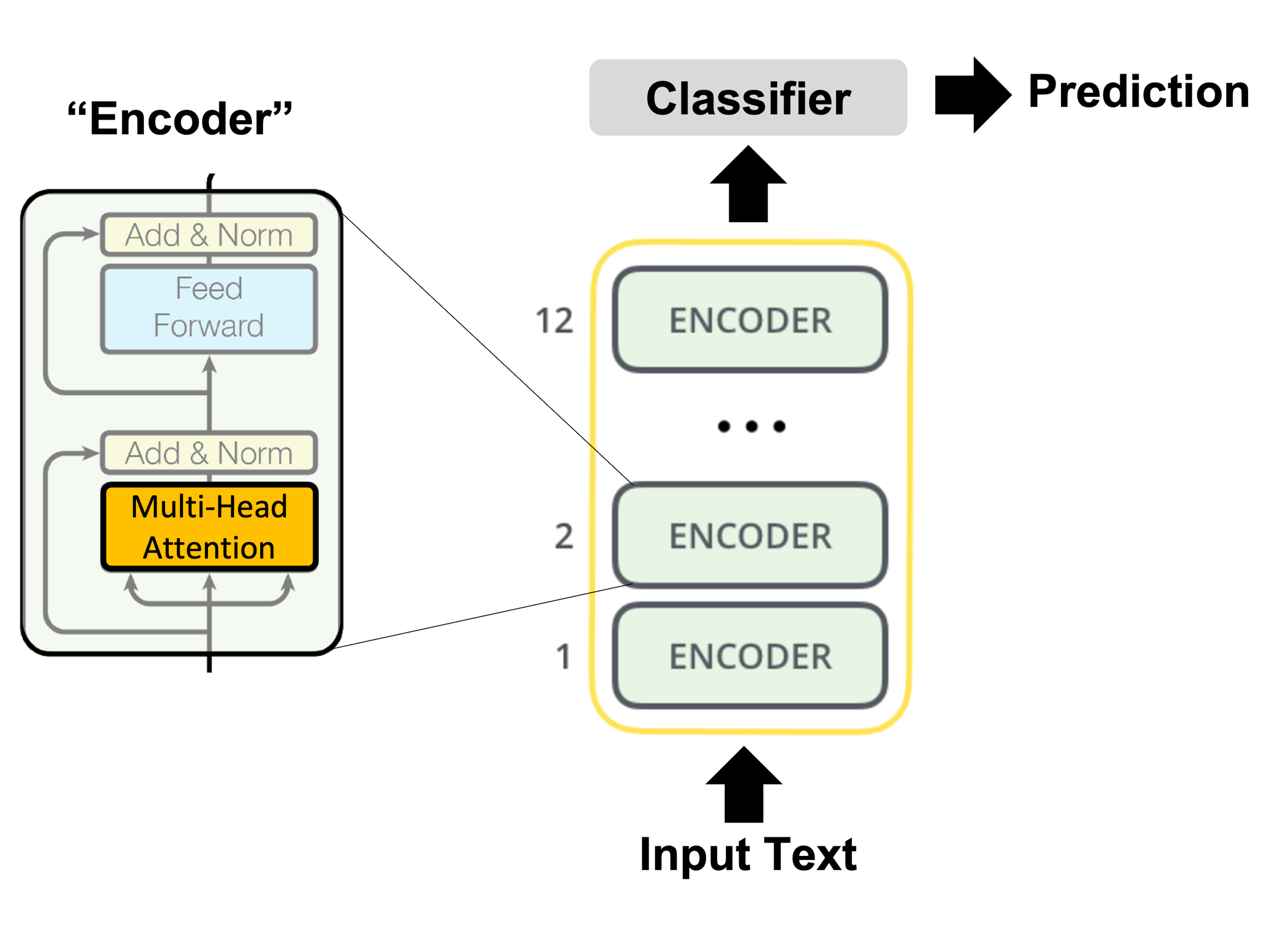}
    \end{center}
    { \hspace*{\fill} \\}
    
    \begin{quote}
Image modified from: http://jalammar.github.io/illustrated-bert/
\end{quote}

    \begin{quote}
Note: The actual self-attention mechanism is more complicated. The
``attention weight'' is computed by three trainable matrices - the
query, key, and value matrices.
\end{quote}

\begin{quote}
Likewise, although the self-attention layers are arguably the most
critical component, it is not the only component in a transformer. As
shown in the figure aboe, there are other building blocks like layer
normalization, residual connection, linear layers, positional encodings
etc. If you are interested in learning more about transformers in
detail, I highly recommend checking out the interactive tutorial on
transformers (by Jay Alammar, the same author of the BERT tutorial
linked above): http://jalammar.github.io/illustrated-transformer/. This
tutorial provides a comprehensive and visually engaging explanation of
the transformer architecture. Some plots in my tutorial are borrowed
from this resource. Also note that there are different variants of BERT
with different sizes of the transformer architecture. For example,
BERT-Base has 12 self-attention layers, while BERT-Large has 24
self-attention layers. In this tutorial, we will be using BERT-Base as a
running example.
\end{quote}

    \begin{center}\rule{0.5\linewidth}{0.5pt}\end{center}

    \hypertarget{pre-training-and-fine-tuning-learning-from-lots-of-text-and-adapting-to-specific-tasks}{%
\subsubsection{Pre-training and Fine-tuning: Learning from Lots of Text
and Adapting to Specific
Tasks}\label{pre-training-and-fine-tuning-learning-from-lots-of-text-and-adapting-to-specific-tasks}}

Now we know the architecture of BERT, which is a transformer model with
12 self-attention layers. But how does BERT learn to understand the
meaning of words? And how can we use BERT to solve specific NLP tasks,
say, stance detection?

Note that the BERT model contains around 110 million parameters,
necessitating a substantial amount of data for training. So, how can we
effectively train BERT when dealing with a specific task that has a
limited dataset? For instance, the Abortion dataset we used in this
tutorial comprises only 933 labeled tweets.

One of the key secrets behind BERT's success is its ability to 1) learn
from vast amounts of ``unlabled text'' and then 2) adapt that knowledge
to specific tasks with labels. These two components correspond to the
two stages when training a BERT model: 1) pre-training and 2)
fine-tuning.

    \hypertarget{pre-training-phase}{%
\paragraph{\texorpdfstring{1) \textbf{Pre-training
phase}}{1) Pre-training phase}}\label{pre-training-phase}}

During the initial pre-training phase, BERT is exposed to massive
amounts of unlabeled text (the raw text itself without any annotation
about sentiment, stance etc.). The standard BERT model was pretrained on
the entire English Wikipedia and 11k+ online books, which in total
contains about 3.3B words.

Why do we want to pre-train BERT on these corpora, even though they are
not related to the specific tasks we want to solve (i.e., the Abortion
tweet dataset)? The answer is that the pre-training phase allows BERT to
learn the general language understanding, for example, the meaning of
words, the relationships between words, and the context of words.

    \begin{tcolorbox}[breakable, size=fbox, boxrule=1pt, pad at break*=1mm,colback=cellbackground, colframe=cellborder]
\prompt{In}{incolor}{ }{\boxspacing}
\begin{Verbatim}[commandchars=\\\{\}]
\PY{n}{display\PYZus{}resized\PYZus{}image\PYZus{}in\PYZus{}notebook}\PY{p}{(}\PY{l+s+s2}{\PYZdq{}}\PY{l+s+s2}{bert\PYZus{}pretrain.png}\PY{l+s+s2}{\PYZdq{}}\PY{p}{,}\PY{l+m+mf}{0.5}\PY{p}{)}
\end{Verbatim}
\end{tcolorbox}

    \begin{center}
    \adjustimage{max size={0.9\linewidth}{0.9\paperheight}}{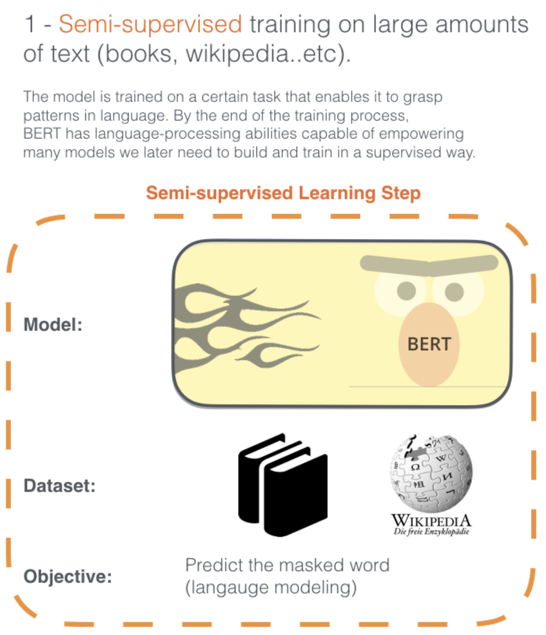}
    \end{center}
    { \hspace*{\fill} \\}
    
    \begin{quote}
Image modified from: http://jalammar.github.io/illustrated-bert/
\end{quote}

    In order to learn these general language understanding, in the
pre-training phase, BERT uses two different tasks: 1) masked language
modeling and 2) next sentence prediction. This phase allows BERT to
learn the relationships between words even without any task-specific
labels (e.g., stance labels are not needed for pre-training).

The masked language modeling task is a simple task where BERT is asked
to predict some ``masked out'' word in a sentence. For example, given
the sentence ``The animal didn't cross the street because it was too
tired'', when pre-training BERT, the word ``it'' may be masked out and
the model is asked to predict this missing word.

    \begin{tcolorbox}[breakable, size=fbox, boxrule=1pt, pad at break*=1mm,colback=cellbackground, colframe=cellborder]
\prompt{In}{incolor}{ }{\boxspacing}
\begin{Verbatim}[commandchars=\\\{\}]
\PY{n}{display\PYZus{}resized\PYZus{}image\PYZus{}in\PYZus{}notebook}\PY{p}{(}\PY{l+s+s2}{\PYZdq{}}\PY{l+s+s2}{bert\PYZus{}pretrain\PYZus{}mlm.png}\PY{l+s+s2}{\PYZdq{}}\PY{p}{,}\PY{l+m+mf}{0.7}\PY{p}{)}
\end{Verbatim}
\end{tcolorbox}

    \begin{center}
    \adjustimage{max size={0.9\linewidth}{0.9\paperheight}}{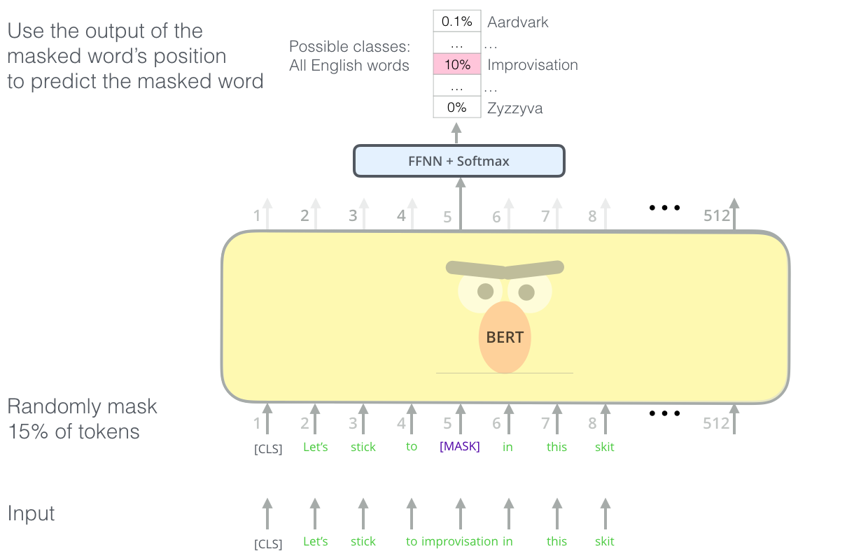}
    \end{center}
    { \hspace*{\fill} \\}
    
    \begin{quote}
Image modified from: http://jalammar.github.io/illustrated-bert/
\end{quote}

    The next sentence prediction task is a binary classification task where
BERT is asked to predict whether the second sentence is a continuation
of the first sentence. For example, if there is a paragraph in the
training data, where ``\emph{It was an sleepy dog.}'' is the second
sentence that follows the first sentence ``\emph{The animal didn't cross
the street because it was too tired.}'', then we say that the second
sentence is a continuation of the first sentence.

In this task, BERT is asked to decide whether a random sentence is a
continuation of another sentence.

    \begin{tcolorbox}[breakable, size=fbox, boxrule=1pt, pad at break*=1mm,colback=cellbackground, colframe=cellborder]
\prompt{In}{incolor}{ }{\boxspacing}
\begin{Verbatim}[commandchars=\\\{\}]
\PY{n}{display\PYZus{}resized\PYZus{}image\PYZus{}in\PYZus{}notebook}\PY{p}{(}\PY{l+s+s2}{\PYZdq{}}\PY{l+s+s2}{bert\PYZus{}pretrain\PYZus{}next\PYZus{}sentence\PYZus{}prediction.png}\PY{l+s+s2}{\PYZdq{}}\PY{p}{,}\PY{l+m+mf}{0.7}\PY{p}{)}
\end{Verbatim}
\end{tcolorbox}

    \begin{center}
    \adjustimage{max size={0.9\linewidth}{0.9\paperheight}}{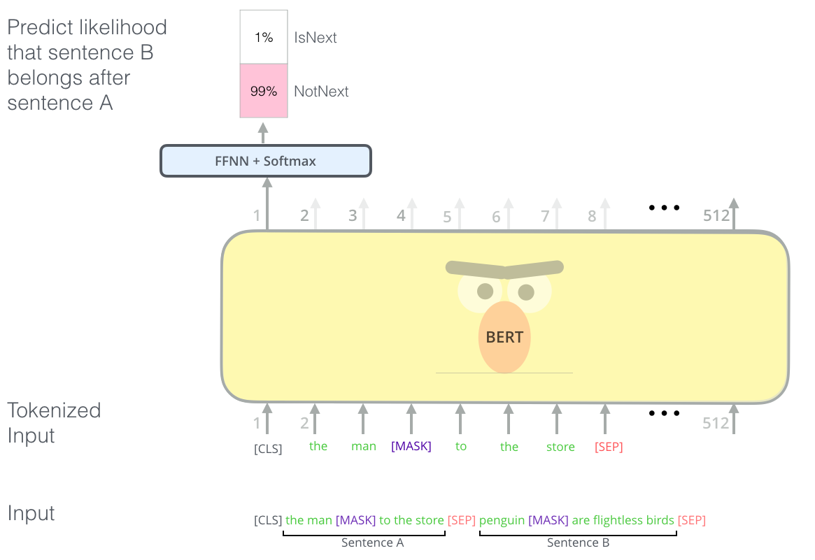}
    \end{center}
    { \hspace*{\fill} \\}
    
    \begin{quote}
Image modified from: http://jalammar.github.io/illustrated-bert/
\end{quote}

    \begin{quote}
Note: One caveat about pre-training is that, the more similar the
pre-training corpus is to the task-specific corpus, the better the
performance of BERT. For example, if you want to use BERT to solve a
stance detection task on tweets about abortion, it is better to
pre-train BERT on a corpus that is similar to this dataset. For example,
you can pre-train BERT on a corpus that contains tweets (rather than the
original Wikipedia and online books corpus). This makes sense because
the style of tweets is different from the style of Wikipedia and online
books (e.g., they are shorter and more informal). More about this in the
next section
\hyperref[considering-more-appropriate-pre-trained-models]{Considering More Appropriate Pre-trained Models}.
\end{quote}

    \hypertarget{fine-tuning-phase}{%
\paragraph{\texorpdfstring{2) \textbf{Fine-tuning
phase}}{2) Fine-tuning phase}}\label{fine-tuning-phase}}

After pre-training, BERT can be fine-tuned for a specific task with a
smaller labeled dataset (e.g., the Abortion tweet dataset). Fine-tuning
involves updating the model's weights using the labeled data, allowing
BERT to adapt its general language understanding to the specific task.
This process is relatively fast and requires less training data compared
to training a model from scratch.

    \begin{tcolorbox}[breakable, size=fbox, boxrule=1pt, pad at break*=1mm,colback=cellbackground, colframe=cellborder]
\prompt{In}{incolor}{ }{\boxspacing}
\begin{Verbatim}[commandchars=\\\{\}]
\PY{n}{display\PYZus{}resized\PYZus{}image\PYZus{}in\PYZus{}notebook}\PY{p}{(}\PY{l+s+s2}{\PYZdq{}}\PY{l+s+s2}{bert\PYZus{}fine\PYZus{}tune\PYZus{}stance.png}\PY{l+s+s2}{\PYZdq{}}\PY{p}{,}\PY{l+m+mf}{0.35}\PY{p}{)}
\end{Verbatim}
\end{tcolorbox}

    \begin{center}
    \adjustimage{max size={0.9\linewidth}{0.9\paperheight}}{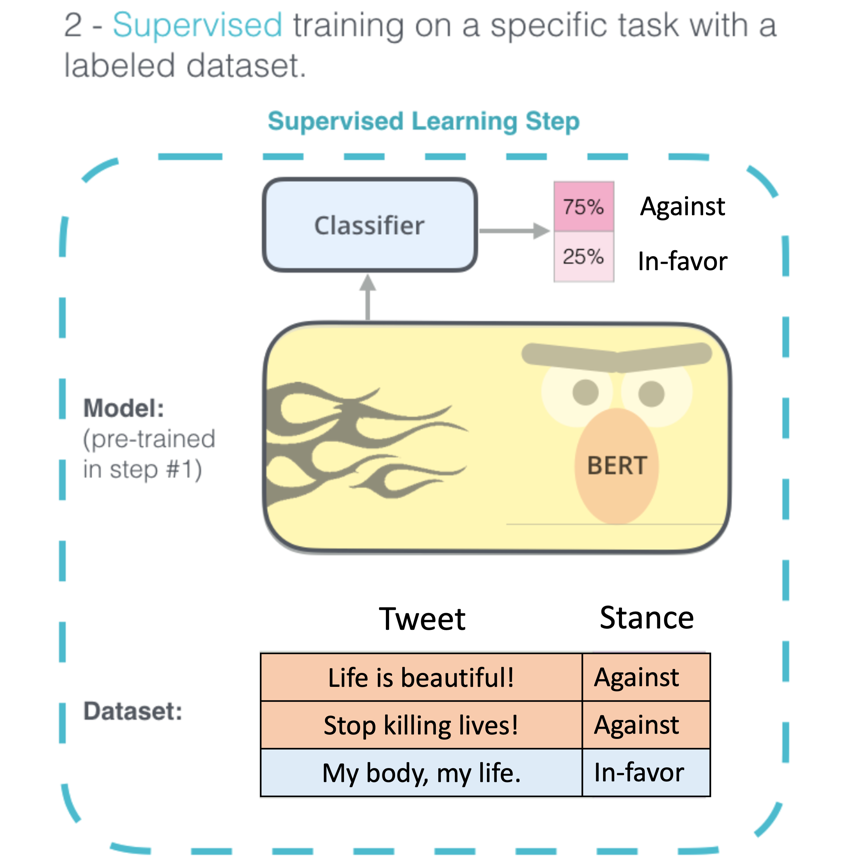}
    \end{center}
    { \hspace*{\fill} \\}
    
    \begin{quote}
Image modified from: http://jalammar.github.io/illustrated-bert/
\end{quote}

    \begin{center}\rule{0.5\linewidth}{0.5pt}\end{center}

    \hypertarget{berts-sub-word-tokenization}{%
\subsubsection{\texorpdfstring{\textbf{BERT's Sub-word
Tokenization}}{BERT's Sub-word Tokenization}}\label{berts-sub-word-tokenization}}

One caveat of BERT is that it requires a special
``subword-tokenization'' process (i.e., WordPiece tokenization). That
is, it does not directly encode each individual word, but rather encode
each word as a sequence of ``sub-word tokens''. For example, the word
``university'' can be broken down into the subwords ``uni'' and
``versity,'' which are more likely to appear in the corpus than the word
``university'' itself. This process of breaking down words into subwords
is called sub-word tokenization.

Sub-word tokenization is important for several reasons. Just to name two
important ones:

\hypertarget{consistent-representation-of-similar-words}{%
\paragraph{\texorpdfstring{\textbf{Consistent Representation of Similar
Words}}{Consistent Representation of Similar Words}}\label{consistent-representation-of-similar-words}}

Tokenization ensures that the text is represented in a consistent
manner, making it easier for the model to learn and identify patterns in
the data. By breaking the text into tokens, the model can focus on the
essential units of meaning, allowing it to better understand and analyze
the input. For an example, let us consider the following two words that
are commonly used in the abortion debate.: ``\textbf{pro-life}'' and
``\textbf{pro-choice}''.

Tokenization can help standardize the text by breaking them down into
smaller, overlapping tokens, i.e., \texttt{{[}"pro",\ "-",\ "life"{]}}
and \texttt{{[}"pro",\ "-",\ "choice"{]}}.

By representing the words as a sequence of tokens, the model can more
effectively identify the commonality between them (the shared ``pro-''
prefix) while also distinguishing the unique parts (``life'' and
``choice''). This approach helps the model learn the relationships
between word parts and the context (i.e., other words in the sentence)
in a more generalizable way. For example, sub-word tokenization enable
the model handle out-of-vocabulary words more effectively, as we will
see below.

\hypertarget{handling-out-of-vocabulary-words}{%
\paragraph{\texorpdfstring{\textbf{Handling Out-of-Vocabulary
Words}}{Handling Out-of-Vocabulary Words}}\label{handling-out-of-vocabulary-words}}

One of the challenges in NLP is dealing with words that the model has
not encountered during training, also known as out-of-vocabulary (OOV)
words. By using tokenization, BERT can handle OOV words more
effectively.

For example, suppose we have a sentence containing a relatively
newly-coined word: ``pro-birth''.

Here, the word ``\textbf{pro-birth}'' is a neologism that may not be
present in the model's vocabulary during pre-training, particularly if
the model was trained on older data. If we used a simple word-based
tokenization, the model would struggle to understand this word. However,
using a subword tokenization approach, the word can be broken down into
smaller parts that the model has likely seen before:

\texttt{{[}"pro",\ "-",\ "birth"{]}}

This breakdown allows the model to infer the meaning of the previously
unseen word based on the subword components it has encountered during
training. The model can recognize the ``pro'' prefix and the suffix
``birth''. This enables BERT to better understand these
out-of-vocabulary words, especially those that are relatively new or
coined, making it more robust and adaptable to a wide range of text
inputs.

    \begin{center}\rule{0.5\linewidth}{0.5pt}\end{center}

    \hypertarget{programming-exercise-fine-tuning-a-bert-model-with-huggingface}{%
\section{Programming Exercise: Fine-tuning a BERT Model with
HuggingFace}\label{programming-exercise-fine-tuning-a-bert-model-with-huggingface}}

Now, let's fine-tune a standard BERT model using the HuggingFace
Transformers library.

    \begin{quote}
Hugging Face, often called the ``GitHub'' for NLP models, provides an
extensive open-source Transformers library and a model hub, making it
easy to access, share, and implement state-of-the-art NLP models like
BERT (and other open-source LLMs, more on this in the second tutorial).
\end{quote}

    First, you need to decide whether you want to train the models on your
own or use the predictions I made and uploaded to my GitHub repo. If
you're running this notebook for the first time, I recommend setting
\texttt{DO\_TRAIN\_MODELS\ =\ False} (the default setting below) to save
time. This will load the precomputed predictions from my GitHub repo.

However, if you want to try training the models yourself, which I
encourage, you can set \texttt{DO\_TRAIN\_MODELS\ =\ True} and rerun the
notebook. If you're running this on Google Colab, ensure you're using
the GPU runtime for a more efficient and faster experience. To enable
this, go to \texttt{Runtime} -\textgreater{}
\texttt{Change\ runtime\ type} -\textgreater{}
\texttt{Hardware\ accelerator} -\textgreater{} \texttt{GPU}. Note that
even with the GPU runtime, running this entire notebook on Colab will
take about 10-15 minutes.

    \begin{quote}
Note: I have attempted to minimize randomness in the notebook by using a
random seed. However, the results you obtain may still vary slightly
from those in this notebook due to factors such as different library
versions or hardware configurations. To obtain the exact same results as
presented in this notebook, keep \texttt{DO\_TRAIN\_MODELS\ =\ False}
and use the precomputed predictions I made.
\end{quote}

    \begin{tcolorbox}[breakable, size=fbox, boxrule=1pt, pad at break*=1mm,colback=cellbackground, colframe=cellborder]
\prompt{In}{incolor}{ }{\boxspacing}
\begin{Verbatim}[commandchars=\\\{\}]
\PY{n}{DO\PYZus{}TRAIN\PYZus{}MODELS} \PY{o}{=} \PY{k+kc}{False}
\end{Verbatim}
\end{tcolorbox}

    \hypertarget{read-the-raw-data}{%
\subsection{Read the Raw Data}\label{read-the-raw-data}}

    \begin{tcolorbox}[breakable, size=fbox, boxrule=1pt, pad at break*=1mm,colback=cellbackground, colframe=cellborder]
\prompt{In}{incolor}{ }{\boxspacing}
\begin{Verbatim}[commandchars=\\\{\}]
\PY{o}{\PYZpc{}}\PY{k}{load\PYZus{}ext} autoreload
\PY{o}{\PYZpc{}}\PY{k}{autoreload} 2

\PY{k+kn}{from} \PY{n+nn}{data\PYZus{}processor} \PY{k+kn}{import} \PY{n}{SemEvalDataProcessor}
\PY{k+kn}{from} \PY{n+nn}{utils} \PY{k+kn}{import} \PY{n}{get\PYZus{}parameters\PYZus{}for\PYZus{}dataset}\PY{p}{,} \PY{n}{tidy\PYZus{}name}\PY{p}{,} \PY{n}{display\PYZus{}resized\PYZus{}image\PYZus{}in\PYZus{}notebook}
\end{Verbatim}
\end{tcolorbox}

    \begin{Verbatim}[commandchars=\\\{\}]
/home/sean/miniconda3/envs/prelim/lib/python3.10/site-packages/tqdm/auto.py:21:
TqdmWarning: IProgress not found. Please update jupyter and ipywidgets. See
https://ipywidgets.readthedocs.io/en/stable/user\_install.html
  from .autonotebook import tqdm as notebook\_tqdm
    \end{Verbatim}

    \begin{tcolorbox}[breakable, size=fbox, boxrule=1pt, pad at break*=1mm,colback=cellbackground, colframe=cellborder]
\prompt{In}{incolor}{ }{\boxspacing}
\begin{Verbatim}[commandchars=\\\{\}]
\PY{c+c1}{\PYZsh{} set up}
\PY{n}{SEED} \PY{o}{=} \PY{l+m+mi}{42}
\PY{n}{TOPIC\PYZus{}OF\PYZus{}INTEREST} \PY{o}{=} \PY{l+s+s2}{\PYZdq{}}\PY{l+s+s2}{Abortion}\PY{l+s+s2}{\PYZdq{}}
\PY{n}{DATASET} \PY{o}{=} \PY{l+s+s2}{\PYZdq{}}\PY{l+s+s2}{SEM\PYZus{}EVAL}\PY{l+s+s2}{\PYZdq{}}
\PY{n}{par} \PY{o}{=} \PY{n}{get\PYZus{}parameters\PYZus{}for\PYZus{}dataset}\PY{p}{(}\PY{n}{DATASET}\PY{p}{)}
\end{Verbatim}
\end{tcolorbox}

    \begin{tcolorbox}[breakable, size=fbox, boxrule=1pt, pad at break*=1mm,colback=cellbackground, colframe=cellborder]
\prompt{In}{incolor}{ }{\boxspacing}
\begin{Verbatim}[commandchars=\\\{\}]
\PY{c+c1}{\PYZsh{} read the raw data}
\PY{n}{sem\PYZus{}eval\PYZus{}data} \PY{o}{=} \PY{n}{SemEvalDataProcessor}\PY{p}{(}\PY{p}{)}
\PY{n}{df\PYZus{}raw\PYZus{}train} \PY{o}{=} \PY{n}{sem\PYZus{}eval\PYZus{}data}\PY{o}{.}\PY{n}{\PYZus{}read\PYZus{}raw\PYZus{}data}\PY{p}{(}\PY{n}{read\PYZus{}train}\PY{o}{=}\PY{k+kc}{True}\PY{p}{,}\PY{n}{read\PYZus{}test}\PY{o}{=}\PY{k+kc}{False}\PY{p}{,}\PY{n}{topic}\PY{o}{=}\PY{n}{TOPIC\PYZus{}OF\PYZus{}INTEREST}\PY{p}{)}
\PY{n}{df\PYZus{}raw\PYZus{}test} \PY{o}{=} \PY{n}{sem\PYZus{}eval\PYZus{}data}\PY{o}{.}\PY{n}{\PYZus{}read\PYZus{}raw\PYZus{}data}\PY{p}{(}\PY{n}{read\PYZus{}train}\PY{o}{=}\PY{k+kc}{False}\PY{p}{,}\PY{n}{read\PYZus{}test}\PY{o}{=}\PY{k+kc}{True}\PY{p}{,}\PY{n}{topic}\PY{o}{=}\PY{n}{TOPIC\PYZus{}OF\PYZus{}INTEREST}\PY{p}{)}
\end{Verbatim}
\end{tcolorbox}

    Let's look at the raw data first. The raw data is in the format below.

Each line contains a ID of the text, a target topic (which is
``Legalization of Abortion''), the raw tweet content, and a stance label
(i.e., \texttt{FAVOR}, \texttt{AGAINST}, \texttt{NONE}).

    \begin{tcolorbox}[breakable, size=fbox, boxrule=1pt, pad at break*=1mm,colback=cellbackground, colframe=cellborder]
\prompt{In}{incolor}{ }{\boxspacing}
\begin{Verbatim}[commandchars=\\\{\}]
\PY{n}{df\PYZus{}raw\PYZus{}train}\PY{p}{[}\PY{p}{[}\PY{l+s+s2}{\PYZdq{}}\PY{l+s+s2}{ID}\PY{l+s+s2}{\PYZdq{}}\PY{p}{,}\PY{l+s+s2}{\PYZdq{}}\PY{l+s+s2}{Target}\PY{l+s+s2}{\PYZdq{}}\PY{p}{,}\PY{l+s+s2}{\PYZdq{}}\PY{l+s+s2}{Tweet}\PY{l+s+s2}{\PYZdq{}}\PY{p}{,}\PY{l+s+s2}{\PYZdq{}}\PY{l+s+s2}{Stance}\PY{l+s+s2}{\PYZdq{}}\PY{p}{]}\PY{p}{]}\PY{o}{.}\PY{n}{head}\PY{p}{(}\PY{p}{)}
\end{Verbatim}
\end{tcolorbox}

            \begin{tcolorbox}[breakable, size=fbox, boxrule=.5pt, pad at break*=1mm, opacityfill=0]
\prompt{Out}{outcolor}{ }{\boxspacing}
\begin{Verbatim}[commandchars=\\\{\}]
        ID                    Target  \textbackslash{}
2211  2312  Legalization of Abortion
2212  2313  Legalization of Abortion
2213  2314  Legalization of Abortion
2214  2315  Legalization of Abortion
2215  2316  Legalization of Abortion

                                                  Tweet   Stance
2211  I really don't understand how some people are {\ldots}  AGAINST
2212  Let's agree that it's not ok to kill a 7lbs ba{\ldots}  AGAINST
2213  @glennbeck I would like to see poll: How many {\ldots}  AGAINST
2214  Democrats are always AGAINST "Personhood" or w{\ldots}  AGAINST
2215  @CultureShifting "If you don't draw the line w{\ldots}     NONE
\end{Verbatim}
\end{tcolorbox}
        
    \begin{tcolorbox}[breakable, size=fbox, boxrule=1pt, pad at break*=1mm,colback=cellbackground, colframe=cellborder]
\prompt{In}{incolor}{ }{\boxspacing}
\begin{Verbatim}[commandchars=\\\{\}]
\PY{n+nb}{print}\PY{p}{(}\PY{l+s+s2}{\PYZdq{}}\PY{l+s+s2}{number of tweets in the training data: }\PY{l+s+s2}{\PYZdq{}}\PY{p}{,}\PY{n+nb}{len}\PY{p}{(}\PY{n}{df\PYZus{}raw\PYZus{}train}\PY{p}{)}\PY{p}{)}
\end{Verbatim}
\end{tcolorbox}

    \begin{Verbatim}[commandchars=\\\{\}]
number of tweets in the training data:  603
    \end{Verbatim}

    The testing data below has the same format. Note that this set is not
used for training, but for evaluating a trained model's performance on
unseen data.

    \begin{tcolorbox}[breakable, size=fbox, boxrule=1pt, pad at break*=1mm,colback=cellbackground, colframe=cellborder]
\prompt{In}{incolor}{ }{\boxspacing}
\begin{Verbatim}[commandchars=\\\{\}]
\PY{n}{df\PYZus{}raw\PYZus{}test}\PY{p}{[}\PY{p}{[}\PY{l+s+s2}{\PYZdq{}}\PY{l+s+s2}{ID}\PY{l+s+s2}{\PYZdq{}}\PY{p}{,}\PY{l+s+s2}{\PYZdq{}}\PY{l+s+s2}{Target}\PY{l+s+s2}{\PYZdq{}}\PY{p}{,}\PY{l+s+s2}{\PYZdq{}}\PY{l+s+s2}{Tweet}\PY{l+s+s2}{\PYZdq{}}\PY{p}{,}\PY{l+s+s2}{\PYZdq{}}\PY{l+s+s2}{Stance}\PY{l+s+s2}{\PYZdq{}}\PY{p}{]}\PY{p}{]}\PY{o}{.}\PY{n}{head}\PY{p}{(}\PY{p}{)}
\end{Verbatim}
\end{tcolorbox}

            \begin{tcolorbox}[breakable, size=fbox, boxrule=.5pt, pad at break*=1mm, opacityfill=0]
\prompt{Out}{outcolor}{ }{\boxspacing}
\begin{Verbatim}[commandchars=\\\{\}]
        ID                    Target  \textbackslash{}
969  10970  Legalization of Abortion
970  10971  Legalization of Abortion
971  10972  Legalization of Abortion
972  10973  Legalization of Abortion
973  10974  Legalization of Abortion

                                                 Tweet   Stance
969  Need a ProLife R.E. Agent? - Support a ProLife{\ldots}  AGAINST
970  Where is the childcare program @joanburton whi{\ldots}  AGAINST
971  I get several requests with petitions to save {\ldots}  AGAINST
972  we must always see others as Christ sees us,we{\ldots}  AGAINST
973  PRAYERS FOR BABIES Urgent prayer one in Lexing{\ldots}  AGAINST
\end{Verbatim}
\end{tcolorbox}
        
    \begin{tcolorbox}[breakable, size=fbox, boxrule=1pt, pad at break*=1mm,colback=cellbackground, colframe=cellborder]
\prompt{In}{incolor}{ }{\boxspacing}
\begin{Verbatim}[commandchars=\\\{\}]
\PY{n+nb}{print}\PY{p}{(}\PY{l+s+s2}{\PYZdq{}}\PY{l+s+s2}{number of tweets in the test data: }\PY{l+s+s2}{\PYZdq{}}\PY{p}{,}\PY{n+nb}{len}\PY{p}{(}\PY{n}{df\PYZus{}raw\PYZus{}test}\PY{p}{)}\PY{p}{)}
\end{Verbatim}
\end{tcolorbox}

    \begin{Verbatim}[commandchars=\\\{\}]
number of tweets in the test data:  280
    \end{Verbatim}

    Let's look at some examples of the raw tweets.

    \begin{tcolorbox}[breakable, size=fbox, boxrule=1pt, pad at break*=1mm,colback=cellbackground, colframe=cellborder]
\prompt{In}{incolor}{ }{\boxspacing}
\begin{Verbatim}[commandchars=\\\{\}]
\PY{c+c1}{\PYZsh{} convert to list}
\PY{n}{df\PYZus{}raw\PYZus{}train}\PY{p}{[}\PY{p}{[}\PY{l+s+s2}{\PYZdq{}}\PY{l+s+s2}{Tweet}\PY{l+s+s2}{\PYZdq{}}\PY{p}{]}\PY{p}{]}\PY{o}{.}\PY{n}{values}\PY{p}{[}\PY{p}{[}\PY{l+m+mi}{7}\PY{p}{,}\PY{l+m+mi}{21}\PY{p}{]}\PY{p}{]}\PY{o}{.}\PY{n}{tolist}\PY{p}{(}\PY{p}{)}
\end{Verbatim}
\end{tcolorbox}

            \begin{tcolorbox}[breakable, size=fbox, boxrule=.5pt, pad at break*=1mm, opacityfill=0]
\prompt{Out}{outcolor}{ }{\boxspacing}
\begin{Verbatim}[commandchars=\\\{\}]
[['RT @createdequalorg: "We\textbackslash{}'re all human, aren\textbackslash{}'t we? Every human life is worth
the same, and worth saving." -J.K. Rowling \#{\ldots} \#SemST'],
 ['Follow \#Patriot --> @Enuffis2Much.  Thanks for following back!!  \#Truth
\#Liberty \#Justice \#ProIsrael \#WakeUpAmerica \#FreeAmirNow \#SemST']]
\end{Verbatim}
\end{tcolorbox}
        
    We can see that the tweets are not very clean.

For example, the first tweet contains a retweet tag (i.e., ``RT
@createdequalorg''). This tag entails that the tweet is a retweet of
another tweet. This messsage is not part of the content of the original
tweet, and thus should be removed.

    Aside from the retweet tag, the tweets also contain some other noise,
such as some special characters (e.g., /'). We will also remove these
special characters.

In addition, the mentions (e.g., ``@Enuffis2Much'') contains the
reference to other users. These mentions may confuse the model and
should be removed as well.

    \begin{quote}
Note: In practice, these non-language features can be leveraged to
improve the model's performance for various text sources. However, we
will not be exploring that approach in this tutorial to maintain a
general focus on the core techniques and to accommodate a wide range of
text types.
\end{quote}

    Finally, all tweets end with a special hashtag (e.g., ``\#SemST'').
These hashtags are added by the owners of the SemST dataset to indicate
the stance of the tweet, and are not part of the original tweet content.
We will also remove these special hashtags.

    \hypertarget{preprocess-the-raw-data}{%
\subsection{Preprocess the Raw Data}\label{preprocess-the-raw-data}}

We will preprocess the raw data to address the issues mentioned above.

Aside from preprocessing the raw tweets, we will also partition the
training data into a training set and a validation set (with a 4:1
ratio). The validation set will be used to evaluate the model's
performance during training.

    \begin{tcolorbox}[breakable, size=fbox, boxrule=1pt, pad at break*=1mm,colback=cellbackground, colframe=cellborder]
\prompt{In}{incolor}{ }{\boxspacing}
\begin{Verbatim}[commandchars=\\\{\}]
\PY{c+c1}{\PYZsh{} preprocess the raw tweets}
\PY{n}{sem\PYZus{}eval\PYZus{}data}\PY{o}{.}\PY{n}{preprocess}\PY{p}{(}\PY{p}{)}
\PY{n}{df\PYZus{}processed} \PY{o}{=} \PY{n}{sem\PYZus{}eval\PYZus{}data}\PY{o}{.}\PY{n}{\PYZus{}read\PYZus{}preprocessed\PYZus{}data}\PY{p}{(}\PY{n}{topic}\PY{o}{=}\PY{n}{TOPIC\PYZus{}OF\PYZus{}INTEREST}\PY{p}{)}\PY{o}{.}\PY{n}{reset\PYZus{}index}\PY{p}{(}\PY{n}{drop}\PY{o}{=}\PY{k+kc}{True}\PY{p}{)}
\end{Verbatim}
\end{tcolorbox}

    \begin{tcolorbox}[breakable, size=fbox, boxrule=1pt, pad at break*=1mm,colback=cellbackground, colframe=cellborder]
\prompt{In}{incolor}{ }{\boxspacing}
\begin{Verbatim}[commandchars=\\\{\}]
\PY{c+c1}{\PYZsh{} partition the data into train, vali, test sets}
\PY{n}{df\PYZus{}partitions} \PY{o}{=} \PY{n}{sem\PYZus{}eval\PYZus{}data}\PY{o}{.}\PY{n}{partition\PYZus{}processed\PYZus{}data}\PY{p}{(}\PY{n}{seed}\PY{o}{=}\PY{n}{SEED}\PY{p}{,}\PY{n}{verbose}\PY{o}{=}\PY{k+kc}{False}\PY{p}{)}
\end{Verbatim}
\end{tcolorbox}

    Let's look at the preprocessed data. The first thing to notice is that
there is a new column called ``partitions''. This column indicates
whether the tweet belongs to the training set, validation set, or
testing set.

    \begin{tcolorbox}[breakable, size=fbox, boxrule=1pt, pad at break*=1mm,colback=cellbackground, colframe=cellborder]
\prompt{In}{incolor}{ }{\boxspacing}
\begin{Verbatim}[commandchars=\\\{\}]
\PY{n}{df\PYZus{}processed}\PY{o}{.}\PY{n}{head}\PY{p}{(}\PY{p}{)}
\end{Verbatim}
\end{tcolorbox}

            \begin{tcolorbox}[breakable, size=fbox, boxrule=.5pt, pad at break*=1mm, opacityfill=0]
\prompt{Out}{outcolor}{ }{\boxspacing}
\begin{Verbatim}[commandchars=\\\{\}]
     ID                                              tweet     topic    label  \textbackslash{}
0  2312  i really don't understand how some people are {\ldots}  Abortion  AGAINST
1  2313  let's agree that it's not ok to kill a 7lbs ba{\ldots}  Abortion  AGAINST
2  2314  @USERNAME i would like to see poll: how many a{\ldots}  Abortion  AGAINST
3  2315  democrats are always against 'personhood' or w{\ldots}  Abortion  AGAINST
4  2316  @USERNAME 'if you don't draw the line where i'{\ldots}  Abortion     NONE

  partition
0     train
1     train
2     train
3     train
4     train
\end{Verbatim}
\end{tcolorbox}
        
    Let's look at the preprocessed tweets.

We can see that the tweets are now much cleaner. For example, the
retweet tag, special characters, and special hashtags have been removed.
The mention tags are replaced by a sentinel token (i.e., ``@USERNAME'').

    \begin{tcolorbox}[breakable, size=fbox, boxrule=1pt, pad at break*=1mm,colback=cellbackground, colframe=cellborder]
\prompt{In}{incolor}{ }{\boxspacing}
\begin{Verbatim}[commandchars=\\\{\}]
\PY{n}{df\PYZus{}processed}\PY{p}{[}\PY{p}{[}\PY{l+s+s2}{\PYZdq{}}\PY{l+s+s2}{tweet}\PY{l+s+s2}{\PYZdq{}}\PY{p}{]}\PY{p}{]}\PY{o}{.}\PY{n}{values}\PY{p}{[}\PY{p}{[}\PY{l+m+mi}{7}\PY{p}{,}\PY{l+m+mi}{21}\PY{p}{]}\PY{p}{]}\PY{o}{.}\PY{n}{tolist}\PY{p}{(}\PY{p}{)}
\end{Verbatim}
\end{tcolorbox}

            \begin{tcolorbox}[breakable, size=fbox, boxrule=.5pt, pad at break*=1mm, opacityfill=0]
\prompt{Out}{outcolor}{ }{\boxspacing}
\begin{Verbatim}[commandchars=\\\{\}]
[["'we're all human, aren't we? every human life is worth the same, and worth
saving.' -j.k. rowling \#{\ldots}"],
 ['follow \#patriot --> @USERNAME. thanks for following back!! \#truth \#liberty
\#justice \#proisrael \#wakeupamerica \#freeamirnow']]
\end{Verbatim}
\end{tcolorbox}
        
    Let's look at the distribution of the stance labels across the training,
validation, and testing sets.

    \begin{tcolorbox}[breakable, size=fbox, boxrule=1pt, pad at break*=1mm,colback=cellbackground, colframe=cellborder]
\prompt{In}{incolor}{ }{\boxspacing}
\begin{Verbatim}[commandchars=\\\{\}]
\PY{c+c1}{\PYZsh{} add a \PYZdq{}count\PYZdq{} column to count the number of tweets in each partition}
\PY{n}{df\PYZus{}label\PYZus{}dist} \PY{o}{=} \PY{n}{df\PYZus{}partitions}\PY{p}{[}\PY{n}{df\PYZus{}partitions}\PY{o}{.}\PY{n}{topic} \PY{o}{==} \PY{n}{TOPIC\PYZus{}OF\PYZus{}INTEREST}\PY{p}{]}\PY{o}{.}\PY{n}{value\PYZus{}counts}\PY{p}{(}\PY{p}{[}\PY{l+s+s1}{\PYZsq{}}\PY{l+s+s1}{partition}\PY{l+s+s1}{\PYZsq{}}\PY{p}{,}\PY{l+s+s1}{\PYZsq{}}\PY{l+s+s1}{label}\PY{l+s+s1}{\PYZsq{}}\PY{p}{]}\PY{p}{)}\PY{o}{.}\PY{n}{sort\PYZus{}index}\PY{p}{(}\PY{p}{)}
\end{Verbatim}
\end{tcolorbox}

    \begin{tcolorbox}[breakable, size=fbox, boxrule=1pt, pad at break*=1mm,colback=cellbackground, colframe=cellborder]
\prompt{In}{incolor}{ }{\boxspacing}
\begin{Verbatim}[commandchars=\\\{\}]
\PY{k+kn}{import} \PY{n+nn}{seaborn} \PY{k}{as} \PY{n+nn}{sns}
\PY{k+kn}{import} \PY{n+nn}{matplotlib}\PY{n+nn}{.}\PY{n+nn}{pyplot} \PY{k}{as} \PY{n+nn}{plt}

\PY{n}{df\PYZus{}label\PYZus{}dist\PYZus{}plot} \PY{o}{=} \PY{n}{df\PYZus{}label\PYZus{}dist}\PY{o}{.}\PY{n}{reset\PYZus{}index}\PY{p}{(}\PY{p}{)}


\PY{c+c1}{\PYZsh{} Create the bar plot}
\PY{n}{plt}\PY{o}{.}\PY{n}{figure}\PY{p}{(}\PY{n}{figsize}\PY{o}{=}\PY{p}{(}\PY{l+m+mi}{8}\PY{p}{,} \PY{l+m+mi}{6}\PY{p}{)}\PY{p}{)}
\PY{n}{ax} \PY{o}{=} \PY{n}{sns}\PY{o}{.}\PY{n}{barplot}\PY{p}{(}\PY{n}{data}\PY{o}{=}\PY{n}{df\PYZus{}label\PYZus{}dist\PYZus{}plot}\PY{p}{,} \PY{n}{x}\PY{o}{=}\PY{l+s+s2}{\PYZdq{}}\PY{l+s+s2}{partition}\PY{l+s+s2}{\PYZdq{}}\PY{p}{,} \PY{n}{y}\PY{o}{=}\PY{l+m+mi}{0}\PY{p}{,} \PY{n}{hue}\PY{o}{=}\PY{l+s+s2}{\PYZdq{}}\PY{l+s+s2}{label}\PY{l+s+s2}{\PYZdq{}}\PY{p}{,} \PY{n}{order}\PY{o}{=}\PY{p}{[}\PY{l+s+s2}{\PYZdq{}}\PY{l+s+s2}{train}\PY{l+s+s2}{\PYZdq{}}\PY{p}{,} \PY{l+s+s2}{\PYZdq{}}\PY{l+s+s2}{vali}\PY{l+s+s2}{\PYZdq{}}\PY{p}{,} \PY{l+s+s2}{\PYZdq{}}\PY{l+s+s2}{test}\PY{l+s+s2}{\PYZdq{}}\PY{p}{]}\PY{p}{)}

\PY{c+c1}{\PYZsh{} Customize the plot}
\PY{n}{plt}\PY{o}{.}\PY{n}{xlabel}\PY{p}{(}\PY{l+s+s2}{\PYZdq{}}\PY{l+s+s2}{Partition}\PY{l+s+s2}{\PYZdq{}}\PY{p}{)}
\PY{n}{plt}\PY{o}{.}\PY{n}{ylabel}\PY{p}{(}\PY{l+s+s2}{\PYZdq{}}\PY{l+s+s2}{Count}\PY{l+s+s2}{\PYZdq{}}\PY{p}{)}
\PY{n}{plt}\PY{o}{.}\PY{n}{title}\PY{p}{(}\PY{l+s+s2}{\PYZdq{}}\PY{l+s+s2}{Label Distribution}\PY{l+s+s2}{\PYZdq{}}\PY{p}{)}

\PY{c+c1}{\PYZsh{} Add count on top of each bar}
\PY{k}{for} \PY{n}{p} \PY{o+ow}{in} \PY{n}{ax}\PY{o}{.}\PY{n}{patches}\PY{p}{:}
    \PY{n}{ax}\PY{o}{.}\PY{n}{annotate}\PY{p}{(}
        \PY{l+s+sa}{f}\PY{l+s+s1}{\PYZsq{}}\PY{l+s+si}{\PYZob{}}\PY{n}{p}\PY{o}{.}\PY{n}{get\PYZus{}height}\PY{p}{(}\PY{p}{)}\PY{l+s+si}{:}\PY{l+s+s1}{.0f}\PY{l+s+si}{\PYZcb{}}\PY{l+s+s1}{\PYZsq{}}\PY{p}{,}
        \PY{p}{(}\PY{n}{p}\PY{o}{.}\PY{n}{get\PYZus{}x}\PY{p}{(}\PY{p}{)} \PY{o}{+} \PY{n}{p}\PY{o}{.}\PY{n}{get\PYZus{}width}\PY{p}{(}\PY{p}{)} \PY{o}{/} \PY{l+m+mf}{2.}\PY{p}{,} \PY{n}{p}\PY{o}{.}\PY{n}{get\PYZus{}height}\PY{p}{(}\PY{p}{)}\PY{p}{)}\PY{p}{,}
        \PY{n}{ha}\PY{o}{=}\PY{l+s+s1}{\PYZsq{}}\PY{l+s+s1}{center}\PY{l+s+s1}{\PYZsq{}}\PY{p}{,}
        \PY{n}{va}\PY{o}{=}\PY{l+s+s1}{\PYZsq{}}\PY{l+s+s1}{baseline}\PY{l+s+s1}{\PYZsq{}}\PY{p}{,}
        \PY{n}{fontsize}\PY{o}{=}\PY{l+m+mi}{12}\PY{p}{,}
        \PY{n}{color}\PY{o}{=}\PY{l+s+s1}{\PYZsq{}}\PY{l+s+s1}{black}\PY{l+s+s1}{\PYZsq{}}\PY{p}{,}
        \PY{n}{xytext}\PY{o}{=}\PY{p}{(}\PY{l+m+mi}{0}\PY{p}{,} \PY{l+m+mi}{5}\PY{p}{)}\PY{p}{,}
        \PY{n}{textcoords}\PY{o}{=}\PY{l+s+s1}{\PYZsq{}}\PY{l+s+s1}{offset points}\PY{l+s+s1}{\PYZsq{}}
    \PY{p}{)}
\PY{c+c1}{\PYZsh{} Show the plot}
\PY{n}{plt}\PY{o}{.}\PY{n}{show}\PY{p}{(}\PY{p}{)}
\end{Verbatim}
\end{tcolorbox}

    \begin{center}
    \adjustimage{max size={0.9\linewidth}{0.9\paperheight}}{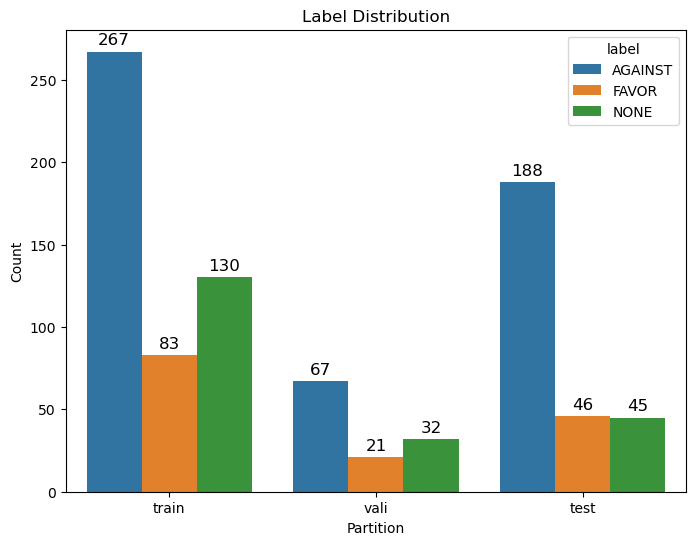}
    \end{center}
    { \hspace*{\fill} \\}
    
    \hypertarget{train-a-stadard-bert-model}{%
\section{Train a stadard BERT model}\label{train-a-stadard-bert-model}}

\begin{itemize}
\tightlist
\item
  Here, I use the BERT-base-uncased model, which is a standard BERT
  model with 12 self-attention layers and 110 million parameters.
\end{itemize}

    \begin{tcolorbox}[breakable, size=fbox, boxrule=1pt, pad at break*=1mm,colback=cellbackground, colframe=cellborder]
\prompt{In}{incolor}{ }{\boxspacing}
\begin{Verbatim}[commandchars=\\\{\}]
\PY{k+kn}{import} \PY{n+nn}{pandas} \PY{k}{as} \PY{n+nn}{pd}
\PY{k+kn}{from} \PY{n+nn}{os}\PY{n+nn}{.}\PY{n+nn}{path} \PY{k+kn}{import} \PY{n}{join}

\PY{k+kn}{from} \PY{n+nn}{transformers} \PY{k+kn}{import} \PY{n}{TrainingArguments}
\PY{k+kn}{from} \PY{n+nn}{transformers} \PY{k+kn}{import} \PY{n}{Trainer}
\PY{k+kn}{from} \PY{n+nn}{transformers} \PY{k+kn}{import} \PY{n}{AutoTokenizer}
\PY{k+kn}{from} \PY{n+nn}{transformers} \PY{k+kn}{import} \PY{n}{AutoModelForSequenceClassification}
\PY{k+kn}{import} \PY{n+nn}{torch} \PY{k}{as} \PY{n+nn}{th}


\PY{k+kn}{from} \PY{n+nn}{data\PYZus{}processor} \PY{k+kn}{import} \PY{n}{SemEvalDataProcessor}
\PY{k+kn}{from} \PY{n+nn}{utils} \PY{k+kn}{import} \PY{n}{process\PYZus{}dataframe}\PY{p}{,} \PY{n}{preprocess\PYZus{}dataset}\PY{p}{,} \PY{n}{partition\PYZus{}and\PYZus{}resample\PYZus{}df}\PY{p}{,} \PY{n}{evaluate\PYZus{}trained\PYZus{}trainer\PYZus{}over\PYZus{}sets}\PY{p}{,} \PY{n}{func\PYZus{}compute\PYZus{}metrics\PYZus{}sem\PYZus{}eval}\PY{p}{,} \PY{n}{remove\PYZus{}saved\PYZus{}models\PYZus{}in\PYZus{}checkpoint}\PY{p}{,} \PY{n}{remove\PYZus{}checkpoint\PYZus{}dir}\PY{p}{,} \PY{n}{seed\PYZus{}all}\PY{p}{,}\PY{n}{convert\PYZus{}stance\PYZus{}code\PYZus{}to\PYZus{}text}
\end{Verbatim}
\end{tcolorbox}

    \hypertarget{set-up}{%
\subsection{Set up}\label{set-up}}

    \begin{tcolorbox}[breakable, size=fbox, boxrule=1pt, pad at break*=1mm,colback=cellbackground, colframe=cellborder]
\prompt{In}{incolor}{ }{\boxspacing}
\begin{Verbatim}[commandchars=\\\{\}]
\PY{c+c1}{\PYZsh{} use the standard bert model}
\PY{n}{ENCODER} \PY{o}{=} \PY{l+s+s2}{\PYZdq{}}\PY{l+s+s2}{bert\PYZhy{}base\PYZhy{}uncased}\PY{l+s+s2}{\PYZdq{}}
\PY{n}{PATH\PYZus{}OUTPUT\PYZus{}ROOT} \PY{o}{=} \PY{n}{par}\PY{o}{.}\PY{n}{PATH\PYZus{}RESULT\PYZus{}SEM\PYZus{}EVAL\PYZus{}TUNING}
\end{Verbatim}
\end{tcolorbox}

    Use GPU if available

    \begin{tcolorbox}[breakable, size=fbox, boxrule=1pt, pad at break*=1mm,colback=cellbackground, colframe=cellborder]
\prompt{In}{incolor}{ }{\boxspacing}
\begin{Verbatim}[commandchars=\\\{\}]
\PY{n}{device} \PY{o}{=} \PY{n}{th}\PY{o}{.}\PY{n}{device}\PY{p}{(}\PY{l+s+s1}{\PYZsq{}}\PY{l+s+s1}{cuda}\PY{l+s+s1}{\PYZsq{}} \PY{k}{if} \PY{n}{th}\PY{o}{.}\PY{n}{cuda}\PY{o}{.}\PY{n}{is\PYZus{}available}\PY{p}{(}\PY{p}{)} \PY{k}{else} \PY{l+s+s1}{\PYZsq{}}\PY{l+s+s1}{cpu}\PY{l+s+s1}{\PYZsq{}}\PY{p}{)}
\PY{n+nb}{print}\PY{p}{(}\PY{l+s+s1}{\PYZsq{}}\PY{l+s+s1}{Using device:}\PY{l+s+s1}{\PYZsq{}}\PY{p}{,} \PY{n}{device}\PY{p}{)}
\PY{n}{seed\PYZus{}all}\PY{p}{(}\PY{n}{SEED}\PY{p}{)}
\end{Verbatim}
\end{tcolorbox}

    \begin{Verbatim}[commandchars=\\\{\}]
Using device: cuda
    \end{Verbatim}

    \begin{tcolorbox}[breakable, size=fbox, boxrule=1pt, pad at break*=1mm,colback=cellbackground, colframe=cellborder]
\prompt{In}{incolor}{ }{\boxspacing}
\begin{Verbatim}[commandchars=\\\{\}]
\PY{c+c1}{\PYZsh{} specify the output path}
\PY{n}{path\PYZus{}run\PYZus{}this} \PY{o}{=} \PY{n}{join}\PY{p}{(}\PY{n}{PATH\PYZus{}OUTPUT\PYZus{}ROOT}\PY{p}{,} \PY{n}{ENCODER}\PY{p}{)}
\PY{n}{file\PYZus{}metrics} \PY{o}{=} \PY{n}{join}\PY{p}{(}\PY{n}{path\PYZus{}run\PYZus{}this}\PY{p}{,} \PY{l+s+s2}{\PYZdq{}}\PY{l+s+s2}{metrics.csv}\PY{l+s+s2}{\PYZdq{}}\PY{p}{)}
\PY{n}{file\PYZus{}confusion\PYZus{}matrix} \PY{o}{=} \PY{n}{join}\PY{p}{(}\PY{n}{path\PYZus{}run\PYZus{}this}\PY{p}{,} \PY{l+s+s2}{\PYZdq{}}\PY{l+s+s2}{confusion\PYZus{}matrix.csv}\PY{l+s+s2}{\PYZdq{}}\PY{p}{)}
\PY{n}{file\PYZus{}predictions} \PY{o}{=} \PY{n}{join}\PY{p}{(}\PY{n}{path\PYZus{}run\PYZus{}this}\PY{p}{,} \PY{l+s+s2}{\PYZdq{}}\PY{l+s+s2}{predictions.csv}\PY{l+s+s2}{\PYZdq{}}\PY{p}{)}
\PY{n+nb}{print}\PY{p}{(}\PY{l+s+s2}{\PYZdq{}}\PY{l+s+s2}{path\PYZus{}run\PYZus{}this:}\PY{l+s+s2}{\PYZdq{}}\PY{p}{,} \PY{n}{path\PYZus{}run\PYZus{}this}\PY{p}{)}
\PY{n+nb}{print}\PY{p}{(}\PY{l+s+s2}{\PYZdq{}}\PY{l+s+s2}{file\PYZus{}metrics:}\PY{l+s+s2}{\PYZdq{}}\PY{p}{,} \PY{n}{file\PYZus{}metrics}\PY{p}{)}
\PY{n+nb}{print}\PY{p}{(}\PY{l+s+s2}{\PYZdq{}}\PY{l+s+s2}{file\PYZus{}confusion\PYZus{}matrix:}\PY{l+s+s2}{\PYZdq{}}\PY{p}{,} \PY{n}{file\PYZus{}confusion\PYZus{}matrix}\PY{p}{)}
\PY{n+nb}{print}\PY{p}{(}\PY{l+s+s2}{\PYZdq{}}\PY{l+s+s2}{file\PYZus{}predictions:}\PY{l+s+s2}{\PYZdq{}}\PY{p}{,} \PY{n}{file\PYZus{}predictions}\PY{p}{)}
\end{Verbatim}
\end{tcolorbox}

    \begin{Verbatim}[commandchars=\\\{\}]
path\_run\_this:
/home/sean/prelim\_stance\_detection/results/semeval\_2016/tuning/bert-base-uncased
file\_metrics:
/home/sean/prelim\_stance\_detection/results/semeval\_2016/tuning/bert-base-
uncased/metrics.csv
file\_confusion\_matrix:
/home/sean/prelim\_stance\_detection/results/semeval\_2016/tuning/bert-base-
uncased/confusion\_matrix.csv
file\_predictions:
/home/sean/prelim\_stance\_detection/results/semeval\_2016/tuning/bert-base-
uncased/predictions.csv
    \end{Verbatim}

    \begin{tcolorbox}[breakable, size=fbox, boxrule=1pt, pad at break*=1mm,colback=cellbackground, colframe=cellborder]
\prompt{In}{incolor}{ }{\boxspacing}
\begin{Verbatim}[commandchars=\\\{\}]
\PY{c+c1}{\PYZsh{} Load the preprocessed data and the partitions}
\PY{n}{data\PYZus{}processor} \PY{o}{=} \PY{n}{SemEvalDataProcessor}\PY{p}{(}\PY{p}{)}
\PY{n}{func\PYZus{}compute\PYZus{}metrics} \PY{o}{=} \PY{n}{func\PYZus{}compute\PYZus{}metrics\PYZus{}sem\PYZus{}eval}\PY{p}{(}\PY{p}{)}
\PY{n}{file\PYZus{}processed} \PY{o}{=} \PY{n}{data\PYZus{}processor}\PY{o}{.}\PY{n}{\PYZus{}get\PYZus{}file\PYZus{}processed\PYZus{}default}\PY{p}{(}\PY{n}{topic}\PY{o}{=}\PY{n}{TOPIC\PYZus{}OF\PYZus{}INTEREST}\PY{p}{)}
\PY{n}{df\PYZus{}partitions} \PY{o}{=} \PY{n}{data\PYZus{}processor}\PY{o}{.}\PY{n}{read\PYZus{}partitions}\PY{p}{(}\PY{n}{topic}\PY{o}{=}\PY{n}{TOPIC\PYZus{}OF\PYZus{}INTEREST}\PY{p}{)}

\PY{n}{df} \PY{o}{=} \PY{n}{process\PYZus{}dataframe}\PY{p}{(}\PY{n}{input\PYZus{}csv}\PY{o}{=}\PY{n}{file\PYZus{}processed}\PY{p}{,}
                        \PY{n}{dataset}\PY{o}{=}\PY{n}{DATASET}\PY{p}{)}
\end{Verbatim}
\end{tcolorbox}

    Note that there is a label imbalance issue in the dataset. There are way
more tweets with the \texttt{AGAINST} label than the other two labels.
This may cause the model to be biased towards predicting the
\texttt{AGAINST} label.

To address this issue, we will first upsample the training set to make
the number of tweets with each label equal.

    \begin{quote}
To learn more about the data imbalance issue, I recommend taking a look
at this tutorial
https://towardsdatascience.com/5-techniques-to-work-with-imbalanced-data-in-machine-learning-80836d45d30c
\end{quote}

    \begin{tcolorbox}[breakable, size=fbox, boxrule=1pt, pad at break*=1mm,colback=cellbackground, colframe=cellborder]
\prompt{In}{incolor}{ }{\boxspacing}
\begin{Verbatim}[commandchars=\\\{\}]
\PY{c+c1}{\PYZsh{} upsample the minority class}
\PY{n}{dict\PYZus{}df} \PY{o}{=} \PY{n}{partition\PYZus{}and\PYZus{}resample\PYZus{}df}\PY{p}{(}\PY{n}{df}\PY{p}{,} \PY{n}{SEED}\PY{p}{,} \PY{l+s+s2}{\PYZdq{}}\PY{l+s+s2}{single\PYZus{}domain}\PY{l+s+s2}{\PYZdq{}}\PY{p}{,}
                                    \PY{n}{factor\PYZus{}upsample}\PY{o}{=}\PY{l+m+mi}{1}\PY{p}{,}
                                    \PY{n}{read\PYZus{}partition\PYZus{}from\PYZus{}df}\PY{o}{=}\PY{k+kc}{True}\PY{p}{,}
                                    \PY{n}{df\PYZus{}partitions}\PY{o}{=}\PY{n}{df\PYZus{}partitions}\PY{p}{)}
\end{Verbatim}
\end{tcolorbox}

    Let's check if the upsampled data is balanced now.

    \begin{tcolorbox}[breakable, size=fbox, boxrule=1pt, pad at break*=1mm,colback=cellbackground, colframe=cellborder]
\prompt{In}{incolor}{ }{\boxspacing}
\begin{Verbatim}[commandchars=\\\{\}]
\PY{c+c1}{\PYZsh{} before upsampling}
\PY{n}{dict\PYZus{}df}\PY{p}{[}\PY{l+s+s2}{\PYZdq{}}\PY{l+s+s2}{train\PYZus{}raw}\PY{l+s+s2}{\PYZdq{}}\PY{p}{]}\PY{p}{[}\PY{l+s+s2}{\PYZdq{}}\PY{l+s+s2}{label}\PY{l+s+s2}{\PYZdq{}}\PY{p}{]}\PY{o}{.}\PY{n}{apply}\PY{p}{(}\PY{k}{lambda} \PY{n}{x}\PY{p}{:} \PY{n}{convert\PYZus{}stance\PYZus{}code\PYZus{}to\PYZus{}text}\PY{p}{(}\PY{n}{x}\PY{p}{,} \PY{n}{DATASET}\PY{p}{)}\PY{p}{)}\PY{o}{.}\PY{n}{value\PYZus{}counts}\PY{p}{(}\PY{p}{)}\PY{o}{.}\PY{n}{sort\PYZus{}index}\PY{p}{(}\PY{p}{)}
\end{Verbatim}
\end{tcolorbox}

            \begin{tcolorbox}[breakable, size=fbox, boxrule=.5pt, pad at break*=1mm, opacityfill=0]
\prompt{Out}{outcolor}{ }{\boxspacing}
\begin{Verbatim}[commandchars=\\\{\}]
AGAINST    267
FAVOR       83
NONE       130
Name: label, dtype: int64
\end{Verbatim}
\end{tcolorbox}
        
    \begin{tcolorbox}[breakable, size=fbox, boxrule=1pt, pad at break*=1mm,colback=cellbackground, colframe=cellborder]
\prompt{In}{incolor}{ }{\boxspacing}
\begin{Verbatim}[commandchars=\\\{\}]
\PY{c+c1}{\PYZsh{} after upsampling}
\PY{n}{dict\PYZus{}df}\PY{p}{[}\PY{l+s+s2}{\PYZdq{}}\PY{l+s+s2}{train\PYZus{}upsampled}\PY{l+s+s2}{\PYZdq{}}\PY{p}{]}\PY{p}{[}\PY{l+s+s2}{\PYZdq{}}\PY{l+s+s2}{label}\PY{l+s+s2}{\PYZdq{}}\PY{p}{]}\PY{o}{.}\PY{n}{apply}\PY{p}{(}\PY{k}{lambda} \PY{n}{x}\PY{p}{:} \PY{n}{convert\PYZus{}stance\PYZus{}code\PYZus{}to\PYZus{}text}\PY{p}{(}\PY{n}{x}\PY{p}{,} \PY{n}{DATASET}\PY{p}{)}\PY{p}{)}\PY{o}{.}\PY{n}{value\PYZus{}counts}\PY{p}{(}\PY{p}{)}\PY{o}{.}\PY{n}{sort\PYZus{}index}\PY{p}{(}\PY{p}{)}
\end{Verbatim}
\end{tcolorbox}

            \begin{tcolorbox}[breakable, size=fbox, boxrule=.5pt, pad at break*=1mm, opacityfill=0]
\prompt{Out}{outcolor}{ }{\boxspacing}
\begin{Verbatim}[commandchars=\\\{\}]
AGAINST    267
FAVOR      267
NONE       267
Name: label, dtype: int64
\end{Verbatim}
\end{tcolorbox}
        
    \hypertarget{load-the-tokenizer-and-tokenize-the-tweets}{%
\subsection{Load the tokenizer and tokenize the
tweets}\label{load-the-tokenizer-and-tokenize-the-tweets}}

Recall that BERT requires a special ``subword-tokenization'' process
(i.e., WordPiece tokenization). That is, it does not directly encode
each individual word, but rather encode each word as a sequence of
``sub-word tokens''. For example, the word ``university'' can be broken
down into the subwords ``uni'' and ``versity,'' which are more likely to
appear in the corpus than the word ``university'' itself. This process
of breaking down words into subwords is called sub-word tokenization.

    \begin{tcolorbox}[breakable, size=fbox, boxrule=1pt, pad at break*=1mm,colback=cellbackground, colframe=cellborder]
\prompt{In}{incolor}{ }{\boxspacing}
\begin{Verbatim}[commandchars=\\\{\}]
\PY{c+c1}{\PYZsh{} load the tokenizer and preprocess the data}
\PY{n}{tokenizer} \PY{o}{=} \PY{n}{AutoTokenizer}\PY{o}{.}\PY{n}{from\PYZus{}pretrained}\PY{p}{(}\PY{n}{ENCODER}\PY{p}{)}
\PY{n}{dict\PYZus{}dataset} \PY{o}{=} \PY{n+nb}{dict}\PY{p}{(}\PY{p}{)}
\PY{k}{for} \PY{n}{data\PYZus{}set} \PY{o+ow}{in} \PY{n}{dict\PYZus{}df}\PY{p}{:}
    \PY{n}{dict\PYZus{}dataset}\PY{p}{[}\PY{n}{data\PYZus{}set}\PY{p}{]} \PY{o}{=} \PY{n}{preprocess\PYZus{}dataset}\PY{p}{(}\PY{n}{dict\PYZus{}df}\PY{p}{[}\PY{n}{data\PYZus{}set}\PY{p}{]}\PY{p}{,}
                                                \PY{n}{tokenizer}\PY{p}{,}
                                                \PY{n}{keep\PYZus{}tweet\PYZus{}id}\PY{o}{=}\PY{k+kc}{True}\PY{p}{,}
                                                \PY{n}{col\PYZus{}name\PYZus{}tweet\PYZus{}id}\PY{o}{=}\PY{n}{par}\PY{o}{.}\PY{n}{TEXT\PYZus{}ID}\PY{p}{)}
\end{Verbatim}
\end{tcolorbox}

    \begin{Verbatim}[commandchars=\\\{\}]

    \end{Verbatim}

    \begin{tcolorbox}[breakable, size=fbox, boxrule=1pt, pad at break*=1mm,colback=cellbackground, colframe=cellborder]
\prompt{In}{incolor}{ }{\boxspacing}
\begin{Verbatim}[commandchars=\\\{\}]
\PY{c+c1}{\PYZsh{} the processed data set has the following structure}
\PY{c+c1}{\PYZsh{} \PYZhy{} text: the text of each tweet}
\PY{c+c1}{\PYZsh{} \PYZhy{} label: the label of each stance}
\PY{c+c1}{\PYZsh{} \PYZhy{} input\PYZus{}ids: the \PYZdq{}token ids\PYZdq{} of each tweet}
\PY{n}{dict\PYZus{}dataset}\PY{p}{[}\PY{l+s+s2}{\PYZdq{}}\PY{l+s+s2}{train\PYZus{}upsampled}\PY{l+s+s2}{\PYZdq{}}\PY{p}{]}
\end{Verbatim}
\end{tcolorbox}

            \begin{tcolorbox}[breakable, size=fbox, boxrule=.5pt, pad at break*=1mm, opacityfill=0]
\prompt{Out}{outcolor}{ }{\boxspacing}
\begin{Verbatim}[commandchars=\\\{\}]
Dataset(\{
    features: ['text', 'ID', 'label', '\_\_index\_level\_0\_\_', 'input\_ids',
'token\_type\_ids', 'attention\_mask'],
    num\_rows: 801
\})
\end{Verbatim}
\end{tcolorbox}
        
    \hypertarget{lets-look-at-one-example-tweet-after-tokenization}{%
\subsubsection{Let's look at one example tweet after
tokenization}\label{lets-look-at-one-example-tweet-after-tokenization}}

    The sentence ``\emph{i really don't understand how some people are
pro-choice. a life is a life no matter if it's 2 weeks old or 20 years
old.}'' is converted into the following tokens ID:

\emph{{[}101, 1045, 2428, 2123, 1005, 1056, 3305, 2129, 2070, 2111,
2024, 4013, 1011, 3601, 1012, 1037, 2166, 2003, 1037, 2166, 2053, 3043,
2065, 2009, 1005, 1055, 1016, 3134, 2214, 2030, 2322, 2086, 2214, 1012,
102, 0, 0, \ldots, 0{]}}

    \begin{quote}
The ``0'' at the end are the padding tokens. They not used to train the
model. Rather, they are used to make all the tweets within a batch have
the same length. This is a common practice when training neural network
models using batches.
\end{quote}

    \begin{tcolorbox}[breakable, size=fbox, boxrule=1pt, pad at break*=1mm,colback=cellbackground, colframe=cellborder]
\prompt{In}{incolor}{ }{\boxspacing}
\begin{Verbatim}[commandchars=\\\{\}]
\PY{n+nb}{print}\PY{p}{(}\PY{l+s+s2}{\PYZdq{}}\PY{l+s+s2}{The original text of this tweet: }\PY{l+s+se}{\PYZbs{}n}\PY{l+s+s2}{ }\PY{l+s+si}{\PYZob{}\PYZcb{}}\PY{l+s+se}{\PYZbs{}n}\PY{l+s+s2}{\PYZdq{}}\PY{o}{.}\PY{n}{format}\PY{p}{(}\PY{n}{dict\PYZus{}dataset}\PY{p}{[}\PY{l+s+s2}{\PYZdq{}}\PY{l+s+s2}{train\PYZus{}upsampled}\PY{l+s+s2}{\PYZdq{}}\PY{p}{]}\PY{p}{[}\PY{l+s+s2}{\PYZdq{}}\PY{l+s+s2}{text}\PY{l+s+s2}{\PYZdq{}}\PY{p}{]}\PY{p}{[}\PY{l+m+mi}{0}\PY{p}{]}\PY{p}{)}\PY{p}{)}
\PY{n+nb}{print}\PY{p}{(}\PY{l+s+s2}{\PYZdq{}}\PY{l+s+s2}{The label of this tweet: }\PY{l+s+se}{\PYZbs{}n}\PY{l+s+s2}{ }\PY{l+s+si}{\PYZob{}\PYZcb{}}\PY{l+s+se}{\PYZbs{}n}\PY{l+s+s2}{\PYZdq{}}\PY{o}{.}\PY{n}{format}\PY{p}{(}\PY{n}{convert\PYZus{}stance\PYZus{}code\PYZus{}to\PYZus{}text}\PY{p}{(}
    \PY{n}{dict\PYZus{}dataset}\PY{p}{[}\PY{l+s+s2}{\PYZdq{}}\PY{l+s+s2}{train\PYZus{}upsampled}\PY{l+s+s2}{\PYZdq{}}\PY{p}{]}\PY{p}{[}\PY{l+s+s2}{\PYZdq{}}\PY{l+s+s2}{label}\PY{l+s+s2}{\PYZdq{}}\PY{p}{]}\PY{p}{[}\PY{l+m+mi}{0}\PY{p}{]}\PY{o}{.}\PY{n}{item}\PY{p}{(}\PY{p}{)}\PY{p}{,} \PY{n}{DATASET}\PY{p}{)}\PY{p}{)}\PY{p}{)}
\PY{n+nb}{print}\PY{p}{(}\PY{l+s+s2}{\PYZdq{}}\PY{l+s+s2}{The token ids of this tweet: }\PY{l+s+se}{\PYZbs{}n}\PY{l+s+s2}{ }\PY{l+s+si}{\PYZob{}\PYZcb{}}\PY{l+s+se}{\PYZbs{}n}\PY{l+s+s2}{\PYZdq{}}\PY{o}{.}\PY{n}{format}\PY{p}{(}\PY{n}{dict\PYZus{}dataset}\PY{p}{[}\PY{l+s+s2}{\PYZdq{}}\PY{l+s+s2}{train\PYZus{}upsampled}\PY{l+s+s2}{\PYZdq{}}\PY{p}{]}\PY{p}{[}\PY{l+s+s2}{\PYZdq{}}\PY{l+s+s2}{input\PYZus{}ids}\PY{l+s+s2}{\PYZdq{}}\PY{p}{]}\PY{p}{[}\PY{l+m+mi}{0}\PY{p}{]}\PY{p}{)}\PY{p}{)}
\end{Verbatim}
\end{tcolorbox}

    \begin{Verbatim}[commandchars=\\\{\}]
The original text of this tweet:
 i really don't understand how some people are pro-choice. a life is a life no
matter if it's 2 weeks old or 20 years old.

The label of this tweet:
 AGAINST

The token ids of this tweet:
 tensor([ 101, 1045, 2428, 2123, 1005, 1056, 3305, 2129, 2070, 2111, 2024, 4013,
        1011, 3601, 1012, 1037, 2166, 2003, 1037, 2166, 2053, 3043, 2065, 2009,
        1005, 1055, 1016, 3134, 2214, 2030, 2322, 2086, 2214, 1012,  102,    0,
           0,    0,    0,    0,    0,    0,    0,    0,    0,    0,    0,    0,
           0,    0,    0,    0,    0,    0,    0,    0,    0,    0,    0,    0,
           0,    0,    0,    0,    0,    0,    0,    0,    0,    0,    0,    0,
           0,    0,    0,    0,    0,    0,    0,    0,    0,    0,    0,    0,
           0,    0,    0,    0,    0,    0,    0,    0,    0,    0,    0,    0,
           0,    0,    0,    0,    0,    0,    0,    0,    0,    0,    0,    0,
           0,    0,    0,    0,    0,    0,    0,    0,    0,    0,    0,    0,
           0,    0,    0,    0,    0,    0,    0,    0,    0,    0,    0,    0,
           0,    0,    0,    0,    0,    0,    0,    0,    0,    0,    0,    0,
           0,    0,    0,    0,    0,    0,    0,    0,    0,    0,    0,    0,
           0,    0,    0,    0,    0,    0,    0,    0,    0,    0,    0,    0,
           0,    0,    0,    0,    0,    0,    0,    0,    0,    0,    0,    0,
           0,    0,    0,    0,    0,    0,    0,    0,    0,    0,    0,    0,
           0,    0,    0,    0,    0,    0,    0,    0,    0,    0,    0,    0,
           0,    0,    0,    0,    0,    0,    0,    0,    0,    0,    0,    0,
           0,    0,    0,    0,    0,    0,    0,    0,    0,    0,    0,    0,
           0,    0,    0,    0,    0,    0,    0,    0,    0,    0,    0,    0,
           0,    0,    0,    0,    0,    0,    0,    0,    0,    0,    0,    0,
           0,    0,    0,    0])

    \end{Verbatim}

    The tokens ID can be converted back to the original tokens, also using
the tokenizer.

    Let look at the first 10 tokens of the first tweet.

Notice that the word ``pro-choice'' is broken down into the subwords
``pro'', ``-'', and ``choice'', as explained above.

    \begin{tcolorbox}[breakable, size=fbox, boxrule=1pt, pad at break*=1mm,colback=cellbackground, colframe=cellborder]
\prompt{In}{incolor}{ }{\boxspacing}
\begin{Verbatim}[commandchars=\\\{\}]
\PY{k}{for} \PY{n}{i} \PY{o+ow}{in} \PY{n+nb}{range}\PY{p}{(}\PY{l+m+mi}{15}\PY{p}{)}\PY{p}{:}
    \PY{n}{token\PYZus{}id\PYZus{}this} \PY{o}{=} \PY{n}{dict\PYZus{}dataset}\PY{p}{[}\PY{l+s+s2}{\PYZdq{}}\PY{l+s+s2}{train\PYZus{}upsampled}\PY{l+s+s2}{\PYZdq{}}\PY{p}{]}\PY{p}{[}\PY{l+s+s2}{\PYZdq{}}\PY{l+s+s2}{input\PYZus{}ids}\PY{l+s+s2}{\PYZdq{}}\PY{p}{]}\PY{p}{[}\PY{l+m+mi}{0}\PY{p}{]}\PY{p}{[}\PY{n}{i}\PY{p}{]}\PY{o}{.}\PY{n}{item}\PY{p}{(}\PY{p}{)}
    \PY{n}{token\PYZus{}this} \PY{o}{=} \PY{n}{tokenizer}\PY{o}{.}\PY{n}{decode}\PY{p}{(}\PY{n}{token\PYZus{}id\PYZus{}this}\PY{p}{)}
    \PY{n+nb}{print}\PY{p}{(}\PY{l+s+s2}{\PYZdq{}}\PY{l+s+s2}{token\PYZus{}id: }\PY{l+s+si}{\PYZob{}\PYZcb{}}\PY{l+s+s2}{; token: }\PY{l+s+si}{\PYZob{}\PYZcb{}}\PY{l+s+s2}{\PYZdq{}}\PY{o}{.}\PY{n}{format}\PY{p}{(}
        \PY{n}{token\PYZus{}id\PYZus{}this}\PY{p}{,}
        \PY{n}{token\PYZus{}this}\PY{p}{)}
    \PY{p}{)}
\end{Verbatim}
\end{tcolorbox}

    \begin{Verbatim}[commandchars=\\\{\}]
token\_id: 101; token: [CLS]
token\_id: 1045; token: i
token\_id: 2428; token: really
token\_id: 2123; token: don
token\_id: 1005; token: '
token\_id: 1056; token: t
token\_id: 3305; token: understand
token\_id: 2129; token: how
token\_id: 2070; token: some
token\_id: 2111; token: people
token\_id: 2024; token: are
token\_id: 4013; token: pro
token\_id: 1011; token: -
token\_id: 3601; token: choice
token\_id: 1012; token: .
    \end{Verbatim}

    \begin{quote}
The ``{[}CLS{]}'' token is another special token (just like the padding
token aboe) that is added to the beginning of each tweet. It is how BERT
knows that the tweet is the beginning of a new sentence.
\end{quote}

    \begin{tcolorbox}[breakable, size=fbox, boxrule=1pt, pad at break*=1mm,colback=cellbackground, colframe=cellborder]
\prompt{In}{incolor}{ }{\boxspacing}
\begin{Verbatim}[commandchars=\\\{\}]
\PY{c+c1}{\PYZsh{} \PYZdq{}decode\PYZdq{} the entire sequence}
\PY{n}{tokenizer}\PY{o}{.}\PY{n}{decode}\PY{p}{(}\PY{n}{dict\PYZus{}dataset}\PY{p}{[}\PY{l+s+s2}{\PYZdq{}}\PY{l+s+s2}{train\PYZus{}upsampled}\PY{l+s+s2}{\PYZdq{}}\PY{p}{]}\PY{p}{[}\PY{l+s+s2}{\PYZdq{}}\PY{l+s+s2}{input\PYZus{}ids}\PY{l+s+s2}{\PYZdq{}}\PY{p}{]}\PY{p}{[}\PY{l+m+mi}{0}\PY{p}{]}\PY{p}{,}\PY{n}{skip\PYZus{}special\PYZus{}tokens}\PY{o}{=}\PY{k+kc}{True}\PY{p}{)}
\end{Verbatim}
\end{tcolorbox}

            \begin{tcolorbox}[breakable, size=fbox, boxrule=.5pt, pad at break*=1mm, opacityfill=0]
\prompt{Out}{outcolor}{ }{\boxspacing}
\begin{Verbatim}[commandchars=\\\{\}]
"i really don't understand how some people are pro - choice. a life is a life no
matter if it's 2 weeks old or 20 years old."
\end{Verbatim}
\end{tcolorbox}
        
    Next, we want to load a pre-trained BERT model, which will be used to
initialize the weights of our model. We will use
\texttt{bert-base-uncased} model, which is a standard BERT model with 12
self-attention layers and 110 million parameters.

    \begin{tcolorbox}[breakable, size=fbox, boxrule=1pt, pad at break*=1mm,colback=cellbackground, colframe=cellborder]
\prompt{In}{incolor}{ }{\boxspacing}
\begin{Verbatim}[commandchars=\\\{\}]
\PY{n+nb}{print}\PY{p}{(}\PY{l+s+s2}{\PYZdq{}}\PY{l+s+s2}{The BERT model to use: }\PY{l+s+si}{\PYZob{}\PYZcb{}}\PY{l+s+s2}{\PYZdq{}}\PY{o}{.}\PY{n}{format}\PY{p}{(}\PY{n}{ENCODER}\PY{p}{)}\PY{p}{)}
\end{Verbatim}
\end{tcolorbox}

    \begin{Verbatim}[commandchars=\\\{\}]
The BERT model to use: bert-base-uncased
    \end{Verbatim}

    \begin{tcolorbox}[breakable, size=fbox, boxrule=1pt, pad at break*=1mm,colback=cellbackground, colframe=cellborder]
\prompt{In}{incolor}{ }{\boxspacing}
\begin{Verbatim}[commandchars=\\\{\}]
\PY{c+c1}{\PYZsh{} load the pretrained model}
\PY{n}{model} \PY{o}{=} \PY{n}{AutoModelForSequenceClassification}\PY{o}{.}\PY{n}{from\PYZus{}pretrained}\PY{p}{(}\PY{n}{ENCODER}\PY{p}{,}
                                                          \PY{n}{num\PYZus{}labels} \PY{o}{=} \PY{n}{par}\PY{o}{.}\PY{n}{DICT\PYZus{}NUM\PYZus{}CLASS}\PY{p}{[}\PY{n}{DATASET}\PY{p}{]}\PY{p}{)}
\end{Verbatim}
\end{tcolorbox}

    \begin{Verbatim}[commandchars=\\\{\}]
Some weights of the model checkpoint at bert-base-uncased were not used when
initializing BertForSequenceClassification:
['cls.predictions.transform.LayerNorm.weight', 'cls.seq\_relationship.bias',
'cls.predictions.transform.LayerNorm.bias',
'cls.predictions.transform.dense.bias',
'cls.predictions.transform.dense.weight', 'cls.predictions.decoder.weight',
'cls.seq\_relationship.weight', 'cls.predictions.bias']
- This IS expected if you are initializing BertForSequenceClassification from
the checkpoint of a model trained on another task or with another architecture
(e.g. initializing a BertForSequenceClassification model from a
BertForPreTraining model).
- This IS NOT expected if you are initializing BertForSequenceClassification
from the checkpoint of a model that you expect to be exactly identical
(initializing a BertForSequenceClassification model from a
BertForSequenceClassification model).
Some weights of BertForSequenceClassification were not initialized from the
model checkpoint at bert-base-uncased and are newly initialized:
['classifier.weight', 'classifier.bias']
You should probably TRAIN this model on a down-stream task to be able to use it
for predictions and inference.
    \end{Verbatim}

    Great! After loading the pre-trained BERT model, now we are ready to
fine-tune the BERT model. We are going to use the classes
\texttt{Trainer} and \texttt{TrainingArguments} provided by the
HuggingFace library.

    Let's specify the training arguments, including the number of epochs,
the batch size, and the learning rate etc.

While the model is being trained, we retain the best model at each epoch
based on the macro F1 score on the validation set. The macro F1 score is
the average of the F1 scores across all three stance classes.

    \begin{quote}
To learn more about macro-F1 score, I recommend taking a look at this
tutorial
https://towardsdatascience.com/micro-macro-weighted-averages-of-f1-score-clearly-explained-b603420b292f\#:\textasciitilde:text=The\%20macro\%2Daveraged\%20F1\%20score,regardless\%20of\%20their\%20support\%20values.
\end{quote}

    \begin{tcolorbox}[breakable, size=fbox, boxrule=1pt, pad at break*=1mm,colback=cellbackground, colframe=cellborder]
\prompt{In}{incolor}{ }{\boxspacing}
\begin{Verbatim}[commandchars=\\\{\}]
\PY{c+c1}{\PYZsh{} specify the training arguments}
\PY{n}{training\PYZus{}args} \PY{o}{=} \PY{n}{TrainingArguments}\PY{p}{(}
    \PY{c+c1}{\PYZsh{} dir to save the model checkpoints}
    \PY{n}{output\PYZus{}dir}\PY{o}{=}\PY{n}{path\PYZus{}run\PYZus{}this}\PY{p}{,}
    \PY{c+c1}{\PYZsh{} how often to evaluate the model on the eval set}
    \PY{c+c1}{\PYZsh{} \PYZhy{} logs the metrics on vali set}
    \PY{n}{evaluation\PYZus{}strategy}\PY{o}{=}\PY{l+s+s2}{\PYZdq{}}\PY{l+s+s2}{epoch}\PY{l+s+s2}{\PYZdq{}}\PY{p}{,}
    \PY{c+c1}{\PYZsh{} how often to log the training process to tensorboard}
    \PY{c+c1}{\PYZsh{} \PYZhy{} only log the train loss , lr, epoch etc info and not the metrics}
    \PY{n}{logging\PYZus{}strategy}\PY{o}{=}\PY{l+s+s2}{\PYZdq{}}\PY{l+s+s2}{epoch}\PY{l+s+s2}{\PYZdq{}}\PY{p}{,}
    \PY{c+c1}{\PYZsh{} how often to save the model on the eval set}
    \PY{c+c1}{\PYZsh{} \PYZhy{} load\PYZus{}best\PYZus{}model\PYZus{}at\PYZus{}end requires the save and eval strategy to match}
    \PY{n}{save\PYZus{}strategy}\PY{o}{=}\PY{l+s+s2}{\PYZdq{}}\PY{l+s+s2}{epoch}\PY{l+s+s2}{\PYZdq{}}\PY{p}{,}
    \PY{c+c1}{\PYZsh{} limit the total amount of checkpoints. deletes the older checkpoints in output\PYZus{}dir.}
    \PY{n}{save\PYZus{}total\PYZus{}limit}\PY{o}{=}\PY{l+m+mi}{1}\PY{p}{,}
    \PY{c+c1}{\PYZsh{} initial learning rate for the adamw optimizer}
    \PY{n}{learning\PYZus{}rate}\PY{o}{=}\PY{l+m+mf}{2e\PYZhy{}5}\PY{p}{,}
    \PY{n}{per\PYZus{}device\PYZus{}train\PYZus{}batch\PYZus{}size}\PY{o}{=}\PY{l+m+mi}{16}\PY{p}{,}
    \PY{n}{per\PYZus{}device\PYZus{}eval\PYZus{}batch\PYZus{}size}\PY{o}{=}\PY{l+m+mi}{16}\PY{p}{,}
    \PY{n}{num\PYZus{}train\PYZus{}epochs}\PY{o}{=}\PY{l+m+mi}{6}\PY{p}{,}
    \PY{n}{weight\PYZus{}decay}\PY{o}{=}\PY{l+m+mf}{0.01}\PY{p}{,}
    \PY{n}{seed}\PY{o}{=}\PY{n}{SEED}\PY{p}{,}
    \PY{n}{data\PYZus{}seed}\PY{o}{=}\PY{n}{SEED}\PY{p}{,}
    \PY{c+c1}{\PYZsh{} retain the best model (evaluated by the metric on the eval set)}
    \PY{n}{load\PYZus{}best\PYZus{}model\PYZus{}at\PYZus{}end}\PY{o}{=}\PY{k+kc}{True}\PY{p}{,}
    \PY{n}{metric\PYZus{}for\PYZus{}best\PYZus{}model}\PY{o}{=}\PY{l+s+s2}{\PYZdq{}}\PY{l+s+s2}{f1\PYZus{}macro}\PY{l+s+s2}{\PYZdq{}}\PY{p}{,}
    \PY{c+c1}{\PYZsh{} number of updates steps to accumulate the gradients for, before performing a backward/update pass}
    \PY{n}{gradient\PYZus{}accumulation\PYZus{}steps}\PY{o}{=}\PY{l+m+mi}{1}\PY{p}{,}
    \PY{n}{optim}\PY{o}{=}\PY{l+s+s2}{\PYZdq{}}\PY{l+s+s2}{adamw\PYZus{}torch}\PY{l+s+s2}{\PYZdq{}}\PY{p}{,}
    \PY{n}{report\PYZus{}to}\PY{o}{=}\PY{l+s+s2}{\PYZdq{}}\PY{l+s+s2}{none}\PY{l+s+s2}{\PYZdq{}}
\PY{p}{)}
\end{Verbatim}
\end{tcolorbox}

    \begin{tcolorbox}[breakable, size=fbox, boxrule=1pt, pad at break*=1mm,colback=cellbackground, colframe=cellborder]
\prompt{In}{incolor}{ }{\boxspacing}
\begin{Verbatim}[commandchars=\\\{\}]
\PY{c+c1}{\PYZsh{} Specify the trainer}
\PY{n}{trainer} \PY{o}{=} \PY{n}{Trainer}\PY{p}{(}\PY{n}{model}\PY{o}{=}\PY{n}{model}\PY{p}{,} \PY{n}{args}\PY{o}{=}\PY{n}{training\PYZus{}args}\PY{p}{,}
                  \PY{n}{train\PYZus{}dataset}\PY{o}{=}\PY{n}{dict\PYZus{}dataset}\PY{p}{[}\PY{l+s+s2}{\PYZdq{}}\PY{l+s+s2}{train\PYZus{}upsampled}\PY{l+s+s2}{\PYZdq{}}\PY{p}{]}\PY{p}{,}
                  \PY{n}{eval\PYZus{}dataset}\PY{o}{=}\PY{n}{dict\PYZus{}dataset}\PY{p}{[}\PY{l+s+s2}{\PYZdq{}}\PY{l+s+s2}{vali\PYZus{}raw}\PY{l+s+s2}{\PYZdq{}}\PY{p}{]}\PY{p}{,}
                  \PY{n}{compute\PYZus{}metrics}\PY{o}{=}\PY{n}{func\PYZus{}compute\PYZus{}metrics}
                  \PY{p}{)}
\end{Verbatim}
\end{tcolorbox}

    \hypertarget{fine-tune-the-bert-model}{%
\subsubsection{Fine-tune the BERT
model!}\label{fine-tune-the-bert-model}}

    \begin{tcolorbox}[breakable, size=fbox, boxrule=1pt, pad at break*=1mm,colback=cellbackground, colframe=cellborder]
\prompt{In}{incolor}{ }{\boxspacing}
\begin{Verbatim}[commandchars=\\\{\}]
\PY{c+c1}{\PYZsh{} Train the model}
\PY{k}{if} \PY{n}{DO\PYZus{}TRAIN\PYZus{}MODELS}\PY{p}{:}
  \PY{n}{trainer}\PY{o}{.}\PY{n}{train}\PY{p}{(}\PY{p}{)}
\end{Verbatim}
\end{tcolorbox}

    \begin{Verbatim}[commandchars=\\\{\}]
/home/sean/miniconda3/envs/prelim/lib/python3.10/site-
packages/torch/nn/parallel/\_functions.py:68: UserWarning: Was asked to gather
along dimension 0, but all input tensors were scalars; will instead unsqueeze
and return a vector.
  warnings.warn('Was asked to gather along dimension 0, but all '
    \end{Verbatim}

    \begin{Verbatim}[commandchars=\\\{\}]
<IPython.core.display.HTML object>
    \end{Verbatim}

    \begin{Verbatim}[commandchars=\\\{\}]
/home/sean/prelim\_stance\_detection/scripts/utils.py:667: FutureWarning:
load\_metric is deprecated and will be removed in the next major version of
datasets. Use 'evaluate.load' instead, from the new library Hugging Face Evaluate:
https://huggingface.co/docs/evaluate
  metric\_computer[name\_metric] = load\_metric(name\_metric)
/home/sean/miniconda3/envs/prelim/lib/python3.10/site-
packages/torch/nn/parallel/\_functions.py:68: UserWarning: Was asked to gather
along dimension 0, but all input tensors were scalars; will instead unsqueeze
and return a vector.
  warnings.warn('Was asked to gather along dimension 0, but all '
/home/sean/miniconda3/envs/prelim/lib/python3.10/site-
packages/torch/nn/parallel/\_functions.py:68: UserWarning: Was asked to gather
along dimension 0, but all input tensors were scalars; will instead unsqueeze
and return a vector.
  warnings.warn('Was asked to gather along dimension 0, but all '
/home/sean/miniconda3/envs/prelim/lib/python3.10/site-
packages/torch/nn/parallel/\_functions.py:68: UserWarning: Was asked to gather
along dimension 0, but all input tensors were scalars; will instead unsqueeze
and return a vector.
  warnings.warn('Was asked to gather along dimension 0, but all '
/home/sean/miniconda3/envs/prelim/lib/python3.10/site-
packages/torch/nn/parallel/\_functions.py:68: UserWarning: Was asked to gather
along dimension 0, but all input tensors were scalars; will instead unsqueeze
and return a vector.
  warnings.warn('Was asked to gather along dimension 0, but all '
/home/sean/miniconda3/envs/prelim/lib/python3.10/site-
packages/torch/nn/parallel/\_functions.py:68: UserWarning: Was asked to gather
along dimension 0, but all input tensors were scalars; will instead unsqueeze
and return a vector.
  warnings.warn('Was asked to gather along dimension 0, but all '
    \end{Verbatim}

    \hypertarget{evaluate-the-model-performance}{%
\subsubsection{Evaluate the model
performance}\label{evaluate-the-model-performance}}

    \begin{tcolorbox}[breakable, size=fbox, boxrule=1pt, pad at break*=1mm,colback=cellbackground, colframe=cellborder]
\prompt{In}{incolor}{ }{\boxspacing}
\begin{Verbatim}[commandchars=\\\{\}]
\PY{o}{\PYZpc{}\PYZpc{}capture} \PYZhy{}\PYZhy{}no\PYZhy{}stderr
\PY{k}{if} \PY{n}{DO\PYZus{}TRAIN\PYZus{}MODELS}\PY{p}{:}
    \PY{c+c1}{\PYZsh{} evaluate on each partition}
    \PY{n}{df\PYZus{}metrics}\PY{p}{,} \PY{n}{df\PYZus{}confusion\PYZus{}matrix}\PY{p}{,} \PY{n}{dict\PYZus{}predictions} \PY{o}{=} \PYZbs{}
        \PY{n}{evaluate\PYZus{}trained\PYZus{}trainer\PYZus{}over\PYZus{}sets}\PY{p}{(}\PY{n}{trainer}\PY{p}{,}
                                            \PY{n}{DATASET}\PY{p}{,}
                                            \PY{n}{dict\PYZus{}dataset}\PY{p}{,} \PY{l+s+s2}{\PYZdq{}}\PY{l+s+s2}{set}\PY{l+s+s2}{\PYZdq{}}\PY{p}{,}
                                            \PY{n}{return\PYZus{}predicted\PYZus{}labels}\PY{o}{=}\PY{k+kc}{True}\PY{p}{,}
                                            \PY{n}{keep\PYZus{}tweet\PYZus{}id}\PY{o}{=}\PY{k+kc}{True}\PY{p}{,}
                                            \PY{n}{col\PYZus{}name\PYZus{}tweet\PYZus{}id}\PY{o}{=}\PY{n}{par}\PY{o}{.}\PY{n}{TEXT\PYZus{}ID}\PY{p}{)}
\end{Verbatim}
\end{tcolorbox}

    \begin{Verbatim}[commandchars=\\\{\}]
/home/sean/miniconda3/envs/prelim/lib/python3.10/site-
packages/torch/nn/parallel/\_functions.py:68: UserWarning: Was asked to gather
along dimension 0, but all input tensors were scalars; will instead unsqueeze
and return a vector.
  warnings.warn('Was asked to gather along dimension 0, but all '
/home/sean/miniconda3/envs/prelim/lib/python3.10/site-
packages/torch/nn/parallel/\_functions.py:68: UserWarning: Was asked to gather
along dimension 0, but all input tensors were scalars; will instead unsqueeze
and return a vector.
  warnings.warn('Was asked to gather along dimension 0, but all '
/home/sean/miniconda3/envs/prelim/lib/python3.10/site-
packages/torch/nn/parallel/\_functions.py:68: UserWarning: Was asked to gather
along dimension 0, but all input tensors were scalars; will instead unsqueeze
and return a vector.
  warnings.warn('Was asked to gather along dimension 0, but all '
/home/sean/miniconda3/envs/prelim/lib/python3.10/site-
packages/torch/nn/parallel/\_functions.py:68: UserWarning: Was asked to gather
along dimension 0, but all input tensors were scalars; will instead unsqueeze
and return a vector.
  warnings.warn('Was asked to gather along dimension 0, but all '
/home/sean/miniconda3/envs/prelim/lib/python3.10/site-
packages/torch/nn/parallel/\_functions.py:68: UserWarning: Was asked to gather
along dimension 0, but all input tensors were scalars; will instead unsqueeze
and return a vector.
  warnings.warn('Was asked to gather along dimension 0, but all '
/home/sean/miniconda3/envs/prelim/lib/python3.10/site-
packages/torch/nn/parallel/\_functions.py:68: UserWarning: Was asked to gather
along dimension 0, but all input tensors were scalars; will instead unsqueeze
and return a vector.
  warnings.warn('Was asked to gather along dimension 0, but all '
    \end{Verbatim}

    \begin{tcolorbox}[breakable, size=fbox, boxrule=1pt, pad at break*=1mm,colback=cellbackground, colframe=cellborder]
\prompt{In}{incolor}{ }{\boxspacing}
\begin{Verbatim}[commandchars=\\\{\}]
\PY{k}{if} \PY{n}{DO\PYZus{}TRAIN\PYZus{}MODELS}\PY{p}{:}
  \PY{c+c1}{\PYZsh{} save the evaluation results}
  \PY{n}{df\PYZus{}metrics}\PY{o}{.}\PY{n}{to\PYZus{}csv}\PY{p}{(}\PY{n}{file\PYZus{}metrics}\PY{p}{)}

  \PY{c+c1}{\PYZsh{} save the confusion matrix}
  \PY{n}{df\PYZus{}confusion\PYZus{}matrix}\PY{o}{.}\PY{n}{to\PYZus{}csv}\PY{p}{(}\PY{n}{file\PYZus{}confusion\PYZus{}matrix}\PY{p}{)}

  \PY{c+c1}{\PYZsh{} save the predictions}
  \PY{c+c1}{\PYZsh{} concatenate the predictions}
  \PY{c+c1}{\PYZsh{} \PYZhy{} \PYZdq{}train\PYZus{}raw\PYZdq{}, \PYZdq{}val\PYZus{}raw\PYZdq{}, \PYZdq{}test\PYZus{}raw\PYZdq{}}
  \PY{n}{df\PYZus{}predictions} \PY{o}{=} \PY{n}{pd}\PY{o}{.}\PY{n}{concat}\PY{p}{(}\PY{p}{[}\PY{n}{dict\PYZus{}predictions}\PY{p}{[}\PY{n}{key}\PY{p}{]} \PY{k}{for} \PY{n}{key} \PY{o+ow}{in} \PY{p}{[}\PY{l+s+s2}{\PYZdq{}}\PY{l+s+s2}{train\PYZus{}raw}\PY{l+s+s2}{\PYZdq{}}\PY{p}{,} \PY{l+s+s2}{\PYZdq{}}\PY{l+s+s2}{vali\PYZus{}raw}\PY{l+s+s2}{\PYZdq{}}\PY{p}{,} \PY{l+s+s2}{\PYZdq{}}\PY{l+s+s2}{test\PYZus{}raw}\PY{l+s+s2}{\PYZdq{}}\PY{p}{]}\PY{p}{]}\PY{p}{,} \PY{n}{axis}\PY{o}{=}\PY{l+m+mi}{0}\PY{p}{)}
  \PY{c+c1}{\PYZsh{} save only the tweet\PYZus{}id and the predicted label}
  \PY{n}{df\PYZus{}predictions} \PY{o}{=} \PY{n}{df\PYZus{}predictions}\PY{p}{[}\PY{p}{[}\PY{l+s+s2}{\PYZdq{}}\PY{l+s+s2}{ID}\PY{l+s+s2}{\PYZdq{}}\PY{p}{,} \PY{l+s+s2}{\PYZdq{}}\PY{l+s+s2}{predicted\PYZus{}label}\PY{l+s+s2}{\PYZdq{}}\PY{p}{]}\PY{p}{]}
  \PY{c+c1}{\PYZsh{} convert the predicted label to the stance code}
  \PY{n}{df\PYZus{}predictions}\PY{p}{[}\PY{l+s+s2}{\PYZdq{}}\PY{l+s+s2}{predicted\PYZus{}label}\PY{l+s+s2}{\PYZdq{}}\PY{p}{]} \PY{o}{=} \PY{n}{df\PYZus{}predictions}\PY{p}{[}\PY{l+s+s2}{\PYZdq{}}\PY{l+s+s2}{predicted\PYZus{}label}\PY{l+s+s2}{\PYZdq{}}\PY{p}{]}\PY{o}{.}\PY{n}{apply}\PY{p}{(}\PY{k}{lambda} \PY{n}{x}\PY{p}{:} \PY{n}{convert\PYZus{}stance\PYZus{}code\PYZus{}to\PYZus{}text}\PY{p}{(}\PY{n}{x}\PY{p}{,} \PY{n}{DATASET}\PY{p}{)}\PY{p}{)}
  \PY{n}{df\PYZus{}predictions}\PY{o}{.}\PY{n}{to\PYZus{}csv}\PY{p}{(}\PY{n}{file\PYZus{}predictions}\PY{p}{,} \PY{n}{index}\PY{o}{=}\PY{k+kc}{False}\PY{p}{)}

  \PY{c+c1}{\PYZsh{} remove the saved model}
  \PY{n}{remove\PYZus{}saved\PYZus{}models\PYZus{}in\PYZus{}checkpoint}\PY{p}{(}\PY{n}{path\PYZus{}run\PYZus{}this}\PY{p}{)}
  \PY{c+c1}{\PYZsh{} remove the checkpoints}
  \PY{n}{remove\PYZus{}checkpoint\PYZus{}dir}\PY{p}{(}\PY{n}{path\PYZus{}run\PYZus{}this}\PY{p}{)}
\end{Verbatim}
\end{tcolorbox}

    \begin{center}\rule{0.5\linewidth}{0.5pt}\end{center}

    \hypertarget{considering-more-appropriate-pre-trained-domain-specific-models}{%
\section{Considering More Appropriate Pre-trained Domain-specific
Models}\label{considering-more-appropriate-pre-trained-domain-specific-models}}

While the \texttt{bert-base-uncased} model serves as a good starting
point, there are other pre-trained models specifically designed for
social media text analysis that use more relevant pre-training data. Two
such models are BERTweet and polibertweet-mlm.

As mentioned earlier, when aiming to use BERT for a stance detection
task on tweets about abortion, it is more effective to pre-train BERT on
a corpus that is more similar to the Abortion tweet dataset. Given that
tweets are generally shorter and more informal than Wikipedia and online
books, it makes sense to pre-train BERT on a corpus that primarily
consists of tweets (rather than the original Wikipedia and online books
corpus). This is precisely the approach taken by domain-specific models
like \texttt{BERTweet} and \texttt{polibertweet-mlm}, which focus on
capturing the nuances and characteristics of social media text, making
them better suited for stance detection tasks in this context.

    \begin{quote}
Note for advanced readers: It is important to mention that both BERTweet
and polibertweet-mlm are based on the roberta-base architecture, a
variant of BERT that has been optimized for improved performance. As a
result, these domain-specific models not only benefit from more
appropriate pre-training data but also from the enhancements offered by
the roberta-base architecture. To understand the differences between
BERT and roberta-base, I recommend taking a look at
\href{https://appliedsingularity.com/2022/03/29/nlp-tutorials-part-14-roberta/}{this
tutorial}
\end{quote}

    \hypertarget{bertweet}{%
\subsection{BERTweet}\label{bertweet}}

\texttt{BERTweet} is a pre-trained model specifically designed for
processing and understanding Twitter data. It is trained on a large
corpus of 850 million English tweets. As Twitter text contains unique
language patterns, slang, and abbreviations, BERTweet is expected to
perform better on stance detection tasks involving tweets compared to
the general-purpose BERT model. For more details, see:
https://huggingface.co/docs/transformers/model\_doc/bertweet

    \hypertarget{polibertweet-mlm}{%
\subsection{Polibertweet-mlm}\label{polibertweet-mlm}}

    \texttt{Polibertweet-mlm} is a pre-trained model specifically designed
for Twitter data, with a focus on political discourse. The dataset used
for pretraining contains over 83 million English tweets related to the
2020 US Presidential Election.

As the Abortion stance dataset may also involve political topics,
polibertweet-mlm can also to be a suitable model for this stance
detection task. For more details, see:
https://huggingface.co/kornosk/polibertweet-political-twitter-roberta-mlm

    \hypertarget{helper-function}{%
\subsubsection{Helper Function}\label{helper-function}}

    Before we proceed with fine-tuning the models, we first create a wrapper
function for each type of pretrained model. Using a wrapper function for
the training pipeline streamlines the process of experimenting with
different models, ensuring consistency, reproducibility, and
maintainability.

    \begin{tcolorbox}[breakable, size=fbox, boxrule=1pt, pad at break*=1mm,colback=cellbackground, colframe=cellborder]
\prompt{In}{incolor}{ }{\boxspacing}
\begin{Verbatim}[commandchars=\\\{\}]
\PY{k}{def} \PY{n+nf}{train}\PY{p}{(}\PY{n}{encoder}\PY{p}{)}\PY{p}{:}
    \PY{c+c1}{\PYZsh{} specify the output paths}
    \PY{n}{encoder\PYZus{}name\PYZus{}tidy} \PY{o}{=} \PY{n}{tidy\PYZus{}name}\PY{p}{(}\PY{n}{encoder}\PY{p}{)}
    \PY{n}{path\PYZus{}run\PYZus{}this} \PY{o}{=} \PY{n}{join}\PY{p}{(}\PY{n}{PATH\PYZus{}OUTPUT\PYZus{}ROOT}\PY{p}{,} \PY{n}{encoder\PYZus{}name\PYZus{}tidy}\PY{p}{)}
    \PY{n}{file\PYZus{}metrics} \PY{o}{=} \PY{n}{join}\PY{p}{(}\PY{n}{path\PYZus{}run\PYZus{}this}\PY{p}{,} \PY{l+s+s2}{\PYZdq{}}\PY{l+s+s2}{metrics.csv}\PY{l+s+s2}{\PYZdq{}}\PY{p}{)}
    \PY{n}{file\PYZus{}confusion\PYZus{}matrix} \PY{o}{=} \PY{n}{join}\PY{p}{(}\PY{n}{path\PYZus{}run\PYZus{}this}\PY{p}{,} \PY{l+s+s2}{\PYZdq{}}\PY{l+s+s2}{confusion\PYZus{}matrix.csv}\PY{l+s+s2}{\PYZdq{}}\PY{p}{)}
    \PY{n}{file\PYZus{}predictions} \PY{o}{=} \PY{n}{join}\PY{p}{(}\PY{n}{path\PYZus{}run\PYZus{}this}\PY{p}{,} \PY{l+s+s2}{\PYZdq{}}\PY{l+s+s2}{predictions.csv}\PY{l+s+s2}{\PYZdq{}}\PY{p}{)}

    \PY{c+c1}{\PYZsh{} Load the preprocessed data and the partitions}
    \PY{n}{df} \PY{o}{=} \PY{n}{process\PYZus{}dataframe}\PY{p}{(}\PY{n}{input\PYZus{}csv}\PY{o}{=}\PY{n}{file\PYZus{}processed}\PY{p}{,} \PY{n}{dataset}\PY{o}{=}\PY{n}{DATASET}\PY{p}{)}
    \PY{c+c1}{\PYZsh{} upsample the minority class}
    \PY{n}{dict\PYZus{}df} \PY{o}{=} \PY{n}{partition\PYZus{}and\PYZus{}resample\PYZus{}df}\PY{p}{(}\PY{n}{df}\PY{p}{,} \PY{n}{SEED}\PY{p}{,} \PY{l+s+s2}{\PYZdq{}}\PY{l+s+s2}{single\PYZus{}domain}\PY{l+s+s2}{\PYZdq{}}\PY{p}{,}
                                        \PY{n}{factor\PYZus{}upsample}\PY{o}{=}\PY{l+m+mi}{1}\PY{p}{,}
                                        \PY{n}{read\PYZus{}partition\PYZus{}from\PYZus{}df}\PY{o}{=}\PY{k+kc}{True}\PY{p}{,}
                                        \PY{n}{df\PYZus{}partitions}\PY{o}{=}\PY{n}{df\PYZus{}partitions}\PY{p}{)}
    \PY{c+c1}{\PYZsh{} load the tokenizer and preprocess the data}
    \PY{n}{tokenizer} \PY{o}{=} \PY{n}{AutoTokenizer}\PY{o}{.}\PY{n}{from\PYZus{}pretrained}\PY{p}{(}\PY{n}{encoder}\PY{p}{)}
    \PY{n}{dict\PYZus{}dataset} \PY{o}{=} \PY{n+nb}{dict}\PY{p}{(}\PY{p}{)}
    \PY{k}{for} \PY{n}{data\PYZus{}set} \PY{o+ow}{in} \PY{n}{dict\PYZus{}df}\PY{p}{:}
        \PY{n}{dict\PYZus{}dataset}\PY{p}{[}\PY{n}{data\PYZus{}set}\PY{p}{]} \PY{o}{=} \PY{n}{preprocess\PYZus{}dataset}\PY{p}{(}\PY{n}{dict\PYZus{}df}\PY{p}{[}\PY{n}{data\PYZus{}set}\PY{p}{]}\PY{p}{,}
                                                    \PY{n}{tokenizer}\PY{p}{,}
                                                    \PY{n}{keep\PYZus{}tweet\PYZus{}id}\PY{o}{=}\PY{k+kc}{True}\PY{p}{,}
                                                    \PY{n}{col\PYZus{}name\PYZus{}tweet\PYZus{}id}\PY{o}{=}\PY{n}{par}\PY{o}{.}\PY{n}{TEXT\PYZus{}ID}\PY{p}{)}
    \PY{c+c1}{\PYZsh{} specify the training arguments}
    \PY{n}{training\PYZus{}args} \PY{o}{=} \PY{n}{TrainingArguments}\PY{p}{(}
        \PY{c+c1}{\PYZsh{} dir to save the model checkpoints}
        \PY{n}{output\PYZus{}dir}\PY{o}{=}\PY{n}{path\PYZus{}run\PYZus{}this}\PY{p}{,}
        \PY{n}{evaluation\PYZus{}strategy}\PY{o}{=}\PY{l+s+s2}{\PYZdq{}}\PY{l+s+s2}{epoch}\PY{l+s+s2}{\PYZdq{}}\PY{p}{,}
        \PY{n}{logging\PYZus{}strategy}\PY{o}{=}\PY{l+s+s2}{\PYZdq{}}\PY{l+s+s2}{epoch}\PY{l+s+s2}{\PYZdq{}}\PY{p}{,}
        \PY{n}{save\PYZus{}strategy}\PY{o}{=}\PY{l+s+s2}{\PYZdq{}}\PY{l+s+s2}{epoch}\PY{l+s+s2}{\PYZdq{}}\PY{p}{,}
        \PY{n}{save\PYZus{}total\PYZus{}limit}\PY{o}{=}\PY{l+m+mi}{1}\PY{p}{,}
        \PY{n}{learning\PYZus{}rate}\PY{o}{=}\PY{l+m+mf}{2e\PYZhy{}5}\PY{p}{,}
        \PY{n}{per\PYZus{}device\PYZus{}train\PYZus{}batch\PYZus{}size}\PY{o}{=}\PY{l+m+mi}{16}\PY{p}{,}
        \PY{n}{per\PYZus{}device\PYZus{}eval\PYZus{}batch\PYZus{}size}\PY{o}{=}\PY{l+m+mi}{16}\PY{p}{,}
        \PY{n}{num\PYZus{}train\PYZus{}epochs}\PY{o}{=}\PY{l+m+mi}{6}\PY{p}{,}
        \PY{n}{weight\PYZus{}decay}\PY{o}{=}\PY{l+m+mf}{0.01}\PY{p}{,}
        \PY{n}{seed}\PY{o}{=}\PY{n}{SEED}\PY{p}{,}
        \PY{n}{data\PYZus{}seed}\PY{o}{=}\PY{n}{SEED}\PY{p}{,}
        \PY{n}{load\PYZus{}best\PYZus{}model\PYZus{}at\PYZus{}end}\PY{o}{=}\PY{k+kc}{True}\PY{p}{,}
        \PY{n}{metric\PYZus{}for\PYZus{}best\PYZus{}model}\PY{o}{=}\PY{l+s+s2}{\PYZdq{}}\PY{l+s+s2}{f1\PYZus{}macro}\PY{l+s+s2}{\PYZdq{}}\PY{p}{,}
        \PY{n}{gradient\PYZus{}accumulation\PYZus{}steps}\PY{o}{=}\PY{l+m+mi}{1}\PY{p}{,}
        \PY{n}{optim}\PY{o}{=}\PY{l+s+s2}{\PYZdq{}}\PY{l+s+s2}{adamw\PYZus{}torch}\PY{l+s+s2}{\PYZdq{}}\PY{p}{,}
        \PY{n}{report\PYZus{}to}\PY{o}{=}\PY{l+s+s2}{\PYZdq{}}\PY{l+s+s2}{none}\PY{l+s+s2}{\PYZdq{}}
    \PY{p}{)}

    \PY{c+c1}{\PYZsh{} load the pretrained model}
    \PY{n}{model} \PY{o}{=} \PY{n}{AutoModelForSequenceClassification}\PY{o}{.}\PY{n}{from\PYZus{}pretrained}\PY{p}{(}\PY{n}{encoder}\PY{p}{,}
                                                               \PY{n}{num\PYZus{}labels}\PY{o}{=}\PY{n}{par}\PY{o}{.}\PY{n}{DICT\PYZus{}NUM\PYZus{}CLASS}\PY{p}{[}\PY{n}{DATASET}\PY{p}{]}\PY{p}{)}

    \PY{c+c1}{\PYZsh{} specify the trainer}
    \PY{n}{trainer} \PY{o}{=} \PY{n}{Trainer}\PY{p}{(}\PY{n}{model}\PY{o}{=}\PY{n}{model}\PY{p}{,} \PY{n}{args}\PY{o}{=}\PY{n}{training\PYZus{}args}\PY{p}{,}
                      \PY{n}{train\PYZus{}dataset}\PY{o}{=}\PY{n}{dict\PYZus{}dataset}\PY{p}{[}\PY{l+s+s2}{\PYZdq{}}\PY{l+s+s2}{train\PYZus{}upsampled}\PY{l+s+s2}{\PYZdq{}}\PY{p}{]}\PY{p}{,}
                      \PY{n}{eval\PYZus{}dataset}\PY{o}{=}\PY{n}{dict\PYZus{}dataset}\PY{p}{[}\PY{l+s+s2}{\PYZdq{}}\PY{l+s+s2}{vali\PYZus{}raw}\PY{l+s+s2}{\PYZdq{}}\PY{p}{]}\PY{p}{,}
                      \PY{n}{compute\PYZus{}metrics}\PY{o}{=}\PY{n}{func\PYZus{}compute\PYZus{}metrics}
                      \PY{p}{)}

    \PY{c+c1}{\PYZsh{} train the model}
    \PY{n}{trainer}\PY{o}{.}\PY{n}{train}\PY{p}{(}\PY{p}{)}

    \PY{c+c1}{\PYZsh{} evaluate on each partition}
    \PY{n}{df\PYZus{}metrics}\PY{p}{,} \PY{n}{df\PYZus{}confusion\PYZus{}matrix}\PY{p}{,} \PY{n}{dict\PYZus{}predictions} \PY{o}{=} \PYZbs{}
        \PY{n}{evaluate\PYZus{}trained\PYZus{}trainer\PYZus{}over\PYZus{}sets}\PY{p}{(}\PY{n}{trainer}\PY{p}{,}
                                           \PY{n}{DATASET}\PY{p}{,}
                                           \PY{n}{dict\PYZus{}dataset}\PY{p}{,} \PY{l+s+s2}{\PYZdq{}}\PY{l+s+s2}{set}\PY{l+s+s2}{\PYZdq{}}\PY{p}{,}
                                           \PY{n}{return\PYZus{}predicted\PYZus{}labels}\PY{o}{=}\PY{k+kc}{True}\PY{p}{,}
                                           \PY{n}{keep\PYZus{}tweet\PYZus{}id}\PY{o}{=}\PY{k+kc}{True}\PY{p}{,}
                                           \PY{n}{col\PYZus{}name\PYZus{}tweet\PYZus{}id}\PY{o}{=}\PY{n}{par}\PY{o}{.}\PY{n}{TEXT\PYZus{}ID}\PY{p}{)}
    \PY{c+c1}{\PYZsh{} save the evaluation results}
    \PY{n}{df\PYZus{}metrics}\PY{o}{.}\PY{n}{to\PYZus{}csv}\PY{p}{(}\PY{n}{file\PYZus{}metrics}\PY{p}{)}

    \PY{c+c1}{\PYZsh{} save the confusion matrix}
    \PY{n}{df\PYZus{}confusion\PYZus{}matrix}\PY{o}{.}\PY{n}{to\PYZus{}csv}\PY{p}{(}\PY{n}{file\PYZus{}confusion\PYZus{}matrix}\PY{p}{)}

    \PY{c+c1}{\PYZsh{} save the predictions}
    \PY{c+c1}{\PYZsh{} concatenate the predictions}
    \PY{c+c1}{\PYZsh{} \PYZhy{} \PYZdq{}train\PYZus{}raw\PYZdq{}, \PYZdq{}val\PYZus{}raw\PYZdq{}, \PYZdq{}test\PYZus{}raw\PYZdq{}}
    \PY{n}{df\PYZus{}predictions} \PY{o}{=} \PY{n}{pd}\PY{o}{.}\PY{n}{concat}\PY{p}{(}\PY{p}{[}\PY{n}{dict\PYZus{}predictions}\PY{p}{[}\PY{n}{key}\PY{p}{]} \PY{k}{for} \PY{n}{key} \PY{o+ow}{in} \PY{p}{[}\PY{l+s+s2}{\PYZdq{}}\PY{l+s+s2}{train\PYZus{}raw}\PY{l+s+s2}{\PYZdq{}}\PY{p}{,} \PY{l+s+s2}{\PYZdq{}}\PY{l+s+s2}{vali\PYZus{}raw}\PY{l+s+s2}{\PYZdq{}}\PY{p}{,} \PY{l+s+s2}{\PYZdq{}}\PY{l+s+s2}{test\PYZus{}raw}\PY{l+s+s2}{\PYZdq{}}\PY{p}{]}\PY{p}{]}\PY{p}{,} \PY{n}{axis}\PY{o}{=}\PY{l+m+mi}{0}\PY{p}{)}
    \PY{c+c1}{\PYZsh{} save only the tweet\PYZus{}id and the predicted label}
    \PY{n}{df\PYZus{}predictions} \PY{o}{=} \PY{n}{df\PYZus{}predictions}\PY{p}{[}\PY{p}{[}\PY{l+s+s2}{\PYZdq{}}\PY{l+s+s2}{ID}\PY{l+s+s2}{\PYZdq{}}\PY{p}{,} \PY{l+s+s2}{\PYZdq{}}\PY{l+s+s2}{predicted\PYZus{}label}\PY{l+s+s2}{\PYZdq{}}\PY{p}{]}\PY{p}{]}
    \PY{c+c1}{\PYZsh{} convert the predicted label to the stance code}
    \PY{n}{df\PYZus{}predictions}\PY{p}{[}\PY{l+s+s2}{\PYZdq{}}\PY{l+s+s2}{predicted\PYZus{}label}\PY{l+s+s2}{\PYZdq{}}\PY{p}{]} \PY{o}{=} \PY{n}{df\PYZus{}predictions}\PY{p}{[}\PY{l+s+s2}{\PYZdq{}}\PY{l+s+s2}{predicted\PYZus{}label}\PY{l+s+s2}{\PYZdq{}}\PY{p}{]}\PY{o}{.}\PY{n}{apply}\PY{p}{(}\PY{k}{lambda} \PY{n}{x}\PY{p}{:} \PY{n}{convert\PYZus{}stance\PYZus{}code\PYZus{}to\PYZus{}text}\PY{p}{(}\PY{n}{x}\PY{p}{,} \PY{n}{DATASET}\PY{p}{)}\PY{p}{)}
    \PY{n}{df\PYZus{}predictions}\PY{o}{.}\PY{n}{to\PYZus{}csv}\PY{p}{(}\PY{n}{file\PYZus{}predictions}\PY{p}{,} \PY{n}{index}\PY{o}{=}\PY{k+kc}{False}\PY{p}{)}

    \PY{c+c1}{\PYZsh{} remove the saved model}
    \PY{n}{remove\PYZus{}saved\PYZus{}models\PYZus{}in\PYZus{}checkpoint}\PY{p}{(}\PY{n}{path\PYZus{}run\PYZus{}this}\PY{p}{)}
    \PY{c+c1}{\PYZsh{} remove the checkpoints}
    \PY{n}{remove\PYZus{}checkpoint\PYZus{}dir}\PY{p}{(}\PY{n}{path\PYZus{}run\PYZus{}this}\PY{p}{)}
\end{Verbatim}
\end{tcolorbox}

    \begin{tcolorbox}[breakable, size=fbox, boxrule=1pt, pad at break*=1mm,colback=cellbackground, colframe=cellborder]
\prompt{In}{incolor}{ }{\boxspacing}
\begin{Verbatim}[commandchars=\\\{\}]
\PY{k}{if} \PY{n}{DO\PYZus{}TRAIN\PYZus{}MODELS}\PY{p}{:}
  \PY{c+c1}{\PYZsh{} BERTweet, having the same architecture as BERT\PYZhy{}base (Devlin et al., 2019), is trained using the RoBERTa pre\PYZhy{}training procedure}
  \PY{c+c1}{\PYZsh{} \PYZhy{} https://huggingface.co/docs/transformers/model\PYZus{}doc/bertweet}
  \PY{n}{train}\PY{p}{(}\PY{l+s+s2}{\PYZdq{}}\PY{l+s+s2}{vinai/bertweet\PYZhy{}base}\PY{l+s+s2}{\PYZdq{}}\PY{p}{)}
\end{Verbatim}
\end{tcolorbox}

    \begin{Verbatim}[commandchars=\\\{\}]
Special tokens have been added in the vocabulary, make sure the associated word
embeddings are fine-tuned or trained.
Some weights of the model checkpoint at vinai/bertweet-base were not used when
initializing RobertaForSequenceClassification: ['roberta.pooler.dense.weight',
'lm\_head.layer\_norm.bias', 'lm\_head.layer\_norm.weight', 'lm\_head.dense.bias',
'lm\_head.decoder.bias', 'lm\_head.dense.weight', 'roberta.pooler.dense.bias',
'lm\_head.decoder.weight', 'lm\_head.bias']
- This IS expected if you are initializing RobertaForSequenceClassification from
the checkpoint of a model trained on another task or with another architecture
(e.g. initializing a BertForSequenceClassification model from a
BertForPreTraining model).
- This IS NOT expected if you are initializing RobertaForSequenceClassification
from the checkpoint of a model that you expect to be exactly identical
(initializing a BertForSequenceClassification model from a
BertForSequenceClassification model).
Some weights of RobertaForSequenceClassification were not initialized from the
model checkpoint at vinai/bertweet-base and are newly initialized:
['classifier.out\_proj.bias', 'classifier.dense.weight',
'classifier.out\_proj.weight', 'classifier.dense.bias']
You should probably TRAIN this model on a down-stream task to be able to use it
for predictions and inference.
/home/sean/miniconda3/envs/prelim/lib/python3.10/site-
packages/torch/nn/parallel/\_functions.py:68: UserWarning: Was asked to gather
along dimension 0, but all input tensors were scalars; will instead unsqueeze
and return a vector.
  warnings.warn('Was asked to gather along dimension 0, but all '
    \end{Verbatim}

    \begin{Verbatim}[commandchars=\\\{\}]
<IPython.core.display.HTML object>
    \end{Verbatim}

    \begin{Verbatim}[commandchars=\\\{\}]
/home/sean/miniconda3/envs/prelim/lib/python3.10/site-
packages/torch/nn/parallel/\_functions.py:68: UserWarning: Was asked to gather
along dimension 0, but all input tensors were scalars; will instead unsqueeze
and return a vector.
  warnings.warn('Was asked to gather along dimension 0, but all '
/home/sean/miniconda3/envs/prelim/lib/python3.10/site-
packages/torch/nn/parallel/\_functions.py:68: UserWarning: Was asked to gather
along dimension 0, but all input tensors were scalars; will instead unsqueeze
and return a vector.
  warnings.warn('Was asked to gather along dimension 0, but all '
/home/sean/miniconda3/envs/prelim/lib/python3.10/site-
packages/torch/nn/parallel/\_functions.py:68: UserWarning: Was asked to gather
along dimension 0, but all input tensors were scalars; will instead unsqueeze
and return a vector.
  warnings.warn('Was asked to gather along dimension 0, but all '
/home/sean/miniconda3/envs/prelim/lib/python3.10/site-
packages/torch/nn/parallel/\_functions.py:68: UserWarning: Was asked to gather
along dimension 0, but all input tensors were scalars; will instead unsqueeze
and return a vector.
  warnings.warn('Was asked to gather along dimension 0, but all '
/home/sean/miniconda3/envs/prelim/lib/python3.10/site-
packages/torch/nn/parallel/\_functions.py:68: UserWarning: Was asked to gather
along dimension 0, but all input tensors were scalars; will instead unsqueeze
and return a vector.
  warnings.warn('Was asked to gather along dimension 0, but all '
/home/sean/miniconda3/envs/prelim/lib/python3.10/site-
packages/torch/nn/parallel/\_functions.py:68: UserWarning: Was asked to gather
along dimension 0, but all input tensors were scalars; will instead unsqueeze
and return a vector.
  warnings.warn('Was asked to gather along dimension 0, but all '
    \end{Verbatim}

    \begin{Verbatim}[commandchars=\\\{\}]
<IPython.core.display.HTML object>
    \end{Verbatim}

    \begin{Verbatim}[commandchars=\\\{\}]
/home/sean/miniconda3/envs/prelim/lib/python3.10/site-
packages/torch/nn/parallel/\_functions.py:68: UserWarning: Was asked to gather
along dimension 0, but all input tensors were scalars; will instead unsqueeze
and return a vector.
  warnings.warn('Was asked to gather along dimension 0, but all '
    \end{Verbatim}

    \begin{Verbatim}[commandchars=\\\{\}]
<IPython.core.display.HTML object>
    \end{Verbatim}

    \begin{Verbatim}[commandchars=\\\{\}]
/home/sean/miniconda3/envs/prelim/lib/python3.10/site-
packages/torch/nn/parallel/\_functions.py:68: UserWarning: Was asked to gather
along dimension 0, but all input tensors were scalars; will instead unsqueeze
and return a vector.
  warnings.warn('Was asked to gather along dimension 0, but all '
    \end{Verbatim}

    \begin{Verbatim}[commandchars=\\\{\}]
<IPython.core.display.HTML object>
    \end{Verbatim}

    \begin{Verbatim}[commandchars=\\\{\}]
/home/sean/miniconda3/envs/prelim/lib/python3.10/site-
packages/torch/nn/parallel/\_functions.py:68: UserWarning: Was asked to gather
along dimension 0, but all input tensors were scalars; will instead unsqueeze
and return a vector.
  warnings.warn('Was asked to gather along dimension 0, but all '
    \end{Verbatim}

    \begin{Verbatim}[commandchars=\\\{\}]
<IPython.core.display.HTML object>
    \end{Verbatim}

    \begin{Verbatim}[commandchars=\\\{\}]
/home/sean/miniconda3/envs/prelim/lib/python3.10/site-
packages/torch/nn/parallel/\_functions.py:68: UserWarning: Was asked to gather
along dimension 0, but all input tensors were scalars; will instead unsqueeze
and return a vector.
  warnings.warn('Was asked to gather along dimension 0, but all '
    \end{Verbatim}

    \begin{Verbatim}[commandchars=\\\{\}]
<IPython.core.display.HTML object>
    \end{Verbatim}

    \begin{Verbatim}[commandchars=\\\{\}]
/home/sean/miniconda3/envs/prelim/lib/python3.10/site-
packages/torch/nn/parallel/\_functions.py:68: UserWarning: Was asked to gather
along dimension 0, but all input tensors were scalars; will instead unsqueeze
and return a vector.
  warnings.warn('Was asked to gather along dimension 0, but all '
    \end{Verbatim}

    \begin{Verbatim}[commandchars=\\\{\}]
<IPython.core.display.HTML object>
    \end{Verbatim}

    \begin{tcolorbox}[breakable, size=fbox, boxrule=1pt, pad at break*=1mm,colback=cellbackground, colframe=cellborder]
\prompt{In}{incolor}{ }{\boxspacing}
\begin{Verbatim}[commandchars=\\\{\}]
\PY{k}{if} \PY{n}{DO\PYZus{}TRAIN\PYZus{}MODELS}\PY{p}{:}
  \PY{c+c1}{\PYZsh{} polibertweet\PYZhy{}mlm}
  \PY{c+c1}{\PYZsh{} \PYZhy{} https://huggingface.co/kornosk/polibertweet\PYZhy{}political\PYZhy{}twitter\PYZhy{}roberta\PYZhy{}mlm}
  \PY{n}{train}\PY{p}{(}\PY{l+s+s2}{\PYZdq{}}\PY{l+s+s2}{kornosk/polibertweet\PYZhy{}mlm}\PY{l+s+s2}{\PYZdq{}}\PY{p}{)}
\end{Verbatim}
\end{tcolorbox}

    \begin{Verbatim}[commandchars=\\\{\}]
Some weights of the model checkpoint at kornosk/polibertweet-mlm were not used
when initializing RobertaForSequenceClassification: ['lm\_head.layer\_norm.bias',
'lm\_head.layer\_norm.weight', 'lm\_head.dense.bias', 'lm\_head.decoder.bias',
'lm\_head.dense.weight', 'lm\_head.decoder.weight', 'lm\_head.bias']
- This IS expected if you are initializing RobertaForSequenceClassification from
the checkpoint of a model trained on another task or with another architecture
(e.g. initializing a BertForSequenceClassification model from a
BertForPreTraining model).
- This IS NOT expected if you are initializing RobertaForSequenceClassification
from the checkpoint of a model that you expect to be exactly identical
(initializing a BertForSequenceClassification model from a
BertForSequenceClassification model).
Some weights of RobertaForSequenceClassification were not initialized from the
model checkpoint at kornosk/polibertweet-mlm and are newly initialized:
['classifier.out\_proj.bias', 'classifier.dense.weight',
'classifier.out\_proj.weight', 'classifier.dense.bias']
You should probably TRAIN this model on a down-stream task to be able to use it
for predictions and inference.
/home/sean/miniconda3/envs/prelim/lib/python3.10/site-
packages/torch/nn/parallel/\_functions.py:68: UserWarning: Was asked to gather
along dimension 0, but all input tensors were scalars; will instead unsqueeze
and return a vector.
  warnings.warn('Was asked to gather along dimension 0, but all '
    \end{Verbatim}

    \begin{Verbatim}[commandchars=\\\{\}]
<IPython.core.display.HTML object>
    \end{Verbatim}

    \begin{Verbatim}[commandchars=\\\{\}]
/home/sean/miniconda3/envs/prelim/lib/python3.10/site-
packages/torch/nn/parallel/\_functions.py:68: UserWarning: Was asked to gather
along dimension 0, but all input tensors were scalars; will instead unsqueeze
and return a vector.
  warnings.warn('Was asked to gather along dimension 0, but all '
/home/sean/miniconda3/envs/prelim/lib/python3.10/site-
packages/torch/nn/parallel/\_functions.py:68: UserWarning: Was asked to gather
along dimension 0, but all input tensors were scalars; will instead unsqueeze
and return a vector.
  warnings.warn('Was asked to gather along dimension 0, but all '
/home/sean/miniconda3/envs/prelim/lib/python3.10/site-
packages/torch/nn/parallel/\_functions.py:68: UserWarning: Was asked to gather
along dimension 0, but all input tensors were scalars; will instead unsqueeze
and return a vector.
  warnings.warn('Was asked to gather along dimension 0, but all '
/home/sean/miniconda3/envs/prelim/lib/python3.10/site-
packages/torch/nn/parallel/\_functions.py:68: UserWarning: Was asked to gather
along dimension 0, but all input tensors were scalars; will instead unsqueeze
and return a vector.
  warnings.warn('Was asked to gather along dimension 0, but all '
/home/sean/miniconda3/envs/prelim/lib/python3.10/site-
packages/torch/nn/parallel/\_functions.py:68: UserWarning: Was asked to gather
along dimension 0, but all input tensors were scalars; will instead unsqueeze
and return a vector.
  warnings.warn('Was asked to gather along dimension 0, but all '
/home/sean/miniconda3/envs/prelim/lib/python3.10/site-
packages/torch/nn/parallel/\_functions.py:68: UserWarning: Was asked to gather
along dimension 0, but all input tensors were scalars; will instead unsqueeze
and return a vector.
  warnings.warn('Was asked to gather along dimension 0, but all '
    \end{Verbatim}

    \begin{Verbatim}[commandchars=\\\{\}]
<IPython.core.display.HTML object>
    \end{Verbatim}

    \begin{Verbatim}[commandchars=\\\{\}]
/home/sean/miniconda3/envs/prelim/lib/python3.10/site-
packages/torch/nn/parallel/\_functions.py:68: UserWarning: Was asked to gather
along dimension 0, but all input tensors were scalars; will instead unsqueeze
and return a vector.
  warnings.warn('Was asked to gather along dimension 0, but all '
    \end{Verbatim}

    \begin{Verbatim}[commandchars=\\\{\}]
<IPython.core.display.HTML object>
    \end{Verbatim}

    \begin{Verbatim}[commandchars=\\\{\}]
/home/sean/miniconda3/envs/prelim/lib/python3.10/site-
packages/torch/nn/parallel/\_functions.py:68: UserWarning: Was asked to gather
along dimension 0, but all input tensors were scalars; will instead unsqueeze
and return a vector.
  warnings.warn('Was asked to gather along dimension 0, but all '
    \end{Verbatim}

    \begin{Verbatim}[commandchars=\\\{\}]
<IPython.core.display.HTML object>
    \end{Verbatim}

    \begin{Verbatim}[commandchars=\\\{\}]
/home/sean/miniconda3/envs/prelim/lib/python3.10/site-
packages/torch/nn/parallel/\_functions.py:68: UserWarning: Was asked to gather
along dimension 0, but all input tensors were scalars; will instead unsqueeze
and return a vector.
  warnings.warn('Was asked to gather along dimension 0, but all '
    \end{Verbatim}

    \begin{Verbatim}[commandchars=\\\{\}]
<IPython.core.display.HTML object>
    \end{Verbatim}

    \begin{Verbatim}[commandchars=\\\{\}]
/home/sean/miniconda3/envs/prelim/lib/python3.10/site-
packages/torch/nn/parallel/\_functions.py:68: UserWarning: Was asked to gather
along dimension 0, but all input tensors were scalars; will instead unsqueeze
and return a vector.
  warnings.warn('Was asked to gather along dimension 0, but all '
    \end{Verbatim}

    \begin{Verbatim}[commandchars=\\\{\}]
<IPython.core.display.HTML object>
    \end{Verbatim}

    \begin{Verbatim}[commandchars=\\\{\}]
/home/sean/miniconda3/envs/prelim/lib/python3.10/site-
packages/torch/nn/parallel/\_functions.py:68: UserWarning: Was asked to gather
along dimension 0, but all input tensors were scalars; will instead unsqueeze
and return a vector.
  warnings.warn('Was asked to gather along dimension 0, but all '
    \end{Verbatim}

    \begin{Verbatim}[commandchars=\\\{\}]
<IPython.core.display.HTML object>
    \end{Verbatim}

    \hypertarget{compare-the-performance-of-the-different-models}{%
\section{Compare the performance of the different
models}\label{compare-the-performance-of-the-different-models}}

    \begin{tcolorbox}[breakable, size=fbox, boxrule=1pt, pad at break*=1mm,colback=cellbackground, colframe=cellborder]
\prompt{In}{incolor}{ }{\boxspacing}
\begin{Verbatim}[commandchars=\\\{\}]
\PY{k+kn}{from} \PY{n+nn}{result\PYZus{}summarizer} \PY{k+kn}{import} \PY{n}{ResultSummarizer}
\end{Verbatim}
\end{tcolorbox}

    \begin{tcolorbox}[breakable, size=fbox, boxrule=1pt, pad at break*=1mm,colback=cellbackground, colframe=cellborder]
\prompt{In}{incolor}{ }{\boxspacing}
\begin{Verbatim}[commandchars=\\\{\}]
\PY{n}{LIST\PYZus{}MODEL\PYZus{}NAME} \PY{o}{=} \PY{p}{[}\PY{l+s+s2}{\PYZdq{}}\PY{l+s+s2}{bert\PYZhy{}base\PYZhy{}uncased}\PY{l+s+s2}{\PYZdq{}}\PY{p}{,}\PY{l+s+s2}{\PYZdq{}}\PY{l+s+s2}{vinai\PYZus{}bertweet\PYZus{}base}\PY{l+s+s2}{\PYZdq{}}\PY{p}{,}\PY{l+s+s2}{\PYZdq{}}\PY{l+s+s2}{kornosk\PYZus{}polibertweet\PYZus{}mlm}\PY{l+s+s2}{\PYZdq{}}\PY{p}{]}
\PY{c+c1}{\PYZsh{} summarize the results}
\PY{c+c1}{\PYZsh{} \PYZhy{} bert\PYZhy{}base\PYZhy{}uncased}
\PY{n}{result\PYZus{}summarizer} \PY{o}{=} \PY{n}{ResultSummarizer}\PY{p}{(}\PY{n}{dataset}\PY{o}{=}\PY{n}{DATASET}\PY{p}{,}
                                     \PY{n}{list\PYZus{}version\PYZus{}output}\PY{o}{=}\PY{n}{LIST\PYZus{}MODEL\PYZus{}NAME}\PY{p}{,}
                                     \PY{n}{eval\PYZus{}mode}\PY{o}{=}\PY{l+s+s2}{\PYZdq{}}\PY{l+s+s2}{single\PYZus{}domain}\PY{l+s+s2}{\PYZdq{}}\PY{p}{,}
                                     \PY{n}{model\PYZus{}type}\PY{o}{=}\PY{l+s+s2}{\PYZdq{}}\PY{l+s+s2}{single\PYZus{}domain\PYZus{}baseline}\PY{l+s+s2}{\PYZdq{}}\PY{p}{,}
                                     \PY{n}{task}\PY{o}{=}\PY{k+kc}{None}\PY{p}{,}
                                     \PY{n}{file\PYZus{}name\PYZus{}metrics}\PY{o}{=}\PY{l+s+s2}{\PYZdq{}}\PY{l+s+s2}{metrics.csv}\PY{l+s+s2}{\PYZdq{}}\PY{p}{,}
                                     \PY{n}{file\PYZus{}name\PYZus{}confusion\PYZus{}mat}\PY{o}{=}\PY{l+s+s2}{\PYZdq{}}\PY{l+s+s2}{confusion\PYZus{}matrix.csv}\PY{l+s+s2}{\PYZdq{}}\PY{p}{,}
                                     \PY{n}{path\PYZus{}input\PYZus{}root}\PY{o}{=}\PY{n}{PATH\PYZus{}OUTPUT\PYZus{}ROOT}\PY{p}{,}
                                     \PY{n}{path\PYZus{}output}\PY{o}{=}\PY{n}{join}\PY{p}{(}\PY{n}{PATH\PYZus{}OUTPUT\PYZus{}ROOT}\PY{p}{,} \PY{l+s+s2}{\PYZdq{}}\PY{l+s+s2}{summary}\PY{l+s+s2}{\PYZdq{}}\PY{p}{)}\PY{p}{)}
\PY{c+c1}{\PYZsh{} write the summary to a csv file}
\PY{n}{df\PYZus{}hightlight\PYZus{}metrics} \PY{o}{=} \PY{n}{result\PYZus{}summarizer}\PY{o}{.}\PY{n}{write\PYZus{}hightlight\PYZus{}metrics\PYZus{}to\PYZus{}summary\PYZus{}csv}\PY{p}{(}
                        \PY{n}{list\PYZus{}metrics\PYZus{}highlight}\PY{o}{=}\PY{p}{[}\PY{l+s+s1}{\PYZsq{}}\PY{l+s+s1}{f1\PYZus{}macro}\PY{l+s+s1}{\PYZsq{}}\PY{p}{,} \PY{l+s+s1}{\PYZsq{}}\PY{l+s+s1}{f1\PYZus{}NONE}\PY{l+s+s1}{\PYZsq{}}\PY{p}{,} \PY{l+s+s1}{\PYZsq{}}\PY{l+s+s1}{f1\PYZus{}FAVOR}\PY{l+s+s1}{\PYZsq{}}\PY{p}{,} \PY{l+s+s1}{\PYZsq{}}\PY{l+s+s1}{f1\PYZus{}AGAINST}\PY{l+s+s1}{\PYZsq{}}\PY{p}{]}\PY{p}{,}
                        \PY{n}{list\PYZus{}sets\PYZus{}highlight}\PY{o}{=}\PY{p}{[}\PY{l+s+s1}{\PYZsq{}}\PY{l+s+s1}{train\PYZus{}raw}\PY{l+s+s1}{\PYZsq{}}\PY{p}{,} \PY{l+s+s1}{\PYZsq{}}\PY{l+s+s1}{vali\PYZus{}raw}\PY{l+s+s1}{\PYZsq{}}\PY{p}{,} \PY{l+s+s1}{\PYZsq{}}\PY{l+s+s1}{test\PYZus{}raw}\PY{l+s+s1}{\PYZsq{}}\PY{p}{]}\PY{p}{,}
                        \PY{n}{col\PYZus{}name\PYZus{}set}\PY{o}{=}\PY{l+s+s2}{\PYZdq{}}\PY{l+s+s2}{set}\PY{l+s+s2}{\PYZdq{}}\PY{p}{)}
\PY{c+c1}{\PYZsh{} reorder the rows}
\PY{n}{df\PYZus{}hightlight\PYZus{}metrics}\PY{p}{[}\PY{l+s+s1}{\PYZsq{}}\PY{l+s+s1}{version}\PY{l+s+s1}{\PYZsq{}}\PY{p}{]} \PY{o}{=} \PY{n}{pd}\PY{o}{.}\PY{n}{Categorical}\PY{p}{(}\PY{n}{df\PYZus{}hightlight\PYZus{}metrics}\PY{p}{[}\PY{l+s+s1}{\PYZsq{}}\PY{l+s+s1}{version}\PY{l+s+s1}{\PYZsq{}}\PY{p}{]}\PY{p}{,} \PY{n}{categories}\PY{o}{=}\PY{n}{LIST\PYZus{}MODEL\PYZus{}NAME}\PY{p}{,} \PY{n}{ordered}\PY{o}{=}\PY{k+kc}{True}\PY{p}{)}
\PY{n}{df\PYZus{}hightlight\PYZus{}metrics} \PY{o}{=} \PY{n}{df\PYZus{}hightlight\PYZus{}metrics}\PY{o}{.}\PY{n}{sort\PYZus{}values}\PY{p}{(}\PY{l+s+s1}{\PYZsq{}}\PY{l+s+s1}{version}\PY{l+s+s1}{\PYZsq{}}\PY{p}{)}\PY{o}{.}\PY{n}{reset\PYZus{}index}\PY{p}{(}\PY{n}{drop}\PY{o}{=}\PY{k+kc}{True}\PY{p}{)}

\PY{c+c1}{\PYZsh{} visualize the results and save the figures}
\PY{n}{plot\PYZus{}con\PYZus{}mat} \PY{o}{=} \PY{n}{result\PYZus{}summarizer}\PY{o}{.}\PY{n}{visualize\PYZus{}confusion\PYZus{}metrices\PYZus{}over\PYZus{}domains\PYZus{}comb}\PY{p}{(}
    \PY{p}{[}\PY{l+s+s2}{\PYZdq{}}\PY{l+s+s2}{train\PYZus{}raw}\PY{l+s+s2}{\PYZdq{}}\PY{p}{,} \PY{l+s+s2}{\PYZdq{}}\PY{l+s+s2}{vali\PYZus{}raw}\PY{l+s+s2}{\PYZdq{}}\PY{p}{,} \PY{l+s+s2}{\PYZdq{}}\PY{l+s+s2}{test\PYZus{}raw}\PY{l+s+s2}{\PYZdq{}}\PY{p}{]}\PY{p}{,}
    \PY{n}{preserve\PYZus{}order\PYZus{}list\PYZus{}sets}\PY{o}{=}\PY{k+kc}{True}\PY{p}{)}
\end{Verbatim}
\end{tcolorbox}

    \begin{Verbatim}[commandchars=\\\{\}]
<Figure size 1500x750 with 0 Axes>
    \end{Verbatim}

    \begin{Verbatim}[commandchars=\\\{\}]
<Figure size 1500x750 with 0 Axes>
    \end{Verbatim}

    \begin{Verbatim}[commandchars=\\\{\}]
<Figure size 1500x750 with 0 Axes>
    \end{Verbatim}

    \begin{tcolorbox}[breakable, size=fbox, boxrule=1pt, pad at break*=1mm,colback=cellbackground, colframe=cellborder]
\prompt{In}{incolor}{ }{\boxspacing}
\begin{Verbatim}[commandchars=\\\{\}]
\PY{c+c1}{\PYZsh{} print the summary table}
\PY{n}{df\PYZus{}hightlight\PYZus{}metrics}\PY{p}{[}\PY{p}{[}\PY{l+s+s2}{\PYZdq{}}\PY{l+s+s2}{version}\PY{l+s+s2}{\PYZdq{}}\PY{p}{,}\PY{l+s+s2}{\PYZdq{}}\PY{l+s+s2}{set}\PY{l+s+s2}{\PYZdq{}}\PY{p}{,}\PY{l+s+s2}{\PYZdq{}}\PY{l+s+s2}{f1\PYZus{}macro}\PY{l+s+s2}{\PYZdq{}}\PY{p}{,}\PY{l+s+s2}{\PYZdq{}}\PY{l+s+s2}{f1\PYZus{}NONE}\PY{l+s+s2}{\PYZdq{}}\PY{p}{,}\PY{l+s+s2}{\PYZdq{}}\PY{l+s+s2}{f1\PYZus{}FAVOR}\PY{l+s+s2}{\PYZdq{}}\PY{p}{,}\PY{l+s+s2}{\PYZdq{}}\PY{l+s+s2}{f1\PYZus{}AGAINST}\PY{l+s+s2}{\PYZdq{}}\PY{p}{]}\PY{p}{]}\PY{p}{[}\PY{n}{df\PYZus{}hightlight\PYZus{}metrics}\PY{o}{.}\PY{n}{set}\PY{o}{.}\PY{n}{isin}\PY{p}{(}\PY{p}{[}\PY{l+s+s2}{\PYZdq{}}\PY{l+s+s2}{test\PYZus{}raw}\PY{l+s+s2}{\PYZdq{}}\PY{p}{]}\PY{p}{)}\PY{p}{]}
\end{Verbatim}
\end{tcolorbox}

            \begin{tcolorbox}[breakable, size=fbox, boxrule=.5pt, pad at break*=1mm, opacityfill=0]
\prompt{Out}{outcolor}{ }{\boxspacing}
\begin{Verbatim}[commandchars=\\\{\}]
                    version       set  f1\_macro  f1\_NONE  f1\_FAVOR  f1\_AGAINST
2         bert-base-uncased  test\_raw    0.4748   0.4196    0.4275      0.5775
5       vinai\_bertweet\_base  test\_raw    0.5797   0.5323    0.5401      0.6667
8  kornosk\_polibertweet\_mlm  test\_raw    0.5616   0.5440    0.4762      0.6645
\end{Verbatim}
\end{tcolorbox}
        
    In this tutorial, we compared the performance of three different BERT
models for stance detection: \texttt{bert-base-uncased},
\texttt{vinai\_bertweet\_base}, and \texttt{kornosk\_polibertweet\_mlm}.
The latter two models are domain-specific, designed specifically for
tweets. The results show that both domain-specific models outperform the
general \texttt{bert-base-uncased} model in terms of macro-F1 scores.
The \texttt{vinai\_bertweet\_base} model achieves the best performance
with an macro-F1 score of 0.5797, followed by
\texttt{kornosk\_polibertweet\_mlm} with a score of 0.5616, and finally,
the \texttt{bert-base-uncased} model with a score of 0.4748. This
demonstrates the advantage of using domain-specific models when dealing
with tasks that involve specific types of data, such as social media
text.

    \hypertarget{analyzing-the-confusion-matrix-for-deeper-insights}{%
\subsection{Analyzing the Confusion Matrix for Deeper
Insights}\label{analyzing-the-confusion-matrix-for-deeper-insights}}

    Given the performance differences observed among the three models, it is
valuable to investigate their ``confusion matrices'' for each model
Examining these matrices can also help identify potential biases,
challenges, and opportunities for improvement.

    \begin{quote}
Note: Examining the confusion matrix is essential because it provides a
detailed overview of the model's performance across different classes.
It reveals not only the correct predictions (true positives) but also
the instances where the model made errors (false positives and false
negatives). By analyzing the confusion matrix, we can identify patterns
in misclassifications and gain insights into the strengths and
weaknesses of the model. Here is a great tutorial on how to interpret
the confusion matrix and its relationships with macro-F1 scores:
https://towardsdatascience.com/confusion-matrix-for-your-multi-class-machine-learning-model-ff9aa3bf7826
\end{quote}

    Below are the confusion matrices for the standard
\texttt{bert-base-uncased} model.

The second row are the matrices for the training set, the second row are
for the validation set, and the third row are for the test set.

Each row of matrices consists of three types:

\begin{enumerate}
\def\labelenumi{\arabic{enumi}.}
\tightlist
\item
  The leftmost matrices are the raw confusion matrices.
\item
  The middle matrices show the confusion matrices normalized by row
  (i.e., the sum of each row equals 100). In these matrices, the
  diagonal values correspond to the recall value of each class.
\item
  The rightmost matrices illustrate the confusion matrices normalized by
  column (i.e., the sum of each column equals 100). In these matrices,
  the diagonal values are the precision value for each class.
\end{enumerate}

In each matrix, the rows represent the true labels, and the columns
represent the predicted labels. The diagonal elements denote correct
predictions, while the off-diagonal elements indicate incorrect
predictions.

    \begin{tcolorbox}[breakable, size=fbox, boxrule=1pt, pad at break*=1mm,colback=cellbackground, colframe=cellborder]
\prompt{In}{incolor}{ }{\boxspacing}
\begin{Verbatim}[commandchars=\\\{\}]
\PY{n}{display\PYZus{}resized\PYZus{}image\PYZus{}in\PYZus{}notebook}\PY{p}{(}\PY{n}{join}\PY{p}{(}\PY{n}{PATH\PYZus{}OUTPUT\PYZus{}ROOT}\PY{p}{,}\PY{l+s+s2}{\PYZdq{}}\PY{l+s+s2}{summary}\PY{l+s+s2}{\PYZdq{}}\PY{p}{,}\PY{l+s+s2}{\PYZdq{}}\PY{l+s+s2}{bert\PYZhy{}base\PYZhy{}uncased\PYZus{}comb\PYZus{}confusion\PYZus{}mat.png}\PY{l+s+s2}{\PYZdq{}}\PY{p}{)}\PY{p}{)}
\end{Verbatim}
\end{tcolorbox}

    \begin{center}
    \adjustimage{max size={0.9\linewidth}{0.9\paperheight}}{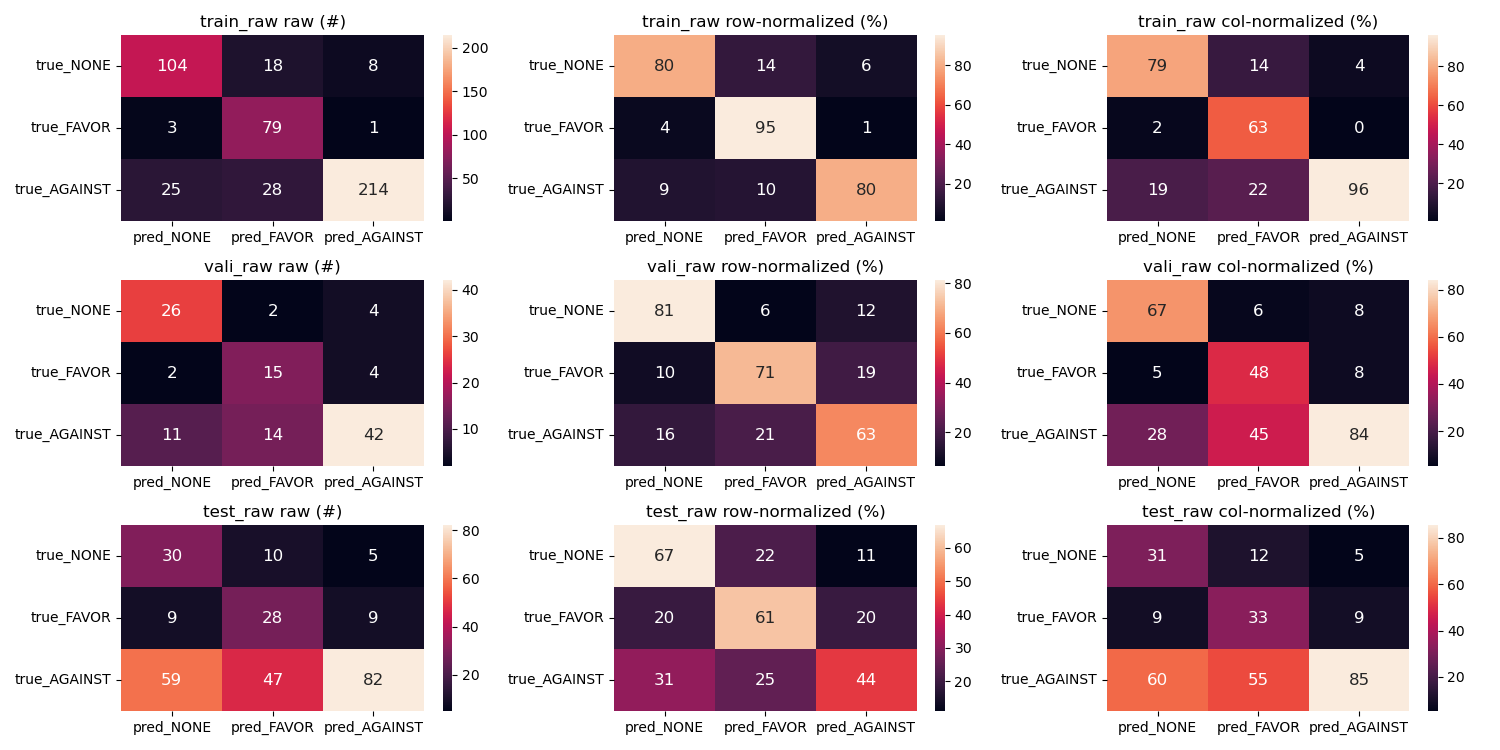}
    \end{center}
    { \hspace*{\fill} \\}
    
    As shown in the test set confusion matrices, the model demonstrates
proficiency in distinguishing between the \texttt{FAVOR} and
\texttt{NONE} stances. However, it faces challenges in accurately
predicting the \texttt{AGAINST} stance, frequently misclassifying them
as \texttt{NONE}.

    Now, let's examine the confusion matrix for the domain-specific
\texttt{vinai\_bertweet\_base} model, which has the highest macro-F1
score.

The confusion matrices below reveal that the model is slight better at
classifying all three stance classes compared to the standard
\texttt{bert-base-uncased} model.

    \begin{tcolorbox}[breakable, size=fbox, boxrule=1pt, pad at break*=1mm,colback=cellbackground, colframe=cellborder]
\prompt{In}{incolor}{ }{\boxspacing}
\begin{Verbatim}[commandchars=\\\{\}]
\PY{n}{display\PYZus{}resized\PYZus{}image\PYZus{}in\PYZus{}notebook}\PY{p}{(}\PY{n}{join}\PY{p}{(}\PY{n}{PATH\PYZus{}OUTPUT\PYZus{}ROOT}\PY{p}{,}\PY{l+s+s2}{\PYZdq{}}\PY{l+s+s2}{summary}\PY{l+s+s2}{\PYZdq{}}\PY{p}{,}\PY{l+s+s2}{\PYZdq{}}\PY{l+s+s2}{vinai\PYZus{}bertweet\PYZus{}base\PYZus{}comb\PYZus{}confusion\PYZus{}mat.png}\PY{l+s+s2}{\PYZdq{}}\PY{p}{)}\PY{p}{)}
\end{Verbatim}
\end{tcolorbox}

    \begin{center}
    \adjustimage{max size={0.9\linewidth}{0.9\paperheight}}{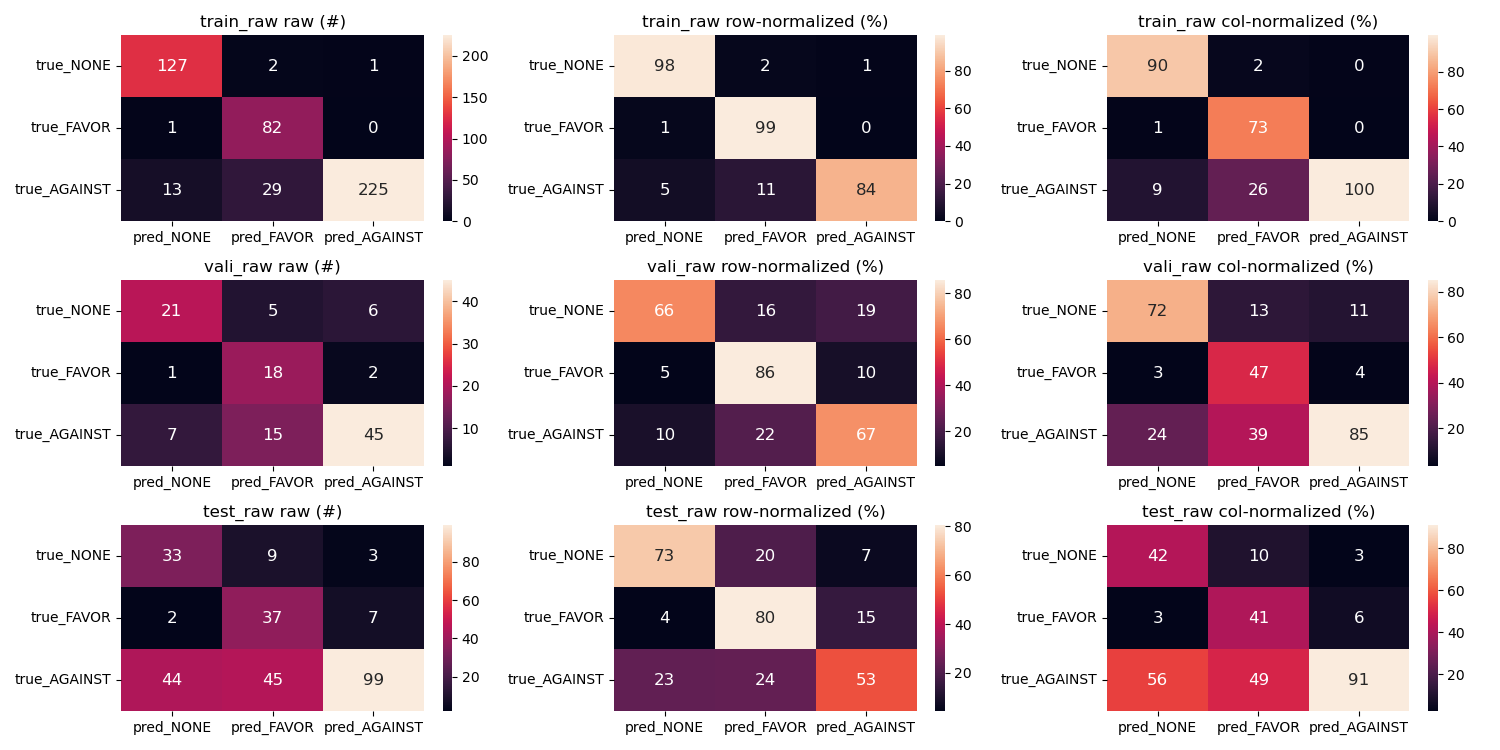}
    \end{center}
    { \hspace*{\fill} \\}
    
    \begin{center}\rule{0.5\linewidth}{0.5pt}\end{center}

    \hypertarget{conclusion}{%
\section{Conclusion}\label{conclusion}}

    In this first part of the tutorial series, we have explored stance
detection, its importance, and one of the two distinct paradigms for
approaching the task: tuning BERT. We discussed the concept of BERT, its
architecture, and the fundamentals of transfer learning with pre-trained
models. Additionally, we covered the process of tokenization in BERT,
which is crucial for preparing input data.

We then walked through the process of fine-tuning a BERT model using the
HuggingFace Transformers library. This involved installing the library,
loading a pre-trained BERT model, preparing and processing the labeled
dataset, fine-tuning the model for stance detection, and finally
evaluating the model and analyzing the results.

Throughout the tutorial, we emphasized the benefits of using
domain-specific models for tasks involving specific types of data, such
as social media text. In the upcoming
\href{https://colab.research.google.com/drive/1IFr6Iz1YH9XBWUKcWZyTU-1QtxgYqrmX?usp=sharing}{second
part} of the tutorial series, we will cover the second paradigm,
prompting LLMs, and demonstrate its effectiveness in stance detection.

%    \hypertarget{stance-detection-on-tweets-using-nlp-methods---part-2}

    \begin{center}\rule{\linewidth}{1pt}\end{center}
    
\section{Part 2: Stance Detection on Tweets using Large Language Models (LLMs)}
\label{stance-detection-on-tweets-using-nlp-methods---part-2}

Author: Yun-Shiuan Chuang (yunshiuan.chuang@gmail.com)

Note: This tutorial consists of two separate Python notebooks. This
notebook is the second one. The first notebook can be found
\href{https://colab.research.google.com/drive/1nxziaKStwRnSyOLI6pLNBaAnB_aB6IsE?usp=sharing}{here}.
I recommend that you go through the first notebook before the second one
as the second notebook builds on top of the first one.

\begin{enumerate}
\def\labelenumi{\arabic{enumi}.}
\tightlist
\item
  First notebook: Fine-tuning BERT models: include standard BERT and
  domain-specific BERT
\end{enumerate}

\begin{itemize}
\tightlist
\item
  https://colab.research.google.com/drive/1nxziaKStwRnSyOLI6pLNBaAnB\_aB6IsE?usp=sharing
\end{itemize}

\begin{enumerate}
\def\labelenumi{\arabic{enumi}.}
\setcounter{enumi}{1}
\tightlist
\item
  Second notebook (this one): Prompting large language models (LLMs):
  include ChatGPT, FLAN-T5 and different prompt types (zero-shot,
  few-shot, chain-of-thought)
\end{enumerate}

\begin{itemize}
\tightlist
\item
  https://colab.research.google.com/drive/1IFr6Iz1YH9XBWUKcWZyTU-1QtxgYqrmX?usp=sharing
\end{itemize}

    \hypertarget{getting-started-overview-prerequisites-and-setup}{%
\subsection{Getting Started: Overview, Prerequisites, and
Setup}\label{getting-started-overview-prerequisites-and-setup}}

\textbf{Objective of the tutorial}: This tutorial will guide you through
the process of stance detection on tweets using two main approaches:
fine-tuning a BERT model and using large language models (LLMs).

\textbf{Prerequisites}:

\begin{itemize}
\tightlist
\item
  If you want to run the tutorial without editting the codes but want to
  understand the content

  \begin{itemize}
  \tightlist
  \item
    Understand what transformer is and what BERT is. If you are not
    familiar with these concepts, I strongly recommend that you go
    through the first tutorial
    \href{https://colab.research.google.com/drive/1nxziaKStwRnSyOLI6pLNBaAnB_aB6IsE?usp=sharing}{here}
    before this one.
  \item
    Basic Python skills: functions, classes, pandas, etc.
  \item
    Basic ML knowledge: train-validation-test split, F1 score, forward
    pass, backpropagation etc.
  \end{itemize}
\end{itemize}

    \textbf{Acknowledgements}

\begin{itemize}
\tightlist
\item
  While the application of LLMs on stance detection is my own work, some
  part of this tutorials, e.g., GPT-3, are inspired by the following
  tutorials. Some of the figures are also modified from the images in
  these tutorials. I highly recommend you check them out if you want to
  learn more about LLMs.

  \begin{itemize}
  \tightlist
  \item
    https://jalammar.github.io/illustrated-gpt2/\#part-1-got-and-language-modeling
  \item
    https://jalammar.github.io/how-gpt3-works-visualizations-animations/
  \end{itemize}
\item
  This tutorial was created with the assistance of ChatGPT (GPT-4), a
  cutting-edge language model developed by OpenAI. The AI-aided writing
  process involved an iterative approach, where I provided the model
  with ideas for each section and GPT-4 transformed those ideas into
  well-structured paragraphs. Even the outline itself underwent a
  similar iterative process to refine and improve the tutorial
  structure. Following this, I fact-checked and revised the generated
  content, asking GPT-4 to make further revisions based on my
  evaluation, until I took over and finalized the content.
\end{itemize}

    \textbf{Setup}

    \begin{enumerate}
\def\labelenumi{\arabic{enumi}.}
\tightlist
\item
  Before we begin with Google Colab, please ensure that you have
  selected the GPU runtime. To do this, go to \texttt{Runtime}
  -\textgreater{} \texttt{Change\ runtime\ type} -\textgreater{}
  \texttt{Hardware\ accelerator} -\textgreater{} \texttt{GPU}. This will
  ensure that the note will run more efficiently and quickly.
\end{enumerate}

    \begin{enumerate}
\def\labelenumi{\arabic{enumi}.}
\setcounter{enumi}{1}
\tightlist
\item
  Now, let's download the content of this tutorial and install the
  necessary libraries by running the following cell.
\end{enumerate}

    \begin{tcolorbox}[breakable, size=fbox, boxrule=1pt, pad at break*=1mm,colback=cellbackground, colframe=cellborder]
\prompt{In}{incolor}{1}{\boxspacing}
\begin{Verbatim}[commandchars=\\\{\}]
\PY{k+kn}{from} \PY{n+nn}{os}\PY{n+nn}{.}\PY{n+nn}{path} \PY{k+kn}{import} \PY{n}{join}
\PY{n}{ON\PYZus{}COLAB} \PY{o}{=} \PY{k+kc}{True}
\PY{k}{if} \PY{n}{ON\PYZus{}COLAB}\PY{p}{:}
  \PY{o}{!}git\PY{+w}{ }clone\PY{+w}{ }\PYZhy{}\PYZhy{}single\PYZhy{}branch\PY{+w}{ }\PYZhy{}\PYZhy{}branch\PY{+w}{ }colab\PY{+w}{ }https://github.com/yunshiuan/prelim\PYZus{}stance\PYZus{}detection.git
  \PY{o}{!}python\PY{+w}{ }\PYZhy{}m\PY{+w}{ }pip\PY{+w}{ }install\PY{+w}{ }pandas\PY{+w}{ }datasets\PY{+w}{ }openai\PY{+w}{ }tiktoken\PY{+w}{ }accelerate\PY{+w}{ }transformers\PY{+w}{ }transformers\PY{o}{[}sentencepiece\PY{o}{]}\PY{+w}{ }\PY{n+nv}{torch}\PY{o}{=}\PY{o}{=}\PY{l+m}{1}.12.1+cu113\PY{+w}{ }\PYZhy{}f\PY{+w}{ }https://download.pytorch.org/whl/torch\PYZus{}stable.html\PY{+w}{ }emoji\PY{+w}{ }\PYZhy{}q
  \PY{o}{\PYZpc{}}\PY{k}{cd} /content/prelim\PYZus{}stance\PYZus{}detection/scripts
\PY{k}{else}\PY{p}{:}
  \PY{c+c1}{\PYZsh{} if you are not on colab, you have to set up the environment by yourself. You would also need a machine with GPU.}
  \PY{o}{\PYZpc{}}\PY{k}{cd} scripts
\end{Verbatim}
\end{tcolorbox}

    \begin{Verbatim}[commandchars=\\\{\}]
Cloning into 'prelim\_stance\_detection'{\ldots}
remote: Enumerating objects: 513, done.
remote: Counting objects: 100\% (36/36), done.
remote: Compressing objects: 100\% (24/24), done.
remote: Total 513 (delta 21), reused 24 (delta 12), pack-reused 477
Receiving objects: 100\% (513/513), 58.56 MiB | 18.24 MiB/s, done.
Resolving deltas: 100\% (254/254), done.
     \textcolor{ansi-black-intense}{------------------------------------------------------------------------------------------------------------------------} \textcolor{ansi-green}{492.4/492.4
kB} \textcolor{ansi-red}{4.4 MB/s} eta \textcolor{ansi-cyan}{0:00:00}
     \textcolor{ansi-black-intense}{------------------------------------------------------------------------------------------------------------------------} \textcolor{ansi-green}{73.6/73.6 kB}
\textcolor{ansi-red}{6.3 MB/s} eta \textcolor{ansi-cyan}{0:00:00}
     \textcolor{ansi-black-intense}{------------------------------------------------------------------------------------------------------------------------} \textcolor{ansi-green}{1.7/1.7 MB}
\textcolor{ansi-red}{10.9 MB/s} eta \textcolor{ansi-cyan}{0:00:00}
     \textcolor{ansi-black-intense}{------------------------------------------------------------------------------------------------------------------------} \textcolor{ansi-green}{244.2/244.2
kB} \textcolor{ansi-red}{7.5 MB/s} eta \textcolor{ansi-cyan}{0:00:00}
     \textcolor{ansi-black-intense}{------------------------------------------------------------------------------------------------------------------------} \textcolor{ansi-green}{7.4/7.4 MB}
\textcolor{ansi-red}{21.6 MB/s} eta \textcolor{ansi-cyan}{0:00:00}
     \textcolor{ansi-black-intense}{------------------------------------------------------------------------------------------------------------------------} \textcolor{ansi-green}{1.8/1.8 GB}
\textcolor{ansi-red}{555.8 kB/s} eta \textcolor{ansi-cyan}{0:00:00}
     \textcolor{ansi-black-intense}{------------------------------------------------------------------------------------------------------------------------} \textcolor{ansi-green}{361.8/361.8 kB}
\textcolor{ansi-red}{24.3 MB/s} eta \textcolor{ansi-cyan}{0:00:00}
  Installing build dependencies {\ldots} done
  Getting requirements to build wheel {\ldots} done
  Preparing metadata (pyproject.toml) {\ldots} done
     \textcolor{ansi-black-intense}{------------------------------------------------------------------------------------------------------------------------} \textcolor{ansi-green}{115.3/115.3
kB} \textcolor{ansi-red}{9.9 MB/s} eta \textcolor{ansi-cyan}{0:00:00}
     \textcolor{ansi-black-intense}{------------------------------------------------------------------------------------------------------------------------} \textcolor{ansi-green}{212.5/212.5 kB}
\textcolor{ansi-red}{17.4 MB/s} eta \textcolor{ansi-cyan}{0:00:00}
     \textcolor{ansi-black-intense}{------------------------------------------------------------------------------------------------------------------------} \textcolor{ansi-green}{134.8/134.8 kB}
\textcolor{ansi-red}{12.4 MB/s} eta \textcolor{ansi-cyan}{0:00:00}
     \textcolor{ansi-black-intense}{------------------------------------------------------------------------------------------------------------------------} \textcolor{ansi-green}{268.8/268.8 kB}
\textcolor{ansi-red}{22.7 MB/s} eta \textcolor{ansi-cyan}{0:00:00}
     \textcolor{ansi-black-intense}{-----------------------------------------------------------------------------------------------------------------------} \textcolor{ansi-green}{7.8/7.8 MB}
\textcolor{ansi-red}{43.8 MB/s} eta \textcolor{ansi-cyan}{0:00:00}
     \textcolor{ansi-black-intense}{------------------------------------------------------------------------------------------------------------------------} \textcolor{ansi-green}{1.3/1.3 MB}
\textcolor{ansi-red}{49.7 MB/s} eta \textcolor{ansi-cyan}{0:00:00}
     \textcolor{ansi-black-intense}{------------------------------------------------------------------------------------------------------------------------} \textcolor{ansi-green}{1.3/1.3 MB}
\textcolor{ansi-red}{53.6 MB/s} eta \textcolor{ansi-cyan}{0:00:00}
  Building wheel for emoji (pyproject.toml) {\ldots} done
\textcolor{ansi-red}{ERROR: pip's dependency resolver does not currently take into account all
the packages that are installed. This behaviour is the source of the following
dependency conflicts.
torchaudio 2.0.2+cu118 requires torch==2.0.1, but you have torch 1.12.1+cu113
which is incompatible.
torchdata 0.6.1 requires torch==2.0.1, but you have torch 1.12.1+cu113 which is
incompatible.
torchtext 0.15.2 requires torch==2.0.1, but you have torch 1.12.1+cu113 which is
incompatible.
torchvision 0.15.2+cu118 requires torch==2.0.1, but you have torch 1.12.1+cu113
which is incompatible.}\textcolor{ansi-red}{
}/content/prelim\_stance\_detection/scripts
    \end{Verbatim}

    \begin{tcolorbox}[breakable, size=fbox, boxrule=1pt, pad at break*=1mm,colback=cellbackground, colframe=cellborder]
\prompt{In}{incolor}{2}{\boxspacing}
\begin{Verbatim}[commandchars=\\\{\}]
\PY{c+c1}{\PYZsh{} a helper function to load images in the notebook}
\PY{k+kn}{from} \PY{n+nn}{PIL} \PY{k+kn}{import} \PY{n}{Image} \PY{k}{as} \PY{n}{PILImage}
\PY{k+kn}{from} \PY{n+nn}{IPython}\PY{n+nn}{.}\PY{n+nn}{display} \PY{k+kn}{import} \PY{n}{display}\PY{p}{,} \PY{n}{Image}

\PY{k+kn}{from} \PY{n+nn}{parameters\PYZus{}meta} \PY{k+kn}{import} \PY{n}{ParametersMeta} \PY{k}{as} \PY{n}{par}
\PY{n}{PATH\PYZus{}IMAGES} \PY{o}{=} \PY{n}{join}\PY{p}{(}\PY{n}{par}\PY{o}{.}\PY{n}{PATH\PYZus{}ROOT}\PY{p}{,} \PY{l+s+s2}{\PYZdq{}}\PY{l+s+s2}{images}\PY{l+s+s2}{\PYZdq{}}\PY{p}{)}

\PY{k}{def} \PY{n+nf}{display\PYZus{}resized\PYZus{}image\PYZus{}in\PYZus{}notebook}\PY{p}{(}\PY{n}{file\PYZus{}image}\PY{p}{,} \PY{n}{scale}\PY{o}{=}\PY{l+m+mi}{1}\PY{p}{,} \PY{n}{use\PYZus{}default\PYZus{}path}\PY{o}{=}\PY{k+kc}{True}\PY{p}{)}\PY{p}{:}
\PY{+w}{    }\PY{l+s+sd}{\PYZdq{}\PYZdq{}\PYZdq{}Display an image in a notebook.\PYZdq{}\PYZdq{}\PYZdq{}}
    \PY{c+c1}{\PYZsh{} \PYZhy{} https://stackoverflow.com/questions/69654877/how\PYZhy{}to\PYZhy{}set\PYZhy{}image\PYZhy{}size\PYZhy{}to\PYZhy{}display\PYZhy{}in\PYZhy{}ipython\PYZhy{}display}
    \PY{k}{if} \PY{n}{use\PYZus{}default\PYZus{}path}\PY{p}{:}
        \PY{n}{file\PYZus{}image} \PY{o}{=} \PY{n}{join}\PY{p}{(}\PY{n}{PATH\PYZus{}IMAGES}\PY{p}{,} \PY{n}{file\PYZus{}image}\PY{p}{)}
    \PY{n}{image} \PY{o}{=} \PY{n}{PILImage}\PY{o}{.}\PY{n}{open}\PY{p}{(}\PY{n}{file\PYZus{}image}\PY{p}{)}
    \PY{n}{display}\PY{p}{(}\PY{n}{image}\PY{o}{.}\PY{n}{resize}\PY{p}{(}\PY{p}{(}\PY{n+nb}{int}\PY{p}{(}\PY{n}{image}\PY{o}{.}\PY{n}{width} \PY{o}{*} \PY{n}{scale}\PY{p}{)}\PY{p}{,} \PY{n+nb}{int}\PY{p}{(}\PY{n}{image}\PY{o}{.}\PY{n}{height} \PY{o}{*} \PY{n}{scale}\PY{p}{)}\PY{p}{)}\PY{p}{)}\PY{p}{)}

\PY{k}{def} \PY{n+nf}{display\PYZus{}gif\PYZus{}in\PYZus{}notebook}\PY{p}{(}\PY{n}{file\PYZus{}git}\PY{p}{,} \PY{n}{use\PYZus{}default\PYZus{}path}\PY{o}{=}\PY{k+kc}{True}\PY{p}{)}\PY{p}{:}
    \PY{k}{if} \PY{n}{use\PYZus{}default\PYZus{}path}\PY{p}{:}
        \PY{n}{file\PYZus{}git} \PY{o}{=} \PY{n}{join}\PY{p}{(}\PY{n}{PATH\PYZus{}IMAGES}\PY{p}{,} \PY{n}{file\PYZus{}git}\PY{p}{)}
    \PY{k}{return} \PY{n}{Image}\PY{p}{(}\PY{n}{file\PYZus{}git}\PY{p}{)}
\end{Verbatim}
\end{tcolorbox}

    \begin{center}\rule{0.5\linewidth}{0.5pt}\end{center}

    \hypertarget{two-stance-detection-paradigms}{%
\subsection{Two Stance Detection
Paradigms}\label{two-stance-detection-paradigms}}

    \begin{tcolorbox}[breakable, size=fbox, boxrule=1pt, pad at break*=1mm,colback=cellbackground, colframe=cellborder]
\prompt{In}{incolor}{3}{\boxspacing}
\begin{Verbatim}[commandchars=\\\{\}]
\PY{n}{display\PYZus{}resized\PYZus{}image\PYZus{}in\PYZus{}notebook}\PY{p}{(}\PY{l+s+s2}{\PYZdq{}}\PY{l+s+s2}{stance\PYZus{}detection\PYZus{}two\PYZus{}paradigm.png}\PY{l+s+s2}{\PYZdq{}}\PY{p}{,} \PY{l+m+mf}{0.7}\PY{p}{)}
\end{Verbatim}
\end{tcolorbox}

    \begin{center}
    \adjustimage{max size={0.9\linewidth}{0.9\paperheight}}{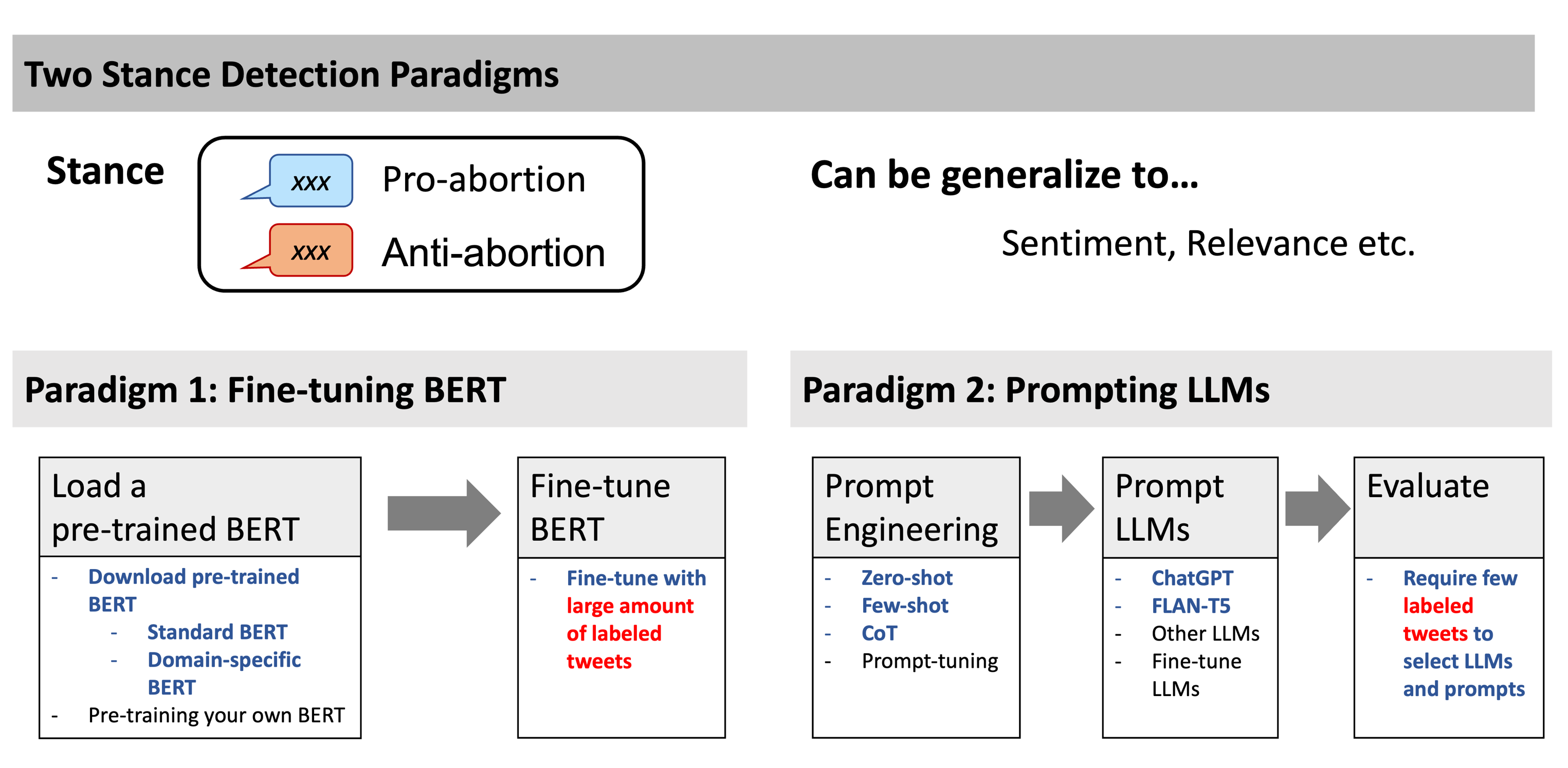}
    \end{center}
    { \hspace*{\fill} \\}
    
    \begin{quote}
The diagram above illustrates the two paradigms for stance detection:
(1) Fine-tuning a BERT model and (2) Prompting Large Language Models
(LLMs). The red text highlights the key practical difference between the
two approaches, which is the need for large labeled data when
fine-tuning a BERT model. The blue texts indicates the parts covered in
these two tutorials. While the black parts are not covered in these
tutorials, they are important to consider when applying these two
paradigms in practice.
\end{quote}

    In this tutorial and the previous tutorial, we are exploring two
different paradigms for stance detection: 1) fine-tuning a BERT model,
and 2) prompting large language models (LLMs) like ChatGPT.

Fine-tuning a BERT model involves training the model on a specific task
using a labeled dataset, which adapts the model's pre-existing knowledge
to the nuances of the task. This approach can yield strong performance
but typically requires a substantial amount of labeled data for the
target task.

On the other hand, prompting LLMs involves crafting carefully designed
input prompts that guide the model to generate desired outputs based on
its pre-trained knowledge. This method does not require additional
training, thus significantly reducing the amount of labeled data needed.
Note that some labeled data is still required to evaluate the
performance.

In this second tutorial, we will focus on the second paradigm: prompting
LLMs. We will explore two classes of LLMs: ChatGPT and FLAN-T5. We will
also explore different prompt types: zero-shot, few-shot, and
chain-of-thought.

    \begin{quote}
For an in-depth exploration of the first paradigm, I invite you to refer
to my previous tutorial, which can be found here:
\href{https://colab.research.google.com/drive/1nxziaKStwRnSyOLI6pLNBaAnB_aB6IsE?usp=sharing}{Fine-tuning
BERT for Stance Detection}.
\end{quote}

    \begin{center}\rule{0.5\linewidth}{0.5pt}\end{center}

    \hypertarget{paradigm-2-using-large-language-models-llms-for-stance-detection}{%
\section{Paradigm 2: Using Large Language Models (LLMs) for Stance
Detection}\label{paradigm-2-using-large-language-models-llms-for-stance-detection}}

Large Language Models (LLMs) like
\href{https://arxiv.org/abs/2005.14165}{GPT-3} are gaining significant
attention in recent years. These models are designed to understand and
generate human-like text by learning from vast amounts of data. In the
context of stance detection, LLMs can be used to classify text based on
the stance towards a particular topic.

    \begin{quote}
Strictly speaking, BERT is also a type of LLMs. The ``LLMs'' covered in
this notebook actually refer to a special type of LLMs, the ``generative
models'', which means they can be used to generate text. In contrast,
BERT is a discriminative model, which means it can only be used to
classify text. I am using the term ``LLMs'' to refer to these generative
models for the sake of simplicity.
\end{quote}

    Although there are many variants of LLMs, I will explain LLMs with GPT-3
as an example. Later in this tutorial, I will then point out the
differences between GPT-3 and other LLMs, such as ChatGPT, FLAN-T5, etc.

    Because GPT-3 and BERT are both based on transformers, they share many
similarities. I will explain GPT-3 by contrasting it with BERT, assuming
that you alreadt know how BERT works, which is covered in the first
notebook.

    \hypertarget{contrast-gpt-3-with-bert}{%
\subsection{Contrast GPT-3 with BERT}\label{contrast-gpt-3-with-bert}}

    \hypertarget{model-architecture-encoder-vs.-decoder}{%
\subsubsection{Model Architecture: Encoder
vs.~Decoder}\label{model-architecture-encoder-vs.-decoder}}

    Both BERT and GPT-3 are transformer-based models, which means they both
employ self-attention layers to learn the relationships between words in
a text. However, they utilize self-attention layers differently.

Notably, in their model architecture, BERT uses ``encoder blocks,''
while GPT-3 employs ``decoder blocks.'' Due to this distinction, BERT is
often referred to as an ``encoder'' model, while GPT-3 is commonly known
as a ``decoder'' model.

    \textbf{Encoder (BERT)}

In simple terms, an encoder model like BERT encodes an input sequence
into a fixed-length vector (after 12 self-attention layers for BERT).
This vector, or, representation, is then used to classify the input
sequence.

    Let's look at a concrete example from the Abortion dataset.

\begin{quote}
``\emph{It's so brilliant that \#lovewins - now extend the equality to
women's rights \#abortionrights}''
\end{quote}

    Recall from the previous tutorial that when an encoder model like BERT
processes a sentence, it utilizes bidirectional context to accurately
capture the meaning of the sentence. Afterward, we can fine-tune the
model using a labeled dataset to adapt it to our specific task.

    \textbf{Decoder (GPT-3)}

On the other hand, a decoder model like GPT-3 are designed to
\textbf{generate} (rather than encode) a sequence from left to right,
one token at a time. If we provide a partially complete sequence of
words (also known as ``prompt'') to GPT-3, it will help complete the
sequence (also known as ``\textbf{conditional text generation}'').

So, if we rephrase the same sentence into the following format, and
provide GPT-3 with this partially complete sequece (also known is
``\textbf{prompt}''), we can use GPT-3 to generate its prediction of the
stance based on the prompt.

    Let's rephrase the sentence into the following ``\textbf{prompt}'':

\begin{quote}
``What is the stance of the tweet below with respect to `Legalization of
Abortion'? Please use exactly one word from the following 3 categories
to label it: `in-favor', `against', `neutral-or-unclear'. Here is the
tweet: `\emph{It's so brilliant that \#lovewins - now extend the
equality to women's rights \#abortionrights}.' The stance of the tweet
is:''
\end{quote}

    With this rephrased sequence, we convert the stance detection task - a
classification task, into a text generation task. We can then use GPT-3
to generate the stance of the tweet.

    Note that GPT-3, like BERT, still uses self-attention layers to learn
the relationships between words. The critical distinction is that when
GPT-3 generates a sequence, it can only look at the words before the
word to be generated (in this example, the prompt), rather than the
bidirectional context like BERT.

The primary difference in the usege of self-attention mechanism is shown
in the following figure. On the left, we have the encoder model (BERT),
where the self-attention layers are bidirectional, i.e., covering both
the left and right context.

On the right, we have the decoder model (GPT-3), where the
self-attention layers are unidirectional, i.e., only covering the left
context when evaluating the word to predict.

    \begin{tcolorbox}[breakable, size=fbox, boxrule=1pt, pad at break*=1mm,colback=cellbackground, colframe=cellborder]
\prompt{In}{incolor}{4}{\boxspacing}
\begin{Verbatim}[commandchars=\\\{\}]
\PY{n}{display\PYZus{}resized\PYZus{}image\PYZus{}in\PYZus{}notebook}\PY{p}{(}\PY{l+s+s2}{\PYZdq{}}\PY{l+s+s2}{encoder\PYZus{}vs\PYZus{}decoder.png}\PY{l+s+s2}{\PYZdq{}}\PY{p}{,}\PY{l+m+mf}{0.3}\PY{p}{)}
\end{Verbatim}
\end{tcolorbox}

    \begin{center}
    \adjustimage{max size={0.9\linewidth}{0.9\paperheight}}{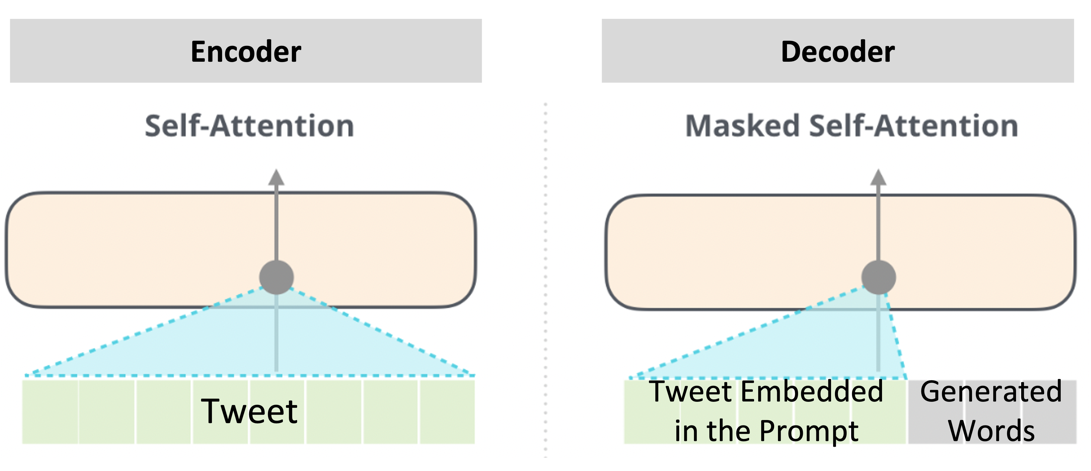}
    \end{center}
    { \hspace*{\fill} \\}
    
    \begin{quote}
Image modified from:
https://jalammar.github.io/illustrated-gpt2/\#part-1-got-and-language-modeling
\end{quote}

    \begin{quote}
Note: Another important difference between BERT and GPT-3 is the size of
the context they can handle. BERT uses a context of 512 tokens, while
GPT-3 can handle a much larger context depending on the specific model
variant, with the largest version handling up to 4097 tokens. This means
that GPT-3 can take into account a larger number of words when capturing
the meaning of a word, which allows GPT-3 to consider more context
surrounding a word and potentially leads to better understanding and
generation capabilitie
\end{quote}

    \hypertarget{different-pre-training-methods}{%
\subsubsection{Different Pre-training
Methods}\label{different-pre-training-methods}}

Because BERT and GPT-3 are different types of transformer models
(encoder vs.~decoder), they use different pre-training methods.

\textbf{BERT}

As mentioned in the
\href{https://colab.research.google.com/drive/1nxziaKStwRnSyOLI6pLNBaAnB_aB6IsE?usp=sharing}{previous
tutorial}, BERT is pre-trained using the 1) masked language modeling
task and the 2) next sentence prediction task. These tasks enable the
BERT model to learn the relationships between words. However, these
tasks alone is not aligned with a classification task like stance
detection, which is why we need to fine-tune BERT using a labeled
dataset to adapt it to our specific task.

\textbf{GPT-3}

In contrast, GPT-3 is pre-trained using the \textbf{unidirectional
language modeling task}. In this task, the model is tasked with
predicting the next word in a sequence, given the previous words in the
sequence.

The animation below shows how GPT-3 is pre-trained using this task. In
the example, the partially complete sequence is ``a robot must'', and
the model is tasked with predicting the next word ``obey'' based on the
previous words. This pretraining task enables the GPT-3 model to predict
the stance of a given prompt with the need of fine-tuning (more on this
later).

    \begin{tcolorbox}[breakable, size=fbox, boxrule=1pt, pad at break*=1mm,colback=cellbackground, colframe=cellborder]
\prompt{In}{incolor}{5}{\boxspacing}
\begin{Verbatim}[commandchars=\\\{\}]
\PY{n}{display\PYZus{}gif\PYZus{}in\PYZus{}notebook}\PY{p}{(}\PY{l+s+s2}{\PYZdq{}}\PY{l+s+s2}{gpt3\PYZus{}pretraining.gif}\PY{l+s+s2}{\PYZdq{}}\PY{p}{)}
\end{Verbatim}
\end{tcolorbox}

            \begin{tcolorbox}[breakable, size=fbox, boxrule=.5pt, pad at break*=1mm, opacityfill=0]
\prompt{Out}{outcolor}{5}{\boxspacing}
\begin{Verbatim}[commandchars=\\\{\}]
<IPython.core.display.Image object>
\end{Verbatim}
\end{tcolorbox}
        
    \begin{quote}
Image modified from:
https://jalammar.github.io/how-gpt3-works-visualizations-animations/
\end{quote}

    \hypertarget{model-size}{%
\subsubsection{Model Size}\label{model-size}}

In addition to the differences in their model architecture, a crucial
distinction between BERT and GPT-3 lies in their scale. GPT-3 primarily
derives its power from its immense model size and the substantial amount
of data utilized for pre-training.

While BERT indeed boasts a massive number of model parameters (hundreds
of millions), GPT-3 surpasses it with an even more colossal number of
parameters, reaching hundreds of billions!

Below is a table showing the number of parameters for different LLMs. As
you can see, GPT-3 has about 500 times more parameters than the largest
BERT model.

Here is the reordered table as requested:

\begin{longtable}[]{@{}
  >{\raggedright\arraybackslash}p{(\columnwidth - 8\tabcolsep) * \real{0.17}}
  >{\raggedright\arraybackslash}p{(\columnwidth - 8\tabcolsep) * \real{0.25}}
  >{\raggedright\arraybackslash}p{(\columnwidth - 8\tabcolsep) * \real{0.23}}
  >{\raggedright\arraybackslash}p{(\columnwidth - 8\tabcolsep) * \real{0.20}}
  >{\raggedright\arraybackslash}p{(\columnwidth - 8\tabcolsep) * \real{0.15}}@{}}
\toprule
Model Variant & Number of Parameters & Transformer Layers & Attention
Heads & Hidden Size \\
\midrule
\endhead
BERT-Large & 340 million & 24 & 16 & 1024 \\
GPT-3 & 175 billion & 96 & 96 & 12288 \\
\bottomrule
\end{longtable}

    \begin{quote}
There are different variants of GPT-3 and different variants of BERT,
with different number of layers, hidden size, and attention heads etc.
Here, I am using the largest variant of both models for comparison.
\end{quote}

    \hypertarget{pre-traing-data-size}{%
\subsubsection{Pre-Traing Data Size}\label{pre-traing-data-size}}

Because GPT-3 has a much larger number of parameters than BERT, it also
requires a much larger corpus of text to pre-train.

Specifically, GPT-3 is trained on about \textbf{400 billion tokens},
while BERT is trained on about \textbf{3.3 billion tokens}. This means
that GPT-3 has about access to about 100 times more information than
BERT, which can lead to better performance.

\begin{itemize}
\tightlist
\item
  BERT's training data: English Wikipedia + about 11k books = about 2.5
  billion English words = 3.3 billion tokens
\item
  GPT-3's training data: English Wikipedia + books + webpages crawled
  from the internet since 2008 with about 1 trillion words = about 400
  billion tokens
\end{itemize}

    \begin{quote}
Note: Because BERT is trained on English corpus, and GPT-3 is trained on
multiple languages, we can't directly compare the number of the words.
Instead, we can compare the number of tokens after tokenization. Here,
the tokenzier is the Byte Pair Encoding (BPE) method, which is the
tokenizer used by GPT-3. If you want to know more about this tokenizer,
I recommend this tutorial.:
https://huggingface.co/learn/nlp-course/chapter6/5?fw=pt
\end{quote}

    \hypertarget{the-necessity-of-fine-tuning-and-labeled-data}{%
\subsubsection{The necessity of fine-tuning and labeled
data}\label{the-necessity-of-fine-tuning-and-labeled-data}}

The role of fine-tuning in BERT and GPT-3 differs significantly.

As mentioned earlier, after the pre-training phase, BERT requires
fine-tuning on specific tasks of interest (such as stance detection)
using labeled data. The reason for this is that BERT is pre-trained on a
masked language modeling task and a next sentence prediction task, which
are not directly related to stance detection.

In contrast, GPT-3 is pre-trained on the unidirectional language
modeling task, which enables it to directly generate a prediction for
any given task of interest. In our use case, this means that using ahe
GPT-3 model for stance detection task does not necessarily require
fine-tuning with labeled data. This implication is huge as collecting
labeled data is often a time-consuming and expensive process.

    \begin{quote}
Note: While it is not necessary for the model to perform well on a given
task without fine-tuning on labeled data (see the prompting section
below), OpenAI does offer
\href{https://platform.openai.com/docs/guides/fine-tuning}{the option to
fine-tune} the GPT-3 model using labeled data, Note that, however,
fine-tining the GPT-3 model and using a fine-tuned model can be
expensive. Please see the \href{https://openai.com/pricing}{OpenAI's
pricing page} for more details.
\end{quote}

    \hypertarget{the-potential-of-prompting}{%
\paragraph{The Potential of
Prompting}\label{the-potential-of-prompting}}

Because GPT-3 model is pre-trained in a way to predict the next word
given a sequence (the ``prompt''), it matter critically what the actual
prompt is.

For instance, the example earlier demonstrates a ``zero-shot prompt'',
which asks GPT-3 for stance prediction without providing any example
tweets.

\begin{quote}
``What is the stance of the tweet below with respect to `Legalization of
Abortion'? Please use exactly one word from the following 3 categories
to label it: `in-favor', `against', `neutral-or-unclear'. Here is the
tweet: `\emph{It's so brilliant that \#lovewins - now extend the
equality to women's rights \#abortionrights}.' The stance of the tweet
is:''
\end{quote}

The name zero-shot prompt comes from the fact that no example tweets are
given, and GPT-3 is expected to generate the stance based on the prompt
alone.

Prompt choice significantly impacts the model's performance, an I will
dedicate a separate section below to discuss different types of prompts
and their implications.

    \hypertarget{conclusion-gpt-3-vs.-bert-on-stance-detection}{%
\subsubsection{Conclusion: GPT-3 vs.~BERT on stance
detection}\label{conclusion-gpt-3-vs.-bert-on-stance-detection}}

In summary, BERT and GPT-3 are both powerful transformer-based models,
but with significant differences in their architecture, pre-training
methods, model size, and use of fine-tuning.

BERT, an encoder model, relies on bidirectional context and fine-tuning
to perform well on specific tasks, whereas GPT-3, a decoder model,
leverages unidirectional context, massive model size, and the art of
prompting to perform a wide range of tasks without the need for
fine-tuning.

The choice between BERT and GPT-3 for a specific task depends on several
factors, such as the availability of labeled data, computational
resources, and the desired level of customization.

\textbf{Computational Resource}

Training and prompting a large model like GPT-3 requires powerful GPUs
and a significant amount of memory. However, you don't need a powerful
computer to use GPT-3. GPT-3 can be used directly through OpenAI's API,
making it more accessible to users without high-performance hardware
through
\href{https://platform.openai.com/docs/api-reference/completions/create}{OpenAI's
API}, which handles the computational load for you. In contrast, BERT is
much smaller, and can be trained, fine-tuned, and loaded on modern
cunsumer-level hardware.

\textbf{Monetary Cost}

Using the OpenAI's API comes with its own monetary cost, as the usage of
the API is billed based on the number of tokens processed, and GPT-3's
large model size can result in higher costs for extensive usage. That
said, the cost is considered low. For the 933 tweets considered in this
tutorial, the cost for using ChatGPT (an extension of the GPT-3 model)
with zero-shot prompting is about \$0.39 and \$0.65 with few-shot
prompting (more below). On the other hand, fine-tuning BERT may require
more time and effort to train but can run with low monetary cost if you
have the necessary hardware.

\textbf{Desired Level of Customization and the Need of Labeled Data}

BERT is often more suitable for tasks that require a deeper
understanding of context and benefit from fine-tuning on domain-specific
labeled data.

On the other hand, GPT-3 is a powerful option for tasks that can
leverage its vast knowledge and benefit from zero-shot or few-shot
learning approaches without large amounts of labeled data. For example,
when labeled data is scarce or the task is more general in nature. GPT-3
may perform well across various tasks without fine-tuning can be
advantageous.

\textbf{Open-source vs Closed-source}

While BERT is an open-source model, GPT-3 is a closed-source model. This
means that you can't access the source code of GPT-3, and you can't
train your own GPT-3 model or modify the mode (and you also have to pay
for the usage). However, not all LLMs are closed-source. For example,
you can train your own GPT-2 model using the
\href{https://huggingface.co/transformers/model_doc/gpt2.html}{Hugging
Face's implementation}. In this tutorial, I will also show you another
state-of-the-art open-source LLM, the
\href{https://huggingface.co/docs/transformers/model_doc/FLAN-T5}{FLAN-T5
model}, which is suitable for stance detection tasks.

Now that we have a better understanding of the differences between BERT
and GPT-3, let's dive deeper into prompting techniques for large
language models in the next section.

    \begin{center}\rule{0.5\linewidth}{0.5pt}\end{center}

    \hypertarget{prompting-techniques-for-llms}{%
\subsection{Prompting Techniques for
LLMs}\label{prompting-techniques-for-llms}}

To effectively utilize LLMs like ChatGPT-3 for stance detection, we can
employ various prompting techniques that guide the model's response
without the need for fine-tuning.

    \begin{quote}
If you want to learn more about prompting techniques, I recommend this
online tutorial: https://learnprompting.org/docs/intro
\end{quote}

    \hypertarget{zero-shot-prompting}{%
\subsubsection{Zero-shot Prompting}\label{zero-shot-prompting}}

In the zero-shot prompting technique, for stance detection, GPT-3 is
provided with a task description and with the tweet of interest, without
any specific examples.

    \begin{tcolorbox}[breakable, size=fbox, boxrule=1pt, pad at break*=1mm,colback=cellbackground, colframe=cellborder]
\prompt{In}{incolor}{6}{\boxspacing}
\begin{Verbatim}[commandchars=\\\{\}]
\PY{n}{display\PYZus{}resized\PYZus{}image\PYZus{}in\PYZus{}notebook}\PY{p}{(}\PY{l+s+s2}{\PYZdq{}}\PY{l+s+s2}{prompt\PYZus{}zero\PYZus{}shot\PYZus{}v1.png}\PY{l+s+s2}{\PYZdq{}}\PY{p}{,}\PY{l+m+mf}{0.3}\PY{p}{)}
\end{Verbatim}
\end{tcolorbox}

    \begin{center}
    \adjustimage{max size={0.9\linewidth}{0.9\paperheight}}{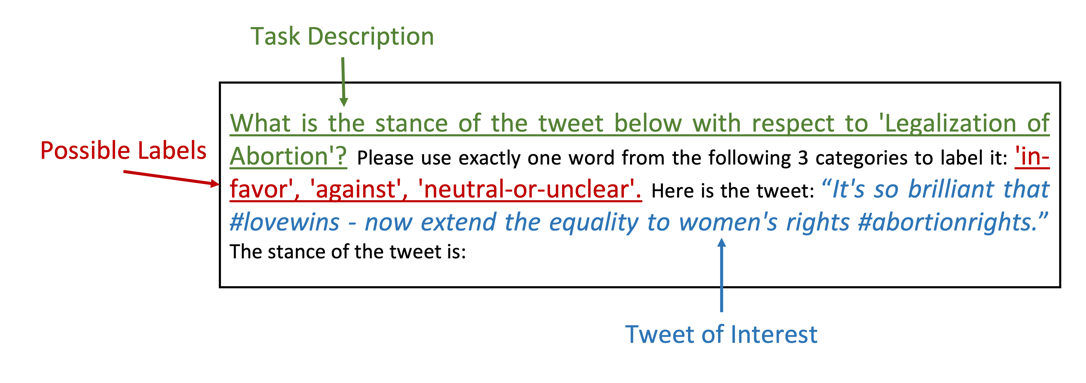}
    \end{center}
    { \hspace*{\fill} \\}
    
    We can further improve this prompt with the definition of each stance
category. Note that when human annotators are asked to label a tweet,
they are usually provided with the definition of each stance category.
To this end, I extract the definitions of each stance category from
``codebook'' of the
\href{https://alt.qcri.org/semeval2016/task6/data/uploads/stance-question.pdf}{SemEval
2016 Task 6 dataset} and include them in the prompt.

    \begin{tcolorbox}[breakable, size=fbox, boxrule=1pt, pad at break*=1mm,colback=cellbackground, colframe=cellborder]
\prompt{In}{incolor}{7}{\boxspacing}
\begin{Verbatim}[commandchars=\\\{\}]
\PY{n}{display\PYZus{}resized\PYZus{}image\PYZus{}in\PYZus{}notebook}\PY{p}{(}\PY{l+s+s2}{\PYZdq{}}\PY{l+s+s2}{prompt\PYZus{}zero\PYZus{}shot\PYZus{}v2.png}\PY{l+s+s2}{\PYZdq{}}\PY{p}{,}\PY{l+m+mf}{0.3}\PY{p}{)}
\end{Verbatim}
\end{tcolorbox}

    \begin{center}
    \adjustimage{max size={0.9\linewidth}{0.9\paperheight}}{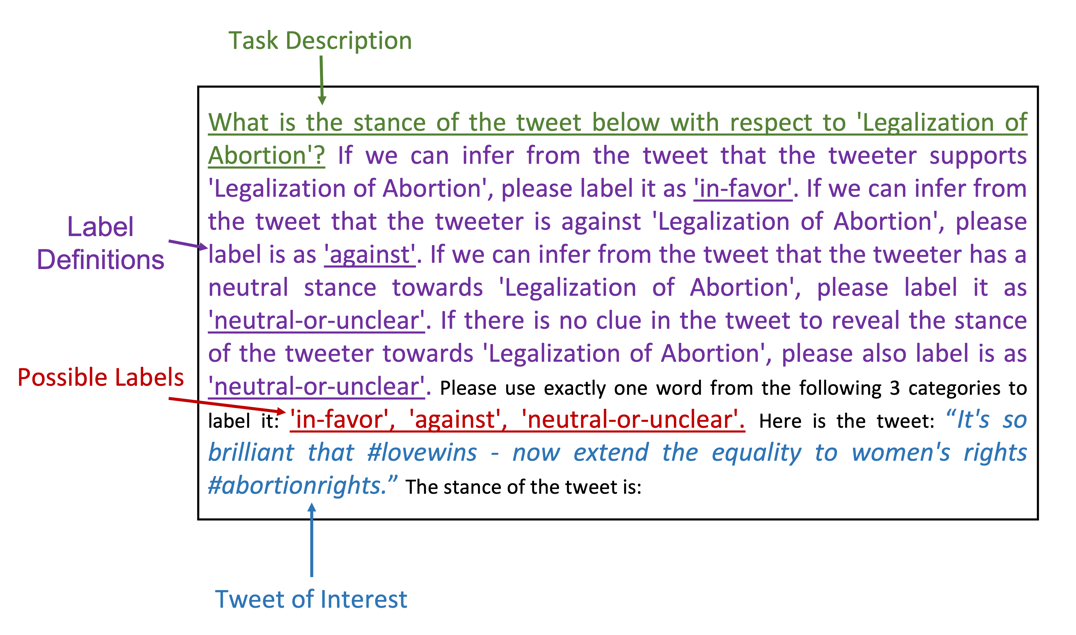}
    \end{center}
    { \hspace*{\fill} \\}
    
    \hypertarget{few-shot-prompting}{%
\subsubsection{Few-shot Prompting}\label{few-shot-prompting}}

The few-shot prompting technique involves providing the LLM with a small
number of examples to guide its response. This allows the model to learn
from the provided examples and adapt its output accordingly. Below is an
example of a few-shot prompt for stance detection, with 1 example tweet
for each stance category.

Note that I append the example tweets to the zero-shot prompt I made
above.

    \begin{tcolorbox}[breakable, size=fbox, boxrule=1pt, pad at break*=1mm,colback=cellbackground, colframe=cellborder]
\prompt{In}{incolor}{8}{\boxspacing}
\begin{Verbatim}[commandchars=\\\{\}]
\PY{n}{display\PYZus{}resized\PYZus{}image\PYZus{}in\PYZus{}notebook}\PY{p}{(}\PY{l+s+s2}{\PYZdq{}}\PY{l+s+s2}{prompt\PYZus{}few\PYZus{}shot.png}\PY{l+s+s2}{\PYZdq{}}\PY{p}{,}\PY{l+m+mf}{0.3}\PY{p}{)}
\end{Verbatim}
\end{tcolorbox}

    \begin{center}
    \adjustimage{max size={0.9\linewidth}{0.9\paperheight}}{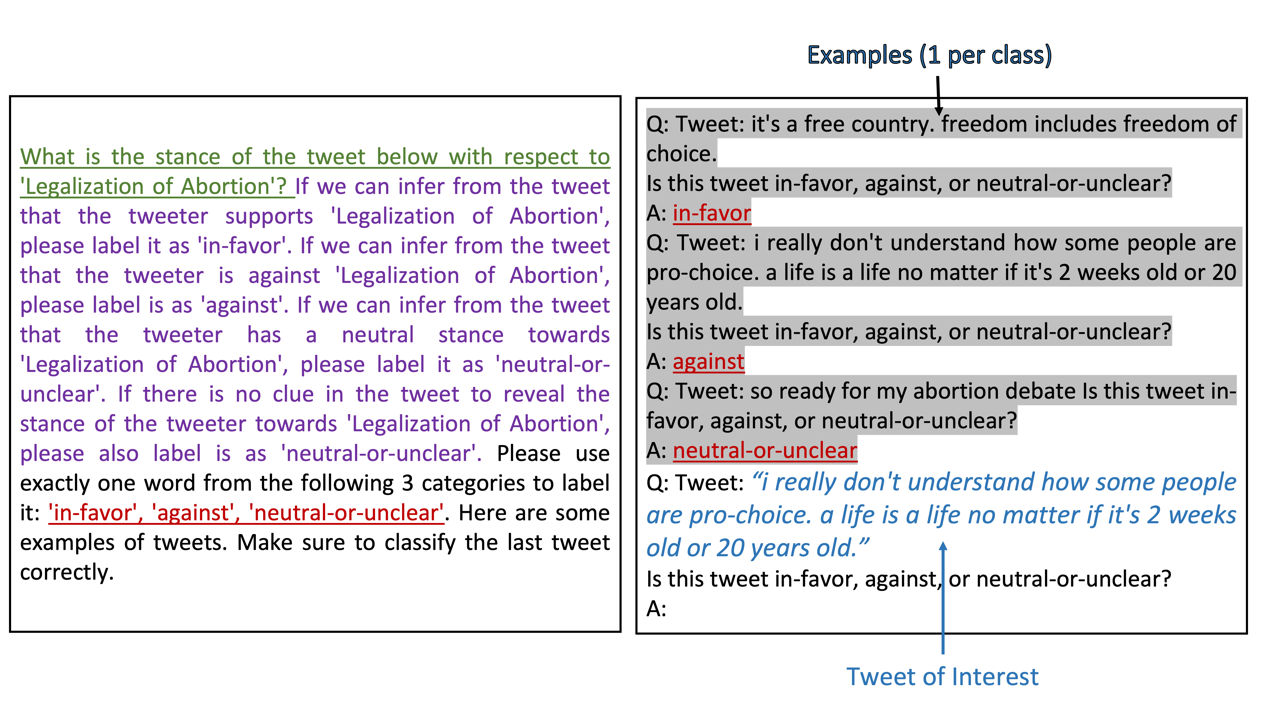}
    \end{center}
    { \hspace*{\fill} \\}
    
    \begin{quote}
Note: It may sound trivial, but when evaluating the performance of the
LLMs with few-shot prompting, it is important to make sure that the
examples provided to the model are not included in the test set to avoid
data leakage.
\end{quote}

    Note: A comprehensive evaluattion of the zero-shot and few-shot
performance of GPT-3 is available in
\href{https://arxiv.org/abs/2005.14165}{Brown, T. et al (2020)}.

    \hypertarget{zero-shot-chain-of-thoughts-cot-prompting}{%
\subsubsection{Zero-shot Chain-of-thoughts (CoT)
Prompting}\label{zero-shot-chain-of-thoughts-cot-prompting}}

By appending the words ``let's think step-by-step'' to the zero-shot
prompt, we can ask GPT-3 to generate a chain of thoughts that lead to
the stance prediction. This has been shown to improve the performance of
LLMs in task that involves reasoning. This particular prompt is called
zero-shot chain-of-thoughts (CoT) prompting.

    One example task where the zero-shot chain-of-thoughts prompting
technique has been shown to be effective is in task involves numerical
reasoning. For instance, in the example below,, GPT-3 is able to
generate a chain of thoughts that lead to the correct answer, as opposd
to the standard zero-shot prompting technique, which fails to generate
the correct answer.

    \begin{tcolorbox}[breakable, size=fbox, boxrule=1pt, pad at break*=1mm,colback=cellbackground, colframe=cellborder]
\prompt{In}{incolor}{9}{\boxspacing}
\begin{Verbatim}[commandchars=\\\{\}]
\PY{n}{display\PYZus{}resized\PYZus{}image\PYZus{}in\PYZus{}notebook}\PY{p}{(}\PY{l+s+s2}{\PYZdq{}}\PY{l+s+s2}{prompt\PYZus{}cot\PYZus{}math.png}\PY{l+s+s2}{\PYZdq{}}\PY{p}{,}\PY{l+m+mf}{0.3}\PY{p}{)}
\end{Verbatim}
\end{tcolorbox}

    \begin{center}
    \adjustimage{max size={0.9\linewidth}{0.9\paperheight}}{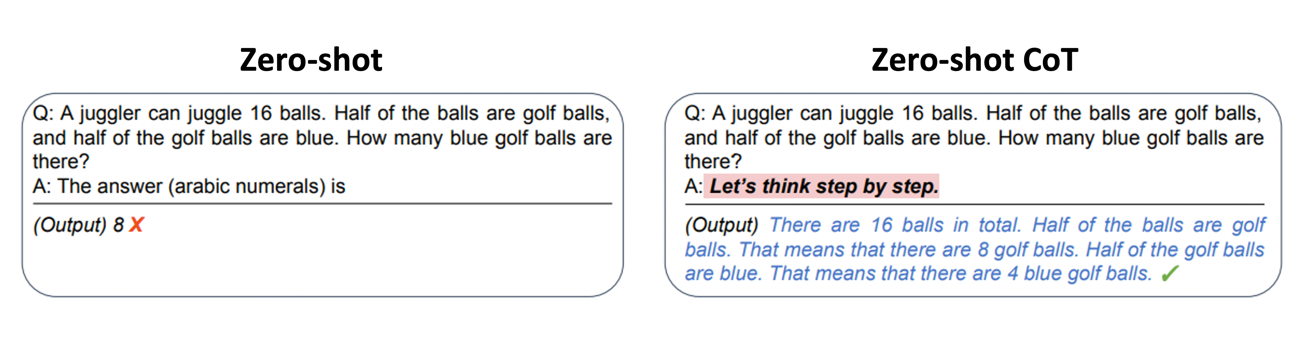}
    \end{center}
    { \hspace*{\fill} \\}
    
    \begin{quote}
The zero-shot CoT is proposed by
\href{https://arxiv.org/pdf/2205.11916.pdf}{Kojima, T. et al (2022)},
and the figure is also modified from their paper.
\end{quote}

    Since stance detection often requires reasoning, it makes sense to
consider zero-shot CoT prompting technique for our task.

Below is an example of a zero-shot chain-of-thoughts prompt for stance
detection. Notice that I append the words ``let's think step-by-step''
to the zero-shot prompt I made above with minor modifications.

    \begin{tcolorbox}[breakable, size=fbox, boxrule=1pt, pad at break*=1mm,colback=cellbackground, colframe=cellborder]
\prompt{In}{incolor}{10}{\boxspacing}
\begin{Verbatim}[commandchars=\\\{\}]
\PY{n}{display\PYZus{}resized\PYZus{}image\PYZus{}in\PYZus{}notebook}\PY{p}{(}\PY{l+s+s2}{\PYZdq{}}\PY{l+s+s2}{prompt\PYZus{}cot.png}\PY{l+s+s2}{\PYZdq{}}\PY{p}{,}\PY{l+m+mf}{0.3}\PY{p}{)}
\end{Verbatim}
\end{tcolorbox}

    \begin{center}
    \adjustimage{max size={0.9\linewidth}{0.9\paperheight}}{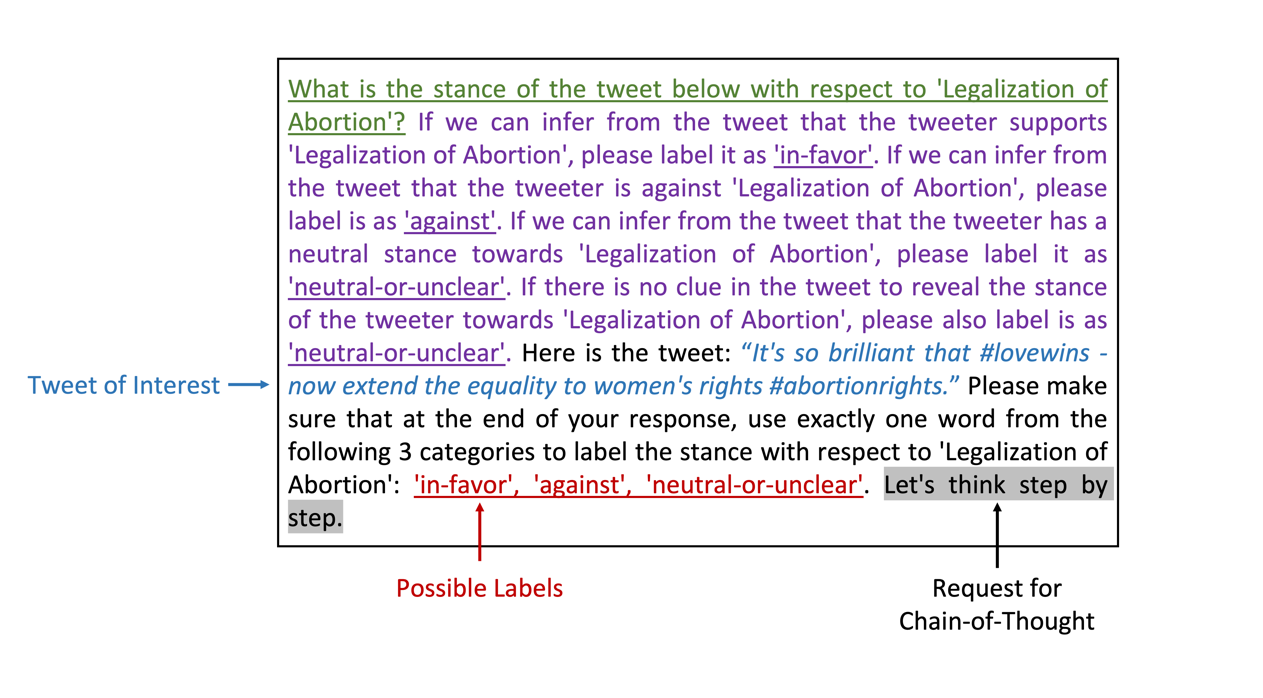}
    \end{center}
    { \hspace*{\fill} \\}
    
    One important caveat about the chain-of-thought (CoT) prompt is that,
according to
\href{https://arxiv.org/pdf/2201.11903.pdf?trk=public_post_comment-text}{Wei
et al.~(2022)}, CoT tends to yield performance gains only when used with
larger models of around 100 billion parameters or more. For smaller
language models, the CoT prompt might actually harm the performance, as
the model may generate incorrect chains of thoughts, leading to an
inaccurate prediction.

    \begin{quote}
Note: The original chain-of-thoughts (CoT) prompt includes some examples
of reasoning in the prompt, like the few-shot prompt. However, in this
tutorial, for the sake of simplicity, I will only use the zero-shot CoT
prompt. I will also use the term ``CoT'' to refer to the zero-shot CoT
prompt.
\end{quote}

    \begin{center}\rule{0.5\linewidth}{0.5pt}\end{center}

    \hypertarget{two-state-of-the-art-llms-chatgpt-and-flan-t5}{%
\subsection{Two state-of-the-art LLMs: ChatGPT and
FLAN-T5}\label{two-state-of-the-art-llms-chatgpt-and-flan-t5}}

Before I move on to the programming part, I want to introduce two
state-of-the-art LLMs that are suitable for stance detection tasks:
ChatGPT and FLAN-T5.

\hypertarget{chatgpt}{%
\subsubsection{ChatGPT}\label{chatgpt}}

\href{https://openai.com/blog/chatgpt}{ChatGPT} (\texttt{gpt-3.5-turbo})
is an extension of the GPT-3 model. One significant advantage of
ChatGPT, compared to GPT-3 (\texttt{gpt-3-davinci}, the most powerful
variant), is its lower cost. According to
\href{https://openai.com/pricing/}{OpenAI's pricing page}, using ChatGPT
costs \$0.002 per 1k tokens (approximately 750 English words, including
both words in the prompt and generated sequence), which is about 10
times cheaper than GPT-3 at \$0.02 per 1k tokens (as of 04/22/2023).

Besides the difference in pricing, the main distinction between ChatGPT
and GPT-3 lies in their training approaches. While both models are
pre-trained using the next word prediction task, ChatGPT is further
trained with human feedback to generate more coherent and contextually
relevant responses in conversational settings. This additional training
is crucial since ChatGPT is designed to function as a chatbot, expected
to provide responses that are coherent and contextually relevant to the
user's input.

Finally, it is worth noting that there is a newer, more powerful variant
of the model called \href{https://openai.com/research/gpt-4}{GPT-4}.
This advanced version has demonstrated remarkable performance on various
sophisticated exams, including AP exams, GRE tests, and the Law Bar
Exam, among others. However, due to the higher cost and accessibility
(as there is currently a waitlist to use it), this tutorial will focus
on using \texttt{gpt-3.5-turbo}.

    \hypertarget{fine-tuning-with-human-feedback-in-chatgpt}{%
\paragraph{Fine-tuning with human feedback in
ChatGPT}\label{fine-tuning-with-human-feedback-in-chatgpt}}

During the fine-tuning of ChatGPT, a technique called
\textbf{reinforcement learning from human feedback (RLHF)} is critical.
The goal is to have ChatGPT generate texts that sound ``human-like'' in
a conversation. This method includes making an initial dataset with the
help of human AI trainers (who give conversations or answers to
different prompts). Then, these human trainers compare and rank several
responses created by Chat-GPT-3.5. Using these rankings, a reward model
is trained to predict human's ranking. The ChatGPT model is then
optimized to maximize the reward evluated by the reward model. By
repeatedly using human feedback, the model gets better and its answers
become more human-like.

The diagram below (copied from
\href{https://openai.com/blog/chatgpt}{OpenAI's post on ChatGPT}) shows
the process of fine-tuning ChatGPT with RLHF.

    \begin{quote}
The details of the RLHF method is beyond the scope of this tutorial. If
you are interested in learning more about RLHF, I recommend reading
\href{https://openai.com/blog/chatgpt}{OpenAI's post on ChatGPT}.
\end{quote}

    \begin{tcolorbox}[breakable, size=fbox, boxrule=1pt, pad at break*=1mm,colback=cellbackground, colframe=cellborder]
\prompt{In}{incolor}{11}{\boxspacing}
\begin{Verbatim}[commandchars=\\\{\}]
\PY{n}{display\PYZus{}resized\PYZus{}image\PYZus{}in\PYZus{}notebook}\PY{p}{(}\PY{l+s+s2}{\PYZdq{}}\PY{l+s+s2}{chatgpt\PYZus{}finetuning.png}\PY{l+s+s2}{\PYZdq{}}\PY{p}{,}\PY{n}{scale} \PY{o}{=} \PY{l+m+mi}{1}\PY{p}{)}
\end{Verbatim}
\end{tcolorbox}

    \begin{center}
    \adjustimage{max size={0.9\linewidth}{0.9\paperheight}}{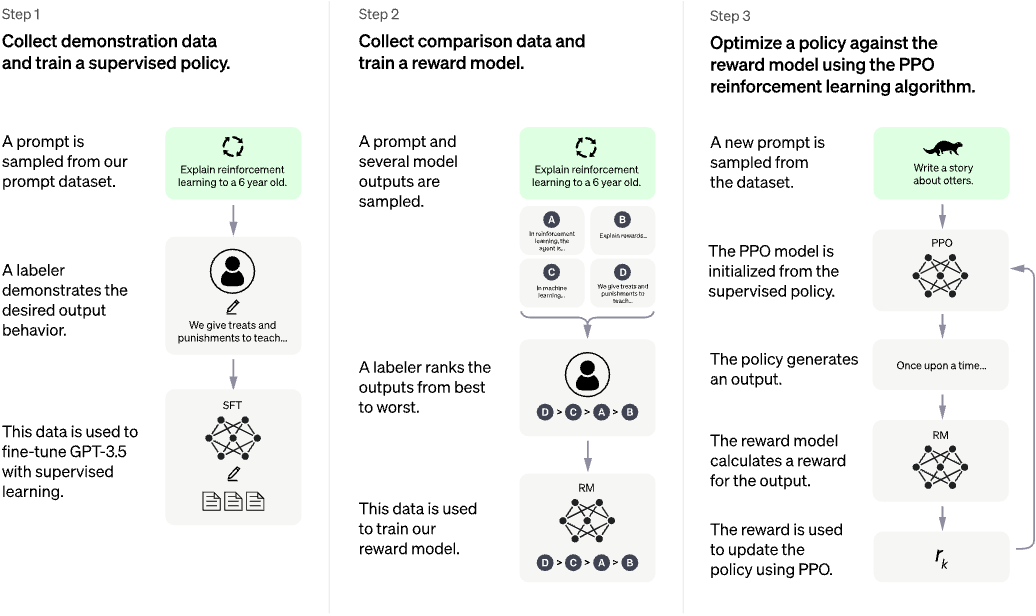}
    \end{center}
    { \hspace*{\fill} \\}
    
    \hypertarget{open-source-model-flan-t5}{%
\subsubsection{Open-source model
FLAN-T5}\label{open-source-model-flan-t5}}

The
\href{https://huggingface.co/docs/transformers/model_doc/flan-t5}{FLAN-T5}
model is another LLM. A significant advantage of FLAN-T5 is that it is
open-source, meaning that it is free to use (if you have access to
GPUs)! This is in contrast to ChatGPT, which is a proprietary model.

It is based on the T5 (Text-to-Text Transfer Transformer) architecture
developed by Google Research. Like GPT-3, FLAN-T5 also has the decoder
component, which enables it to generate sequence given a prompt.

    \hypertarget{instruction-fine-tuning}{%
\paragraph{Instruction Fine-tuning}\label{instruction-fine-tuning}}

One critical difference between FLAN-T5 and GPT-3 is the ``instruction
fine-tuning'' procedure. After pre-training, FLAN-T5 is fine-tuned with
over 1.8k text-to-text tasks, like summarization, translation,
question-answering, among many, where data are fed into the model in
\texttt{(input\_text,\ output\_text)} pairs to predict the output text
given the input text.

This means that FLAN-T5 is more suitable for instruction-based
question-answering tasks, such as stance detection, but is less suitable
for long-form text generation tasks, such as story generation.

Below is a diagram of how FLAN-T5 is fine-tuned.

    \begin{tcolorbox}[breakable, size=fbox, boxrule=1pt, pad at break*=1mm,colback=cellbackground, colframe=cellborder]
\prompt{In}{incolor}{12}{\boxspacing}
\begin{Verbatim}[commandchars=\\\{\}]
\PY{n}{display\PYZus{}resized\PYZus{}image\PYZus{}in\PYZus{}notebook}\PY{p}{(}\PY{l+s+s2}{\PYZdq{}}\PY{l+s+s2}{flan\PYZus{}t5\PYZus{}xxl.png}\PY{l+s+s2}{\PYZdq{}}\PY{p}{,}\PY{l+m+mf}{0.7}\PY{p}{)}
\end{Verbatim}
\end{tcolorbox}

    \begin{center}
    \adjustimage{max size={0.9\linewidth}{0.9\paperheight}}{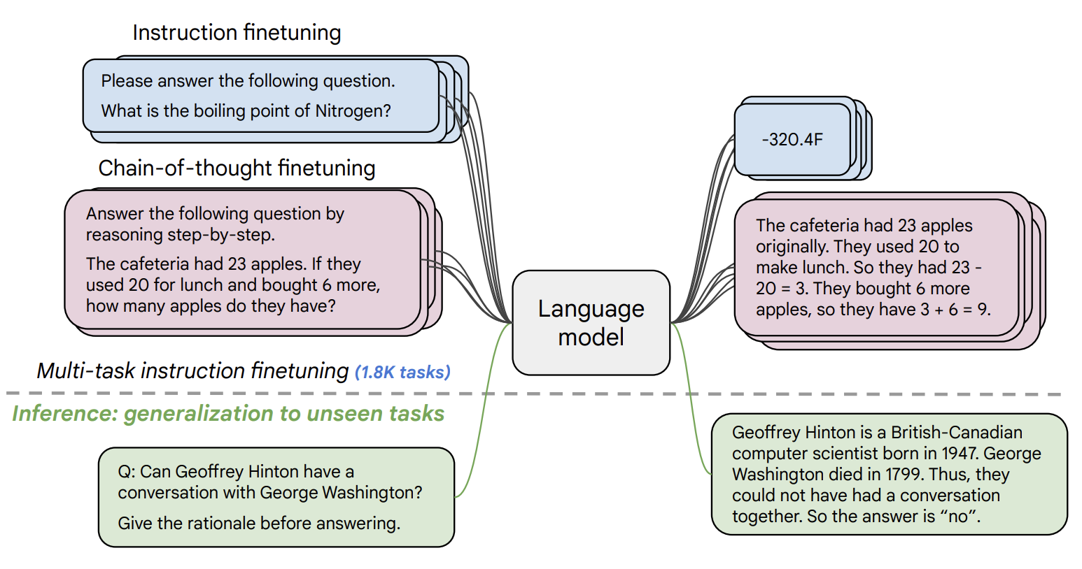}
    \end{center}
    { \hspace*{\fill} \\}
    
    \begin{quote}
Image copied from:
https://arxiv.org/pdf/2210.11416.pdf?trk=public\_post\_comment-text
\end{quote}

    \begin{quote}
Note for advanced readers: The intricate details of the FLAN-T5 model
architecture and training procedure are beyond the scope of this
tutorial. If you are interested in learning more about FLAN-T5, I
recommend reading the
\href{https://towardsdatascience.com/t5-text-to-text-transfer-transformer-643f89e8905e}{blog
post on T5}. Although the blog post is about T5 model and not FLAN-T5
model, T5 model is the foundation of the FLAN-T5 model. Unlike GPT-3, a
pure decoder model, FLAN-T5 is a ``encoder-decoder model'', which means
it has both the encoder and the decoder. Because of the encoder
component, FLAN-T5 is not pre-trained with the next word prediction task
like GPT-3. Instead, FLAN-T5 is pre-trained using the ``denoising
autoencoder framework'', which involves reconstructing the original
input from a corrupted version of it. For curious readers who are eager
to learn more, please check out the
\href{https://arxiv.org/pdf/2210.11416.pdf?trk=public_post_comment-text}{paper}.
\end{quote}

    \hypertarget{critical-distinctions-between-chatgpt-and-flan-t5}{%
\subsubsection{Critical Distinctions between ChatGPT and
FLAN-T5}\label{critical-distinctions-between-chatgpt-and-flan-t5}}

    \hypertarget{closed-source-vs.-open-source}{%
\paragraph{Closed-source
vs.~Open-source}\label{closed-source-vs.-open-source}}

ChatGPT is a closed-source proprietary model developed by OpenAI, and
using it comes with costs with API access. The implementation details
and source code are not openly available, which limits users' ability to
modify or understand the underlying workings of the model. On the other
hand, FLAN-T5 is an open-source model, which means it is free to use,
and the source code is publicly available for anyone to explore, use,
and modify as needed.

\hypertarget{use-case}{%
\paragraph{Use case}\label{use-case}}

ChatGPT focuses on improving general conversational abilities and
controllability and is fine-tuned with human feedback, while FLAN-T5 is
designed to handle a wide range of short question-answering NLP tasks
and is fine-tuned on text-to-text tasks. As a result, FLAN-T5 may be
good at short question-answering task, but may not be as effective as
ChatGPT for tasks that require longer responses, such as writing an
essay.

\hypertarget{model-size}{%
\paragraph{Model size}\label{model-size}}

While OpenAI has not released the exact model size of ChatGPT, the GPT-3
model that is built upon has 175 billion parameters. On the other hand,
the larges variant of FLAN-T5 (\texttt{flan-t5-xxl}) has about 11
billion parameters. The difference in model size entails that ChatGPT
may capture more subtle meanings in language.

\begin{longtable}[]{@{}ll@{}}
\toprule
Model Variant & Number of Parameters \\
\midrule
\endhead
BERT-Large & 340 million \\
FLAN-T5-Large & 780 million \\
FLAN-T5-XXL & 11 billion \\
ChatGPT-3.5 & \textgreater175 billion (approx.) \\
\bottomrule
\end{longtable}

    \begin{quote}
Note: FLAN-T5, like BERT, also comes in different variants, with
different number of parameters. While FLAN-T5-XXL is the most powerful
variant, it can not be run on Google Colab due to GPU memory
limitations. In this tutorial, we will prompt the \texttt{flan-t5-large}
variant. I have prompt the \texttt{flan-t5-xxl} variant elsewhere and
upload the predictions so you can still view and evaluate the
performance.
\end{quote}

    \begin{center}\rule{0.5\linewidth}{0.5pt}\end{center}

    \hypertarget{programming-exercise-implementing-stance-detection-with-flan-t5-and-chatgpt}{%
\subsection{Programming Exercise: Implementing Stance Detection with
FLAN-T5 and
ChatGPT}\label{programming-exercise-implementing-stance-detection-with-flan-t5-and-chatgpt}}

Now that we have a basic understanding of ChatGPT and FLAN-T5, it's time
to explore them through hands-on programming exercises in the context of
stance detection on the Abortion dataset, which we used in the previous
tutorial with BERT. These activities will allow you to gain some
practical experience with both models, helping you to use them
effectively for various NLP tasks while also revealing their differences
in performance and implementation.

Don't worry if you're not an expert yet --- let's just dive in and learn
by doing as we implement stance detection using ChatGPT and FLAN-T5 and.

    \hypertarget{read-and-preprocess-the-raw-data}{%
\subsection{Read and Preprocess the Raw
Data}\label{read-and-preprocess-the-raw-data}}

Note that I am using the same preprocessed procedure as I did when I
fine-tuned BERT in my previous tutorial, including removing retweet tags
etc.

If you are interested in learning more about the preprocessing
procedure, please refer to the previous tutorial.

    \begin{tcolorbox}[breakable, size=fbox, boxrule=1pt, pad at break*=1mm,colback=cellbackground, colframe=cellborder]
\prompt{In}{incolor}{13}{\boxspacing}
\begin{Verbatim}[commandchars=\\\{\}]
\PY{o}{\PYZpc{}}\PY{k}{load\PYZus{}ext} autoreload
\PY{o}{\PYZpc{}}\PY{k}{autoreload} 2
\PY{k+kn}{import} \PY{n+nn}{pandas} \PY{k}{as} \PY{n+nn}{pd}
\PY{k+kn}{from} \PY{n+nn}{os}\PY{n+nn}{.}\PY{n+nn}{path} \PY{k+kn}{import} \PY{n}{join}

\PY{k+kn}{from} \PY{n+nn}{data\PYZus{}processor} \PY{k+kn}{import} \PY{n}{SemEvalDataProcessor}
\PY{k+kn}{from} \PY{n+nn}{utils} \PY{k+kn}{import} \PY{n}{convert\PYZus{}time\PYZus{}unit\PYZus{}into\PYZus{}name}\PY{p}{,} \PY{n}{get\PYZus{}parameters\PYZus{}for\PYZus{}dataset}\PY{p}{,} \PY{n}{glob\PYZus{}re}\PY{p}{,} \PY{n}{list\PYZus{}full\PYZus{}paths}\PY{p}{,} \PY{n}{creat\PYZus{}dir\PYZus{}for\PYZus{}a\PYZus{}file\PYZus{}if\PYZus{}not\PYZus{}exists}\PY{p}{,} \PY{n}{check\PYZus{}if\PYZus{}item\PYZus{}exist\PYZus{}in\PYZus{}nested\PYZus{}list}\PY{p}{,} \PY{n}{get\PYZus{}dir\PYZus{}of\PYZus{}a\PYZus{}file}\PY{p}{,}\PY{n}{func\PYZus{}compute\PYZus{}metrics\PYZus{}sem\PYZus{}eval}\PY{p}{,}\PY{n}{partition\PYZus{}and\PYZus{}resample\PYZus{}df}\PY{p}{,} \PY{n}{process\PYZus{}dataframe}\PY{p}{,} \PY{n}{tidy\PYZus{}name}
\end{Verbatim}
\end{tcolorbox}

    \begin{tcolorbox}[breakable, size=fbox, boxrule=1pt, pad at break*=1mm,colback=cellbackground, colframe=cellborder]
\prompt{In}{incolor}{14}{\boxspacing}
\begin{Verbatim}[commandchars=\\\{\}]
\PY{c+c1}{\PYZsh{} set up}
\PY{n}{SEED} \PY{o}{=} \PY{l+m+mi}{42}
\PY{n}{TOPIC\PYZus{}OF\PYZus{}INTEREST} \PY{o}{=} \PY{l+s+s2}{\PYZdq{}}\PY{l+s+s2}{Abortion}\PY{l+s+s2}{\PYZdq{}}
\PY{n}{DATASET} \PY{o}{=} \PY{l+s+s2}{\PYZdq{}}\PY{l+s+s2}{SEM\PYZus{}EVAL}\PY{l+s+s2}{\PYZdq{}}
\PY{n}{par} \PY{o}{=} \PY{n}{get\PYZus{}parameters\PYZus{}for\PYZus{}dataset}\PY{p}{(}\PY{n}{DATASET}\PY{p}{)}

\PY{n}{PATH\PYZus{}OUTPUT\PYZus{}ROOT} \PY{o}{=} \PY{n}{join}\PY{p}{(}\PY{n}{par}\PY{o}{.}\PY{n}{PATH\PYZus{}RESULT\PYZus{}SEM\PYZus{}EVAL}\PY{p}{,} \PY{l+s+s2}{\PYZdq{}}\PY{l+s+s2}{llm}\PY{l+s+s2}{\PYZdq{}}\PY{p}{)}
\end{Verbatim}
\end{tcolorbox}

    \begin{tcolorbox}[breakable, size=fbox, boxrule=1pt, pad at break*=1mm,colback=cellbackground, colframe=cellborder]
\prompt{In}{incolor}{15}{\boxspacing}
\begin{Verbatim}[commandchars=\\\{\}]
\PY{c+c1}{\PYZsh{} preprocess the data}
\PY{n}{sem\PYZus{}eval\PYZus{}data} \PY{o}{=} \PY{n}{SemEvalDataProcessor}\PY{p}{(}\PY{p}{)}
\PY{n}{sem\PYZus{}eval\PYZus{}data}\PY{o}{.}\PY{n}{preprocess}\PY{p}{(}\PY{p}{)}

\PY{n}{df\PYZus{}processed} \PY{o}{=} \PY{n}{sem\PYZus{}eval\PYZus{}data}\PY{o}{.}\PY{n}{\PYZus{}read\PYZus{}preprocessed\PYZus{}data}\PY{p}{(}\PY{n}{topic}\PY{o}{=}\PY{n}{TOPIC\PYZus{}OF\PYZus{}INTEREST}\PY{p}{)}\PY{o}{.}\PY{n}{reset\PYZus{}index}\PY{p}{(}\PY{n}{drop}\PY{o}{=}\PY{k+kc}{True}\PY{p}{)}
\PY{c+c1}{\PYZsh{} save the partitions (train, dev, test) for later use}
\PY{n}{df\PYZus{}partitions} \PY{o}{=} \PY{n}{sem\PYZus{}eval\PYZus{}data}\PY{o}{.}\PY{n}{partition\PYZus{}processed\PYZus{}data}\PY{p}{(}\PY{n}{seed}\PY{o}{=}\PY{n}{SEED}\PY{p}{,}\PY{n}{verbose}\PY{o}{=}\PY{k+kc}{False}\PY{p}{)}
\end{Verbatim}
\end{tcolorbox}

    \begin{tcolorbox}[breakable, size=fbox, boxrule=1pt, pad at break*=1mm,colback=cellbackground, colframe=cellborder]
\prompt{In}{incolor}{16}{\boxspacing}
\begin{Verbatim}[commandchars=\\\{\}]
\PY{n}{df\PYZus{}processed}\PY{o}{.}\PY{n}{head}\PY{p}{(}\PY{p}{)}
\end{Verbatim}
\end{tcolorbox}

            \begin{tcolorbox}[breakable, size=fbox, boxrule=.5pt, pad at break*=1mm, opacityfill=0]
\prompt{Out}{outcolor}{16}{\boxspacing}
\begin{Verbatim}[commandchars=\\\{\}]
     ID                                              tweet     topic    label  \textbackslash{}
0  2312  i really don't understand how some people are {\ldots}  Abortion  AGAINST
1  2313  let's agree that it's not ok to kill a 7lbs ba{\ldots}  Abortion  AGAINST
2  2314  @USERNAME i would like to see poll: how many a{\ldots}  Abortion  AGAINST
3  2315  democrats are always against 'personhood' or w{\ldots}  Abortion  AGAINST
4  2316  @USERNAME 'if you don't draw the line where i'{\ldots}  Abortion     NONE

  partition
0     train
1     train
2     train
3     train
4     train
\end{Verbatim}
\end{tcolorbox}
        
    Let's look at the distribution of the stance labels across the training,
testdation, and testing sets.

    \begin{tcolorbox}[breakable, size=fbox, boxrule=1pt, pad at break*=1mm,colback=cellbackground, colframe=cellborder]
\prompt{In}{incolor}{17}{\boxspacing}
\begin{Verbatim}[commandchars=\\\{\}]
\PY{c+c1}{\PYZsh{} add a \PYZdq{}count\PYZdq{} column to count the number of tweets in each partition}
\PY{n}{df\PYZus{}label\PYZus{}dist} \PY{o}{=} \PY{n}{df\PYZus{}partitions}\PY{p}{[}\PY{n}{df\PYZus{}partitions}\PY{o}{.}\PY{n}{topic} \PY{o}{==} \PY{n}{TOPIC\PYZus{}OF\PYZus{}INTEREST}\PY{p}{]}\PY{o}{.}\PY{n}{value\PYZus{}counts}\PY{p}{(}\PY{p}{[}\PY{l+s+s1}{\PYZsq{}}\PY{l+s+s1}{partition}\PY{l+s+s1}{\PYZsq{}}\PY{p}{,}\PY{l+s+s1}{\PYZsq{}}\PY{l+s+s1}{label}\PY{l+s+s1}{\PYZsq{}}\PY{p}{]}\PY{p}{)}\PY{o}{.}\PY{n}{sort\PYZus{}index}\PY{p}{(}\PY{p}{)}
\PY{n}{df\PYZus{}label\PYZus{}dist}
\end{Verbatim}
\end{tcolorbox}

            \begin{tcolorbox}[breakable, size=fbox, boxrule=.5pt, pad at break*=1mm, opacityfill=0]
\prompt{Out}{outcolor}{17}{\boxspacing}
\begin{Verbatim}[commandchars=\\\{\}]
partition  label
test       AGAINST    188
           FAVOR       46
           NONE        45
train      AGAINST    267
           FAVOR       83
           NONE       130
vali       AGAINST     67
           FAVOR       21
           NONE        32
dtype: int64
\end{Verbatim}
\end{tcolorbox}
        
    Critically, since we are using LLMs without fine-tuning, there is no
need for a training set. We will only use the validation and testing
sets for evaluation. The validation set can be employed to choose the
right prompt and the right LLM, while the testing set is utilized to
evaluate the final prompt and model.

From a practical standpoint, this approach significantly reduces the
amount of labeled data required.

    \begin{tcolorbox}[breakable, size=fbox, boxrule=1pt, pad at break*=1mm,colback=cellbackground, colframe=cellborder]
\prompt{In}{incolor}{18}{\boxspacing}
\begin{Verbatim}[commandchars=\\\{\}]
\PY{k+kn}{import} \PY{n+nn}{seaborn} \PY{k}{as} \PY{n+nn}{sns}
\PY{k+kn}{import} \PY{n+nn}{matplotlib}\PY{n+nn}{.}\PY{n+nn}{pyplot} \PY{k}{as} \PY{n+nn}{plt}
\PY{c+c1}{\PYZsh{} only keep the vali and test partitions}
\PY{n}{df\PYZus{}label\PYZus{}dist\PYZus{}plot} \PY{o}{=} \PY{n}{df\PYZus{}label\PYZus{}dist}\PY{o}{.}\PY{n}{reset\PYZus{}index}\PY{p}{(}\PY{p}{)}
\PY{n}{df\PYZus{}label\PYZus{}dist\PYZus{}plot} \PY{o}{=} \PY{n}{df\PYZus{}label\PYZus{}dist\PYZus{}plot}\PY{p}{[}\PY{n}{df\PYZus{}label\PYZus{}dist\PYZus{}plot}\PY{o}{.}\PY{n}{partition}\PY{o}{.}\PY{n}{isin}\PY{p}{(}\PY{p}{[}\PY{l+s+s2}{\PYZdq{}}\PY{l+s+s2}{vali}\PY{l+s+s2}{\PYZdq{}}\PY{p}{,} \PY{l+s+s2}{\PYZdq{}}\PY{l+s+s2}{test}\PY{l+s+s2}{\PYZdq{}}\PY{p}{]}\PY{p}{)}\PY{p}{]}


\PY{c+c1}{\PYZsh{} Create the bar plot}
\PY{n}{plt}\PY{o}{.}\PY{n}{figure}\PY{p}{(}\PY{n}{figsize}\PY{o}{=}\PY{p}{(}\PY{l+m+mi}{8}\PY{p}{,} \PY{l+m+mi}{6}\PY{p}{)}\PY{p}{)}
\PY{n}{ax} \PY{o}{=} \PY{n}{sns}\PY{o}{.}\PY{n}{barplot}\PY{p}{(}\PY{n}{data}\PY{o}{=}\PY{n}{df\PYZus{}label\PYZus{}dist\PYZus{}plot}\PY{p}{,} \PY{n}{x}\PY{o}{=}\PY{l+s+s2}{\PYZdq{}}\PY{l+s+s2}{partition}\PY{l+s+s2}{\PYZdq{}}\PY{p}{,} \PY{n}{y}\PY{o}{=}\PY{l+m+mi}{0}\PY{p}{,} \PY{n}{hue}\PY{o}{=}\PY{l+s+s2}{\PYZdq{}}\PY{l+s+s2}{label}\PY{l+s+s2}{\PYZdq{}}\PY{p}{,} \PY{n}{order}\PY{o}{=}\PY{p}{[}\PY{l+s+s2}{\PYZdq{}}\PY{l+s+s2}{vali}\PY{l+s+s2}{\PYZdq{}}\PY{p}{,} \PY{l+s+s2}{\PYZdq{}}\PY{l+s+s2}{test}\PY{l+s+s2}{\PYZdq{}}\PY{p}{]}\PY{p}{)}

\PY{c+c1}{\PYZsh{} Customize the plot}
\PY{n}{plt}\PY{o}{.}\PY{n}{xlabel}\PY{p}{(}\PY{l+s+s2}{\PYZdq{}}\PY{l+s+s2}{Partition}\PY{l+s+s2}{\PYZdq{}}\PY{p}{)}
\PY{n}{plt}\PY{o}{.}\PY{n}{ylabel}\PY{p}{(}\PY{l+s+s2}{\PYZdq{}}\PY{l+s+s2}{Count}\PY{l+s+s2}{\PYZdq{}}\PY{p}{)}
\PY{n}{plt}\PY{o}{.}\PY{n}{title}\PY{p}{(}\PY{l+s+s2}{\PYZdq{}}\PY{l+s+s2}{Label Distribution}\PY{l+s+s2}{\PYZdq{}}\PY{p}{)}

\PY{c+c1}{\PYZsh{} Add count on top of each bar}
\PY{k}{for} \PY{n}{p} \PY{o+ow}{in} \PY{n}{ax}\PY{o}{.}\PY{n}{patches}\PY{p}{:}
    \PY{n}{ax}\PY{o}{.}\PY{n}{annotate}\PY{p}{(}
        \PY{l+s+sa}{f}\PY{l+s+s1}{\PYZsq{}}\PY{l+s+si}{\PYZob{}}\PY{n}{p}\PY{o}{.}\PY{n}{get\PYZus{}height}\PY{p}{(}\PY{p}{)}\PY{l+s+si}{:}\PY{l+s+s1}{.0f}\PY{l+s+si}{\PYZcb{}}\PY{l+s+s1}{\PYZsq{}}\PY{p}{,}
        \PY{p}{(}\PY{n}{p}\PY{o}{.}\PY{n}{get\PYZus{}x}\PY{p}{(}\PY{p}{)} \PY{o}{+} \PY{n}{p}\PY{o}{.}\PY{n}{get\PYZus{}width}\PY{p}{(}\PY{p}{)} \PY{o}{/} \PY{l+m+mf}{2.}\PY{p}{,} \PY{n}{p}\PY{o}{.}\PY{n}{get\PYZus{}height}\PY{p}{(}\PY{p}{)}\PY{p}{)}\PY{p}{,}
        \PY{n}{ha}\PY{o}{=}\PY{l+s+s1}{\PYZsq{}}\PY{l+s+s1}{center}\PY{l+s+s1}{\PYZsq{}}\PY{p}{,}
        \PY{n}{va}\PY{o}{=}\PY{l+s+s1}{\PYZsq{}}\PY{l+s+s1}{baseline}\PY{l+s+s1}{\PYZsq{}}\PY{p}{,}
        \PY{n}{fontsize}\PY{o}{=}\PY{l+m+mi}{12}\PY{p}{,}
        \PY{n}{color}\PY{o}{=}\PY{l+s+s1}{\PYZsq{}}\PY{l+s+s1}{black}\PY{l+s+s1}{\PYZsq{}}\PY{p}{,}
        \PY{n}{xytext}\PY{o}{=}\PY{p}{(}\PY{l+m+mi}{0}\PY{p}{,} \PY{l+m+mi}{5}\PY{p}{)}\PY{p}{,}
        \PY{n}{textcoords}\PY{o}{=}\PY{l+s+s1}{\PYZsq{}}\PY{l+s+s1}{offset points}\PY{l+s+s1}{\PYZsq{}}
    \PY{p}{)}
\PY{c+c1}{\PYZsh{} Show the plot}
\PY{n}{plt}\PY{o}{.}\PY{n}{show}\PY{p}{(}\PY{p}{)}
\end{Verbatim}
\end{tcolorbox}

    \begin{center}
    \adjustimage{max size={0.9\linewidth}{0.9\paperheight}}{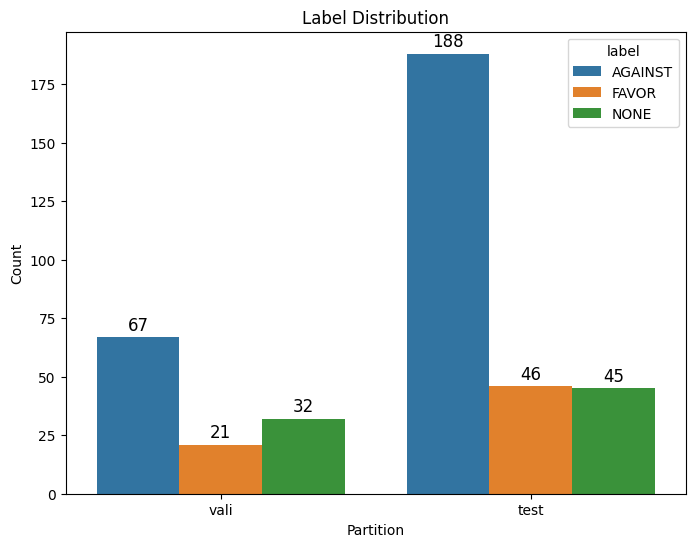}
    \end{center}
    { \hspace*{\fill} \\}
    
    \hypertarget{create-the-prompts}{%
\section{Create the Prompts}\label{create-the-prompts}}

    Let's create the prompts for the stance detection task. We will use the
following three prompts: zero-shot, few-shot (1 example per class), and
zero-shot chain-of-thought (CoT) prompts.

    \begin{tcolorbox}[breakable, size=fbox, boxrule=1pt, pad at break*=1mm,colback=cellbackground, colframe=cellborder]
\prompt{In}{incolor}{19}{\boxspacing}
\begin{Verbatim}[commandchars=\\\{\}]
\PY{o}{\PYZpc{}}\PY{k}{load\PYZus{}ext} autoreload
\PY{o}{\PYZpc{}}\PY{k}{autoreload} 2
\PY{k+kn}{from} \PY{n+nn}{gpt\PYZus{}data\PYZus{}processor} \PY{k+kn}{import} \PY{n}{SemEvalGPTDataProcessor}
\end{Verbatim}
\end{tcolorbox}

    \begin{Verbatim}[commandchars=\\\{\}]
The autoreload extension is already loaded. To reload it, use:
  \%reload\_ext autoreload
    \end{Verbatim}

    \hypertarget{zero-shot-prompt}{%
\subsection{Zero-shot prompt}\label{zero-shot-prompt}}

    \begin{tcolorbox}[breakable, size=fbox, boxrule=1pt, pad at break*=1mm,colback=cellbackground, colframe=cellborder]
\prompt{In}{incolor}{20}{\boxspacing}
\begin{Verbatim}[commandchars=\\\{\}]
\PY{c+c1}{\PYZsh{} zero\PYZhy{}shot (k = 0)}
\PY{n}{VERSION\PYZus{}PROMPT} \PY{o}{=} \PY{l+s+s2}{\PYZdq{}}\PY{l+s+s2}{zero\PYZus{}shot}\PY{l+s+s2}{\PYZdq{}}
\end{Verbatim}
\end{tcolorbox}

    \begin{tcolorbox}[breakable, size=fbox, boxrule=1pt, pad at break*=1mm,colback=cellbackground, colframe=cellborder]
\prompt{In}{incolor}{21}{\boxspacing}
\begin{Verbatim}[commandchars=\\\{\}]
\PY{n}{gpt\PYZus{}data\PYZus{}processor} \PY{o}{=} \PY{n}{SemEvalGPTDataProcessor}\PY{p}{(}\PY{n}{version\PYZus{}prompt} \PY{o}{=} \PY{n}{VERSION\PYZus{}PROMPT}\PY{p}{,} \PY{n}{topic}\PY{o}{=}\PY{n}{TOPIC\PYZus{}OF\PYZus{}INTEREST}\PY{p}{)}
\PY{c+c1}{\PYZsh{} the prompt, which is the input to the LLMs}
\PY{n}{df\PYZus{}input\PYZus{}text} \PY{o}{=} \PY{n}{gpt\PYZus{}data\PYZus{}processor}\PY{o}{.}\PY{n}{embed\PYZus{}prompt}\PY{p}{(}\PY{n}{file\PYZus{}output}\PY{o}{=}\PY{k+kc}{None}\PY{p}{,} \PY{n}{write\PYZus{}csv}\PY{o}{=}\PY{k+kc}{False}\PY{p}{,}
                                                \PY{n}{return\PYZus{}df}\PY{o}{=}\PY{k+kc}{True}\PY{p}{)}
\end{Verbatim}
\end{tcolorbox}

    \begin{tcolorbox}[breakable, size=fbox, boxrule=1pt, pad at break*=1mm,colback=cellbackground, colframe=cellborder]
\prompt{In}{incolor}{22}{\boxspacing}
\begin{Verbatim}[commandchars=\\\{\}]
\PY{n}{df\PYZus{}input\PYZus{}text}\PY{o}{.}\PY{n}{head}\PY{p}{(}\PY{p}{)}
\end{Verbatim}
\end{tcolorbox}

            \begin{tcolorbox}[breakable, size=fbox, boxrule=.5pt, pad at break*=1mm, opacityfill=0]
\prompt{Out}{outcolor}{22}{\boxspacing}
\begin{Verbatim}[commandchars=\\\{\}]
     ID                                              tweet     topic    label  \textbackslash{}
0  2312  i really don't understand how some people are {\ldots}  Abortion  AGAINST
1  2313  let's agree that it's not ok to kill a 7lbs ba{\ldots}  Abortion  AGAINST
2  2314  @USERNAME i would like to see poll: how many a{\ldots}  Abortion  AGAINST
3  2315  democrats are always against 'personhood' or w{\ldots}  Abortion  AGAINST
4  2316  @USERNAME 'if you don't draw the line where i'{\ldots}  Abortion     NONE

  partition                                     tweet\_embedded
0     train  What is the stance of the tweet below with res{\ldots}
1     train  What is the stance of the tweet below with res{\ldots}
2     train  What is the stance of the tweet below with res{\ldots}
3     train  What is the stance of the tweet below with res{\ldots}
4     train  What is the stance of the tweet below with res{\ldots}
\end{Verbatim}
\end{tcolorbox}
        
    Let's look at one example of the zero-shot prompt. As you can see, the
tweet of interest is now embedded in the prompt.

    \begin{tcolorbox}[breakable, size=fbox, boxrule=1pt, pad at break*=1mm,colback=cellbackground, colframe=cellborder]
\prompt{In}{incolor}{23}{\boxspacing}
\begin{Verbatim}[commandchars=\\\{\}]
\PY{n+nb}{print}\PY{p}{(}\PY{l+s+s2}{\PYZdq{}}\PY{l+s+s2}{tweet:}\PY{l+s+se}{\PYZbs{}n}\PY{l+s+si}{\PYZob{}\PYZcb{}}\PY{l+s+se}{\PYZbs{}n}\PY{l+s+s2}{\PYZdq{}}\PY{o}{.}\PY{n}{format}\PY{p}{(}\PY{n}{df\PYZus{}input\PYZus{}text}\PY{p}{[}\PY{l+s+s2}{\PYZdq{}}\PY{l+s+s2}{tweet}\PY{l+s+s2}{\PYZdq{}}\PY{p}{]}\PY{p}{[}\PY{l+m+mi}{0}\PY{p}{]}\PY{p}{)}\PY{p}{)}
\PY{n+nb}{print}\PY{p}{(}\PY{l+s+s2}{\PYZdq{}}\PY{l+s+s2}{label:}\PY{l+s+se}{\PYZbs{}n}\PY{l+s+si}{\PYZob{}\PYZcb{}}\PY{l+s+se}{\PYZbs{}n}\PY{l+s+s2}{\PYZdq{}}\PY{o}{.}\PY{n}{format}\PY{p}{(}\PY{n}{df\PYZus{}input\PYZus{}text}\PY{p}{[}\PY{l+s+s2}{\PYZdq{}}\PY{l+s+s2}{label}\PY{l+s+s2}{\PYZdq{}}\PY{p}{]}\PY{p}{[}\PY{l+m+mi}{0}\PY{p}{]}\PY{p}{)}\PY{p}{)}
\PY{n+nb}{print}\PY{p}{(}\PY{l+s+s2}{\PYZdq{}}\PY{l+s+s2}{prompt:}\PY{l+s+se}{\PYZbs{}n}\PY{l+s+si}{\PYZob{}\PYZcb{}}\PY{l+s+s2}{\PYZdq{}}\PY{o}{.}\PY{n}{format}\PY{p}{(}\PY{n}{df\PYZus{}input\PYZus{}text}\PY{p}{[}\PY{l+s+s2}{\PYZdq{}}\PY{l+s+s2}{tweet\PYZus{}embedded}\PY{l+s+s2}{\PYZdq{}}\PY{p}{]}\PY{p}{[}\PY{l+m+mi}{0}\PY{p}{]}\PY{p}{)}\PY{p}{)}
\end{Verbatim}
\end{tcolorbox}

    \begin{Verbatim}[commandchars=\\\{\}]
tweet:
i really don't understand how some people are pro-choice. a life is a life no
matter if it's 2 weeks old or 20 years old.

label:
AGAINST

prompt:
What is the stance of the tweet below with respect to 'Legalization of
Abortion'?  If we can infer from the tweet that the tweeter supports
'Legalization of Abortion', please label it as 'in-favor'. If we can infer from
the tweet that the tweeter is against 'Legalization of Abortion', please label
is as 'against'. If we can infer from the tweet that the tweeter has a neutral
stance towards 'Legalization of Abortion', please label it as 'neutral-or-
unclear'. If there is no clue in the tweet to reveal the stance of the tweeter
towards 'Legalization of Abortion', please also label is as 'neutral-or-
unclear'. Please use exactly one word from the following 3 categories to label
it: 'in-favor', 'against', 'neutral-or-unclear'. Here is the tweet. 'i really
don't understand how some people are pro-choice. a life is a life no matter if
it's 2 weeks old or 20 years old.' The stance of the tweet is:
    \end{Verbatim}

    \hypertarget{few-shot-prompt}{%
\subsection{Few-shot prompt}\label{few-shot-prompt}}

We repeat the same process for the few-shot prompt.

Note that the 3 examples are manually selected by me from the training
set with the hope that they are representative of the stance class.
These 3 examples are placed in the prompt for all the tweets while
predicting the stance. There may be more effective ways to select the
few-shot examples, but this is beyond the scope of this tutorial.

    \begin{tcolorbox}[breakable, size=fbox, boxrule=1pt, pad at break*=1mm,colback=cellbackground, colframe=cellborder]
\prompt{In}{incolor}{24}{\boxspacing}
\begin{Verbatim}[commandchars=\\\{\}]
\PY{n}{gpt\PYZus{}data\PYZus{}processor} \PY{o}{=} \PY{n}{SemEvalGPTDataProcessor}\PY{p}{(}\PY{n}{version\PYZus{}prompt} \PY{o}{=} \PY{l+s+s2}{\PYZdq{}}\PY{l+s+s2}{few\PYZus{}shot}\PY{l+s+s2}{\PYZdq{}}\PY{p}{,} \PY{n}{topic}\PY{o}{=}\PY{n}{TOPIC\PYZus{}OF\PYZus{}INTEREST}\PY{p}{)}
\PY{c+c1}{\PYZsh{} the prompt, which is the input to the LLMs}
\PY{n}{df\PYZus{}input\PYZus{}text} \PY{o}{=} \PY{n}{gpt\PYZus{}data\PYZus{}processor}\PY{o}{.}\PY{n}{embed\PYZus{}prompt}\PY{p}{(}\PY{n}{file\PYZus{}output}\PY{o}{=}\PY{k+kc}{None}\PY{p}{,} \PY{n}{write\PYZus{}csv}\PY{o}{=}\PY{k+kc}{False}\PY{p}{,}
                                                \PY{n}{return\PYZus{}df}\PY{o}{=}\PY{k+kc}{True}\PY{p}{)}
\end{Verbatim}
\end{tcolorbox}

    \begin{tcolorbox}[breakable, size=fbox, boxrule=1pt, pad at break*=1mm,colback=cellbackground, colframe=cellborder]
\prompt{In}{incolor}{25}{\boxspacing}
\begin{Verbatim}[commandchars=\\\{\}]
\PY{n+nb}{print}\PY{p}{(}\PY{l+s+s2}{\PYZdq{}}\PY{l+s+s2}{tweet:}\PY{l+s+se}{\PYZbs{}n}\PY{l+s+si}{\PYZob{}\PYZcb{}}\PY{l+s+se}{\PYZbs{}n}\PY{l+s+s2}{\PYZdq{}}\PY{o}{.}\PY{n}{format}\PY{p}{(}\PY{n}{df\PYZus{}input\PYZus{}text}\PY{p}{[}\PY{l+s+s2}{\PYZdq{}}\PY{l+s+s2}{tweet}\PY{l+s+s2}{\PYZdq{}}\PY{p}{]}\PY{p}{[}\PY{l+m+mi}{0}\PY{p}{]}\PY{p}{)}\PY{p}{)}
\PY{n+nb}{print}\PY{p}{(}\PY{l+s+s2}{\PYZdq{}}\PY{l+s+s2}{label:}\PY{l+s+se}{\PYZbs{}n}\PY{l+s+si}{\PYZob{}\PYZcb{}}\PY{l+s+se}{\PYZbs{}n}\PY{l+s+s2}{\PYZdq{}}\PY{o}{.}\PY{n}{format}\PY{p}{(}\PY{n}{df\PYZus{}input\PYZus{}text}\PY{p}{[}\PY{l+s+s2}{\PYZdq{}}\PY{l+s+s2}{label}\PY{l+s+s2}{\PYZdq{}}\PY{p}{]}\PY{p}{[}\PY{l+m+mi}{0}\PY{p}{]}\PY{p}{)}\PY{p}{)}
\PY{n+nb}{print}\PY{p}{(}\PY{l+s+s2}{\PYZdq{}}\PY{l+s+s2}{prompt:}\PY{l+s+se}{\PYZbs{}n}\PY{l+s+si}{\PYZob{}\PYZcb{}}\PY{l+s+s2}{\PYZdq{}}\PY{o}{.}\PY{n}{format}\PY{p}{(}\PY{n}{df\PYZus{}input\PYZus{}text}\PY{p}{[}\PY{l+s+s2}{\PYZdq{}}\PY{l+s+s2}{tweet\PYZus{}embedded}\PY{l+s+s2}{\PYZdq{}}\PY{p}{]}\PY{p}{[}\PY{l+m+mi}{0}\PY{p}{]}\PY{p}{)}\PY{p}{)}
\end{Verbatim}
\end{tcolorbox}

    \begin{Verbatim}[commandchars=\\\{\}]
tweet:
i really don't understand how some people are pro-choice. a life is a life no
matter if it's 2 weeks old or 20 years old.

label:
AGAINST

prompt:
What is the stance of the tweet below with respect to 'Legalization of
Abortion'?  If we can infer from the tweet that the tweeter supports
'Legalization of Abortion', please label it as 'in-favor'. If we can infer from
the tweet that the tweeter is against 'Legalization of Abortion', please label
is as 'against'. If we can infer from the tweet that the tweeter has a neutral
stance towards 'Legalization of Abortion', please label it as 'neutral-or-
unclear'. If there is no clue in the tweet to reveal the stance of the tweeter
towards 'Legalization of Abortion', please also label is as 'neutral-or-
unclear'. Please use exactly one word from the following 3 categories to label
it: 'in-favor', 'against', 'neutral-or-unclear'.  Here are some examples of
tweets. Make sure to classify the last tweet correctly.
Q: Tweet: it's a free country. freedom includes freedom of choice.
Is this tweet in-favor, against, or neutral-or-unclear?
A: in-favor
Q: Tweet: i really don't understand how some people are pro-choice. a life is a
life no matter if it's 2 weeks old or 20 years old.
Is this tweet in-favor, against, or neutral-or-unclear?
A: against
Q: Tweet: so ready for my abortion debate
Is this tweet in-favor, against, or neutral-or-unclear?
A: neutral-or-unclear
Q: Tweet: 'i really don't understand how some people are pro-choice. a life is a
life no matter if it's 2 weeks old or 20 years old.'
Is this tweet in-favor, against, or neutral-or-unclear?
A:
    \end{Verbatim}

    \hypertarget{chain-of-thought-prompt-cot}{%
\subsection{Chain-of-thought prompt
(CoT)}\label{chain-of-thought-prompt-cot}}

We repeat the same process for the zero-shot CoT prompt. The critical
sentence ``Let's think step by step.'' is added to the end of prompt.

    \begin{tcolorbox}[breakable, size=fbox, boxrule=1pt, pad at break*=1mm,colback=cellbackground, colframe=cellborder]
\prompt{In}{incolor}{26}{\boxspacing}
\begin{Verbatim}[commandchars=\\\{\}]
\PY{n}{gpt\PYZus{}data\PYZus{}processor} \PY{o}{=} \PY{n}{SemEvalGPTDataProcessor}\PY{p}{(}\PY{n}{version\PYZus{}prompt} \PY{o}{=} \PY{l+s+s2}{\PYZdq{}}\PY{l+s+s2}{CoT}\PY{l+s+s2}{\PYZdq{}}\PY{p}{,} \PY{n}{topic}\PY{o}{=}\PY{n}{TOPIC\PYZus{}OF\PYZus{}INTEREST}\PY{p}{)}
\PY{c+c1}{\PYZsh{} the prompt, which is the input to the LLMs}
\PY{n}{df\PYZus{}input\PYZus{}text} \PY{o}{=} \PY{n}{gpt\PYZus{}data\PYZus{}processor}\PY{o}{.}\PY{n}{embed\PYZus{}prompt}\PY{p}{(}\PY{n}{file\PYZus{}output}\PY{o}{=}\PY{k+kc}{None}\PY{p}{,} \PY{n}{write\PYZus{}csv}\PY{o}{=}\PY{k+kc}{False}\PY{p}{,}
                                                \PY{n}{return\PYZus{}df}\PY{o}{=}\PY{k+kc}{True}\PY{p}{)}
\end{Verbatim}
\end{tcolorbox}

    \begin{tcolorbox}[breakable, size=fbox, boxrule=1pt, pad at break*=1mm,colback=cellbackground, colframe=cellborder]
\prompt{In}{incolor}{27}{\boxspacing}
\begin{Verbatim}[commandchars=\\\{\}]
\PY{n+nb}{print}\PY{p}{(}\PY{l+s+s2}{\PYZdq{}}\PY{l+s+s2}{tweet:}\PY{l+s+se}{\PYZbs{}n}\PY{l+s+si}{\PYZob{}\PYZcb{}}\PY{l+s+se}{\PYZbs{}n}\PY{l+s+s2}{\PYZdq{}}\PY{o}{.}\PY{n}{format}\PY{p}{(}\PY{n}{df\PYZus{}input\PYZus{}text}\PY{p}{[}\PY{l+s+s2}{\PYZdq{}}\PY{l+s+s2}{tweet}\PY{l+s+s2}{\PYZdq{}}\PY{p}{]}\PY{p}{[}\PY{l+m+mi}{0}\PY{p}{]}\PY{p}{)}\PY{p}{)}
\PY{n+nb}{print}\PY{p}{(}\PY{l+s+s2}{\PYZdq{}}\PY{l+s+s2}{label:}\PY{l+s+se}{\PYZbs{}n}\PY{l+s+si}{\PYZob{}\PYZcb{}}\PY{l+s+se}{\PYZbs{}n}\PY{l+s+s2}{\PYZdq{}}\PY{o}{.}\PY{n}{format}\PY{p}{(}\PY{n}{df\PYZus{}input\PYZus{}text}\PY{p}{[}\PY{l+s+s2}{\PYZdq{}}\PY{l+s+s2}{label}\PY{l+s+s2}{\PYZdq{}}\PY{p}{]}\PY{p}{[}\PY{l+m+mi}{0}\PY{p}{]}\PY{p}{)}\PY{p}{)}
\PY{n+nb}{print}\PY{p}{(}\PY{l+s+s2}{\PYZdq{}}\PY{l+s+s2}{prompt:}\PY{l+s+se}{\PYZbs{}n}\PY{l+s+si}{\PYZob{}\PYZcb{}}\PY{l+s+s2}{\PYZdq{}}\PY{o}{.}\PY{n}{format}\PY{p}{(}\PY{n}{df\PYZus{}input\PYZus{}text}\PY{p}{[}\PY{l+s+s2}{\PYZdq{}}\PY{l+s+s2}{tweet\PYZus{}embedded}\PY{l+s+s2}{\PYZdq{}}\PY{p}{]}\PY{p}{[}\PY{l+m+mi}{0}\PY{p}{]}\PY{p}{)}\PY{p}{)}
\end{Verbatim}
\end{tcolorbox}

    \begin{Verbatim}[commandchars=\\\{\}]
tweet:
i really don't understand how some people are pro-choice. a life is a life no
matter if it's 2 weeks old or 20 years old.

label:
AGAINST

prompt:
What is the stance of the tweet below with respect to 'Legalization of
Abortion'?  If we can infer from the tweet that the tweeter supports
'Legalization of Abortion', please label it as 'in-favor'. If we can infer from
the tweet that the tweeter is against 'Legalization of Abortion', please label
is as 'against'. If we can infer from the tweet that the tweeter has a neutral
stance towards 'Legalization of Abortion', please label it as 'neutral-or-
unclear'. If there is no clue in the tweet to reveal the stance of the tweeter
towards 'Legalization of Abortion', please also label is as 'neutral-or-
unclear'. Here is the tweet. 'i really don't understand how some people are pro-
choice. a life is a life no matter if it's 2 weeks old or 20 years old.' What is
the stance of the tweet with respect to 'Legalization of Abortion'? Please make
sure that at the end of your response, use exactly one word from the following 3
categories to label the stance with respect to 'Legalization of Abortion': 'in-
favor', 'against', 'neutral-or-unclear’. Let's think step by step.
    \end{Verbatim}

    \begin{center}\rule{0.5\linewidth}{0.5pt}\end{center}

    \hypertarget{feed-the-prompts-to-chatgpt}{%
\section{Feed the prompts to
ChatGPT}\label{feed-the-prompts-to-chatgpt}}

Note that the specifc version of ChatGPT in used is
\texttt{gpt-3.5-turbo}.

    First, you have to decide whether you want to prompt ChatGPT on your
own.

I recommend keeping \texttt{PROMPT\_CHAT\_GPT\ =\ False} (the default
setting below) if you are running this notebook for the first time. This
will read the predictions I made and uploaded to my GitHub repo, which
will save you time and money.

If you want to try it out on your own, you can set
\texttt{PROMPT\_CHAT\_GPT\ =\ True} and run the code below. In this
case, you should also provide your own API key below. Here is the
\href{https://help.openai.com/en/articles/4936850-where-do-i-find-my-secret-api-key}{OpenAI's
page} on how to find your API key. Note that this will cost you about
\$1 to run this notebook.

    \begin{tcolorbox}[breakable, size=fbox, boxrule=1pt, pad at break*=1mm,colback=cellbackground, colframe=cellborder]
\prompt{In}{incolor}{28}{\boxspacing}
\begin{Verbatim}[commandchars=\\\{\}]
\PY{c+c1}{\PYZsh{} Whether you want to prompt the GPT model yourself (this would cost around \PYZdl{}1 USD for the entire tutorial)}
\PY{n}{PROMPT\PYZus{}CHAT\PYZus{}GPT} \PY{o}{=} \PY{k+kc}{False}
\end{Verbatim}
\end{tcolorbox}

    \begin{tcolorbox}[breakable, size=fbox, boxrule=1pt, pad at break*=1mm,colback=cellbackground, colframe=cellborder]
\prompt{In}{incolor}{29}{\boxspacing}
\begin{Verbatim}[commandchars=\\\{\}]
\PY{k+kn}{import} \PY{n+nn}{os}
\PY{c+c1}{\PYZsh{} You should set the API key for OpenAI}
\PY{n}{OPEN\PYZus{}AI\PYZus{}KEY} \PY{o}{=} \PY{l+s+s2}{\PYZdq{}}\PY{l+s+s2}{FILL YOUR KEY HERE}\PY{l+s+s2}{\PYZdq{}}

\PY{k}{if} \PY{n}{PROMPT\PYZus{}CHAT\PYZus{}GPT}\PY{p}{:}
    \PY{k+kn}{import} \PY{n+nn}{openai}
    \PY{n}{openai}\PY{o}{.}\PY{n}{api\PYZus{}key} \PY{o}{=} \PY{n}{OPEN\PYZus{}AI\PYZus{}KEY}
    \PY{k}{assert} \PY{n}{OPEN\PYZus{}AI\PYZus{}KEY}\PY{o}{!=}\PY{l+s+s2}{\PYZdq{}}\PY{l+s+s2}{FILL YOUR KEY HERE}\PY{l+s+s2}{\PYZdq{}}\PY{p}{,} \PY{l+s+s2}{\PYZdq{}}\PY{l+s+s2}{You should set the API key for OpenAI}\PY{l+s+s2}{\PYZdq{}}

\PY{c+c1}{\PYZsh{} to avoid overwriting the existing prediction file}
\PY{k}{for} \PY{n}{prompt\PYZus{}type} \PY{o+ow}{in} \PY{p}{[}\PY{l+s+s2}{\PYZdq{}}\PY{l+s+s2}{zero\PYZus{}shot}\PY{l+s+s2}{\PYZdq{}}\PY{p}{,} \PY{l+s+s2}{\PYZdq{}}\PY{l+s+s2}{few\PYZus{}shot}\PY{l+s+s2}{\PYZdq{}}\PY{p}{,} \PY{l+s+s2}{\PYZdq{}}\PY{l+s+s2}{CoT}\PY{l+s+s2}{\PYZdq{}}\PY{p}{]}\PY{p}{:}
    \PY{n}{file\PYZus{}prediction} \PY{o}{=} \PY{n}{join}\PY{p}{(}\PY{n}{PATH\PYZus{}OUTPUT\PYZus{}ROOT}\PY{p}{,} \PY{l+s+s2}{\PYZdq{}}\PY{l+s+s2}{chatgpt\PYZus{}turbo\PYZus{}3\PYZus{}5}\PY{l+s+s2}{\PYZdq{}}\PY{p}{,} \PY{n}{prompt\PYZus{}type}\PY{p}{,} \PY{l+s+s2}{\PYZdq{}}\PY{l+s+s2}{predictions.csv}\PY{l+s+s2}{\PYZdq{}}\PY{p}{)}
    \PY{n}{file\PYZus{}prediction\PYZus{}cached} \PY{o}{=} \PY{n}{join}\PY{p}{(}\PY{n}{PATH\PYZus{}OUTPUT\PYZus{}ROOT}\PY{p}{,}\PY{l+s+s2}{\PYZdq{}}\PY{l+s+s2}{chatgpt\PYZus{}turbo\PYZus{}3\PYZus{}5}\PY{l+s+s2}{\PYZdq{}}\PY{p}{,} \PY{n}{prompt\PYZus{}type}\PY{p}{,} \PY{l+s+s2}{\PYZdq{}}\PY{l+s+s2}{predictions\PYZus{}cached.csv}\PY{l+s+s2}{\PYZdq{}}\PY{p}{)}
    \PY{c+c1}{\PYZsh{} suffix the existing prediction file with \PYZdq{}\PYZus{}cached\PYZdq{}}
    \PY{k}{if} \PY{n}{PROMPT\PYZus{}CHAT\PYZus{}GPT}\PY{p}{:}
        \PY{k}{if} \PY{n}{os}\PY{o}{.}\PY{n}{path}\PY{o}{.}\PY{n}{exists}\PY{p}{(}\PY{n}{file\PYZus{}prediction}\PY{p}{)}\PY{p}{:}
            \PY{n}{os}\PY{o}{.}\PY{n}{rename}\PY{p}{(}\PY{n}{file\PYZus{}prediction}\PY{p}{,} \PY{n}{file\PYZus{}prediction\PYZus{}cached}\PY{p}{)}
    \PY{k}{else}\PY{p}{:}
        \PY{k}{if} \PY{n}{os}\PY{o}{.}\PY{n}{path}\PY{o}{.}\PY{n}{exists}\PY{p}{(}\PY{n}{file\PYZus{}prediction\PYZus{}cached}\PY{p}{)}\PY{p}{:}
            \PY{n}{os}\PY{o}{.}\PY{n}{rename}\PY{p}{(}\PY{n}{file\PYZus{}prediction\PYZus{}cached}\PY{p}{,}\PY{n}{file\PYZus{}prediction}\PY{p}{)}
\end{Verbatim}
\end{tcolorbox}

    \begin{tcolorbox}[breakable, size=fbox, boxrule=1pt, pad at break*=1mm,colback=cellbackground, colframe=cellborder]
\prompt{In}{incolor}{30}{\boxspacing}
\begin{Verbatim}[commandchars=\\\{\}]
\PY{k+kn}{from} \PY{n+nn}{gpt\PYZus{}predict\PYZus{}label} \PY{k+kn}{import} \PY{n}{GPTChatTurbo3\PYZus{}5LabelPredictor}
\end{Verbatim}
\end{tcolorbox}

    \begin{tcolorbox}[breakable, size=fbox, boxrule=1pt, pad at break*=1mm,colback=cellbackground, colframe=cellborder]
\prompt{In}{incolor}{31}{\boxspacing}
\begin{Verbatim}[commandchars=\\\{\}]
\PY{k}{class} \PY{n+nc}{PromptChatGPT\PYZus{}3\PYZus{}5}\PY{p}{:}
    \PY{k}{def} \PY{n+nf+fm}{\PYZus{}\PYZus{}init\PYZus{}\PYZus{}}\PY{p}{(}\PY{n+nb+bp}{self}\PY{p}{,}\PY{n}{prompt\PYZus{}version}\PY{p}{)} \PY{o}{\PYZhy{}}\PY{o}{\PYZgt{}} \PY{k+kc}{None}\PY{p}{:}
        \PY{c+c1}{\PYZsh{} get the output paths}
        \PY{c+c1}{\PYZsh{} model\PYZus{}type\PYZus{}name = tidy\PYZus{}name(model\PYZus{}type)}
        \PY{n}{path\PYZus{}output} \PY{o}{=} \PY{n}{join}\PY{p}{(}\PY{n}{PATH\PYZus{}OUTPUT\PYZus{}ROOT}\PY{p}{,} \PY{l+s+s2}{\PYZdq{}}\PY{l+s+s2}{chatgpt\PYZus{}turbo\PYZus{}3\PYZus{}5}\PY{l+s+s2}{\PYZdq{}}\PY{p}{,} \PY{n}{prompt\PYZus{}version}\PY{p}{)}
        \PY{n+nb+bp}{self}\PY{o}{.}\PY{n}{file\PYZus{}output\PYZus{}predictions} \PY{o}{=} \PY{n}{join}\PY{p}{(}\PY{n}{path\PYZus{}output}\PY{p}{,} \PY{l+s+s2}{\PYZdq{}}\PY{l+s+s2}{predictions.csv}\PY{l+s+s2}{\PYZdq{}}\PY{p}{)}

        \PY{n}{gpt\PYZus{}data\PYZus{}processor} \PY{o}{=} \PY{n}{SemEvalGPTDataProcessor}\PY{p}{(}\PY{n}{version\PYZus{}prompt}\PY{o}{=}\PY{n}{prompt\PYZus{}version}\PY{p}{,} \PY{n}{topic}\PY{o}{=}\PY{n}{TOPIC\PYZus{}OF\PYZus{}INTEREST}\PY{p}{)}

        \PY{c+c1}{\PYZsh{} create the prompt, which is the input to the LLMs}
        \PY{n}{df\PYZus{}input\PYZus{}text} \PY{o}{=} \PY{n}{gpt\PYZus{}data\PYZus{}processor}\PY{o}{.}\PY{n}{embed\PYZus{}prompt}\PY{p}{(}\PY{n}{file\PYZus{}output}\PY{o}{=}\PY{k+kc}{None}\PY{p}{,} \PY{n}{write\PYZus{}csv}\PY{o}{=}\PY{k+kc}{False}\PY{p}{,}
                                                        \PY{n}{return\PYZus{}df}\PY{o}{=}\PY{k+kc}{True}\PY{p}{)}

        \PY{c+c1}{\PYZsh{} partition the data into train, vali, test (note that we only need the vali and test set in this approach)}
        \PY{n}{dict\PYZus{}df\PYZus{}single\PYZus{}domain} \PY{o}{=} \PY{n}{partition\PYZus{}and\PYZus{}resample\PYZus{}df}\PY{p}{(}
            \PY{n}{df\PYZus{}input\PYZus{}text}\PY{p}{,} \PY{n}{seed}\PY{o}{=}\PY{k+kc}{None}\PY{p}{,} \PY{n}{partition\PYZus{}type}\PY{o}{=}\PY{l+s+s2}{\PYZdq{}}\PY{l+s+s2}{single\PYZus{}domain}\PY{l+s+s2}{\PYZdq{}}\PY{p}{,}
            \PY{n}{read\PYZus{}partition\PYZus{}from\PYZus{}df}\PY{o}{=}\PY{k+kc}{True}\PY{p}{,}
            \PY{n}{df\PYZus{}partitions}\PY{o}{=}\PY{n}{df\PYZus{}partitions}\PY{p}{)}

        \PY{c+c1}{\PYZsh{} select the partition to be labeled}
        \PY{c+c1}{\PYZsh{} \PYZhy{} vali and test}
        \PY{n}{df\PYZus{}input\PYZus{}text\PYZus{}filtered} \PY{o}{=} \PY{n}{pd}\PY{o}{.}\PY{n}{DataFrame}\PY{p}{(}\PY{p}{)}
        \PY{k}{for} \PY{n}{partition} \PY{o+ow}{in} \PY{p}{[}\PY{l+s+s2}{\PYZdq{}}\PY{l+s+s2}{vali\PYZus{}raw}\PY{l+s+s2}{\PYZdq{}}\PY{p}{,} \PY{l+s+s2}{\PYZdq{}}\PY{l+s+s2}{test\PYZus{}raw}\PY{l+s+s2}{\PYZdq{}}\PY{p}{]}\PY{p}{:}
            \PY{n}{df\PYZus{}input\PYZus{}text\PYZus{}filtered} \PY{o}{=} \PY{n}{pd}\PY{o}{.}\PY{n}{concat}\PY{p}{(}\PY{p}{[}\PY{n}{df\PYZus{}input\PYZus{}text\PYZus{}filtered}\PY{p}{,} \PY{n}{dict\PYZus{}df\PYZus{}single\PYZus{}domain}\PY{p}{[}\PY{n}{partition}\PY{p}{]}\PY{p}{]}\PY{p}{)}

        \PY{n+nb+bp}{self}\PY{o}{.}\PY{n}{df\PYZus{}input\PYZus{}text} \PY{o}{=} \PY{n}{df\PYZus{}input\PYZus{}text\PYZus{}filtered}

        \PY{c+c1}{\PYZsh{} specify the output type (single\PYZhy{}word or multi\PYZhy{}word)}
        \PY{k}{if} \PY{n}{prompt\PYZus{}version} \PY{o+ow}{in} \PY{p}{[}\PY{l+s+s2}{\PYZdq{}}\PY{l+s+s2}{zero\PYZus{}shot}\PY{l+s+s2}{\PYZdq{}}\PY{p}{,}\PY{l+s+s2}{\PYZdq{}}\PY{l+s+s2}{few\PYZus{}shot}\PY{l+s+s2}{\PYZdq{}}\PY{p}{]}\PY{p}{:}
            \PY{n}{mode\PYZus{}output} \PY{o}{=} \PY{l+s+s2}{\PYZdq{}}\PY{l+s+s2}{single\PYZhy{}word}\PY{l+s+s2}{\PYZdq{}}
        \PY{k}{elif} \PY{n}{prompt\PYZus{}version} \PY{o}{==} \PY{l+s+s2}{\PYZdq{}}\PY{l+s+s2}{CoT}\PY{l+s+s2}{\PYZdq{}}\PY{p}{:}
            \PY{n}{mode\PYZus{}output} \PY{o}{=} \PY{l+s+s2}{\PYZdq{}}\PY{l+s+s2}{CoT}\PY{l+s+s2}{\PYZdq{}}

        \PY{n+nb+bp}{self}\PY{o}{.}\PY{n}{llm\PYZus{}label\PYZus{}predictor} \PY{o}{=} \PY{n}{GPTChatTurbo3\PYZus{}5LabelPredictor}\PY{p}{(}\PY{n}{col\PYZus{}name\PYZus{}text}\PY{o}{=}\PY{l+s+s2}{\PYZdq{}}\PY{l+s+s2}{tweet\PYZus{}embedded}\PY{l+s+s2}{\PYZdq{}}\PY{p}{,}
                                                            \PY{n}{col\PYZus{}name\PYZus{}label}\PY{o}{=}\PY{l+s+s2}{\PYZdq{}}\PY{l+s+s2}{stance\PYZus{}predicted}\PY{l+s+s2}{\PYZdq{}}\PY{p}{,}
                                                            \PY{n}{col\PYZus{}name\PYZus{}text\PYZus{}id}\PY{o}{=}\PY{n}{par}\PY{o}{.}\PY{n}{TEXT\PYZus{}ID}\PY{p}{,}
                                                            \PY{n}{mode\PYZus{}output}\PY{o}{=}\PY{n}{mode\PYZus{}output}\PY{p}{)}
    \PY{k}{def} \PY{n+nf}{estimate\PYZus{}cost}\PY{p}{(}\PY{n+nb+bp}{self}\PY{p}{)}\PY{p}{:}
        \PY{n}{total\PYZus{}cost\PYZus{}estimate} \PY{o}{=} \PY{n+nb+bp}{self}\PY{o}{.}\PY{n}{llm\PYZus{}label\PYZus{}predictor}\PY{o}{.}\PY{n}{estimate\PYZus{}total\PYZus{}cost}\PY{p}{(}\PY{n+nb+bp}{self}\PY{o}{.}\PY{n}{df\PYZus{}input\PYZus{}text}\PY{p}{)}
        \PY{n+nb}{print}\PY{p}{(}\PY{l+s+s2}{\PYZdq{}}\PY{l+s+s2}{Estimated total cost: \PYZdl{}}\PY{l+s+si}{\PYZob{}\PYZcb{}}\PY{l+s+s2}{\PYZdq{}}\PY{o}{.}\PY{n}{format}\PY{p}{(}\PY{n}{total\PYZus{}cost\PYZus{}estimate}\PY{p}{)}\PY{p}{)}
    \PY{k}{def} \PY{n+nf}{predict\PYZus{}labels}\PY{p}{(}\PY{n+nb+bp}{self}\PY{p}{)}\PY{p}{:}
        \PY{c+c1}{\PYZsh{} prompt the LLM to make predictions (write the predictions to `file\PYZus{}output\PYZus{}predictions`)}
        \PY{n+nb+bp}{self}\PY{o}{.}\PY{n}{llm\PYZus{}label\PYZus{}predictor}\PY{o}{.}\PY{n}{predict\PYZus{}labels}\PY{p}{(}\PY{n+nb+bp}{self}\PY{o}{.}\PY{n}{df\PYZus{}input\PYZus{}text}\PY{p}{,}
                                           \PY{n+nb+bp}{self}\PY{o}{.}\PY{n}{file\PYZus{}output\PYZus{}predictions}\PY{p}{,}
                                           \PY{n}{keep\PYZus{}tweet\PYZus{}id}\PY{o}{=}\PY{k+kc}{True}\PY{p}{,} \PY{n}{keep\PYZus{}text}\PY{o}{=}\PY{k+kc}{False}\PY{p}{,}
                                           \PY{n}{output\PYZus{}prob\PYZus{}mode}\PY{o}{=}\PY{k+kc}{None}\PY{p}{,}
                                           \PY{n}{list\PYZus{}label\PYZus{}space}\PY{o}{=}\PY{k+kc}{None}\PY{p}{,}
                                           \PY{n}{col\PYZus{}name\PYZus{}tweet\PYZus{}id}\PY{o}{=}\PY{n}{par}\PY{o}{.}\PY{n}{TEXT\PYZus{}ID}\PY{p}{)}
    \PY{k}{def} \PY{n+nf}{load\PYZus{}predicted\PYZus{}labels}\PY{p}{(}\PY{n+nb+bp}{self}\PY{p}{)}\PY{p}{:}
        \PY{c+c1}{\PYZsh{} load the predictions}
        \PY{n}{df\PYZus{}predictions} \PY{o}{=} \PY{n}{pd}\PY{o}{.}\PY{n}{read\PYZus{}csv}\PY{p}{(}\PY{n+nb+bp}{self}\PY{o}{.}\PY{n}{file\PYZus{}output\PYZus{}predictions}\PY{p}{)}
        \PY{k}{return} \PY{n}{df\PYZus{}predictions}
\end{Verbatim}
\end{tcolorbox}

    \begin{tcolorbox}[breakable, size=fbox, boxrule=1pt, pad at break*=1mm,colback=cellbackground, colframe=cellborder]
\prompt{In}{incolor}{32}{\boxspacing}
\begin{Verbatim}[commandchars=\\\{\}]
\PY{n}{prompt\PYZus{}chat\PYZus{}gpt\PYZus{}zero\PYZus{}shot} \PY{o}{=} \PY{n}{PromptChatGPT\PYZus{}3\PYZus{}5}\PY{p}{(}\PY{l+s+s2}{\PYZdq{}}\PY{l+s+s2}{zero\PYZus{}shot}\PY{l+s+s2}{\PYZdq{}}\PY{p}{)}
\end{Verbatim}
\end{tcolorbox}

    \begin{tcolorbox}[breakable, size=fbox, boxrule=1pt, pad at break*=1mm,colback=cellbackground, colframe=cellborder]
\prompt{In}{incolor}{33}{\boxspacing}
\begin{Verbatim}[commandchars=\\\{\}]
\PY{n}{prompt\PYZus{}chat\PYZus{}gpt\PYZus{}zero\PYZus{}shot}\PY{o}{.}\PY{n}{estimate\PYZus{}cost}\PY{p}{(}\PY{p}{)}
\end{Verbatim}
\end{tcolorbox}

    \begin{Verbatim}[commandchars=\\\{\}]
Estimated total cost: \$0.19196200000000002
    \end{Verbatim}

    \begin{tcolorbox}[breakable, size=fbox, boxrule=1pt, pad at break*=1mm,colback=cellbackground, colframe=cellborder]
\prompt{In}{incolor}{34}{\boxspacing}
\begin{Verbatim}[commandchars=\\\{\}]
\PY{k}{if} \PY{n}{PROMPT\PYZus{}CHAT\PYZus{}GPT}\PY{p}{:}
  \PY{n}{prompt\PYZus{}chat\PYZus{}gpt\PYZus{}zero\PYZus{}shot}\PY{o}{.}\PY{n}{predict\PYZus{}labels}\PY{p}{(}\PY{p}{)}
\end{Verbatim}
\end{tcolorbox}

    \begin{tcolorbox}[breakable, size=fbox, boxrule=1pt, pad at break*=1mm,colback=cellbackground, colframe=cellborder]
\prompt{In}{incolor}{35}{\boxspacing}
\begin{Verbatim}[commandchars=\\\{\}]
\PY{n}{prompt\PYZus{}chat\PYZus{}gpt\PYZus{}few\PYZus{}shot} \PY{o}{=} \PY{n}{PromptChatGPT\PYZus{}3\PYZus{}5}\PY{p}{(}\PY{l+s+s2}{\PYZdq{}}\PY{l+s+s2}{few\PYZus{}shot}\PY{l+s+s2}{\PYZdq{}}\PY{p}{)}
\PY{n}{prompt\PYZus{}chat\PYZus{}gpt\PYZus{}few\PYZus{}shot}\PY{o}{.}\PY{n}{estimate\PYZus{}cost}\PY{p}{(}\PY{p}{)}
\PY{k}{if} \PY{n}{PROMPT\PYZus{}CHAT\PYZus{}GPT}\PY{p}{:}
  \PY{n}{prompt\PYZus{}chat\PYZus{}gpt\PYZus{}few\PYZus{}shot}\PY{o}{.}\PY{n}{predict\PYZus{}labels}\PY{p}{(}\PY{p}{)}
\end{Verbatim}
\end{tcolorbox}

    \begin{Verbatim}[commandchars=\\\{\}]
Estimated total cost: \$0.3164880000000003
    \end{Verbatim}

    \begin{tcolorbox}[breakable, size=fbox, boxrule=1pt, pad at break*=1mm,colback=cellbackground, colframe=cellborder]
\prompt{In}{incolor}{36}{\boxspacing}
\begin{Verbatim}[commandchars=\\\{\}]
\PY{n}{prompt\PYZus{}chat\PYZus{}gpt\PYZus{}cot} \PY{o}{=} \PY{n}{PromptChatGPT\PYZus{}3\PYZus{}5}\PY{p}{(}\PY{l+s+s2}{\PYZdq{}}\PY{l+s+s2}{CoT}\PY{l+s+s2}{\PYZdq{}}\PY{p}{)}
\PY{n}{prompt\PYZus{}chat\PYZus{}gpt\PYZus{}cot}\PY{o}{.}\PY{n}{estimate\PYZus{}cost}\PY{p}{(}\PY{p}{)}
\PY{k}{if} \PY{n}{PROMPT\PYZus{}CHAT\PYZus{}GPT}\PY{p}{:}
  \PY{n}{prompt\PYZus{}chat\PYZus{}gpt\PYZus{}cot}\PY{o}{.}\PY{n}{predict\PYZus{}labels}\PY{p}{(}\PY{p}{)}
\end{Verbatim}
\end{tcolorbox}

    \begin{Verbatim}[commandchars=\\\{\}]
Estimated total cost: \$0.3324100000000002
    \end{Verbatim}

    \hypertarget{view-the-predictions}{%
\subsection{View the predictions}\label{view-the-predictions}}

    \begin{tcolorbox}[breakable, size=fbox, boxrule=1pt, pad at break*=1mm,colback=cellbackground, colframe=cellborder]
\prompt{In}{incolor}{37}{\boxspacing}
\begin{Verbatim}[commandchars=\\\{\}]
\PY{n}{TEXT\PYZus{}ID\PYZus{}EXAMPLE} \PY{o}{=} \PY{l+m+mi}{9}
\end{Verbatim}
\end{tcolorbox}

    \begin{tcolorbox}[breakable, size=fbox, boxrule=1pt, pad at break*=1mm,colback=cellbackground, colframe=cellborder]
\prompt{In}{incolor}{38}{\boxspacing}
\begin{Verbatim}[commandchars=\\\{\}]
\PY{n+nb}{print}\PY{p}{(}\PY{l+s+s2}{\PYZdq{}}\PY{l+s+s2}{prompt:}\PY{l+s+se}{\PYZbs{}n}\PY{l+s+s2}{ }\PY{l+s+si}{\PYZob{}\PYZcb{}}\PY{l+s+se}{\PYZbs{}n}\PY{l+s+s2}{\PYZdq{}}\PY{o}{.}\PY{n}{format}\PY{p}{(}\PY{n}{prompt\PYZus{}chat\PYZus{}gpt\PYZus{}zero\PYZus{}shot}\PY{o}{.}\PY{n}{load\PYZus{}predicted\PYZus{}labels}\PY{p}{(}\PY{p}{)}\PY{o}{.}\PY{n}{loc}\PY{p}{[}\PY{n}{TEXT\PYZus{}ID\PYZus{}EXAMPLE}\PY{p}{,}\PY{l+s+s2}{\PYZdq{}}\PY{l+s+s2}{tweet\PYZus{}embedded}\PY{l+s+s2}{\PYZdq{}}\PY{p}{]}\PY{p}{)}\PY{p}{)}
\PY{n+nb}{print}\PY{p}{(}\PY{l+s+s2}{\PYZdq{}}\PY{l+s+s2}{GPT response:}\PY{l+s+se}{\PYZbs{}n}\PY{l+s+s2}{ }\PY{l+s+si}{\PYZob{}\PYZcb{}}\PY{l+s+se}{\PYZbs{}n}\PY{l+s+s2}{\PYZdq{}}\PY{o}{.}\PY{n}{format}\PY{p}{(}\PY{n}{prompt\PYZus{}chat\PYZus{}gpt\PYZus{}zero\PYZus{}shot}\PY{o}{.}\PY{n}{load\PYZus{}predicted\PYZus{}labels}\PY{p}{(}\PY{p}{)}\PY{o}{.}\PY{n}{loc}\PY{p}{[}\PY{n}{TEXT\PYZus{}ID\PYZus{}EXAMPLE}\PY{p}{,}\PY{l+s+s2}{\PYZdq{}}\PY{l+s+s2}{stance\PYZus{}predicted}\PY{l+s+s2}{\PYZdq{}}\PY{p}{]}\PY{p}{)}\PY{p}{)}
\end{Verbatim}
\end{tcolorbox}

    \begin{Verbatim}[commandchars=\\\{\}]
prompt:
 What is the stance of the tweet below with respect to 'Legalization of
Abortion'?  If we can infer from the tweet that the tweeter supports
'Legalization of Abortion', please label it as 'in-favor'. If we can infer from
the tweet that the tweeter is against 'Legalization of Abortion', please label
is as 'against'. If we can infer from the tweet that the tweeter has a neutral
stance towards 'Legalization of Abortion', please label it as 'neutral-or-
unclear'. If there is no clue in the tweet to reveal the stance of the tweeter
towards 'Legalization of Abortion', please also label is as 'neutral-or-
unclear'. Please use exactly one word from the following 3 categories to label
it: 'in-favor', 'against', 'neutral-or-unclear'. Here is the tweet. 'dear
religious right: i keep my uterus out of your church, so keep your church out my
uterus.' The stance of the tweet is:

GPT response:
 in-favor.

    \end{Verbatim}

    \begin{tcolorbox}[breakable, size=fbox, boxrule=1pt, pad at break*=1mm,colback=cellbackground, colframe=cellborder]
\prompt{In}{incolor}{39}{\boxspacing}
\begin{Verbatim}[commandchars=\\\{\}]
\PY{n+nb}{print}\PY{p}{(}\PY{l+s+s2}{\PYZdq{}}\PY{l+s+s2}{prompt:}\PY{l+s+se}{\PYZbs{}n}\PY{l+s+s2}{ }\PY{l+s+si}{\PYZob{}\PYZcb{}}\PY{l+s+se}{\PYZbs{}n}\PY{l+s+s2}{\PYZdq{}}\PY{o}{.}\PY{n}{format}\PY{p}{(}\PY{n}{prompt\PYZus{}chat\PYZus{}gpt\PYZus{}few\PYZus{}shot}\PY{o}{.}\PY{n}{load\PYZus{}predicted\PYZus{}labels}\PY{p}{(}\PY{p}{)}\PY{o}{.}\PY{n}{loc}\PY{p}{[}\PY{n}{TEXT\PYZus{}ID\PYZus{}EXAMPLE}\PY{p}{,}\PY{l+s+s2}{\PYZdq{}}\PY{l+s+s2}{tweet\PYZus{}embedded}\PY{l+s+s2}{\PYZdq{}}\PY{p}{]}\PY{p}{)}\PY{p}{)}
\PY{n+nb}{print}\PY{p}{(}\PY{l+s+s2}{\PYZdq{}}\PY{l+s+s2}{GPT response:}\PY{l+s+se}{\PYZbs{}n}\PY{l+s+s2}{ }\PY{l+s+si}{\PYZob{}\PYZcb{}}\PY{l+s+se}{\PYZbs{}n}\PY{l+s+s2}{\PYZdq{}}\PY{o}{.}\PY{n}{format}\PY{p}{(}\PY{n}{prompt\PYZus{}chat\PYZus{}gpt\PYZus{}few\PYZus{}shot}\PY{o}{.}\PY{n}{load\PYZus{}predicted\PYZus{}labels}\PY{p}{(}\PY{p}{)}\PY{o}{.}\PY{n}{loc}\PY{p}{[}\PY{n}{TEXT\PYZus{}ID\PYZus{}EXAMPLE}\PY{p}{,}\PY{l+s+s2}{\PYZdq{}}\PY{l+s+s2}{stance\PYZus{}predicted}\PY{l+s+s2}{\PYZdq{}}\PY{p}{]}\PY{p}{)}\PY{p}{)}
\end{Verbatim}
\end{tcolorbox}

    \begin{Verbatim}[commandchars=\\\{\}]
prompt:
 What is the stance of the tweet below with respect to 'Legalization of
Abortion'?  If we can infer from the tweet that the tweeter supports
'Legalization of Abortion', please label it as 'in-favor'. If we can infer from
the tweet that the tweeter is against 'Legalization of Abortion', please label
is as 'against'. If we can infer from the tweet that the tweeter has a neutral
stance towards 'Legalization of Abortion', please label it as 'neutral-or-
unclear'. If there is no clue in the tweet to reveal the stance of the tweeter
towards 'Legalization of Abortion', please also label is as 'neutral-or-
unclear'. Please use exactly one word from the following 3 categories to label
it: 'in-favor', 'against', 'neutral-or-unclear'.  Here are some examples of
tweets. Make sure to classify the last tweet correctly.
Q: Tweet: it's a free country. freedom includes freedom of choice.
Is this tweet in-favor, against, or neutral-or-unclear?
A: in-favor
Q: Tweet: i really don't understand how some people are pro-choice. a life is a
life no matter if it's 2 weeks old or 20 years old.
Is this tweet in-favor, against, or neutral-or-unclear?
A: against
Q: Tweet: so ready for my abortion debate
Is this tweet in-favor, against, or neutral-or-unclear?
A: neutral-or-unclear
Q: Tweet: 'dear religious right: i keep my uterus out of your church, so keep
your church out my uterus.'
Is this tweet in-favor, against, or neutral-or-unclear?
A:

GPT response:
 in-favor

    \end{Verbatim}

    \begin{tcolorbox}[breakable, size=fbox, boxrule=1pt, pad at break*=1mm,colback=cellbackground, colframe=cellborder]
\prompt{In}{incolor}{40}{\boxspacing}
\begin{Verbatim}[commandchars=\\\{\}]
\PY{n+nb}{print}\PY{p}{(}\PY{l+s+s2}{\PYZdq{}}\PY{l+s+s2}{prompt:}\PY{l+s+se}{\PYZbs{}n}\PY{l+s+s2}{ }\PY{l+s+si}{\PYZob{}\PYZcb{}}\PY{l+s+se}{\PYZbs{}n}\PY{l+s+s2}{\PYZdq{}}\PY{o}{.}\PY{n}{format}\PY{p}{(}\PY{n}{prompt\PYZus{}chat\PYZus{}gpt\PYZus{}cot}\PY{o}{.}\PY{n}{load\PYZus{}predicted\PYZus{}labels}\PY{p}{(}\PY{p}{)}\PY{o}{.}\PY{n}{loc}\PY{p}{[}\PY{n}{TEXT\PYZus{}ID\PYZus{}EXAMPLE}\PY{p}{,}\PY{l+s+s2}{\PYZdq{}}\PY{l+s+s2}{tweet\PYZus{}embedded}\PY{l+s+s2}{\PYZdq{}}\PY{p}{]}\PY{p}{)}\PY{p}{)}
\PY{n+nb}{print}\PY{p}{(}\PY{l+s+s2}{\PYZdq{}}\PY{l+s+s2}{GPT response:}\PY{l+s+se}{\PYZbs{}n}\PY{l+s+s2}{ }\PY{l+s+si}{\PYZob{}\PYZcb{}}\PY{l+s+se}{\PYZbs{}n}\PY{l+s+s2}{\PYZdq{}}\PY{o}{.}\PY{n}{format}\PY{p}{(}\PY{n}{prompt\PYZus{}chat\PYZus{}gpt\PYZus{}cot}\PY{o}{.}\PY{n}{load\PYZus{}predicted\PYZus{}labels}\PY{p}{(}\PY{p}{)}\PY{o}{.}\PY{n}{loc}\PY{p}{[}\PY{n}{TEXT\PYZus{}ID\PYZus{}EXAMPLE}\PY{p}{,}\PY{l+s+s2}{\PYZdq{}}\PY{l+s+s2}{stance\PYZus{}predicted}\PY{l+s+s2}{\PYZdq{}}\PY{p}{]}\PY{p}{)}\PY{p}{)}
\end{Verbatim}
\end{tcolorbox}

    \begin{Verbatim}[commandchars=\\\{\}]
prompt:
 What is the stance of the tweet below with respect to 'Legalization of
Abortion'?  If we can infer from the tweet that the tweeter supports
'Legalization of Abortion', please label it as 'in-favor'. If we can infer from
the tweet that the tweeter is against 'Legalization of Abortion', please label
is as 'against'. If we can infer from the tweet that the tweeter has a neutral
stance towards 'Legalization of Abortion', please label it as 'neutral-or-
unclear'. If there is no clue in the tweet to reveal the stance of the tweeter
towards 'Legalization of Abortion', please also label is as 'neutral-or-
unclear'. Here is the tweet. 'dear religious right: i keep my uterus out of your
church, so keep your church out my uterus.' What is the stance of the tweet with
respect to 'Legalization of Abortion'? Please make sure that at the end of your
response, use exactly one word from the following 3 categories to label the
stance with respect to 'Legalization of Abortion': 'in-favor', 'against',
'neutral-or-unclear’. Let's think step by step.

GPT response:
 The tweet implies that the tweeter supports the legalization of abortion and
believes that the religious right should not interfere with a woman's right to
choose. Therefore, the stance of the tweet with respect to 'Legalization of
Abortion' is 'in-favor'.

    \end{Verbatim}

    \hypertarget{evaluate-the-predictions-of-different-prompts}{%
\subsection{Evaluate the Predictions of Different
Prompts}\label{evaluate-the-predictions-of-different-prompts}}

    \begin{tcolorbox}[breakable, size=fbox, boxrule=1pt, pad at break*=1mm,colback=cellbackground, colframe=cellborder]
\prompt{In}{incolor}{41}{\boxspacing}
\begin{Verbatim}[commandchars=\\\{\}]
\PY{k+kn}{from} \PY{n+nn}{gpt\PYZus{}evaluate\PYZus{}labels} \PY{k+kn}{import} \PY{n}{SemEvalGPTLabelEvaluator}
\PY{k+kn}{from} \PY{n+nn}{result\PYZus{}summarizer} \PY{k+kn}{import} \PY{n}{ResultSummarizer}
\end{Verbatim}
\end{tcolorbox}

    \begin{tcolorbox}[breakable, size=fbox, boxrule=1pt, pad at break*=1mm,colback=cellbackground, colframe=cellborder]
\prompt{In}{incolor}{42}{\boxspacing}
\begin{Verbatim}[commandchars=\\\{\}]
\PY{k}{def} \PY{n+nf}{evaluate}\PY{p}{(}\PY{n}{model\PYZus{}type}\PY{p}{,} \PY{n}{prompt\PYZus{}version}\PY{p}{)}\PY{p}{:}
    \PY{n}{gpt\PYZus{}data\PYZus{}processor} \PY{o}{=} \PY{n}{SemEvalGPTDataProcessor}\PY{p}{(}\PY{n}{topic}\PY{o}{=}\PY{n}{TOPIC\PYZus{}OF\PYZus{}INTEREST}\PY{p}{,}
                                                \PY{n}{version\PYZus{}prompt}\PY{o}{=}\PY{n}{prompt\PYZus{}version}\PY{p}{)}
    \PY{n}{data\PYZus{}processor} \PY{o}{=} \PY{n}{SemEvalDataProcessor}\PY{p}{(}\PY{p}{)}

    \PY{c+c1}{\PYZsh{} get the data with ground\PYZhy{}truth labels and partition information (only evaluate on the vali and test set)}
    \PY{n}{file\PYZus{}ground\PYZus{}truth} \PY{o}{=} \PY{n}{data\PYZus{}processor}\PY{o}{.}\PY{n}{\PYZus{}get\PYZus{}file\PYZus{}processed\PYZus{}default}\PY{p}{(}\PY{n}{topic}\PY{o}{=}\PY{n}{TOPIC\PYZus{}OF\PYZus{}INTEREST}\PY{p}{)}
    \PY{n}{df} \PY{o}{=} \PY{n}{process\PYZus{}dataframe}\PY{p}{(}
        \PY{n}{input\PYZus{}csv}\PY{o}{=}\PY{n}{file\PYZus{}ground\PYZus{}truth}\PY{p}{,}
        \PY{n}{dataset}\PY{o}{=}\PY{n}{DATASET}\PY{p}{)}
    \PY{n}{df\PYZus{}partitions} \PY{o}{=} \PY{n}{data\PYZus{}processor}\PY{o}{.}\PY{n}{read\PYZus{}partitions}\PY{p}{(}\PY{n}{topic}\PY{o}{=}\PY{n}{TOPIC\PYZus{}OF\PYZus{}INTEREST}\PY{p}{)}
    \PY{n}{dict\PYZus{}df\PYZus{}single\PYZus{}domain} \PY{o}{=} \PY{n}{partition\PYZus{}and\PYZus{}resample\PYZus{}df}\PY{p}{(}
        \PY{n}{df}\PY{p}{,} \PY{n}{seed}\PY{o}{=}\PY{k+kc}{None}\PY{p}{,} \PY{n}{partition\PYZus{}type}\PY{o}{=}\PY{l+s+s2}{\PYZdq{}}\PY{l+s+s2}{single\PYZus{}domain}\PY{l+s+s2}{\PYZdq{}}\PY{p}{,}
        \PY{n}{read\PYZus{}partition\PYZus{}from\PYZus{}df}\PY{o}{=}\PY{k+kc}{True}\PY{p}{,}
        \PY{n}{df\PYZus{}partitions}\PY{o}{=}\PY{n}{df\PYZus{}partitions}\PY{p}{)}
    \PY{k}{del} \PY{n}{dict\PYZus{}df\PYZus{}single\PYZus{}domain}\PY{p}{[}\PY{l+s+s2}{\PYZdq{}}\PY{l+s+s2}{train\PYZus{}raw}\PY{l+s+s2}{\PYZdq{}}\PY{p}{]}


    \PY{c+c1}{\PYZsh{} set the output path}
    \PY{n}{path\PYZus{}model\PYZus{}output} \PY{o}{=}  \PY{n}{join}\PY{p}{(}\PY{n}{PATH\PYZus{}OUTPUT\PYZus{}ROOT}\PY{p}{,} \PY{n}{model\PYZus{}type}\PY{p}{,} \PY{n}{prompt\PYZus{}version}\PY{p}{)}
    \PY{n}{file\PYZus{}output\PYZus{}metrics} \PY{o}{=} \PY{n}{join}\PY{p}{(}\PY{n}{path\PYZus{}model\PYZus{}output}\PY{p}{,}\PY{l+s+s2}{\PYZdq{}}\PY{l+s+s2}{metrics.csv}\PY{l+s+s2}{\PYZdq{}}\PY{p}{)}
    \PY{n}{file\PYZus{}output\PYZus{}confusion\PYZus{}mat} \PY{o}{=} \PY{n}{join}\PY{p}{(}\PY{n}{path\PYZus{}model\PYZus{}output}\PY{p}{,}\PY{l+s+s2}{\PYZdq{}}\PY{l+s+s2}{confusion\PYZus{}matrix.csv}\PY{l+s+s2}{\PYZdq{}}\PY{p}{)}
    \PY{c+c1}{\PYZsh{} also get the predictions}
    \PY{n}{file\PYZus{}input\PYZus{}predictions} \PY{o}{=} \PY{n}{join}\PY{p}{(}\PY{n}{path\PYZus{}model\PYZus{}output}\PY{p}{,}\PY{l+s+s2}{\PYZdq{}}\PY{l+s+s2}{predictions.csv}\PY{l+s+s2}{\PYZdq{}}\PY{p}{)}

    \PY{c+c1}{\PYZsh{} evaluate the predictions}
    \PY{n}{gpt\PYZus{}label\PYZus{}evaluator} \PY{o}{=} \PY{n}{SemEvalGPTLabelEvaluator}\PY{p}{(}
        \PY{n}{file\PYZus{}input\PYZus{}predictions}\PY{o}{=}\PY{n}{file\PYZus{}input\PYZus{}predictions}\PY{p}{,}
        \PY{n}{file\PYZus{}input\PYZus{}ground\PYZus{}truth}\PY{o}{=}\PY{n}{file\PYZus{}ground\PYZus{}truth}\PY{p}{,}
        \PY{n}{topic}\PY{o}{=}\PY{n}{TOPIC\PYZus{}OF\PYZus{}INTEREST}\PY{p}{,} \PY{n}{dataset}\PY{o}{=}\PY{n}{DATASET}\PY{p}{,}
        \PY{n}{model\PYZus{}gpt}\PY{o}{=}\PY{n}{model\PYZus{}type}\PY{p}{,}
        \PY{n}{num\PYZus{}examples\PYZus{}in\PYZus{}prompt}\PY{o}{=}\PY{n}{gpt\PYZus{}data\PYZus{}processor}\PY{o}{.}\PY{n}{\PYZus{}get\PYZus{}num\PYZus{}examples\PYZus{}in\PYZus{}prompt}\PY{p}{(}\PY{p}{)}\PY{p}{,}
        \PY{n}{key\PYZus{}join}\PY{o}{=}\PY{l+s+s2}{\PYZdq{}}\PY{l+s+s2}{ID}\PY{l+s+s2}{\PYZdq{}}\PY{p}{,}
        \PY{n}{list\PYZus{}tweet\PYZus{}id\PYZus{}in\PYZus{}prompt}\PY{o}{=}\PY{n}{gpt\PYZus{}data\PYZus{}processor}\PY{o}{.}\PY{n}{\PYZus{}get\PYZus{}list\PYZus{}tweet\PYZus{}id\PYZus{}in\PYZus{}prompt}\PY{p}{(}\PY{p}{)}\PY{p}{,}
        \PY{n}{full\PYZus{}predictions}\PY{o}{=}\PY{k+kc}{False}\PY{p}{)}
    \PY{n}{gpt\PYZus{}label\PYZus{}evaluator}\PY{o}{.}\PY{n}{evaluate}\PY{p}{(}
        \PY{n}{file\PYZus{}output\PYZus{}metrics}\PY{o}{=}\PY{n}{file\PYZus{}output\PYZus{}metrics}\PY{p}{,}
        \PY{n}{file\PYZus{}output\PYZus{}confusion\PYZus{}mat}\PY{o}{=}\PY{n}{file\PYZus{}output\PYZus{}confusion\PYZus{}mat}\PY{p}{,}
        \PY{n}{dict\PYZus{}df\PYZus{}eval}\PY{o}{=}\PY{n}{dict\PYZus{}df\PYZus{}single\PYZus{}domain}\PY{p}{,}
        \PY{n}{col\PYZus{}name\PYZus{}set}\PY{o}{=}\PY{l+s+s2}{\PYZdq{}}\PY{l+s+s2}{set}\PY{l+s+s2}{\PYZdq{}}\PY{p}{)}
\end{Verbatim}
\end{tcolorbox}

    \begin{tcolorbox}[breakable, size=fbox, boxrule=1pt, pad at break*=1mm,colback=cellbackground, colframe=cellborder]
\prompt{In}{incolor}{43}{\boxspacing}
\begin{Verbatim}[commandchars=\\\{\}]
\PY{k}{def} \PY{n+nf}{summarize\PYZus{}results}\PY{p}{(}\PY{n}{list\PYZus{}model\PYZus{}type}\PY{p}{,} \PY{n}{list\PYZus{}prompt\PYZus{}version}\PY{p}{)}\PY{p}{:}
    \PY{n}{df\PYZus{}hightlight\PYZus{}metrics} \PY{o}{=} \PY{n}{pd}\PY{o}{.}\PY{n}{DataFrame}\PY{p}{(}\PY{p}{)}
    \PY{k}{for} \PY{n}{model\PYZus{}type} \PY{o+ow}{in} \PY{n}{list\PYZus{}model\PYZus{}type}\PY{p}{:}
        \PY{c+c1}{\PYZsh{} summarize the results}
        \PY{n}{result\PYZus{}summarizer} \PY{o}{=} \PY{n}{ResultSummarizer}\PY{p}{(}\PY{n}{dataset}\PY{o}{=}\PY{n}{DATASET}\PY{p}{,}
                                             \PY{n}{list\PYZus{}version\PYZus{}output}\PY{o}{=}\PY{n}{list\PYZus{}prompt\PYZus{}version}\PY{p}{,}
                                             \PY{n}{eval\PYZus{}mode}\PY{o}{=}\PY{l+s+s2}{\PYZdq{}}\PY{l+s+s2}{single\PYZus{}domain}\PY{l+s+s2}{\PYZdq{}}\PY{p}{,}
                                             \PY{n}{model\PYZus{}type}\PY{o}{=}\PY{l+s+s2}{\PYZdq{}}\PY{l+s+s2}{llm\PYZus{}}\PY{l+s+s2}{\PYZdq{}} \PY{o}{+} \PY{n}{model\PYZus{}type}\PY{p}{,}
                                             \PY{n}{task}\PY{o}{=}\PY{k+kc}{None}\PY{p}{,}
                                             \PY{n}{file\PYZus{}name\PYZus{}metrics}\PY{o}{=}\PY{l+s+s2}{\PYZdq{}}\PY{l+s+s2}{metrics.csv}\PY{l+s+s2}{\PYZdq{}}\PY{p}{,}
                                             \PY{n}{file\PYZus{}name\PYZus{}confusion\PYZus{}mat}\PY{o}{=}\PY{l+s+s2}{\PYZdq{}}\PY{l+s+s2}{confusion\PYZus{}matrix.csv}\PY{l+s+s2}{\PYZdq{}}\PY{p}{,}
                                             \PY{n}{path\PYZus{}input\PYZus{}root}\PY{o}{=}\PY{n}{join}\PY{p}{(}\PY{n}{PATH\PYZus{}OUTPUT\PYZus{}ROOT}\PY{p}{,} \PY{n}{model\PYZus{}type}\PY{p}{)}\PY{p}{,}
                                             \PY{n}{path\PYZus{}output}\PY{o}{=}\PY{n}{join}\PY{p}{(}\PY{n}{PATH\PYZus{}OUTPUT\PYZus{}ROOT}\PY{p}{,}\PY{l+s+s2}{\PYZdq{}}\PY{l+s+s2}{summary}\PY{l+s+s2}{\PYZdq{}}\PY{p}{)}\PY{p}{)}
        \PY{c+c1}{\PYZsh{} write the summary to a csv file}
        \PY{n}{df\PYZus{}hightlight\PYZus{}metrics\PYZus{}this} \PY{o}{=} \PY{n}{result\PYZus{}summarizer}\PY{o}{.}\PY{n}{write\PYZus{}hightlight\PYZus{}metrics\PYZus{}to\PYZus{}summary\PYZus{}csv}\PY{p}{(}
            \PY{n}{list\PYZus{}metrics\PYZus{}highlight}\PY{o}{=}\PY{p}{[}\PY{l+s+s1}{\PYZsq{}}\PY{l+s+s1}{f1\PYZus{}macro}\PY{l+s+s1}{\PYZsq{}}\PY{p}{,} \PY{l+s+s1}{\PYZsq{}}\PY{l+s+s1}{f1\PYZus{}NONE}\PY{l+s+s1}{\PYZsq{}}\PY{p}{,} \PY{l+s+s1}{\PYZsq{}}\PY{l+s+s1}{f1\PYZus{}FAVOR}\PY{l+s+s1}{\PYZsq{}}\PY{p}{,} \PY{l+s+s1}{\PYZsq{}}\PY{l+s+s1}{f1\PYZus{}AGAINST}\PY{l+s+s1}{\PYZsq{}}\PY{p}{]}\PY{p}{,}
            \PY{n}{list\PYZus{}sets\PYZus{}highlight}\PY{o}{=}\PY{p}{[}\PY{l+s+s1}{\PYZsq{}}\PY{l+s+s1}{vali\PYZus{}raw}\PY{l+s+s1}{\PYZsq{}}\PY{p}{,} \PY{l+s+s1}{\PYZsq{}}\PY{l+s+s1}{test\PYZus{}raw}\PY{l+s+s1}{\PYZsq{}}\PY{p}{]}\PY{p}{,}
            \PY{n}{col\PYZus{}name\PYZus{}set}\PY{o}{=}\PY{l+s+s2}{\PYZdq{}}\PY{l+s+s2}{set}\PY{l+s+s2}{\PYZdq{}}\PY{p}{)}
        \PY{c+c1}{\PYZsh{} visualize the confusion matrices and save the figures}
        \PY{n}{result\PYZus{}summarizer}\PY{o}{.}\PY{n}{visualize\PYZus{}confusion\PYZus{}metrices\PYZus{}over\PYZus{}domains\PYZus{}comb}\PY{p}{(}
            \PY{p}{[}\PY{l+s+s2}{\PYZdq{}}\PY{l+s+s2}{vali\PYZus{}raw}\PY{l+s+s2}{\PYZdq{}}\PY{p}{,} \PY{l+s+s2}{\PYZdq{}}\PY{l+s+s2}{test\PYZus{}raw}\PY{l+s+s2}{\PYZdq{}}\PY{p}{]}\PY{p}{,}
            \PY{n}{preserve\PYZus{}order\PYZus{}list\PYZus{}sets}\PY{o}{=}\PY{k+kc}{True}\PY{p}{)}

        \PY{n}{df\PYZus{}hightlight\PYZus{}metrics} \PY{o}{=} \PY{n}{pd}\PY{o}{.}\PY{n}{concat}\PY{p}{(}\PY{p}{[}\PY{n}{df\PYZus{}hightlight\PYZus{}metrics}\PY{p}{,} \PY{n}{df\PYZus{}hightlight\PYZus{}metrics\PYZus{}this}\PY{p}{]}\PY{p}{,} \PY{n}{axis}\PY{o}{=}\PY{l+m+mi}{0}\PY{p}{)}
    \PY{k}{return} \PY{n}{df\PYZus{}hightlight\PYZus{}metrics}
\end{Verbatim}
\end{tcolorbox}

    \begin{tcolorbox}[breakable, size=fbox, boxrule=1pt, pad at break*=1mm,colback=cellbackground, colframe=cellborder]
\prompt{In}{incolor}{44}{\boxspacing}
\begin{Verbatim}[commandchars=\\\{\}]
\PY{n}{evaluate}\PY{p}{(}\PY{l+s+s2}{\PYZdq{}}\PY{l+s+s2}{chatgpt\PYZus{}turbo\PYZus{}3\PYZus{}5}\PY{l+s+s2}{\PYZdq{}}\PY{p}{,} \PY{l+s+s2}{\PYZdq{}}\PY{l+s+s2}{zero\PYZus{}shot}\PY{l+s+s2}{\PYZdq{}}\PY{p}{)}
\PY{n}{evaluate}\PY{p}{(}\PY{l+s+s2}{\PYZdq{}}\PY{l+s+s2}{chatgpt\PYZus{}turbo\PYZus{}3\PYZus{}5}\PY{l+s+s2}{\PYZdq{}}\PY{p}{,} \PY{l+s+s2}{\PYZdq{}}\PY{l+s+s2}{few\PYZus{}shot}\PY{l+s+s2}{\PYZdq{}}\PY{p}{)}
\PY{n}{evaluate}\PY{p}{(}\PY{l+s+s2}{\PYZdq{}}\PY{l+s+s2}{chatgpt\PYZus{}turbo\PYZus{}3\PYZus{}5}\PY{l+s+s2}{\PYZdq{}}\PY{p}{,} \PY{l+s+s2}{\PYZdq{}}\PY{l+s+s2}{CoT}\PY{l+s+s2}{\PYZdq{}}\PY{p}{)}
\end{Verbatim}
\end{tcolorbox}

    \begin{Verbatim}[commandchars=\\\{\}]
/content/prelim\_stance\_detection/scripts/utils.py:713: FutureWarning:
load\_metric is deprecated and will be removed in the next major version of
datasets. Use 'evaluate.load' instead, from the new library Hugging Face Evaluate:
https://huggingface.co/docs/evaluate
  metric\_computer[name\_metric] = load\_metric(name\_metric)
    \end{Verbatim}

    \begin{Verbatim}[commandchars=\\\{\}]
Downloading builder script:   0\%|          | 0.00/2.32k [00:00<?, ?B/s]
    \end{Verbatim}

    \begin{Verbatim}[commandchars=\\\{\}]
Downloading builder script:   0\%|          | 0.00/1.65k [00:00<?, ?B/s]
    \end{Verbatim}

    \begin{Verbatim}[commandchars=\\\{\}]
Downloading builder script:   0\%|          | 0.00/2.52k [00:00<?, ?B/s]
    \end{Verbatim}

    \begin{Verbatim}[commandchars=\\\{\}]
Downloading builder script:   0\%|          | 0.00/2.58k [00:00<?, ?B/s]
    \end{Verbatim}

    \begin{tcolorbox}[breakable, size=fbox, boxrule=1pt, pad at break*=1mm,colback=cellbackground, colframe=cellborder]
\prompt{In}{incolor}{45}{\boxspacing}
\begin{Verbatim}[commandchars=\\\{\}]
\PY{n}{df\PYZus{}hightlight\PYZus{}metrics\PYZus{}chatgpt} \PY{o}{=} \PY{n}{summarize\PYZus{}results}\PY{p}{(}\PY{p}{[}\PY{l+s+s2}{\PYZdq{}}\PY{l+s+s2}{chatgpt\PYZus{}turbo\PYZus{}3\PYZus{}5}\PY{l+s+s2}{\PYZdq{}}\PY{p}{]}\PY{p}{,} \PY{p}{[}\PY{l+s+s2}{\PYZdq{}}\PY{l+s+s2}{zero\PYZus{}shot}\PY{l+s+s2}{\PYZdq{}}\PY{p}{,}\PY{l+s+s2}{\PYZdq{}}\PY{l+s+s2}{few\PYZus{}shot}\PY{l+s+s2}{\PYZdq{}}\PY{p}{,}\PY{l+s+s2}{\PYZdq{}}\PY{l+s+s2}{CoT}\PY{l+s+s2}{\PYZdq{}}\PY{p}{]}\PY{p}{)}

\PY{c+c1}{\PYZsh{} reorder the rows}
\PY{n}{df\PYZus{}hightlight\PYZus{}metrics\PYZus{}chatgpt}\PY{p}{[}\PY{l+s+s1}{\PYZsq{}}\PY{l+s+s1}{model\PYZus{}type}\PY{l+s+s1}{\PYZsq{}}\PY{p}{]} \PY{o}{=} \PY{n}{pd}\PY{o}{.}\PY{n}{Categorical}\PY{p}{(}\PY{n}{df\PYZus{}hightlight\PYZus{}metrics\PYZus{}chatgpt}\PY{p}{[}\PY{l+s+s1}{\PYZsq{}}\PY{l+s+s1}{model\PYZus{}type}\PY{l+s+s1}{\PYZsq{}}\PY{p}{]}\PY{p}{,} \PY{n}{categories}\PY{o}{=}\PY{p}{[}\PY{l+s+s2}{\PYZdq{}}\PY{l+s+s2}{llm\PYZus{}chatgpt\PYZus{}turbo\PYZus{}3\PYZus{}5}\PY{l+s+s2}{\PYZdq{}}\PY{p}{]}\PY{p}{,} \PY{n}{ordered}\PY{o}{=}\PY{k+kc}{True}\PY{p}{)}
\PY{n}{df\PYZus{}hightlight\PYZus{}metrics\PYZus{}chatgpt}\PY{p}{[}\PY{l+s+s1}{\PYZsq{}}\PY{l+s+s1}{version}\PY{l+s+s1}{\PYZsq{}}\PY{p}{]} \PY{o}{=} \PY{n}{pd}\PY{o}{.}\PY{n}{Categorical}\PY{p}{(}\PY{n}{df\PYZus{}hightlight\PYZus{}metrics\PYZus{}chatgpt}\PY{p}{[}\PY{l+s+s1}{\PYZsq{}}\PY{l+s+s1}{version}\PY{l+s+s1}{\PYZsq{}}\PY{p}{]}\PY{p}{,} \PY{n}{categories}\PY{o}{=}\PY{p}{[}\PY{l+s+s2}{\PYZdq{}}\PY{l+s+s2}{zero\PYZus{}shot}\PY{l+s+s2}{\PYZdq{}}\PY{p}{,}\PY{l+s+s2}{\PYZdq{}}\PY{l+s+s2}{few\PYZus{}shot}\PY{l+s+s2}{\PYZdq{}}\PY{p}{,}\PY{l+s+s2}{\PYZdq{}}\PY{l+s+s2}{CoT}\PY{l+s+s2}{\PYZdq{}}\PY{p}{]}\PY{p}{,} \PY{n}{ordered}\PY{o}{=}\PY{k+kc}{True}\PY{p}{)}

\PY{c+c1}{\PYZsh{} rename columns}
\PY{n}{df\PYZus{}hightlight\PYZus{}metrics\PYZus{}chatgpt} \PY{o}{=} \PY{n}{df\PYZus{}hightlight\PYZus{}metrics\PYZus{}chatgpt}\PY{p}{[}\PY{p}{[}\PY{l+s+s2}{\PYZdq{}}\PY{l+s+s2}{model\PYZus{}type}\PY{l+s+s2}{\PYZdq{}}\PY{p}{,}\PY{l+s+s2}{\PYZdq{}}\PY{l+s+s2}{version}\PY{l+s+s2}{\PYZdq{}}\PY{p}{,}\PY{l+s+s2}{\PYZdq{}}\PY{l+s+s2}{set}\PY{l+s+s2}{\PYZdq{}}\PY{p}{,}\PY{l+s+s2}{\PYZdq{}}\PY{l+s+s2}{f1\PYZus{}macro}\PY{l+s+s2}{\PYZdq{}}\PY{p}{,}\PY{l+s+s2}{\PYZdq{}}\PY{l+s+s2}{f1\PYZus{}NONE}\PY{l+s+s2}{\PYZdq{}}\PY{p}{,}\PY{l+s+s2}{\PYZdq{}}\PY{l+s+s2}{f1\PYZus{}FAVOR}\PY{l+s+s2}{\PYZdq{}}\PY{p}{,}\PY{l+s+s2}{\PYZdq{}}\PY{l+s+s2}{f1\PYZus{}AGAINST}\PY{l+s+s2}{\PYZdq{}}\PY{p}{]}\PY{p}{]}\PY{o}{.}\PY{n}{sort\PYZus{}values}\PY{p}{(}\PY{n}{by}\PY{o}{=}\PY{p}{[}\PY{l+s+s2}{\PYZdq{}}\PY{l+s+s2}{model\PYZus{}type}\PY{l+s+s2}{\PYZdq{}}\PY{p}{,}\PY{l+s+s2}{\PYZdq{}}\PY{l+s+s2}{version}\PY{l+s+s2}{\PYZdq{}}\PY{p}{]}\PY{p}{)}\PY{o}{.}\PY{n}{rename}\PY{p}{(}\PY{n}{columns}\PY{o}{=}\PY{p}{\PYZob{}}\PY{l+s+s2}{\PYZdq{}}\PY{l+s+s2}{version}\PY{l+s+s2}{\PYZdq{}}\PY{p}{:}\PY{l+s+s2}{\PYZdq{}}\PY{l+s+s2}{prompt\PYZus{}type}\PY{l+s+s2}{\PYZdq{}}\PY{p}{,}\PY{l+s+s2}{\PYZdq{}}\PY{l+s+s2}{set}\PY{l+s+s2}{\PYZdq{}}\PY{p}{:}\PY{l+s+s2}{\PYZdq{}}\PY{l+s+s2}{partition}\PY{l+s+s2}{\PYZdq{}}\PY{p}{\PYZcb{}}\PY{p}{)}
\end{Verbatim}
\end{tcolorbox}

    \begin{Verbatim}[commandchars=\\\{\}]
<Figure size 1500x500 with 0 Axes>
    \end{Verbatim}

    \begin{Verbatim}[commandchars=\\\{\}]
<Figure size 1500x500 with 0 Axes>
    \end{Verbatim}

    \begin{Verbatim}[commandchars=\\\{\}]
<Figure size 1500x500 with 0 Axes>
    \end{Verbatim}

    \begin{tcolorbox}[breakable, size=fbox, boxrule=1pt, pad at break*=1mm,colback=cellbackground, colframe=cellborder]
\prompt{In}{incolor}{46}{\boxspacing}
\begin{Verbatim}[commandchars=\\\{\}]
\PY{n}{df\PYZus{}hightlight\PYZus{}metrics\PYZus{}chatgpt}
\end{Verbatim}
\end{tcolorbox}

            \begin{tcolorbox}[breakable, size=fbox, boxrule=.5pt, pad at break*=1mm, opacityfill=0]
\prompt{Out}{outcolor}{46}{\boxspacing}
\begin{Verbatim}[commandchars=\\\{\}]
              model\_type prompt\_type partition  f1\_macro   f1\_NONE  f1\_FAVOR  \textbackslash{}
1  llm\_chatgpt\_turbo\_3\_5   zero\_shot  vali\_raw  0.631255  0.652632  0.666667
2  llm\_chatgpt\_turbo\_3\_5   zero\_shot  test\_raw  0.507138  0.435644  0.672269
4  llm\_chatgpt\_turbo\_3\_5    few\_shot  vali\_raw  0.777948  0.800000  0.740741
5  llm\_chatgpt\_turbo\_3\_5    few\_shot  test\_raw  0.637211  0.563380  0.676923
7  llm\_chatgpt\_turbo\_3\_5         CoT  vali\_raw  0.450072  0.512397  0.512821
8  llm\_chatgpt\_turbo\_3\_5         CoT  test\_raw  0.387721  0.360000  0.568421

   f1\_AGAINST
1    0.574468
2    0.413502
4    0.793103
5    0.671329
7    0.325000
8    0.234742
\end{Verbatim}
\end{tcolorbox}
        
    \hypertarget{view-the-performance-on-the-validation-set}{%
\subsubsection{View the performance on the validation
set}\label{view-the-performance-on-the-validation-set}}

    \begin{tcolorbox}[breakable, size=fbox, boxrule=1pt, pad at break*=1mm,colback=cellbackground, colframe=cellborder]
\prompt{In}{incolor}{47}{\boxspacing}
\begin{Verbatim}[commandchars=\\\{\}]
\PY{n}{df\PYZus{}hightlight\PYZus{}metrics\PYZus{}chatgpt\PYZus{}vali} \PY{o}{=} \PY{n}{df\PYZus{}hightlight\PYZus{}metrics\PYZus{}chatgpt}\PY{p}{[}\PY{n}{df\PYZus{}hightlight\PYZus{}metrics\PYZus{}chatgpt}\PY{p}{[}\PY{l+s+s2}{\PYZdq{}}\PY{l+s+s2}{partition}\PY{l+s+s2}{\PYZdq{}}\PY{p}{]}\PY{o}{.}\PY{n}{isin}\PY{p}{(}\PY{p}{[}\PY{l+s+s2}{\PYZdq{}}\PY{l+s+s2}{vali\PYZus{}raw}\PY{l+s+s2}{\PYZdq{}}\PY{p}{]}\PY{p}{)}\PY{p}{]}

\PY{c+c1}{\PYZsh{} convert \PYZdq{}vali\PYZus{}raw\PYZdq{} to \PYZdq{}vali\PYZdq{} (in the partition column)}
\PY{n}{df\PYZus{}hightlight\PYZus{}metrics\PYZus{}chatgpt\PYZus{}vali}\PY{o}{.}\PY{n}{loc}\PY{p}{[}\PY{n}{df\PYZus{}hightlight\PYZus{}metrics\PYZus{}chatgpt\PYZus{}vali}\PY{p}{[}\PY{l+s+s2}{\PYZdq{}}\PY{l+s+s2}{partition}\PY{l+s+s2}{\PYZdq{}}\PY{p}{]}\PY{o}{==}\PY{l+s+s2}{\PYZdq{}}\PY{l+s+s2}{vali\PYZus{}raw}\PY{l+s+s2}{\PYZdq{}}\PY{p}{,}\PY{l+s+s2}{\PYZdq{}}\PY{l+s+s2}{partition}\PY{l+s+s2}{\PYZdq{}}\PY{p}{]} \PY{o}{=} \PY{l+s+s2}{\PYZdq{}}\PY{l+s+s2}{vali}\PY{l+s+s2}{\PYZdq{}}

\PY{n}{df\PYZus{}hightlight\PYZus{}metrics\PYZus{}chatgpt\PYZus{}vali}
\end{Verbatim}
\end{tcolorbox}

            \begin{tcolorbox}[breakable, size=fbox, boxrule=.5pt, pad at break*=1mm, opacityfill=0]
\prompt{Out}{outcolor}{47}{\boxspacing}
\begin{Verbatim}[commandchars=\\\{\}]
              model\_type prompt\_type partition  f1\_macro   f1\_NONE  f1\_FAVOR  \textbackslash{}
1  llm\_chatgpt\_turbo\_3\_5   zero\_shot      vali  0.631255  0.652632  0.666667
4  llm\_chatgpt\_turbo\_3\_5    few\_shot      vali  0.777948  0.800000  0.740741
7  llm\_chatgpt\_turbo\_3\_5         CoT      vali  0.450072  0.512397  0.512821

   f1\_AGAINST
1    0.574468
4    0.793103
7    0.325000
\end{Verbatim}
\end{tcolorbox}
        
    The first two columns indicate the model type and the prompt type,
respectively. The third column indicates that the performance is
evaluated on the validation set. The \texttt{f1\_macro} column indicates
the macro-averaged F1 score (across the three stance types). We use this
value to quantify the overall performance of the model. Note that we are
using the \texttt{f1\_macro} metric rather than accuracy because the
dataset is imbalanced, and the \texttt{f1\_macro} metric is more robust
to imbalanced datasets.

    \begin{quote}
To learn more about macro-F1 score, I recommend taking a look at this
tutorial
https://towardsdatascience.com/micro-macro-weighted-averages-of-f1-score-clearly-explained-b603420b292f\#:\textasciitilde:text=The\%20macro\%2Daveraged\%20F1\%20score,regardless\%20of\%20their\%20support\%20values.
\end{quote}

    Based on the macro-F1 scores, the best performing combination is ChatGPT
using the few-shot prompt. On the other hand, the zero-shot CoT prompt
appears to negatively impact performance. In the next section, we will
delve into this further by examining the confusion matrix to better
understand these results.

The last 3 columns indicate the performance of each stance type. For
few-shot prompt, this shows that the model is better at predicting the
\texttt{AGAINST} and \texttt{FAVOR} stance than the \texttt{NONE}
stance.

\begin{quote}
Note: In practice, to avoid data leakage, when selecting the best
combination of prompt type and model type, we should use the performance
on the validation set to choose the best combination, and then use the
performance on the test set to evaluate the final model.
\end{quote}

    \hypertarget{view-the-performance-on-the-test-set}{%
\subsubsection{View the performance on the test
set}\label{view-the-performance-on-the-test-set}}

    \begin{tcolorbox}[breakable, size=fbox, boxrule=1pt, pad at break*=1mm,colback=cellbackground, colframe=cellborder]
\prompt{In}{incolor}{48}{\boxspacing}
\begin{Verbatim}[commandchars=\\\{\}]
\PY{n}{df\PYZus{}hightlight\PYZus{}metrics\PYZus{}chatgpt\PYZus{}test} \PY{o}{=} \PY{n}{df\PYZus{}hightlight\PYZus{}metrics\PYZus{}chatgpt}\PY{p}{[}\PY{n}{df\PYZus{}hightlight\PYZus{}metrics\PYZus{}chatgpt}\PY{p}{[}\PY{l+s+s2}{\PYZdq{}}\PY{l+s+s2}{partition}\PY{l+s+s2}{\PYZdq{}}\PY{p}{]}\PY{o}{.}\PY{n}{isin}\PY{p}{(}\PY{p}{[}\PY{l+s+s2}{\PYZdq{}}\PY{l+s+s2}{test\PYZus{}raw}\PY{l+s+s2}{\PYZdq{}}\PY{p}{]}\PY{p}{)}\PY{p}{]}

\PY{c+c1}{\PYZsh{} convert \PYZdq{}test\PYZus{}raw\PYZdq{} to \PYZdq{}test\PYZdq{} (in the partition column)}
\PY{n}{df\PYZus{}hightlight\PYZus{}metrics\PYZus{}chatgpt\PYZus{}test}\PY{o}{.}\PY{n}{loc}\PY{p}{[}\PY{n}{df\PYZus{}hightlight\PYZus{}metrics\PYZus{}chatgpt\PYZus{}test}\PY{p}{[}\PY{l+s+s2}{\PYZdq{}}\PY{l+s+s2}{partition}\PY{l+s+s2}{\PYZdq{}}\PY{p}{]}\PY{o}{==}\PY{l+s+s2}{\PYZdq{}}\PY{l+s+s2}{test\PYZus{}raw}\PY{l+s+s2}{\PYZdq{}}\PY{p}{,}\PY{l+s+s2}{\PYZdq{}}\PY{l+s+s2}{partition}\PY{l+s+s2}{\PYZdq{}}\PY{p}{]} \PY{o}{=} \PY{l+s+s2}{\PYZdq{}}\PY{l+s+s2}{test}\PY{l+s+s2}{\PYZdq{}}

\PY{n}{df\PYZus{}hightlight\PYZus{}metrics\PYZus{}chatgpt\PYZus{}test}
\end{Verbatim}
\end{tcolorbox}

            \begin{tcolorbox}[breakable, size=fbox, boxrule=.5pt, pad at break*=1mm, opacityfill=0]
\prompt{Out}{outcolor}{48}{\boxspacing}
\begin{Verbatim}[commandchars=\\\{\}]
              model\_type prompt\_type partition  f1\_macro   f1\_NONE  f1\_FAVOR  \textbackslash{}
2  llm\_chatgpt\_turbo\_3\_5   zero\_shot      test  0.507138  0.435644  0.672269
5  llm\_chatgpt\_turbo\_3\_5    few\_shot      test  0.637211  0.563380  0.676923
8  llm\_chatgpt\_turbo\_3\_5         CoT      test  0.387721  0.360000  0.568421

   f1\_AGAINST
2    0.413502
5    0.671329
8    0.234742
\end{Verbatim}
\end{tcolorbox}
        
    The results are similar to the validation set. The best prompt type is
\texttt{ChatGPT-turbo-3.5} with the \texttt{few-shot} prompt. The
zero-shot CoT prompt only seems to hurt the performance.

    \hypertarget{compare-chatgpt-with-bert-on-the-test-set}{%
\subsubsection{Compare ChatGPT with BERT on the test
set}\label{compare-chatgpt-with-bert-on-the-test-set}}

Let's compare the performance of FLAN-T5 with BERT. The results of BERT
are generated from the previous tutorial.

    \begin{tcolorbox}[breakable, size=fbox, boxrule=1pt, pad at break*=1mm,colback=cellbackground, colframe=cellborder]
\prompt{In}{incolor}{49}{\boxspacing}
\begin{Verbatim}[commandchars=\\\{\}]
\PY{n}{df\PYZus{}hightlight\PYZus{}metrics\PYZus{}bert} \PY{o}{=} \PY{n}{pd}\PY{o}{.}\PY{n}{read\PYZus{}csv}\PY{p}{(}\PY{n}{join}\PY{p}{(}\PY{n}{par}\PY{o}{.}\PY{n}{PATH\PYZus{}RESULT\PYZus{}SEM\PYZus{}EVAL\PYZus{}TUNING}\PY{p}{,}\PY{l+s+s2}{\PYZdq{}}\PY{l+s+s2}{summary}\PY{l+s+s2}{\PYZdq{}}\PY{p}{,}\PY{l+s+s2}{\PYZdq{}}\PY{l+s+s2}{metrics\PYZus{}highlights.csv}\PY{l+s+s2}{\PYZdq{}}\PY{p}{)}\PY{p}{)}
\PY{n}{df\PYZus{}hightlight\PYZus{}metrics\PYZus{}bert} \PY{o}{=} \PY{n}{df\PYZus{}hightlight\PYZus{}metrics\PYZus{}bert}\PY{p}{[}\PY{p}{[}\PY{l+s+s2}{\PYZdq{}}\PY{l+s+s2}{version}\PY{l+s+s2}{\PYZdq{}}\PY{p}{,}\PY{l+s+s2}{\PYZdq{}}\PY{l+s+s2}{set}\PY{l+s+s2}{\PYZdq{}}\PY{p}{,}\PY{l+s+s2}{\PYZdq{}}\PY{l+s+s2}{f1\PYZus{}macro}\PY{l+s+s2}{\PYZdq{}}\PY{p}{,}\PY{l+s+s2}{\PYZdq{}}\PY{l+s+s2}{f1\PYZus{}NONE}\PY{l+s+s2}{\PYZdq{}}\PY{p}{,}\PY{l+s+s2}{\PYZdq{}}\PY{l+s+s2}{f1\PYZus{}FAVOR}\PY{l+s+s2}{\PYZdq{}}\PY{p}{,}\PY{l+s+s2}{\PYZdq{}}\PY{l+s+s2}{f1\PYZus{}AGAINST}\PY{l+s+s2}{\PYZdq{}}\PY{p}{]}\PY{p}{]}\PY{p}{[}\PY{n}{df\PYZus{}hightlight\PYZus{}metrics\PYZus{}bert}\PY{o}{.}\PY{n}{set}\PY{o}{.}\PY{n}{isin}\PY{p}{(}\PY{p}{[}\PY{l+s+s2}{\PYZdq{}}\PY{l+s+s2}{test\PYZus{}raw}\PY{l+s+s2}{\PYZdq{}}\PY{p}{]}\PY{p}{)}\PY{p}{]}\PY{o}{.}\PY{n}{rename}\PY{p}{(}\PY{n}{columns}\PY{o}{=}\PY{p}{\PYZob{}}\PY{l+s+s2}{\PYZdq{}}\PY{l+s+s2}{version}\PY{l+s+s2}{\PYZdq{}}\PY{p}{:}\PY{l+s+s2}{\PYZdq{}}\PY{l+s+s2}{model\PYZus{}type}\PY{l+s+s2}{\PYZdq{}}\PY{p}{\PYZcb{}}\PY{p}{)}
\PY{c+c1}{\PYZsh{} reorder the rows}
\PY{n}{df\PYZus{}hightlight\PYZus{}metrics\PYZus{}bert}\PY{p}{[}\PY{l+s+s1}{\PYZsq{}}\PY{l+s+s1}{model\PYZus{}type}\PY{l+s+s1}{\PYZsq{}}\PY{p}{]} \PY{o}{=} \PY{n}{pd}\PY{o}{.}\PY{n}{Categorical}\PY{p}{(}\PY{n}{df\PYZus{}hightlight\PYZus{}metrics\PYZus{}bert}\PY{p}{[}\PY{l+s+s1}{\PYZsq{}}\PY{l+s+s1}{model\PYZus{}type}\PY{l+s+s1}{\PYZsq{}}\PY{p}{]}\PY{p}{,} \PY{n}{categories}\PY{o}{=}\PY{p}{[}\PY{l+s+s2}{\PYZdq{}}\PY{l+s+s2}{bert\PYZhy{}base\PYZhy{}uncased}\PY{l+s+s2}{\PYZdq{}}\PY{p}{,}\PY{l+s+s2}{\PYZdq{}}\PY{l+s+s2}{vinai\PYZus{}bertweet\PYZus{}base}\PY{l+s+s2}{\PYZdq{}}\PY{p}{,}\PY{l+s+s2}{\PYZdq{}}\PY{l+s+s2}{kornosk\PYZus{}polibertweet\PYZus{}mlm}\PY{l+s+s2}{\PYZdq{}}\PY{p}{]}\PY{p}{,} \PY{n}{ordered}\PY{o}{=}\PY{k+kc}{True}\PY{p}{)}
\PY{n}{df\PYZus{}hightlight\PYZus{}metrics\PYZus{}bert} \PY{o}{=} \PY{n}{df\PYZus{}hightlight\PYZus{}metrics\PYZus{}bert}\PY{o}{.}\PY{n}{sort\PYZus{}values}\PY{p}{(}\PY{n}{by}\PY{o}{=}\PY{p}{[}\PY{l+s+s2}{\PYZdq{}}\PY{l+s+s2}{model\PYZus{}type}\PY{l+s+s2}{\PYZdq{}}\PY{p}{]}\PY{p}{)}\PY{o}{.}\PY{n}{reset\PYZus{}index}\PY{p}{(}\PY{n}{drop}\PY{o}{=}\PY{k+kc}{True}\PY{p}{)}
\PY{c+c1}{\PYZsh{} rename columns}
\PY{n}{df\PYZus{}hightlight\PYZus{}metrics\PYZus{}bert} \PY{o}{=} \PY{n}{df\PYZus{}hightlight\PYZus{}metrics\PYZus{}bert}\PY{o}{.}\PY{n}{rename}\PY{p}{(}\PY{n}{columns}\PY{o}{=}\PY{p}{\PYZob{}}\PY{l+s+s2}{\PYZdq{}}\PY{l+s+s2}{set}\PY{l+s+s2}{\PYZdq{}}\PY{p}{:}\PY{l+s+s2}{\PYZdq{}}\PY{l+s+s2}{partition}\PY{l+s+s2}{\PYZdq{}}\PY{p}{\PYZcb{}}\PY{p}{)}
\PY{c+c1}{\PYZsh{} convert \PYZdq{}test\PYZus{}raw\PYZdq{} to \PYZdq{}test\PYZdq{} (in the partition column)}
\PY{n}{df\PYZus{}hightlight\PYZus{}metrics\PYZus{}bert}\PY{o}{.}\PY{n}{loc}\PY{p}{[}\PY{n}{df\PYZus{}hightlight\PYZus{}metrics\PYZus{}bert}\PY{p}{[}\PY{l+s+s2}{\PYZdq{}}\PY{l+s+s2}{partition}\PY{l+s+s2}{\PYZdq{}}\PY{p}{]}\PY{o}{==}\PY{l+s+s2}{\PYZdq{}}\PY{l+s+s2}{test\PYZus{}raw}\PY{l+s+s2}{\PYZdq{}}\PY{p}{,}\PY{l+s+s2}{\PYZdq{}}\PY{l+s+s2}{partition}\PY{l+s+s2}{\PYZdq{}}\PY{p}{]} \PY{o}{=} \PY{l+s+s2}{\PYZdq{}}\PY{l+s+s2}{test}\PY{l+s+s2}{\PYZdq{}}

\PY{n}{df\PYZus{}hightlight\PYZus{}metrics\PYZus{}bert}
\end{Verbatim}
\end{tcolorbox}

            \begin{tcolorbox}[breakable, size=fbox, boxrule=.5pt, pad at break*=1mm, opacityfill=0]
\prompt{Out}{outcolor}{49}{\boxspacing}
\begin{Verbatim}[commandchars=\\\{\}]
                 model\_type partition  f1\_macro  f1\_NONE  f1\_FAVOR  f1\_AGAINST
0         bert-base-uncased      test    0.4748   0.4196    0.4275      0.5775
1       vinai\_bertweet\_base      test    0.5797   0.5323    0.5401      0.6667
2  kornosk\_polibertweet\_mlm      test    0.5616   0.5440    0.4762      0.6645
\end{Verbatim}
\end{tcolorbox}
        
    Based on the macro-F1 scores, the ChatGPT using the zero-shot prompt
outperforms all the BERT variants we examined in the previous tutorial.

    \hypertarget{analyzing-the-confusion-matrix-for-deeper-insights}{%
\subsection{Analyzing the Confusion Matrix for Deeper
Insights}\label{analyzing-the-confusion-matrix-for-deeper-insights}}

    Interestingly, the few-shot prompt emerges as the most effective prompt
type for ChatGPT, while the CoT prompt performs poorly. To gain a deeper
understanding of this phenomenon, let's examine the confusion matrix.

    \begin{quote}
Note: Examining the confusion matrix is essential because it provides a
detailed overview of the model's performance across different classes.
It reveals not only the correct predictions (true positives) but also
the instances where the model made errors (false positives and false
negatives). By analyzing the confusion matrix, we can identify patterns
in misclassifications and gain insights into the strengths and
weaknesses of the model. Here is a great tutorial on how to interpret
the confusion matrix and its relationships with macro-F1 scores:
https://towardsdatascience.com/confusion-matrix-for-your-multi-class-machine-learning-model-ff9aa3bf7826
\end{quote}

    \begin{tcolorbox}[breakable, size=fbox, boxrule=1pt, pad at break*=1mm,colback=cellbackground, colframe=cellborder]
\prompt{In}{incolor}{50}{\boxspacing}
\begin{Verbatim}[commandchars=\\\{\}]
\PY{k+kn}{from} \PY{n+nn}{IPython}\PY{n+nn}{.}\PY{n+nn}{display} \PY{k+kn}{import} \PY{n}{Image}\PY{p}{,} \PY{n}{display}
\end{Verbatim}
\end{tcolorbox}

    Below are the confusion matrices for the few-shot prompt.

The first row are the matrices for the validation set, and the second
row are the matrices for the test set.

Each row of matrices consists of three types:

\begin{enumerate}
\def\labelenumi{\arabic{enumi}.}
\tightlist
\item
  The leftmost matrices are the raw confusion matrices.
\item
  The middle matrices show the confusion matrices normalized by row
  (i.e., the sum of each row equals 100). In these matrices, the
  diagonal values correspond to the recall value of each class.
\item
  The rightmost matrices illustrate the confusion matrices normalized by
  column (i.e., the sum of each column equals 100). In these matrices,
  the diagonal values are the precision value for each class.
\end{enumerate}

In each matrix, the rows represent the true labels, and the columns
represent the predicted labels. The diagonal elements denote correct
predictions, while the off-diagonal elements indicate incorrect
predictions.

    \begin{tcolorbox}[breakable, size=fbox, boxrule=1pt, pad at break*=1mm,colback=cellbackground, colframe=cellborder]
\prompt{In}{incolor}{51}{\boxspacing}
\begin{Verbatim}[commandchars=\\\{\}]
\PY{n}{display\PYZus{}resized\PYZus{}image\PYZus{}in\PYZus{}notebook}\PY{p}{(}\PY{n}{join}\PY{p}{(}\PY{n}{PATH\PYZus{}OUTPUT\PYZus{}ROOT}\PY{p}{,}\PY{l+s+s2}{\PYZdq{}}\PY{l+s+s2}{summary}\PY{l+s+s2}{\PYZdq{}}\PY{p}{,}\PY{l+s+s2}{\PYZdq{}}\PY{l+s+s2}{chatgpt\PYZus{}turbo\PYZus{}3\PYZus{}5\PYZus{}few\PYZus{}shot\PYZus{}comb\PYZus{}confusion\PYZus{}mat.png}\PY{l+s+s2}{\PYZdq{}}\PY{p}{)}\PY{p}{)}
\end{Verbatim}
\end{tcolorbox}

    \begin{center}
    \adjustimage{max size={0.9\linewidth}{0.9\paperheight}}{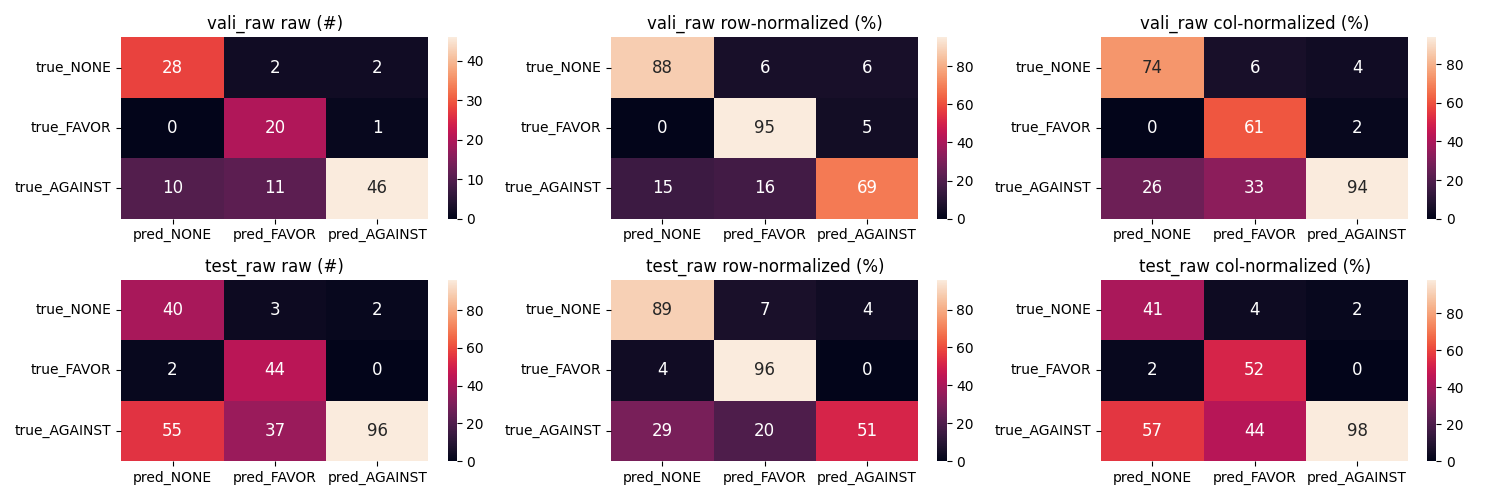}
    \end{center}
    { \hspace*{\fill} \\}
    
    As shown in the test set confusion matrices, the model demonstrates
proficiency in distinguishing between the \texttt{FAVOR} and
\texttt{NONE} stances. However, it faces challenges in accurately
predicting the \texttt{AGAINST} stance, frequently misclassifying them
as \texttt{NONE}.

    Now, let's examine the confusion matrix for the CoT prompt, which has
the worst macro-F1 score.

The confusion matrices below reveal that the model has a strong tendency
to predict the \texttt{NONE} stance, which is defined as the stance
being neutral or unclear. It seems like after reasoning step by step,
the model tends to reach a conclusion that the stance is not clear.

    \begin{tcolorbox}[breakable, size=fbox, boxrule=1pt, pad at break*=1mm,colback=cellbackground, colframe=cellborder]
\prompt{In}{incolor}{52}{\boxspacing}
\begin{Verbatim}[commandchars=\\\{\}]
\PY{n}{display\PYZus{}resized\PYZus{}image\PYZus{}in\PYZus{}notebook}\PY{p}{(}\PY{n}{join}\PY{p}{(}\PY{n}{PATH\PYZus{}OUTPUT\PYZus{}ROOT}\PY{p}{,}\PY{l+s+s2}{\PYZdq{}}\PY{l+s+s2}{summary}\PY{l+s+s2}{\PYZdq{}}\PY{p}{,}\PY{l+s+s2}{\PYZdq{}}\PY{l+s+s2}{chatgpt\PYZus{}turbo\PYZus{}3\PYZus{}5\PYZus{}CoT\PYZus{}comb\PYZus{}confusion\PYZus{}mat.png}\PY{l+s+s2}{\PYZdq{}}\PY{p}{)}\PY{p}{)}
\end{Verbatim}
\end{tcolorbox}

    \begin{center}
    \adjustimage{max size={0.9\linewidth}{0.9\paperheight}}{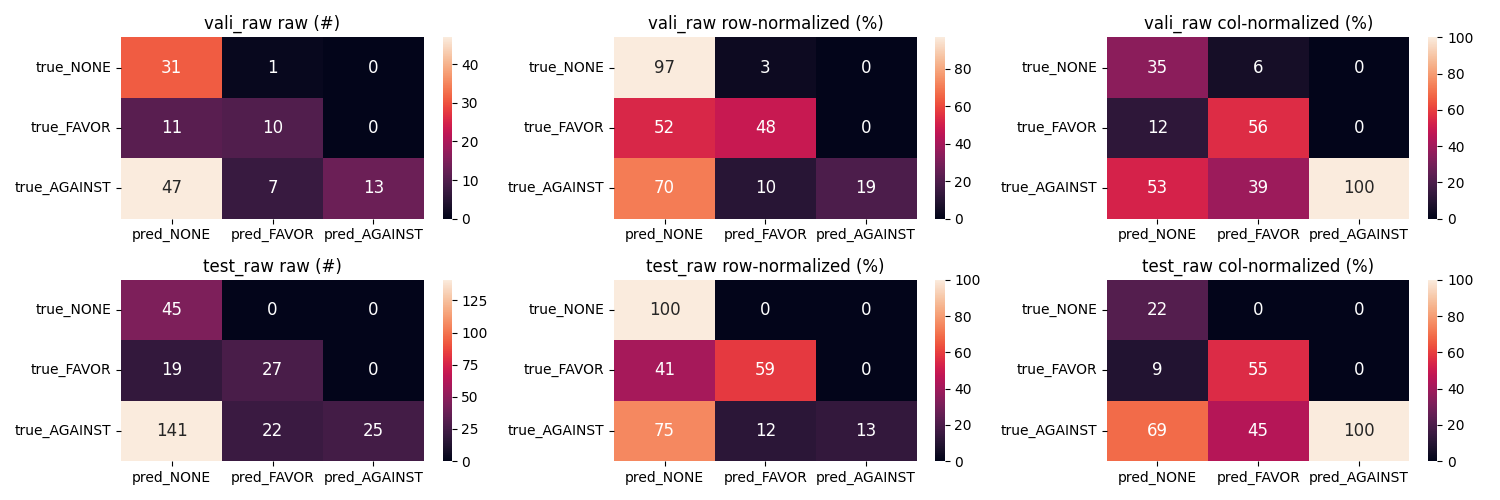}
    \end{center}
    { \hspace*{\fill} \\}
    
    \begin{center}\rule{0.5\linewidth}{0.5pt}\end{center}

    \hypertarget{feed-the-prompts-to-flan-t5-models}{%
\section{Feed the prompts to FLAN-T5
models}\label{feed-the-prompts-to-flan-t5-models}}

    Great! Now we know that ChatGPT with few-shot prompts outperforms
fine-tuned BERT on this dataset. This is promising because prompting the
LLM doesn't require a large amount of labeled data for training.
However, one notable downside is that using ChatGPT incurs a monetary
cost.

Next, let's explore if the open-source FLAN-T5 can achieve similar
results.

    First, you have to decide whether you want to prompt FLAN-T5 on your
own.

I recommend keeping \texttt{PROMPT\_FLAN\_T5\ =\ False} (the default
setting below) if you are running this notebook for the first time. This
will read the predictions I made and uploaded to my GitHub repo, which
will save you time.

If you want to try it out on your own, you can set
\texttt{PROMPT\_CHAT\_GPT\ =\ True} and run the code below. If you are
running this on Google Colab, make sure you have use the GPU run time.
To do this, go to \texttt{Runtime} -\textgreater{}
\texttt{Change\ runtime\ type} -\textgreater{}
\texttt{Hardware\ accelerator} -\textgreater{} \texttt{GPU}. This will
ensure that the note will run more efficiently and quickly.

    \begin{tcolorbox}[breakable, size=fbox, boxrule=1pt, pad at break*=1mm,colback=cellbackground, colframe=cellborder]
\prompt{In}{incolor}{53}{\boxspacing}
\begin{Verbatim}[commandchars=\\\{\}]
\PY{n}{PROMPT\PYZus{}FLAN\PYZus{}T5} \PY{o}{=} \PY{k+kc}{False}
\end{Verbatim}
\end{tcolorbox}

    \begin{tcolorbox}[breakable, size=fbox, boxrule=1pt, pad at break*=1mm,colback=cellbackground, colframe=cellborder]
\prompt{In}{incolor}{54}{\boxspacing}
\begin{Verbatim}[commandchars=\\\{\}]
\PY{o}{\PYZpc{}}\PY{k}{load\PYZus{}ext} autoreload
\PY{o}{\PYZpc{}}\PY{k}{autoreload} 2
\PY{k+kn}{from} \PY{n+nn}{gpt\PYZus{}predict\PYZus{}label} \PY{k+kn}{import} \PY{n}{FlanT5LargeLabelPredictor}\PY{p}{,} \PY{n}{FlanT5XxlLabelPredictor}
\end{Verbatim}
\end{tcolorbox}

    \begin{Verbatim}[commandchars=\\\{\}]
The autoreload extension is already loaded. To reload it, use:
  \%reload\_ext autoreload
    \end{Verbatim}

    \hypertarget{helpfer-function}{%
\subsection{Helpfer function}\label{helpfer-function}}

    \begin{tcolorbox}[breakable, size=fbox, boxrule=1pt, pad at break*=1mm,colback=cellbackground, colframe=cellborder]
\prompt{In}{incolor}{55}{\boxspacing}
\begin{Verbatim}[commandchars=\\\{\}]
\PY{k}{def} \PY{n+nf}{prompt\PYZus{}flan\PYZus{}t5}\PY{p}{(}\PY{n}{model\PYZus{}type}\PY{p}{,} \PY{n}{prompt\PYZus{}version}\PY{p}{)}\PY{p}{:}
    \PY{c+c1}{\PYZsh{} get the output paths}
    \PY{c+c1}{\PYZsh{} model\PYZus{}type\PYZus{}name = tidy\PYZus{}name(model\PYZus{}type)}
    \PY{n}{path\PYZus{}output} \PY{o}{=} \PY{n}{join}\PY{p}{(}\PY{n}{PATH\PYZus{}OUTPUT\PYZus{}ROOT}\PY{p}{,} \PY{n}{model\PYZus{}type}\PY{p}{,} \PY{n}{prompt\PYZus{}version}\PY{p}{)}
    \PY{n}{file\PYZus{}output\PYZus{}predictions} \PY{o}{=} \PY{n}{join}\PY{p}{(}\PY{n}{path\PYZus{}output}\PY{p}{,} \PY{l+s+s2}{\PYZdq{}}\PY{l+s+s2}{predictions.csv}\PY{l+s+s2}{\PYZdq{}}\PY{p}{)}

    \PY{n}{gpt\PYZus{}data\PYZus{}processor} \PY{o}{=} \PY{n}{SemEvalGPTDataProcessor}\PY{p}{(}\PY{n}{version\PYZus{}prompt}\PY{o}{=}\PY{n}{prompt\PYZus{}version}\PY{p}{,} \PY{n}{topic}\PY{o}{=}\PY{n}{TOPIC\PYZus{}OF\PYZus{}INTEREST}\PY{p}{)}

    \PY{c+c1}{\PYZsh{} create the prompt, which is the input to the LLMs}
    \PY{n}{df\PYZus{}input\PYZus{}text} \PY{o}{=} \PY{n}{gpt\PYZus{}data\PYZus{}processor}\PY{o}{.}\PY{n}{embed\PYZus{}prompt}\PY{p}{(}\PY{n}{file\PYZus{}output}\PY{o}{=}\PY{k+kc}{None}\PY{p}{,} \PY{n}{write\PYZus{}csv}\PY{o}{=}\PY{k+kc}{False}\PY{p}{,}
                                                    \PY{n}{return\PYZus{}df}\PY{o}{=}\PY{k+kc}{True}\PY{p}{)}

    \PY{c+c1}{\PYZsh{} partition the data into train, vali, test (note that we only need the vali and test set in this approach)}
    \PY{n}{dict\PYZus{}df\PYZus{}single\PYZus{}domain} \PY{o}{=} \PY{n}{partition\PYZus{}and\PYZus{}resample\PYZus{}df}\PY{p}{(}
        \PY{n}{df\PYZus{}input\PYZus{}text}\PY{p}{,} \PY{n}{seed}\PY{o}{=}\PY{k+kc}{None}\PY{p}{,} \PY{n}{partition\PYZus{}type}\PY{o}{=}\PY{l+s+s2}{\PYZdq{}}\PY{l+s+s2}{single\PYZus{}domain}\PY{l+s+s2}{\PYZdq{}}\PY{p}{,}
        \PY{n}{read\PYZus{}partition\PYZus{}from\PYZus{}df}\PY{o}{=}\PY{k+kc}{True}\PY{p}{,}
        \PY{n}{df\PYZus{}partitions}\PY{o}{=}\PY{n}{df\PYZus{}partitions}\PY{p}{)}

    \PY{c+c1}{\PYZsh{} select the partition to be labeled}
    \PY{c+c1}{\PYZsh{} \PYZhy{} vali and test}
    \PY{n}{df\PYZus{}input\PYZus{}text\PYZus{}filtered} \PY{o}{=} \PY{n}{pd}\PY{o}{.}\PY{n}{DataFrame}\PY{p}{(}\PY{p}{)}
    \PY{k}{for} \PY{n}{partition} \PY{o+ow}{in} \PY{p}{[}\PY{l+s+s2}{\PYZdq{}}\PY{l+s+s2}{vali\PYZus{}raw}\PY{l+s+s2}{\PYZdq{}}\PY{p}{,} \PY{l+s+s2}{\PYZdq{}}\PY{l+s+s2}{test\PYZus{}raw}\PY{l+s+s2}{\PYZdq{}}\PY{p}{]}\PY{p}{:}
        \PY{n}{df\PYZus{}input\PYZus{}text\PYZus{}filtered} \PY{o}{=} \PY{n}{pd}\PY{o}{.}\PY{n}{concat}\PY{p}{(}\PY{p}{[}\PY{n}{df\PYZus{}input\PYZus{}text\PYZus{}filtered}\PY{p}{,} \PY{n}{dict\PYZus{}df\PYZus{}single\PYZus{}domain}\PY{p}{[}\PY{n}{partition}\PY{p}{]}\PY{p}{]}\PY{p}{)}

    \PY{n}{df\PYZus{}input\PYZus{}text} \PY{o}{=} \PY{n}{df\PYZus{}input\PYZus{}text\PYZus{}filtered}

    \PY{c+c1}{\PYZsh{} specify the output type (single\PYZhy{}word or multi\PYZhy{}word)}
    \PY{k}{if} \PY{n}{prompt\PYZus{}version} \PY{o+ow}{in} \PY{p}{[}\PY{l+s+s2}{\PYZdq{}}\PY{l+s+s2}{zero\PYZus{}shot}\PY{l+s+s2}{\PYZdq{}}\PY{p}{,}\PY{l+s+s2}{\PYZdq{}}\PY{l+s+s2}{few\PYZus{}shot}\PY{l+s+s2}{\PYZdq{}}\PY{p}{]}\PY{p}{:}
        \PY{n}{mode\PYZus{}output} \PY{o}{=} \PY{l+s+s2}{\PYZdq{}}\PY{l+s+s2}{single\PYZhy{}word}\PY{l+s+s2}{\PYZdq{}}
    \PY{k}{elif} \PY{n}{prompt\PYZus{}version} \PY{o}{==} \PY{l+s+s2}{\PYZdq{}}\PY{l+s+s2}{CoT}\PY{l+s+s2}{\PYZdq{}}\PY{p}{:}
        \PY{n}{mode\PYZus{}output} \PY{o}{=} \PY{l+s+s2}{\PYZdq{}}\PY{l+s+s2}{CoT}\PY{l+s+s2}{\PYZdq{}}

    \PY{k}{if} \PY{n}{model\PYZus{}type} \PY{o}{==} \PY{l+s+s2}{\PYZdq{}}\PY{l+s+s2}{flan\PYZhy{}t5\PYZhy{}large}\PY{l+s+s2}{\PYZdq{}}\PY{p}{:}
        \PY{n}{llm\PYZus{}label\PYZus{}predictor} \PY{o}{=} \PY{n}{FlanT5LargeLabelPredictor}\PY{p}{(}\PY{n}{col\PYZus{}name\PYZus{}text}\PY{o}{=}\PY{l+s+s2}{\PYZdq{}}\PY{l+s+s2}{tweet\PYZus{}embedded}\PY{l+s+s2}{\PYZdq{}}\PY{p}{,}
                                                        \PY{n}{col\PYZus{}name\PYZus{}label}\PY{o}{=}\PY{l+s+s2}{\PYZdq{}}\PY{l+s+s2}{stance\PYZus{}predicted}\PY{l+s+s2}{\PYZdq{}}\PY{p}{,}
                                                        \PY{n}{col\PYZus{}name\PYZus{}text\PYZus{}id}\PY{o}{=}\PY{n}{par}\PY{o}{.}\PY{n}{TEXT\PYZus{}ID}\PY{p}{,}
                                                        \PY{n}{try\PYZus{}use\PYZus{}gpu}\PY{o}{=}\PY{k+kc}{True}\PY{p}{,}
                                                        \PY{n}{per\PYZus{}device\PYZus{}eval\PYZus{}batch\PYZus{}size}\PY{o}{=}\PY{l+m+mi}{16}\PY{p}{,}
                                                        \PY{n}{mode\PYZus{}output}\PY{o}{=}\PY{n}{mode\PYZus{}output}\PY{p}{,}
                                                        \PY{n}{for\PYZus{}generation\PYZus{}only}\PY{o}{=}\PY{k+kc}{True}\PY{p}{)}
    \PY{k}{elif} \PY{n}{model\PYZus{}type} \PY{o}{==} \PY{l+s+s2}{\PYZdq{}}\PY{l+s+s2}{flan\PYZhy{}t5\PYZhy{}xxl}\PY{l+s+s2}{\PYZdq{}}\PY{p}{:}
        \PY{n}{llm\PYZus{}label\PYZus{}predictor} \PY{o}{=} \PY{n}{FlanT5XxlLabelPredictor}\PY{p}{(}\PY{n}{col\PYZus{}name\PYZus{}text}\PY{o}{=}\PY{l+s+s2}{\PYZdq{}}\PY{l+s+s2}{tweet\PYZus{}embedded}\PY{l+s+s2}{\PYZdq{}}\PY{p}{,}
                                                        \PY{n}{col\PYZus{}name\PYZus{}label}\PY{o}{=}\PY{l+s+s2}{\PYZdq{}}\PY{l+s+s2}{stance\PYZus{}predicted}\PY{l+s+s2}{\PYZdq{}}\PY{p}{,}
                                                        \PY{n}{col\PYZus{}name\PYZus{}text\PYZus{}id}\PY{o}{=}\PY{n}{par}\PY{o}{.}\PY{n}{TEXT\PYZus{}ID}\PY{p}{,}
                                                        \PY{n}{try\PYZus{}use\PYZus{}gpu}\PY{o}{=}\PY{k+kc}{True}\PY{p}{,}
                                                        \PY{n}{per\PYZus{}device\PYZus{}eval\PYZus{}batch\PYZus{}size}\PY{o}{=}\PY{l+m+mi}{16}\PY{p}{,}
                                                        \PY{n}{mode\PYZus{}output}\PY{o}{=}\PY{n}{mode\PYZus{}output}\PY{p}{,}
                                                        \PY{n}{for\PYZus{}generation\PYZus{}only}\PY{o}{=}\PY{k+kc}{True}\PY{p}{)}
    \PY{k}{else}\PY{p}{:}
        \PY{k}{raise} \PY{n+ne}{ValueError}\PY{p}{(}\PY{l+s+s2}{\PYZdq{}}\PY{l+s+s2}{model\PYZus{}type not supported}\PY{l+s+s2}{\PYZdq{}}\PY{p}{)}

    \PY{c+c1}{\PYZsh{} prompt the LLM to make predictions (write the predictions to `file\PYZus{}output\PYZus{}predictions`)}
    \PY{n}{llm\PYZus{}label\PYZus{}predictor}\PY{o}{.}\PY{n}{predict\PYZus{}labels}\PY{p}{(}\PY{n}{df\PYZus{}input\PYZus{}text}\PY{p}{,}
                                       \PY{n}{file\PYZus{}output\PYZus{}predictions}\PY{p}{,}
                                       \PY{n}{keep\PYZus{}tweet\PYZus{}id}\PY{o}{=}\PY{k+kc}{True}\PY{p}{,} \PY{n}{keep\PYZus{}text}\PY{o}{=}\PY{k+kc}{False}\PY{p}{,}
                                       \PY{n}{output\PYZus{}prob\PYZus{}mode}\PY{o}{=}\PY{k+kc}{None}\PY{p}{,}
                                       \PY{n}{list\PYZus{}label\PYZus{}space}\PY{o}{=}\PY{k+kc}{None}\PY{p}{,}
                                       \PY{n}{col\PYZus{}name\PYZus{}tweet\PYZus{}id}\PY{o}{=}\PY{n}{par}\PY{o}{.}\PY{n}{TEXT\PYZus{}ID}\PY{p}{)}
\end{Verbatim}
\end{tcolorbox}

    \hypertarget{use-the-zero-shot-and-few-shot-prompts}{%
\subsection{Use the zero-shot and few-shot
prompts}\label{use-the-zero-shot-and-few-shot-prompts}}

Note that as mentioned earlier, zero-shot CoT prompt only works when the
LLM has more than 100 billion parameters. Even the larges variant of
FLAN-T5 (\texttt{flan-t5-xxl}) has only 11 billion parameters, so we
will not use the zero-shot CoT prompt in this tutorial.

    \begin{tcolorbox}[breakable, size=fbox, boxrule=1pt, pad at break*=1mm,colback=cellbackground, colframe=cellborder]
\prompt{In}{incolor}{56}{\boxspacing}
\begin{Verbatim}[commandchars=\\\{\}]
\PY{n}{model\PYZus{}type} \PY{o}{=} \PY{l+s+s2}{\PYZdq{}}\PY{l+s+s2}{flan\PYZhy{}t5\PYZhy{}large}\PY{l+s+s2}{\PYZdq{}}
\PY{k}{if} \PY{n}{PROMPT\PYZus{}FLAN\PYZus{}T5}\PY{p}{:}
    \PY{k}{for} \PY{n}{prompt\PYZus{}version} \PY{o+ow}{in} \PY{p}{[}\PY{l+s+s2}{\PYZdq{}}\PY{l+s+s2}{zero\PYZus{}shot}\PY{l+s+s2}{\PYZdq{}}\PY{p}{,}\PY{l+s+s2}{\PYZdq{}}\PY{l+s+s2}{few\PYZus{}shot}\PY{l+s+s2}{\PYZdq{}}\PY{p}{]}\PY{p}{:}
        \PY{n+nb}{print}\PY{p}{(}\PY{l+s+s2}{\PYZdq{}}\PY{l+s+s2}{model:}\PY{l+s+si}{\PYZob{}\PYZcb{}}\PY{l+s+s2}{, prompt version: }\PY{l+s+si}{\PYZob{}\PYZcb{}}\PY{l+s+s2}{\PYZdq{}}\PY{o}{.}\PY{n}{format}\PY{p}{(}\PY{n}{model\PYZus{}type}\PY{p}{,} \PY{n}{prompt\PYZus{}version}\PY{p}{)}\PY{p}{)}
        \PY{n}{prompt\PYZus{}flan\PYZus{}t5}\PY{p}{(}\PY{n}{model\PYZus{}type}\PY{p}{,} \PY{n}{prompt\PYZus{}version}\PY{p}{)}
\end{Verbatim}
\end{tcolorbox}

    While FLAN-T5-XXL is the most powerful variant, it can not be run on
Google Colab due to GPU memory limitations. In this tutorial, I have run
the predictions elsewhere and upload the predictions. If you want to
prompt FLAN-T5-XXL on your own and you have access to a large GPU with
more memory (\textgreater30GB), you can prompt the FLAN-T5-XXL model by
setting \texttt{LARGE\_GPU\_AVAILABLE\ =\ True} below.

    \begin{tcolorbox}[breakable, size=fbox, boxrule=1pt, pad at break*=1mm,colback=cellbackground, colframe=cellborder]
\prompt{In}{incolor}{57}{\boxspacing}
\begin{Verbatim}[commandchars=\\\{\}]
\PY{c+c1}{\PYZsh{} run this only if use you a large enough GPU (\PYZgt{}30GB)}
\PY{n}{LARGE\PYZus{}GPU\PYZus{}AVAILABLE} \PY{o}{=} \PY{k+kc}{False}
\PY{k}{if} \PY{n}{LARGE\PYZus{}GPU\PYZus{}AVAILABLE} \PY{o+ow}{and} \PY{n}{PROMPT\PYZus{}FLAN\PYZus{}T5}\PY{p}{:}
    \PY{n}{model\PYZus{}type} \PY{o}{=} \PY{l+s+s2}{\PYZdq{}}\PY{l+s+s2}{flan\PYZhy{}t5\PYZhy{}xxl}\PY{l+s+s2}{\PYZdq{}}
    \PY{k}{for} \PY{n}{prompt\PYZus{}version} \PY{o+ow}{in} \PY{p}{[}\PY{l+s+s2}{\PYZdq{}}\PY{l+s+s2}{zero\PYZus{}shot}\PY{l+s+s2}{\PYZdq{}}\PY{p}{,}\PY{l+s+s2}{\PYZdq{}}\PY{l+s+s2}{few\PYZus{}shot}\PY{l+s+s2}{\PYZdq{}}\PY{p}{]}\PY{p}{:}
        \PY{n+nb}{print}\PY{p}{(}\PY{l+s+s2}{\PYZdq{}}\PY{l+s+s2}{model:}\PY{l+s+si}{\PYZob{}\PYZcb{}}\PY{l+s+s2}{, prompt version: }\PY{l+s+si}{\PYZob{}\PYZcb{}}\PY{l+s+s2}{\PYZdq{}}\PY{o}{.}\PY{n}{format}\PY{p}{(}\PY{n}{model\PYZus{}type}\PY{p}{,} \PY{n}{prompt\PYZus{}version}\PY{p}{)}\PY{p}{)}
        \PY{n}{prompt\PYZus{}flan\PYZus{}t5}\PY{p}{(}\PY{n}{model\PYZus{}type}\PY{p}{,} \PY{n}{prompt\PYZus{}version}\PY{p}{)}
\end{Verbatim}
\end{tcolorbox}

    \hypertarget{view-the-predictions}{%
\subsubsection{View the predictions}\label{view-the-predictions}}

    \begin{tcolorbox}[breakable, size=fbox, boxrule=1pt, pad at break*=1mm,colback=cellbackground, colframe=cellborder]
\prompt{In}{incolor}{58}{\boxspacing}
\begin{Verbatim}[commandchars=\\\{\}]
\PY{k}{def} \PY{n+nf}{read\PYZus{}predictions}\PY{p}{(}\PY{n}{model\PYZus{}type}\PY{p}{,} \PY{n}{prompt\PYZus{}version}\PY{p}{)}\PY{p}{:}
    \PY{c+c1}{\PYZsh{} get the output paths}
    \PY{c+c1}{\PYZsh{} model\PYZus{}type\PYZus{}name = tidy\PYZus{}name(model\PYZus{}type)}
    \PY{n}{path\PYZus{}output} \PY{o}{=} \PY{n}{join}\PY{p}{(}\PY{n}{PATH\PYZus{}OUTPUT\PYZus{}ROOT}\PY{p}{,} \PY{n}{model\PYZus{}type}\PY{p}{,} \PY{n}{prompt\PYZus{}version}\PY{p}{)}
    \PY{n}{file\PYZus{}output\PYZus{}predictions} \PY{o}{=} \PY{n}{join}\PY{p}{(}\PY{n}{path\PYZus{}output}\PY{p}{,} \PY{l+s+s2}{\PYZdq{}}\PY{l+s+s2}{predictions.csv}\PY{l+s+s2}{\PYZdq{}}\PY{p}{)}

    \PY{c+c1}{\PYZsh{} read the predictions}
    \PY{n}{df\PYZus{}predictions} \PY{o}{=} \PY{n}{pd}\PY{o}{.}\PY{n}{read\PYZus{}csv}\PY{p}{(}\PY{n}{file\PYZus{}output\PYZus{}predictions}\PY{p}{)}
    \PY{k}{return} \PY{n}{df\PYZus{}predictions}
\end{Verbatim}
\end{tcolorbox}

    As demonstrated below, using a zero-shot prompt, the FLAN-T5-XXL model
successfully predicts the stance of this example tweet, whereas the
FLAN-T5-Large model struggles to do so.

    \begin{tcolorbox}[breakable, size=fbox, boxrule=1pt, pad at break*=1mm,colback=cellbackground, colframe=cellborder]
\prompt{In}{incolor}{59}{\boxspacing}
\begin{Verbatim}[commandchars=\\\{\}]
\PY{n}{df\PYZus{}predictions\PYZus{}flan\PYZus{}xxl\PYZus{}zero\PYZus{}shot} \PY{o}{=} \PY{n}{read\PYZus{}predictions}\PY{p}{(}\PY{l+s+s2}{\PYZdq{}}\PY{l+s+s2}{flan\PYZhy{}t5\PYZhy{}xxl}\PY{l+s+s2}{\PYZdq{}}\PY{p}{,}\PY{l+s+s2}{\PYZdq{}}\PY{l+s+s2}{zero\PYZus{}shot}\PY{l+s+s2}{\PYZdq{}}\PY{p}{)}
\PY{n}{df\PYZus{}predictions\PYZus{}flan\PYZus{}large\PYZus{}zero\PYZus{}shot} \PY{o}{=} \PY{n}{read\PYZus{}predictions}\PY{p}{(}\PY{l+s+s2}{\PYZdq{}}\PY{l+s+s2}{flan\PYZhy{}t5\PYZhy{}large}\PY{l+s+s2}{\PYZdq{}}\PY{p}{,}\PY{l+s+s2}{\PYZdq{}}\PY{l+s+s2}{zero\PYZus{}shot}\PY{l+s+s2}{\PYZdq{}}\PY{p}{)}
\end{Verbatim}
\end{tcolorbox}

    \begin{tcolorbox}[breakable, size=fbox, boxrule=1pt, pad at break*=1mm,colback=cellbackground, colframe=cellborder]
\prompt{In}{incolor}{60}{\boxspacing}
\begin{Verbatim}[commandchars=\\\{\}]
\PY{c+c1}{\PYZsh{} view the first prediction}
\PY{n+nb}{print}\PY{p}{(}\PY{l+s+s2}{\PYZdq{}}\PY{l+s+s2}{prompt:}\PY{l+s+se}{\PYZbs{}n}\PY{l+s+si}{\PYZob{}\PYZcb{}}\PY{l+s+se}{\PYZbs{}n}\PY{l+s+s2}{\PYZdq{}}\PY{o}{.}\PY{n}{format}\PY{p}{(}\PY{n}{df\PYZus{}predictions\PYZus{}flan\PYZus{}xxl\PYZus{}zero\PYZus{}shot}\PY{p}{[}\PY{l+s+s2}{\PYZdq{}}\PY{l+s+s2}{tweet\PYZus{}embedded}\PY{l+s+s2}{\PYZdq{}}\PY{p}{]}\PY{p}{[}\PY{l+m+mi}{0}\PY{p}{]}\PY{p}{)}\PY{p}{)}
\PY{c+c1}{\PYZsh{} print(\PYZdq{}ID:\PYZbs{}n\PYZob{}\PYZcb{}\PYZbs{}n\PYZdq{}.format(df\PYZus{}predictions\PYZus{}flan\PYZus{}xxl\PYZus{}zero\PYZus{}shot[\PYZdq{}ID\PYZdq{}][0]))}
\PY{n+nb}{print}\PY{p}{(}\PY{l+s+s2}{\PYZdq{}}\PY{l+s+s2}{true label::}\PY{l+s+se}{\PYZbs{}n}\PY{l+s+si}{\PYZob{}\PYZcb{}}\PY{l+s+se}{\PYZbs{}n}\PY{l+s+s2}{\PYZdq{}}\PY{o}{.}\PY{n}{format}\PY{p}{(}\PY{n}{df\PYZus{}input\PYZus{}text}\PY{p}{[}\PY{n}{df\PYZus{}input\PYZus{}text}\PY{p}{[}\PY{l+s+s2}{\PYZdq{}}\PY{l+s+s2}{ID}\PY{l+s+s2}{\PYZdq{}}\PY{p}{]}\PY{o}{==}\PY{l+s+s2}{\PYZdq{}}\PY{l+s+s2}{2313}\PY{l+s+s2}{\PYZdq{}}\PY{p}{]}\PY{p}{[}\PY{p}{[}\PY{l+s+s2}{\PYZdq{}}\PY{l+s+s2}{label}\PY{l+s+s2}{\PYZdq{}}\PY{p}{]}\PY{p}{]}\PY{o}{.}\PY{n}{values}\PY{p}{[}\PY{l+m+mi}{0}\PY{p}{]}\PY{p}{[}\PY{l+m+mi}{0}\PY{p}{]}\PY{o}{.}\PY{n}{lower}\PY{p}{(}\PY{p}{)}\PY{p}{)}\PY{p}{)}

\PY{n+nb}{print}\PY{p}{(}\PY{l+s+s2}{\PYZdq{}}\PY{l+s+s2}{FLAN\PYZhy{}T5\PYZhy{}LARGE}\PY{l+s+s2}{\PYZsq{}}\PY{l+s+s2}{s prediction:}\PY{l+s+se}{\PYZbs{}n}\PY{l+s+si}{\PYZob{}\PYZcb{}}\PY{l+s+s2}{\PYZdq{}}\PY{o}{.}\PY{n}{format}\PY{p}{(}\PY{n}{df\PYZus{}predictions\PYZus{}flan\PYZus{}large\PYZus{}zero\PYZus{}shot}\PY{p}{[}\PY{l+s+s2}{\PYZdq{}}\PY{l+s+s2}{stance\PYZus{}predicted}\PY{l+s+s2}{\PYZdq{}}\PY{p}{]}\PY{p}{[}\PY{l+m+mi}{0}\PY{p}{]}\PY{p}{)}\PY{p}{)}
\PY{n+nb}{print}\PY{p}{(}\PY{l+s+s2}{\PYZdq{}}\PY{l+s+s2}{FLAN\PYZhy{}T5\PYZhy{}XXL}\PY{l+s+s2}{\PYZsq{}}\PY{l+s+s2}{s prediction:}\PY{l+s+se}{\PYZbs{}n}\PY{l+s+si}{\PYZob{}\PYZcb{}}\PY{l+s+s2}{\PYZdq{}}\PY{o}{.}\PY{n}{format}\PY{p}{(}\PY{n}{df\PYZus{}predictions\PYZus{}flan\PYZus{}xxl\PYZus{}zero\PYZus{}shot}\PY{p}{[}\PY{l+s+s2}{\PYZdq{}}\PY{l+s+s2}{stance\PYZus{}predicted}\PY{l+s+s2}{\PYZdq{}}\PY{p}{]}\PY{p}{[}\PY{l+m+mi}{0}\PY{p}{]}\PY{p}{)}\PY{p}{)}
\end{Verbatim}
\end{tcolorbox}

    \begin{Verbatim}[commandchars=\\\{\}]
prompt:
What is the stance of the tweet below with respect to 'Legalization of
Abortion'?  If we can infer from the tweet that the tweeter supports
'Legalization of Abortion', please label it as 'in-favor'. If we can infer from
the tweet that the tweeter is against 'Legalization of Abortion', please label
is as 'against'. If we can infer from the tweet that the tweeter has a neutral
stance towards 'Legalization of Abortion', please label it as 'neutral-or-
unclear'. If there is no clue in the tweet to reveal the stance of the tweeter
towards 'Legalization of Abortion', please also label is as 'neutral-or-
unclear'. Please use exactly one word from the following 3 categories to label
it: 'in-favor', 'against', 'neutral-or-unclear'. Here is the tweet. 'let's agree
that it's not ok to kill a 7lbs baby in the uterus @USERNAME \#dnc \#clinton2016
@USERNAME \#procompromise' The stance of the tweet is:

true label::
against

FLAN-T5-LARGE's prediction:
in-favor
FLAN-T5-XXL's prediction:
against
    \end{Verbatim}

    \hypertarget{evaluate-the-predictions-of-different-prompts-and-different-models}{%
\subsection{Evaluate the Predictions of Different Prompts and Different
Models}\label{evaluate-the-predictions-of-different-prompts-and-different-models}}

    Fantastic! With the predictions in hand from two types of prompts and
two variants of FLAN-T5 models, it's time to evaluate their performance
across all tweets in both the validation and test sets.

    \begin{tcolorbox}[breakable, size=fbox, boxrule=1pt, pad at break*=1mm,colback=cellbackground, colframe=cellborder]
\prompt{In}{incolor}{61}{\boxspacing}
\begin{Verbatim}[commandchars=\\\{\}]
\PY{n}{evaluate}\PY{p}{(}\PY{l+s+s2}{\PYZdq{}}\PY{l+s+s2}{flan\PYZhy{}t5\PYZhy{}large}\PY{l+s+s2}{\PYZdq{}}\PY{p}{,} \PY{l+s+s2}{\PYZdq{}}\PY{l+s+s2}{zero\PYZus{}shot}\PY{l+s+s2}{\PYZdq{}}\PY{p}{)}
\PY{n}{evaluate}\PY{p}{(}\PY{l+s+s2}{\PYZdq{}}\PY{l+s+s2}{flan\PYZhy{}t5\PYZhy{}large}\PY{l+s+s2}{\PYZdq{}}\PY{p}{,} \PY{l+s+s2}{\PYZdq{}}\PY{l+s+s2}{few\PYZus{}shot}\PY{l+s+s2}{\PYZdq{}}\PY{p}{)}
\PY{c+c1}{\PYZsh{} although you may not have run the xxl model due to GPU contraint, you can still evaluate the results because I have already run the model elsewhere and saved the predictions}
\PY{n}{evaluate}\PY{p}{(}\PY{l+s+s2}{\PYZdq{}}\PY{l+s+s2}{flan\PYZhy{}t5\PYZhy{}xxl}\PY{l+s+s2}{\PYZdq{}}\PY{p}{,} \PY{l+s+s2}{\PYZdq{}}\PY{l+s+s2}{zero\PYZus{}shot}\PY{l+s+s2}{\PYZdq{}}\PY{p}{)}
\PY{n}{evaluate}\PY{p}{(}\PY{l+s+s2}{\PYZdq{}}\PY{l+s+s2}{flan\PYZhy{}t5\PYZhy{}xxl}\PY{l+s+s2}{\PYZdq{}}\PY{p}{,} \PY{l+s+s2}{\PYZdq{}}\PY{l+s+s2}{few\PYZus{}shot}\PY{l+s+s2}{\PYZdq{}}\PY{p}{)}
\end{Verbatim}
\end{tcolorbox}

    \begin{tcolorbox}[breakable, size=fbox, boxrule=1pt, pad at break*=1mm,colback=cellbackground, colframe=cellborder]
\prompt{In}{incolor}{62}{\boxspacing}
\begin{Verbatim}[commandchars=\\\{\}]
\PY{c+c1}{\PYZsh{} sumamrize the result acrpss different models and prompt versions}
\PY{c+c1}{\PYZsh{} \PYZhy{} also visualize the confusion matrices}
\PY{n}{df\PYZus{}hightlight\PYZus{}metrics\PYZus{}flan\PYZus{}t5} \PY{o}{=} \PY{n}{summarize\PYZus{}results}\PY{p}{(}\PY{p}{[}\PY{l+s+s2}{\PYZdq{}}\PY{l+s+s2}{flan\PYZhy{}t5\PYZhy{}large}\PY{l+s+s2}{\PYZdq{}}\PY{p}{,}\PY{l+s+s2}{\PYZdq{}}\PY{l+s+s2}{flan\PYZhy{}t5\PYZhy{}xxl}\PY{l+s+s2}{\PYZdq{}}\PY{p}{]}\PY{p}{,}\PY{p}{[}\PY{l+s+s2}{\PYZdq{}}\PY{l+s+s2}{zero\PYZus{}shot}\PY{l+s+s2}{\PYZdq{}}\PY{p}{,}\PY{l+s+s2}{\PYZdq{}}\PY{l+s+s2}{few\PYZus{}shot}\PY{l+s+s2}{\PYZdq{}}\PY{p}{]}\PY{p}{)}

\PY{n}{df\PYZus{}hightlight\PYZus{}metrics\PYZus{}flan\PYZus{}t5} \PY{o}{=} \PY{n}{df\PYZus{}hightlight\PYZus{}metrics\PYZus{}flan\PYZus{}t5}\PY{p}{[}\PY{n}{df\PYZus{}hightlight\PYZus{}metrics\PYZus{}flan\PYZus{}t5}\PY{o}{.}\PY{n}{model\PYZus{}type}\PY{o}{.}\PY{n}{isin}\PY{p}{(}\PY{p}{[}\PY{l+s+s2}{\PYZdq{}}\PY{l+s+s2}{llm\PYZus{}flan\PYZhy{}t5\PYZhy{}large}\PY{l+s+s2}{\PYZdq{}}\PY{p}{,}\PY{l+s+s2}{\PYZdq{}}\PY{l+s+s2}{llm\PYZus{}flan\PYZhy{}t5\PYZhy{}xxl}\PY{l+s+s2}{\PYZdq{}}\PY{p}{]}\PY{p}{)}\PY{p}{]}

\PY{c+c1}{\PYZsh{} reorder the rows}
\PY{n}{df\PYZus{}hightlight\PYZus{}metrics\PYZus{}flan\PYZus{}t5}\PY{p}{[}\PY{l+s+s1}{\PYZsq{}}\PY{l+s+s1}{model\PYZus{}type}\PY{l+s+s1}{\PYZsq{}}\PY{p}{]} \PY{o}{=} \PY{n}{pd}\PY{o}{.}\PY{n}{Categorical}\PY{p}{(}\PY{n}{df\PYZus{}hightlight\PYZus{}metrics\PYZus{}flan\PYZus{}t5}\PY{p}{[}\PY{l+s+s1}{\PYZsq{}}\PY{l+s+s1}{model\PYZus{}type}\PY{l+s+s1}{\PYZsq{}}\PY{p}{]}\PY{p}{,} \PY{n}{categories}\PY{o}{=}\PY{p}{[}\PY{l+s+s2}{\PYZdq{}}\PY{l+s+s2}{llm\PYZus{}flan\PYZhy{}t5\PYZhy{}large}\PY{l+s+s2}{\PYZdq{}}\PY{p}{,}\PY{l+s+s2}{\PYZdq{}}\PY{l+s+s2}{llm\PYZus{}flan\PYZhy{}t5\PYZhy{}xxl}\PY{l+s+s2}{\PYZdq{}}\PY{p}{]}\PY{p}{,} \PY{n}{ordered}\PY{o}{=}\PY{k+kc}{True}\PY{p}{)}
\PY{n}{df\PYZus{}hightlight\PYZus{}metrics\PYZus{}flan\PYZus{}t5}\PY{p}{[}\PY{l+s+s1}{\PYZsq{}}\PY{l+s+s1}{version}\PY{l+s+s1}{\PYZsq{}}\PY{p}{]} \PY{o}{=} \PY{n}{pd}\PY{o}{.}\PY{n}{Categorical}\PY{p}{(}\PY{n}{df\PYZus{}hightlight\PYZus{}metrics\PYZus{}flan\PYZus{}t5}\PY{p}{[}\PY{l+s+s1}{\PYZsq{}}\PY{l+s+s1}{version}\PY{l+s+s1}{\PYZsq{}}\PY{p}{]}\PY{p}{,} \PY{n}{categories}\PY{o}{=}\PY{p}{[}\PY{l+s+s2}{\PYZdq{}}\PY{l+s+s2}{zero\PYZus{}shot}\PY{l+s+s2}{\PYZdq{}}\PY{p}{,}\PY{l+s+s2}{\PYZdq{}}\PY{l+s+s2}{few\PYZus{}shot}\PY{l+s+s2}{\PYZdq{}}\PY{p}{]}\PY{p}{,} \PY{n}{ordered}\PY{o}{=}\PY{k+kc}{True}\PY{p}{)}
\PY{n}{df\PYZus{}hightlight\PYZus{}metrics\PYZus{}flan\PYZus{}t5} \PY{o}{=} \PY{n}{df\PYZus{}hightlight\PYZus{}metrics\PYZus{}flan\PYZus{}t5}\PY{o}{.}\PY{n}{sort\PYZus{}values}\PY{p}{(}\PY{n}{by}\PY{o}{=}\PY{p}{[}\PY{l+s+s2}{\PYZdq{}}\PY{l+s+s2}{model\PYZus{}type}\PY{l+s+s2}{\PYZdq{}}\PY{p}{,}\PY{l+s+s2}{\PYZdq{}}\PY{l+s+s2}{version}\PY{l+s+s2}{\PYZdq{}}\PY{p}{]}\PY{p}{)}\PY{o}{.}\PY{n}{reset\PYZus{}index}\PY{p}{(}\PY{n}{drop}\PY{o}{=}\PY{k+kc}{True}\PY{p}{)}

\PY{c+c1}{\PYZsh{} rename columns}
\PY{n}{df\PYZus{}hightlight\PYZus{}metrics\PYZus{}flan\PYZus{}t5} \PY{o}{=} \PY{n}{df\PYZus{}hightlight\PYZus{}metrics\PYZus{}flan\PYZus{}t5}\PY{p}{[}\PY{p}{[}\PY{l+s+s2}{\PYZdq{}}\PY{l+s+s2}{model\PYZus{}type}\PY{l+s+s2}{\PYZdq{}}\PY{p}{,}\PY{l+s+s2}{\PYZdq{}}\PY{l+s+s2}{version}\PY{l+s+s2}{\PYZdq{}}\PY{p}{,}\PY{l+s+s2}{\PYZdq{}}\PY{l+s+s2}{set}\PY{l+s+s2}{\PYZdq{}}\PY{p}{,}\PY{l+s+s2}{\PYZdq{}}\PY{l+s+s2}{f1\PYZus{}macro}\PY{l+s+s2}{\PYZdq{}}\PY{p}{,}\PY{l+s+s2}{\PYZdq{}}\PY{l+s+s2}{f1\PYZus{}NONE}\PY{l+s+s2}{\PYZdq{}}\PY{p}{,}\PY{l+s+s2}{\PYZdq{}}\PY{l+s+s2}{f1\PYZus{}FAVOR}\PY{l+s+s2}{\PYZdq{}}\PY{p}{,}\PY{l+s+s2}{\PYZdq{}}\PY{l+s+s2}{f1\PYZus{}AGAINST}\PY{l+s+s2}{\PYZdq{}}\PY{p}{]}\PY{p}{]}\PY{o}{.}\PY{n}{sort\PYZus{}values}\PY{p}{(}\PY{n}{by}\PY{o}{=}\PY{p}{[}\PY{l+s+s2}{\PYZdq{}}\PY{l+s+s2}{model\PYZus{}type}\PY{l+s+s2}{\PYZdq{}}\PY{p}{,}\PY{l+s+s2}{\PYZdq{}}\PY{l+s+s2}{version}\PY{l+s+s2}{\PYZdq{}}\PY{p}{]}\PY{p}{)}\PY{o}{.}\PY{n}{rename}\PY{p}{(}\PY{n}{columns}\PY{o}{=}\PY{p}{\PYZob{}}\PY{l+s+s2}{\PYZdq{}}\PY{l+s+s2}{version}\PY{l+s+s2}{\PYZdq{}}\PY{p}{:}\PY{l+s+s2}{\PYZdq{}}\PY{l+s+s2}{prompt\PYZus{}type}\PY{l+s+s2}{\PYZdq{}}\PY{p}{,}\PY{l+s+s2}{\PYZdq{}}\PY{l+s+s2}{set}\PY{l+s+s2}{\PYZdq{}}\PY{p}{:}\PY{l+s+s2}{\PYZdq{}}\PY{l+s+s2}{partition}\PY{l+s+s2}{\PYZdq{}}\PY{p}{\PYZcb{}}\PY{p}{)}
\end{Verbatim}
\end{tcolorbox}

    \begin{Verbatim}[commandchars=\\\{\}]
<Figure size 1500x500 with 0 Axes>
    \end{Verbatim}

    \begin{Verbatim}[commandchars=\\\{\}]
<Figure size 1500x500 with 0 Axes>
    \end{Verbatim}

    \begin{Verbatim}[commandchars=\\\{\}]
<Figure size 1500x500 with 0 Axes>
    \end{Verbatim}

    \begin{Verbatim}[commandchars=\\\{\}]
<Figure size 1500x500 with 0 Axes>
    \end{Verbatim}

    \hypertarget{view-the-performance-on-the-validation-set}{%
\subsubsection{View the performance on the validation
set}\label{view-the-performance-on-the-validation-set}}

    \begin{tcolorbox}[breakable, size=fbox, boxrule=1pt, pad at break*=1mm,colback=cellbackground, colframe=cellborder]
\prompt{In}{incolor}{63}{\boxspacing}
\begin{Verbatim}[commandchars=\\\{\}]
\PY{n}{df\PYZus{}hightlight\PYZus{}metrics\PYZus{}flan\PYZus{}t5\PYZus{}vali} \PY{o}{=} \PY{n}{df\PYZus{}hightlight\PYZus{}metrics\PYZus{}flan\PYZus{}t5}\PY{p}{[}\PY{n}{df\PYZus{}hightlight\PYZus{}metrics\PYZus{}flan\PYZus{}t5}\PY{p}{[}\PY{l+s+s2}{\PYZdq{}}\PY{l+s+s2}{partition}\PY{l+s+s2}{\PYZdq{}}\PY{p}{]}\PY{o}{.}\PY{n}{isin}\PY{p}{(}\PY{p}{[}\PY{l+s+s2}{\PYZdq{}}\PY{l+s+s2}{vali\PYZus{}raw}\PY{l+s+s2}{\PYZdq{}}\PY{p}{]}\PY{p}{)}\PY{p}{]}

\PY{c+c1}{\PYZsh{} convert \PYZdq{}vali\PYZus{}raw\PYZdq{} to \PYZdq{}vali\PYZdq{} (in the partition column)}
\PY{n}{df\PYZus{}hightlight\PYZus{}metrics\PYZus{}flan\PYZus{}t5\PYZus{}vali}\PY{o}{.}\PY{n}{loc}\PY{p}{[}\PY{n}{df\PYZus{}hightlight\PYZus{}metrics\PYZus{}flan\PYZus{}t5\PYZus{}vali}\PY{p}{[}\PY{l+s+s2}{\PYZdq{}}\PY{l+s+s2}{partition}\PY{l+s+s2}{\PYZdq{}}\PY{p}{]}\PY{o}{==}\PY{l+s+s2}{\PYZdq{}}\PY{l+s+s2}{vali\PYZus{}raw}\PY{l+s+s2}{\PYZdq{}}\PY{p}{,}\PY{l+s+s2}{\PYZdq{}}\PY{l+s+s2}{partition}\PY{l+s+s2}{\PYZdq{}}\PY{p}{]} \PY{o}{=} \PY{l+s+s2}{\PYZdq{}}\PY{l+s+s2}{vali}\PY{l+s+s2}{\PYZdq{}}

\PY{n}{df\PYZus{}hightlight\PYZus{}metrics\PYZus{}flan\PYZus{}t5\PYZus{}vali}
\end{Verbatim}
\end{tcolorbox}

            \begin{tcolorbox}[breakable, size=fbox, boxrule=.5pt, pad at break*=1mm, opacityfill=0]
\prompt{Out}{outcolor}{63}{\boxspacing}
\begin{Verbatim}[commandchars=\\\{\}]
          model\_type prompt\_type partition  f1\_macro   f1\_NONE  f1\_FAVOR  \textbackslash{}
0  llm\_flan-t5-large   zero\_shot      vali  0.244478  0.060606  0.215686
2  llm\_flan-t5-large    few\_shot      vali  0.267255  0.060606  0.192771
4    llm\_flan-t5-xxl   zero\_shot      vali  0.693311  0.600000  0.653846
6    llm\_flan-t5-xxl    few\_shot      vali  0.680935  0.653846  0.603774

   f1\_AGAINST
0    0.457143
2    0.548387
4    0.826087
6    0.785185
\end{Verbatim}
\end{tcolorbox}
        
    The first two columns indicate the model type and the prompt type,
respectively. The third column indicates that the performance is
evaluated on the validation set. The \texttt{f1\_macro} column indicates
the macro-averaged F1 score (across the three stance types). We use this
value to quantify the overall performance of the model. Note that we are
using the \texttt{f1\_macro} metric rather than accuracy because the
dataset is imbalanced, and the \texttt{f1\_macro} metric is more robust
to imbalanced datasets.

Based on the macro-F1 scores, the larger FLAN-T5-XXL model performs
better than the smaller FLAN-T5-Large model across two prompt types.
This is expected because the larger model has more parameters and can
capture more subtle meanings in language.

For FLAN-T5-XXL model, using few-shot prompt does not seem to help.

The last 3 columns indicate the performance of each stance type. This
shows that the model is better at predicting the \texttt{AGAINST} stance
than the \texttt{NONE} and \texttt{FAVOR} stances.

\begin{quote}
Note that, in practice, to avoid data leakage, when selecting the best
combination of prompt type and model type, we should use the performance
on the validation set to choose the best combination, and then use the
performance on the test set to evaluate the final model.
\end{quote}

    \begin{quote}
To learn more about macro-F1 score, I recommend taking a look at this
tutorial
https://towardsdatascience.com/micro-macro-weighted-averages-of-f1-score-clearly-explained-b603420b292f\#:\textasciitilde:text=The\%20macro\%2Daveraged\%20F1\%20score,regardless\%20of\%20their\%20support\%20values.
\end{quote}

    \hypertarget{view-the-performance-on-the-test-set}{%
\subsubsection{View the performance on the test
set}\label{view-the-performance-on-the-test-set}}

    \begin{tcolorbox}[breakable, size=fbox, boxrule=1pt, pad at break*=1mm,colback=cellbackground, colframe=cellborder]
\prompt{In}{incolor}{64}{\boxspacing}
\begin{Verbatim}[commandchars=\\\{\}]
\PY{n}{df\PYZus{}hightlight\PYZus{}metrics\PYZus{}flan\PYZus{}t5\PYZus{}test} \PY{o}{=} \PY{n}{df\PYZus{}hightlight\PYZus{}metrics\PYZus{}flan\PYZus{}t5}\PY{p}{[}\PY{n}{df\PYZus{}hightlight\PYZus{}metrics\PYZus{}flan\PYZus{}t5}\PY{p}{[}\PY{l+s+s2}{\PYZdq{}}\PY{l+s+s2}{partition}\PY{l+s+s2}{\PYZdq{}}\PY{p}{]}\PY{o}{.}\PY{n}{isin}\PY{p}{(}\PY{p}{[}\PY{l+s+s2}{\PYZdq{}}\PY{l+s+s2}{test\PYZus{}raw}\PY{l+s+s2}{\PYZdq{}}\PY{p}{]}\PY{p}{)}\PY{p}{]}

\PY{c+c1}{\PYZsh{} convert \PYZdq{}test\PYZus{}raw\PYZdq{} to \PYZdq{}test\PYZdq{} (in the partition column)}
\PY{n}{df\PYZus{}hightlight\PYZus{}metrics\PYZus{}flan\PYZus{}t5\PYZus{}test}\PY{o}{.}\PY{n}{loc}\PY{p}{[}\PY{n}{df\PYZus{}hightlight\PYZus{}metrics\PYZus{}flan\PYZus{}t5\PYZus{}test}\PY{p}{[}\PY{l+s+s2}{\PYZdq{}}\PY{l+s+s2}{partition}\PY{l+s+s2}{\PYZdq{}}\PY{p}{]}\PY{o}{==}\PY{l+s+s2}{\PYZdq{}}\PY{l+s+s2}{test\PYZus{}raw}\PY{l+s+s2}{\PYZdq{}}\PY{p}{,}\PY{l+s+s2}{\PYZdq{}}\PY{l+s+s2}{partition}\PY{l+s+s2}{\PYZdq{}}\PY{p}{]} \PY{o}{=} \PY{l+s+s2}{\PYZdq{}}\PY{l+s+s2}{test}\PY{l+s+s2}{\PYZdq{}}

\PY{n}{df\PYZus{}hightlight\PYZus{}metrics\PYZus{}flan\PYZus{}t5\PYZus{}test}
\end{Verbatim}
\end{tcolorbox}

            \begin{tcolorbox}[breakable, size=fbox, boxrule=.5pt, pad at break*=1mm, opacityfill=0]
\prompt{Out}{outcolor}{64}{\boxspacing}
\begin{Verbatim}[commandchars=\\\{\}]
          model\_type prompt\_type partition  f1\_macro   f1\_NONE  f1\_FAVOR  \textbackslash{}
1  llm\_flan-t5-large   zero\_shot      test  0.212933  0.043478  0.181034
3  llm\_flan-t5-large    few\_shot      test  0.278003  0.083333  0.198953
5    llm\_flan-t5-xxl   zero\_shot      test  0.619033  0.583333  0.531250
7    llm\_flan-t5-xxl    few\_shot      test  0.591850  0.534653  0.519685

   f1\_AGAINST
1    0.414286
3    0.551724
5    0.742515
7    0.721212
\end{Verbatim}
\end{tcolorbox}
        
    The results are similar to the validation set. The larger FLAN-T5-XXL
model performs better than the smaller FLAN-T5-Large model across two
prompt types. For FLAN-T5-XXL model, using few-shot prompt does not seem
to help.

    \hypertarget{confusion-matrix}{%
\subsubsection{Confusion Matrix}\label{confusion-matrix}}

As we did for ChatGPT, let's examine the confusion matrix for best
performing combination of model and prompt (i.e., FLAN-T5-XXL with
zero-shot prompt).

Similar to ChatGPT with few-shot prompt (the best combination for
ChatGPT), the FLAN-T5-XXL model also has trouble classifying the tweets
with the \texttt{AGAINST} stance, misclassifying them as either
\texttt{FAVOR} or \texttt{NONE}. In addition, the model also
misclassifies some \texttt{FAVOR} and \texttt{NONE} tweets as
\texttt{AGAINST}. These collectivly result in a lower macro-F1 score
than ChatGPT with few-shot prompt.

    \begin{tcolorbox}[breakable, size=fbox, boxrule=1pt, pad at break*=1mm,colback=cellbackground, colframe=cellborder]
\prompt{In}{incolor}{65}{\boxspacing}
\begin{Verbatim}[commandchars=\\\{\}]
\PY{n}{display\PYZus{}resized\PYZus{}image\PYZus{}in\PYZus{}notebook}\PY{p}{(}\PY{n}{join}\PY{p}{(}\PY{n}{PATH\PYZus{}OUTPUT\PYZus{}ROOT}\PY{p}{,}\PY{l+s+s2}{\PYZdq{}}\PY{l+s+s2}{summary}\PY{l+s+s2}{\PYZdq{}}\PY{p}{,}\PY{l+s+s2}{\PYZdq{}}\PY{l+s+s2}{flan\PYZhy{}t5\PYZhy{}xxl\PYZus{}zero\PYZus{}shot\PYZus{}comb\PYZus{}confusion\PYZus{}mat.png}\PY{l+s+s2}{\PYZdq{}}\PY{p}{)}\PY{p}{)}
\end{Verbatim}
\end{tcolorbox}

    \begin{center}
    \adjustimage{max size={0.9\linewidth}{0.9\paperheight}}{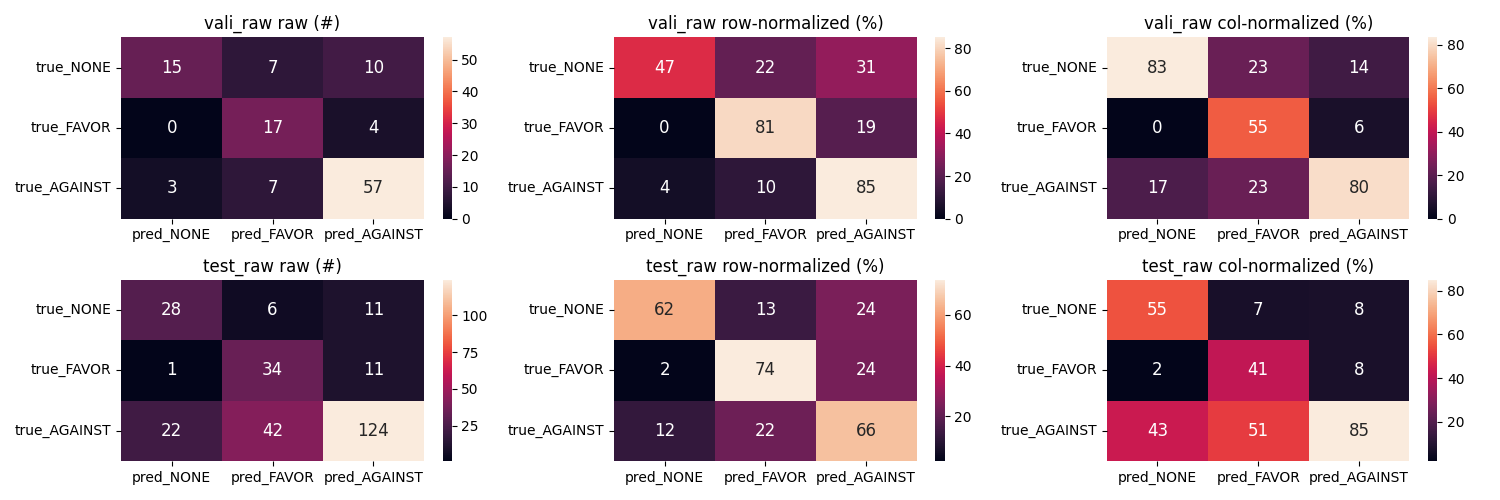}
    \end{center}
    { \hspace*{\fill} \\}
    
    \hypertarget{compare-flan-t5-with-bert}{%
\subsubsection{Compare FLAN-T5 with
BERT}\label{compare-flan-t5-with-bert}}

Let's compare the performance of FLAN-T5 with BERT. The results of BERT
are generated from the previous tutorial.

    \begin{tcolorbox}[breakable, size=fbox, boxrule=1pt, pad at break*=1mm,colback=cellbackground, colframe=cellborder]
\prompt{In}{incolor}{66}{\boxspacing}
\begin{Verbatim}[commandchars=\\\{\}]
\PY{n}{df\PYZus{}hightlight\PYZus{}metrics\PYZus{}bert} \PY{o}{=} \PY{n}{pd}\PY{o}{.}\PY{n}{read\PYZus{}csv}\PY{p}{(}\PY{n}{join}\PY{p}{(}\PY{n}{par}\PY{o}{.}\PY{n}{PATH\PYZus{}RESULT\PYZus{}SEM\PYZus{}EVAL\PYZus{}TUNING}\PY{p}{,}\PY{l+s+s2}{\PYZdq{}}\PY{l+s+s2}{summary}\PY{l+s+s2}{\PYZdq{}}\PY{p}{,}\PY{l+s+s2}{\PYZdq{}}\PY{l+s+s2}{metrics\PYZus{}highlights.csv}\PY{l+s+s2}{\PYZdq{}}\PY{p}{)}\PY{p}{)}
\PY{n}{df\PYZus{}hightlight\PYZus{}metrics\PYZus{}bert} \PY{o}{=} \PY{n}{df\PYZus{}hightlight\PYZus{}metrics\PYZus{}bert}\PY{p}{[}\PY{p}{[}\PY{l+s+s2}{\PYZdq{}}\PY{l+s+s2}{version}\PY{l+s+s2}{\PYZdq{}}\PY{p}{,}\PY{l+s+s2}{\PYZdq{}}\PY{l+s+s2}{set}\PY{l+s+s2}{\PYZdq{}}\PY{p}{,}\PY{l+s+s2}{\PYZdq{}}\PY{l+s+s2}{f1\PYZus{}macro}\PY{l+s+s2}{\PYZdq{}}\PY{p}{,}\PY{l+s+s2}{\PYZdq{}}\PY{l+s+s2}{f1\PYZus{}NONE}\PY{l+s+s2}{\PYZdq{}}\PY{p}{,}\PY{l+s+s2}{\PYZdq{}}\PY{l+s+s2}{f1\PYZus{}FAVOR}\PY{l+s+s2}{\PYZdq{}}\PY{p}{,}\PY{l+s+s2}{\PYZdq{}}\PY{l+s+s2}{f1\PYZus{}AGAINST}\PY{l+s+s2}{\PYZdq{}}\PY{p}{]}\PY{p}{]}\PY{o}{.}\PY{n}{rename}\PY{p}{(}\PY{n}{columns}\PY{o}{=}\PY{p}{\PYZob{}}\PY{l+s+s2}{\PYZdq{}}\PY{l+s+s2}{version}\PY{l+s+s2}{\PYZdq{}}\PY{p}{:}\PY{l+s+s2}{\PYZdq{}}\PY{l+s+s2}{model\PYZus{}type}\PY{l+s+s2}{\PYZdq{}}\PY{p}{\PYZcb{}}\PY{p}{)}
\PY{c+c1}{\PYZsh{} reorder the rows}
\PY{n}{df\PYZus{}hightlight\PYZus{}metrics\PYZus{}bert}\PY{p}{[}\PY{l+s+s1}{\PYZsq{}}\PY{l+s+s1}{model\PYZus{}type}\PY{l+s+s1}{\PYZsq{}}\PY{p}{]} \PY{o}{=} \PY{n}{pd}\PY{o}{.}\PY{n}{Categorical}\PY{p}{(}\PY{n}{df\PYZus{}hightlight\PYZus{}metrics\PYZus{}bert}\PY{p}{[}\PY{l+s+s1}{\PYZsq{}}\PY{l+s+s1}{model\PYZus{}type}\PY{l+s+s1}{\PYZsq{}}\PY{p}{]}\PY{p}{,} \PY{n}{categories}\PY{o}{=}\PY{p}{[}\PY{l+s+s2}{\PYZdq{}}\PY{l+s+s2}{bert\PYZhy{}base\PYZhy{}uncased}\PY{l+s+s2}{\PYZdq{}}\PY{p}{,}\PY{l+s+s2}{\PYZdq{}}\PY{l+s+s2}{vinai\PYZus{}bertweet\PYZus{}base}\PY{l+s+s2}{\PYZdq{}}\PY{p}{,}\PY{l+s+s2}{\PYZdq{}}\PY{l+s+s2}{kornosk\PYZus{}polibertweet\PYZus{}mlm}\PY{l+s+s2}{\PYZdq{}}\PY{p}{]}\PY{p}{,} \PY{n}{ordered}\PY{o}{=}\PY{k+kc}{True}\PY{p}{)}
\PY{n}{df\PYZus{}hightlight\PYZus{}metrics\PYZus{}bert} \PY{o}{=} \PY{n}{df\PYZus{}hightlight\PYZus{}metrics\PYZus{}bert}\PY{o}{.}\PY{n}{sort\PYZus{}values}\PY{p}{(}\PY{n}{by}\PY{o}{=}\PY{p}{[}\PY{l+s+s2}{\PYZdq{}}\PY{l+s+s2}{model\PYZus{}type}\PY{l+s+s2}{\PYZdq{}}\PY{p}{]}\PY{p}{)}\PY{o}{.}\PY{n}{reset\PYZus{}index}\PY{p}{(}\PY{n}{drop}\PY{o}{=}\PY{k+kc}{True}\PY{p}{)}
\PY{c+c1}{\PYZsh{} rename columns}
\PY{n}{df\PYZus{}hightlight\PYZus{}metrics\PYZus{}bert} \PY{o}{=} \PY{n}{df\PYZus{}hightlight\PYZus{}metrics\PYZus{}bert}\PY{o}{.}\PY{n}{rename}\PY{p}{(}\PY{n}{columns}\PY{o}{=}\PY{p}{\PYZob{}}\PY{l+s+s2}{\PYZdq{}}\PY{l+s+s2}{set}\PY{l+s+s2}{\PYZdq{}}\PY{p}{:}\PY{l+s+s2}{\PYZdq{}}\PY{l+s+s2}{partition}\PY{l+s+s2}{\PYZdq{}}\PY{p}{\PYZcb{}}\PY{p}{)}
\PY{c+c1}{\PYZsh{} convert \PYZdq{}test\PYZus{}raw\PYZdq{} to \PYZdq{}test\PYZdq{} (in the partition column)}
\PY{n}{df\PYZus{}hightlight\PYZus{}metrics\PYZus{}bert}\PY{o}{.}\PY{n}{loc}\PY{p}{[}\PY{n}{df\PYZus{}hightlight\PYZus{}metrics\PYZus{}bert}\PY{p}{[}\PY{l+s+s2}{\PYZdq{}}\PY{l+s+s2}{partition}\PY{l+s+s2}{\PYZdq{}}\PY{p}{]}\PY{o}{==}\PY{l+s+s2}{\PYZdq{}}\PY{l+s+s2}{test\PYZus{}raw}\PY{l+s+s2}{\PYZdq{}}\PY{p}{,}\PY{l+s+s2}{\PYZdq{}}\PY{l+s+s2}{partition}\PY{l+s+s2}{\PYZdq{}}\PY{p}{]} \PY{o}{=} \PY{l+s+s2}{\PYZdq{}}\PY{l+s+s2}{test}\PY{l+s+s2}{\PYZdq{}}
\PY{c+c1}{\PYZsh{} subset the test set}
\PY{n}{df\PYZus{}hightlight\PYZus{}metrics\PYZus{}bert\PYZus{}test} \PY{o}{=} \PY{n}{df\PYZus{}hightlight\PYZus{}metrics\PYZus{}bert}\PY{p}{[}\PY{n}{df\PYZus{}hightlight\PYZus{}metrics\PYZus{}bert}\PY{p}{[}\PY{l+s+s2}{\PYZdq{}}\PY{l+s+s2}{partition}\PY{l+s+s2}{\PYZdq{}}\PY{p}{]}\PY{o}{==} \PY{l+s+s2}{\PYZdq{}}\PY{l+s+s2}{test}\PY{l+s+s2}{\PYZdq{}}\PY{p}{]}
\PY{n}{df\PYZus{}hightlight\PYZus{}metrics\PYZus{}bert\PYZus{}test}
\end{Verbatim}
\end{tcolorbox}

            \begin{tcolorbox}[breakable, size=fbox, boxrule=.5pt, pad at break*=1mm, opacityfill=0]
\prompt{Out}{outcolor}{66}{\boxspacing}
\begin{Verbatim}[commandchars=\\\{\}]
                 model\_type partition  f1\_macro  f1\_NONE  f1\_FAVOR  f1\_AGAINST
2         bert-base-uncased      test    0.4748   0.4196    0.4275      0.5775
5       vinai\_bertweet\_base      test    0.5797   0.5323    0.5401      0.6667
8  kornosk\_polibertweet\_mlm      test    0.5616   0.5440    0.4762      0.6645
\end{Verbatim}
\end{tcolorbox}
        
    Based on the macro-F1 scores, the FLAN-T5-XXL model using the zero-shot
prompt outperforms all the BERT variants we examined in the previous
tutorial.

    \begin{center}\rule{0.5\linewidth}{0.5pt}\end{center}

    \hypertarget{compare-chatgpt-with-flan-t5-and-bert-on-the-test-set}{%
\section{Compare ChatGPT with FLAN-T5 and BERT on the test
set}\label{compare-chatgpt-with-flan-t5-and-bert-on-the-test-set}}

    \begin{tcolorbox}[breakable, size=fbox, boxrule=1pt, pad at break*=1mm,colback=cellbackground, colframe=cellborder]
\prompt{In}{incolor}{67}{\boxspacing}
\begin{Verbatim}[commandchars=\\\{\}]
\PY{n}{df\PYZus{}hightlight\PYZus{}metrics\PYZus{}llm\PYZus{}test} \PY{o}{=} \PYZbs{}
  \PY{n}{pd}\PY{o}{.}\PY{n}{concat}\PY{p}{(}\PY{p}{[}\PY{n}{df\PYZus{}hightlight\PYZus{}metrics\PYZus{}flan\PYZus{}t5\PYZus{}test} \PY{p}{,} \PY{n}{df\PYZus{}hightlight\PYZus{}metrics\PYZus{}chatgpt\PYZus{}test}\PY{p}{]}\PY{p}{,}
           \PY{n}{axis}\PY{o}{=}\PY{l+m+mi}{0}\PY{p}{)}

\PY{n}{df\PYZus{}hightlight\PYZus{}metrics\PYZus{}llm\PYZus{}test}\PY{o}{.}\PY{n}{sort\PYZus{}values}\PY{p}{(}\PY{n}{by}\PY{o}{=}\PY{p}{[}\PY{l+s+s2}{\PYZdq{}}\PY{l+s+s2}{f1\PYZus{}macro}\PY{l+s+s2}{\PYZdq{}}\PY{p}{]}\PY{p}{,}\PY{n}{ascending}\PY{o}{=}\PY{k+kc}{False}\PY{p}{)}\PY{o}{.}\PY{n}{reset\PYZus{}index}\PY{p}{(}\PY{n}{drop}\PY{o}{=}\PY{k+kc}{True}\PY{p}{)}
\end{Verbatim}
\end{tcolorbox}

            \begin{tcolorbox}[breakable, size=fbox, boxrule=.5pt, pad at break*=1mm, opacityfill=0]
\prompt{Out}{outcolor}{67}{\boxspacing}
\begin{Verbatim}[commandchars=\\\{\}]
              model\_type prompt\_type partition  f1\_macro   f1\_NONE  f1\_FAVOR  \textbackslash{}
0  llm\_chatgpt\_turbo\_3\_5    few\_shot      test  0.637211  0.563380  0.676923
1        llm\_flan-t5-xxl   zero\_shot      test  0.619033  0.583333  0.531250
2        llm\_flan-t5-xxl    few\_shot      test  0.591850  0.534653  0.519685
3  llm\_chatgpt\_turbo\_3\_5   zero\_shot      test  0.507138  0.435644  0.672269
4  llm\_chatgpt\_turbo\_3\_5         CoT      test  0.387721  0.360000  0.568421
5      llm\_flan-t5-large    few\_shot      test  0.278003  0.083333  0.198953
6      llm\_flan-t5-large   zero\_shot      test  0.212933  0.043478  0.181034

   f1\_AGAINST
0    0.671329
1    0.742515
2    0.721212
3    0.413502
4    0.234742
5    0.551724
6    0.414286
\end{Verbatim}
\end{tcolorbox}
        
    \begin{tcolorbox}[breakable, size=fbox, boxrule=1pt, pad at break*=1mm,colback=cellbackground, colframe=cellborder]
\prompt{In}{incolor}{68}{\boxspacing}
\begin{Verbatim}[commandchars=\\\{\}]
\PY{c+c1}{\PYZsh{} bert}
\PY{n}{df\PYZus{}hightlight\PYZus{}metrics\PYZus{}bert\PYZus{}test}\PY{o}{.}\PY{n}{sort\PYZus{}values}\PY{p}{(}\PY{n}{by}\PY{o}{=}\PY{p}{[}\PY{l+s+s2}{\PYZdq{}}\PY{l+s+s2}{f1\PYZus{}macro}\PY{l+s+s2}{\PYZdq{}}\PY{p}{]}\PY{p}{,}\PY{n}{ascending}\PY{o}{=}\PY{k+kc}{False}\PY{p}{)}\PY{o}{.}\PY{n}{reset\PYZus{}index}\PY{p}{(}\PY{n}{drop}\PY{o}{=}\PY{k+kc}{True}\PY{p}{)}
\end{Verbatim}
\end{tcolorbox}

            \begin{tcolorbox}[breakable, size=fbox, boxrule=.5pt, pad at break*=1mm, opacityfill=0]
\prompt{Out}{outcolor}{68}{\boxspacing}
\begin{Verbatim}[commandchars=\\\{\}]
                 model\_type partition  f1\_macro  f1\_NONE  f1\_FAVOR  f1\_AGAINST
0       vinai\_bertweet\_base      test    0.5797   0.5323    0.5401      0.6667
1  kornosk\_polibertweet\_mlm      test    0.5616   0.5440    0.4762      0.6645
2         bert-base-uncased      test    0.4748   0.4196    0.4275      0.5775
\end{Verbatim}
\end{tcolorbox}
        
    The results of the comparison between ChatGPT, FLAN-T5, and BERT models
across different prompts show that the prompting approach can outperform
a fine-tuned BERT, even when considering domain-specific pretrained
models like kornosk\_polibertweet\_mlm and vinai\_bertweet\_base.

ChatGPT with few-shot prompts achieves the highest macro-F1 scores on
the test set, followed by FLAN-T5-XXL with zero-shot and few-shot
prompts. Notably, even the open-source large language model, FLAN-T5,
reaches decent performance, offering a cost-effective alternative to
ChatGPT. However, it's important to note that using FLAN-T5 requires
access to GPUs with about 30GB of memory to handle its large size.

These large language models with prompting strategies deliver better
performance than the fine-tuned BERT variants, including the
domain-specific models. This demonstrates the potential of using prompts
with large language models, as they offer competitive performance
without requiring extensive labeled data for training.

    However, it is important to note that fine-tuning BERT has a distinct
advantage in that its performance can be straightforwardly improved by
collecting more labeled data. The performance of a fine-tuned BERT model
increases with the amount of labeled training data available. This is
not the case for large language models using prompting strategies, as
the number of examples that can be fit in a prompt is limited.

\begin{quote}
Note for advanced readers: To ensure that the prompting approach
benefits from a larger amount of training data, you can also fine-tune
an LLM on a substantial amount of labeled data. For more information,
refer to the bonus section below.
\end{quote}

    \begin{center}\rule{0.5\linewidth}{0.5pt}\end{center}

    \hypertarget{conclusions}{%
\section{Conclusions}\label{conclusions}}

    In conclusion, this tutorial has demonstrated how to implement stance
detection using ChatGPT and FLAN-T5 on the Abortion dataset. We've
explored the use of different prompt types and compared their
performance to BERT-based models, including domain-specific pre-trained
models. The results show that prompting large language models, like
ChatGPT and FLAN-T5, can outperform fine-tuned BERT models in this task,
even without extensive labeled data for training. However, it's
essential to consider the trade-offs, such as the monetary cost of using
ChatGPT and the memory requirements for utilizing FLAN-T5.

    \begin{center}\rule{0.5\linewidth}{0.5pt}\end{center}

    \hypertarget{bonus-for-eager-readers}{%
\subsection{Bonus for eager readers}\label{bonus-for-eager-readers}}

    In addition to using prompting strategies with large language models,
it's worth noting that fine-tuning LLMs is another viable approach to
improve their performance on specific tasks, including stance detection.
By fine-tuning an LLM on a task-specific dataset, the model can adapt to
the nuances of the data, better understand the domain-specific language,
and potentially yield higher performance.

OpenAI has an
\href{https://platform.openai.com/docs/guides/fine-tuning}{guide on how
to fine-tune GPT-3} on a specific task. It is also possible to fine-tune
open-source LLMs like FLAN-T5 for specific tasks. For example,
\href{https://medium.com/google-cloud/fine-tuning-flan-t5-xxl-with-deepspeed-and-vertex-ai-af499daf694d}{this
tutorial} demonstrates how to fine-tune FLAN-T5 for text classification.
One caveat is that fine-tuning LLMs, like fine-tuning a BERT model, also
requires a large amount of labeled data for training.

Another caveat is that fine-tining the GPT-3 model and using a
fine-tuned model can be expensive. Please see the
\href{https://openai.com/pricing}{OpenAI's pricing page} for more
details. On the other hand,
\href{https://medium.com/google-cloud/fine-tuning-flan-t5-xxl-with-deepspeed-and-vertex-ai-af499daf694d}{fine-tuning
FLAN-T5-XXL is GPU intensive} and requires about 680GB of GPU memory and
few days of training time.

    % Add a bibliography block to the postdoc

\end{document}